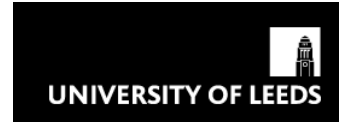

# Statistical Parsing by Machine Learning from a Classical Arabic Treebank

## Kais Dukes

Submitted in accordance with the requirements for the degree of
Doctor of Philosophy

The University of Leeds
School of Computing

September, 2013

# Publications

Chapters 4 to 10 in parts II, III and IV of this thesis are based on jointly-authored publications. I was the lead author and the co-authors acted in an advisory capacity, providing supervision and review. All original contributions presented here are my own.

**Part II - Modelling Classical Arabic**

The formal representations of Classical Arabic orthography, morphology and syntax presented in Chapters 4 to 6 are based on the following papers:

**Part III – Developing the Quranic Arabic Corpus**

The descriptions of the collaborative annotation methodology and online software platform in Chapters 7 and 8 are based on the following publications:

Kais Dukes, Eric Atwell and Nizar Habash (2013). Supervised Collaboration for Syntactic Annotation of Quranic Arabic. *Language Resources and Evaluation Journal (LREJ): Special Issue on Collaboratively Constructed Language Resources,* 47:1 (33-62).

Kais Dukes and Eric Atwell (2012). LAMP: A Multimodal Web Platform for Collaborative Linguistic Analysis. In *Proceedings of the Language Resources and Evaluation Conference (LREC)* (3268-3275). Istanbul.

Kais Dukes, Eric Atwell and Abdul-Baquee Sharaf (2010d). Online Visualization of Traditional Quranic Grammar using Dependency Graphs. In *Proceedings of the Foundations of Arabic Linguistics Conference*. Cambridge.

**Part IV – Statistical Parsing**

Chapters 9 and 10 discuss statistical parsing and machine learning experiments. These chapters form an expanded description of the work summarized in the following paper:

Kais Dukes and Nizar Habash (2011). One-step Statistical Parsing of Hybrid Dependency-Constituency Syntactic Representations. In *Proceedings of the International Conference on Parsing Technologies (IWPT)* (92-103). Dublin, Ireland.

# Acknowledgments

First and foremost, my sincere gratitude is owed to my two PhD supervisors, Eric Atwell at the University of Leeds and Nizar Habash at Columbia University in the City of New York.

During my work on this thesis over the last four years, I benefited immensely from Eric's expert advice on corpus annotation and computational linguistics. I am also deeply indebted to Eric for his encouragement to complete the parts of my work that led to peer-reviewed papers. He allowed me the level of intellectual freedom that I needed to make original contributions to new areas of research.

I owe my deepest appreciation to Nizar, who acted as an external supervisor. He provided expert guidance on Arabic morphological and syntactic theory and how best to approach the problem of Arabic statistical parsing using machine learning. His belief in the direction and quality of my work helped provide the motivation I needed to see my research through to completion.

My sincere gratitude and thanks are also directed to members of the research community who gave me invaluable advice, encouragement and support. Abdul-Rahman Adnan, Imran Alawiye, Mohammed Alyousef, Tim Buckwalter, Michael Carter, Teuku Edward, Lydia Lau, Katja Markert, Mazhar Nurani, Jonathan Owens, Fatma Said, Hind Salhi, Majdi Sawalha, Abdul-Baquee Sharaf, Wajdi Zaghouani and Mai Zaki deserve special mention.

I owe my gratitude to Ahmed El-Helw and Nour Sharabash for kindly donating and administrating the web servers used to host the Quranic Arabic Corpus. I would also like to acknowledge the hard work of the numerous volunteers who contributed their time and effort to continuously improve the annotations online.

Finally, I will be forever grateful for the love, kindness and support shown by my wife, Imen. Without her tireless patience and calming presence, our attempt to combine my part-time PhD with full-time work while raising two young happy children would never have been possible. From the bottom of my heart Imen, thank you for supporting me every step of the way.

# Abstract

Research into statistical parsing for English has enjoyed over a decade of successful results. However, adapting these models to other languages has met with difficulties. Previous comparative work has shown that Modern Arabic is one of the most difficult languages to parse due to rich morphology and free word order. Classical Arabic is the ancient form of Arabic, and is understudied in computational linguistics, relative to its worldwide reach as the language of the Quran. The thesis is based on seven publications that make significant contributions to knowledge relating to annotating and parsing Classical Arabic.

Classical Arabic has been studied in depth by grammarians for over a thousand years using a traditional grammar known as *i'rāb* (اعراب!). Using this grammar to develop a representation for parsing is challenging, as it describes syntax using a hybrid of phrase-structure and dependency relations. This work aims to advance the state-of-the-art for hybrid parsing by introducing a formal representation for annotation and a resource for machine learning. The main contributions are the first treebank for Classical Arabic and the first statistical dependency-based parser in any language for ellipsis, dropped pronouns and hybrid representations.

A central argument of this thesis is that using a hybrid representation closely aligned to traditional grammar leads to improved parsing for Arabic. To test this hypothesis, two approaches are compared. As a reference, a pure dependency parser is adapted using graph transformations, resulting in an 87.47% F1-score. This is compared to an integrated parsing model with an F1-score of 89.03%, demonstrating that joint dependency-constituency parsing is better suited to Classical Arabic.

The Quran was chosen for annotation as a large body of work exists providing detailed syntactic analysis. Volunteer crowdsourcing is used for annotation in combination with expert supervision. A practical result of the annotation effort is the corpus website: http://corpus.quran.com, an educational resource with over two million users per year.

بِسْمِ ٱللَّهِ ٱلرَّحْمَٰنِ ٱلرَّحِيمِ

سُبْحَانَكَ لَا عِلْمَ لَنَا إِلَّا مَا عَلَّمْتَنَا إِنَّكَ أَنتَ ٱلْعَلِيمُ ٱلْحَكِيمُ

'Glory be to thee! We have no knowledge except what you have taught us.
Indeed it is you who is the all-knowing, the all-wise.'

A prayer of the angels
–The Quran, verse (2:32)

# Contents

# List of Figures

# List of Tables

# List of Abbreviations

The following table lists the meanings of the abbreviations used in this thesis. The page on which each abbreviation is defined is also given.

| Abbreviation | Meaning | Page |
|---|---|---|
| ATB | Arabic Treebank | 33 |
| BAMA | Buckwalter Arabic Morphological Analyzer | 14 |
| CATiB | Columbia Arabic Treebank | 28 |
| CoNLL | Computational Natural Language Learning Conference | 34 |
| ELAS | Extended Labelled Attachment Score | 223 |
| FSM | Finite State Machine | 19 |
| HMM | Hidden Markov Model | 233 |
| HSP | Hybrid Statistical Parser | 181 |
| JSP | Java Server Pages | 162 |
| LAMP | Linguistic Analysis Multimodal Platform | 161 |
| LAS | Labelled Attachment Score | 35 |
| LDC | Linguistic Data Consortium | 24 |
| MSA | Modern Standard Arabic | 2 |
| NLG | Natural Language Generation | 166 |
| NLP | Natural Language Processing | 45 |
| NLTK | Natural Language Toolkit | 233 |
| NS | Nominal Sentence | 118 |
| OOP | Object Oriented Programming | 67 |
| POS | Part of Speech | 5 |
| SALMA | Standard Arabic Language Morphological Analysis | 18 |
| SVM | Support Vector Machine | 4 |
| SVO | Subject-Verb-Object | 34 |
| VS | Verbal Sentence | 118 |

# Part I: Introduction and Background

The worthwhile problems are the ones you can really solve or help solve, the ones you can really contribute something to... No problem is too small or too trivial if we can really do something about it.

*– Richard Feynman*

# 1    Introduction

## 1.1    Motivation

The topic of this thesis is statistical parsing for Classical Arabic using machine learning. This work includes constructing a formal grammatical representation and developing the Quranic Arabic Corpus as a dataset to test parsing algorithms.

Parsing is the process of determining the syntactic structure of a sentence. Algorithms for parsing are researched in computational linguistics, an interdisciplinary field that combines computer science, statistical modelling and mathematical logic to process natural language. Analyzing the syntactic structure of a sentence through parsing can be a prerequisite step for deeper processing tasks such as machine translation (Huang et al., 2006; Zollmann and Venugopal, 2006), semantic analysis (Carreras and Màrquez, 2005) and task execution, in which machines execute physical tasks using natural language commands (Kuhlmann et al., 2004).

My own motivation for developing a parser for Classical Arabic is that it is a less-studied language in computational linguistics. Classical Arabic is a 1,600 year-old ancient language that is the direct ancestor of Modern Standard Arabic (MSA) spoken today. Although a variety of parsers exist for Modern Arabic, almost no previous work has been done for statistical parsing of Classical Arabic, the original language of the Quran.

Figure 1.1 shows an example verse (*āyah*) from the Quran, written in Classical Arabic from right-to-left using a connected cursive script. Arabic, together with Hebrew, Turkish and Finnish are examples of languages that are morphologically rich and highly inflected. The complexity of these morphologically rich languages poses special challenges to parsing work.

Figure 1.1:
Verse (6:76) from
the Quran.

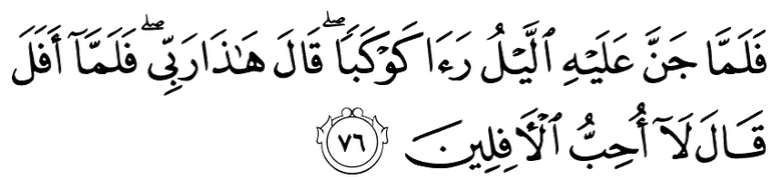

**(6:76)** *When the night covered him, he saw a star. He said, 'This is my Lord.' But when it set, he said, 'I do not love those that disappear.'*

The grammatical system explored in this thesis is *i'rāb* (إعراب), a 1,000 year-old comprehensive linguistic theory that describes Classical Arabic's phonology (the interaction of the units of sound that make up speech), morphology (the study of the substructure of words), syntax (the structure of sentences) and discourse analysis (the study of the discourse structures used in communication). This linguistic theory developed independently of Western thought and has influenced modern theories of syntax (Versteegh, 1997b; Baalbaki, 2008). For example, along with Panini's *Ashtadhyayi* for Classical Sanskrit, *i'rāb* is considered to be one of the origins of modern dependency grammar (Kruijff, 2006; Owens, 1988).

My motivation for this thesis originated in a personal interest in the linguistic structure of the Quran. Classical Arabic grammar is widely studied in the Islamic world due to the importance of the Quran, and several grammatical works exist that provide detailed analysis of its syntax (Salih, 2007; Darwish, 1996). I have often wondered if this analysis could be derived through statistical models using machine learning. Could algorithms learn from example data and reproduce the historical analyses of traditional grammarians? My interest in this idea led me to research statistical methods for parsing Classical Arabic, inspired by Arabic's long and rich grammatical tradition.

## 1.2   Research Questions

### 1.2.1 Is Statistical Parsing Viable for Classical Arabic?

Over the last two decades, statistical parsers have been used as an alternative to, and in combination with, previous rule-based parsers (Marcus et al., 1993; Abney, 1996). In contrast to rule-based parsers, statistical parsers learn a grammatical model from a treebank – a syntactically annotated corpus of example sentences. A variety of methods are used for statistical parsing, ranging from maximum entropy techniques for phrase-structure representations (Charniak, 2000) to support vector machines (SVMs) for dependency grammar (Nivre et al., 2007b).

Most research into statistical parsing has focused on English, with the best models achieving up to 92% accuracy (McClosky, Charniak and Johnson, 2006). Adapting these parsing models to other languages has been less successful. For example, adapting Bikel's parser to Chinese has resulted in an F1-score of 79.9% (Chiang and Bikel, 2002). Similarly, results from the CoNLL shared task on multilingual dependency parsing show that Modern Arabic is one of the most challenging languages to parse (Nivre et al., 2007a). This is in part due to Arabic's complex morphology. As noted by Soudi et al. (2007):

> The morphology of Arabic poses special challenges to computational natural language processing systems. The exceptional degree of ambiguity in the writing system, the rich morphology, and the highly complex word formation process of roots and patterns all contribute to making computational approaches to Arabic very challenging.

It is thus not immediately obvious if parsing Classical Arabic is tractable using purely statistical methods. The primary research question that will be answered in this thesis is to determine whether or not statistical parsing for Classical Arabic is a viable approach for achieving state-of-the-art parsing accuracy.

## 1.2.2 Is a Hybrid Representation Suitable for Parsing?

In modern linguistics, there is no universally accepted grammatical theory for representing syntactic information. Examples of different theories include transformational grammar (Chomsky, 1970), dependency grammar (Mel'čuk, 1988), functional grammar (Halliday and Matthiessen, 2006) and combinatory categorial grammar (Steedman, 2000). For annotation, multiple representations can be used. The two main representations used by treebanks are constituency phrase-structure (using relations between clauses and their constituents), and dependency grammar (using dependency relations between words). This thesis describes a novel hybrid representation, combining aspects of both dependency and constituency syntax. The motivation for using a hybrid approach for Classical Arabic is to remain closely aligned to traditional analyses of Quranic grammar.

This section introduces the hybrid representation by comparing to two existing representations. The following two diagrams annotate the same English sentence. Figure 1.2 is a constituency tree, with preterminal nodes annotated using an example POS (part-of-speech) tagset (PRON = pronoun, MOD = modal, NEG = negative particle, V = verb, PUNC = punctuation). Non-terminals are phrase tags (NP = noun phrase, VP = verb phrase, ADVP = adverb phrase, S = sentence).

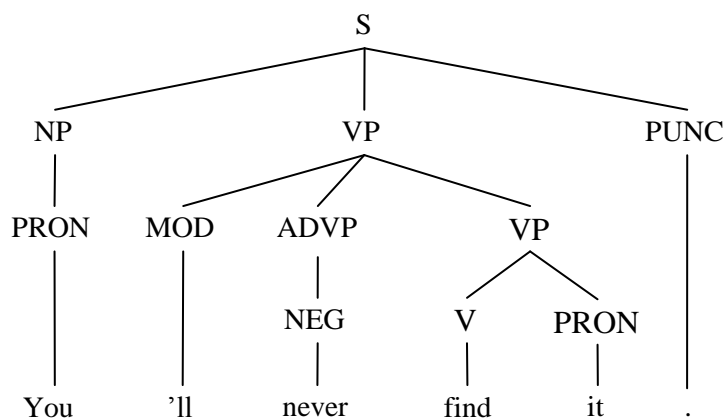

Figure 1.2: Phrase-structure parse tree using a simple grammar.

In contrast to the constituency approach, dependency theory represents sentence structure using binary dependencies between pairs of words. In Figure 1.3, the example sentence has been annotated using the same part-of-speech tags as Figure 1.2, but using an alternative dependency tagset for syntax (subj = subject, obj = object, mod = modal, neg = negation). Unless otherwise stated, dependency diagrams in this thesis follow the convention of dependent nodes pointing to head nodes, the same convention used to annotate Classical Arabic in the Quranic Arabic Corpus.[1]

Although these two diagrams annotate an English sentence, they illustrate a task that is more challenging in Arabic – morphological segmentation. In the diagrams, terminal nodes are not words but segments of words. For example, the word 'you'll' has been segmented into the pronoun 'you' and the modal 'will'. In English, only a minority of words such as contractions require segmentation for treebank construction. This contrasts with Arabic, where morphological analysis is complex, as many words require segmentation into multiple morphemes that each have different syntactic roles in sentence structure.

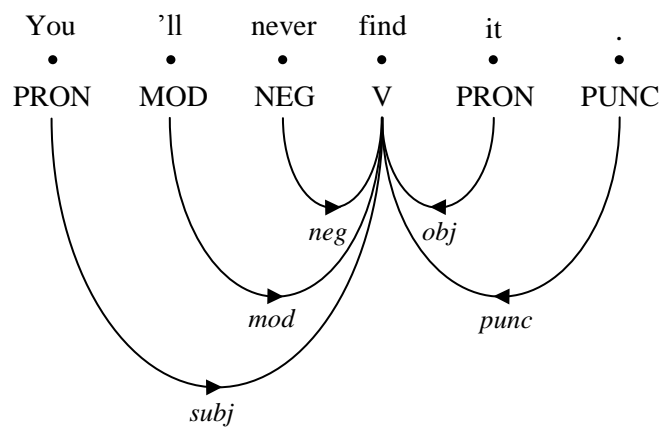

Figure 1.3: Pure dependency graph for an English sentence.

---

[1] Appendix A describes the graph layout algorithm used to produce syntax diagrams in this thesis and for the online Quranic Treebank (http://corpus.quran.com/treebank.jsp).

The previous diagrams illustrated two different representations for syntactic annotation. For parsing, the choice of representation used to model a language is fundamental to the operation of a parser. It constrains possible parsing algorithms and has a direct effect on parsing accuracy. This is highlighted by the recent use of model adaptation, where existing statistical parsers designed for English have been retrained for Modern Arabic (Green and Manning, 2010). Because Arabic contains linguistic constructions not found in English, this has resulted in parsing underperformance (described further in section 2.4).

قَالَ هَٰذَا رَبِّي

*He said, 'This is my Lord.'*

Figure 1.4: Extract from verse (6:76).

In this thesis, Classical Arabic syntax will be described using an alternative representation based on Arabic's grammatical tradition. However, despite its prominence in Arabic linguistic works, the grammatical rules of *i'rāb* have previously lacked a formal representation, making computational modelling of Classical Arabic grammar challenging. In contrast to formal methods, traditional analysis is described by grammarians through prose. For example, the syntax of verse (6:76) shown in Figure 1.4 is described by Salih (2007) using the following analysis (translated from Arabic):

> In this verse, 'said' is a perfective verb, whose subject is a dropped pronoun of the form 'he'. The noun 'lord' is in the nominative case and is the predicate of the demonstrative pronoun 'this'. The suffixed pronoun 'my' attached to the noun is a possessive clitic. The nominal sentence, headed by the demonstrative pronoun, is governed by the verb 'said' as a direct object.

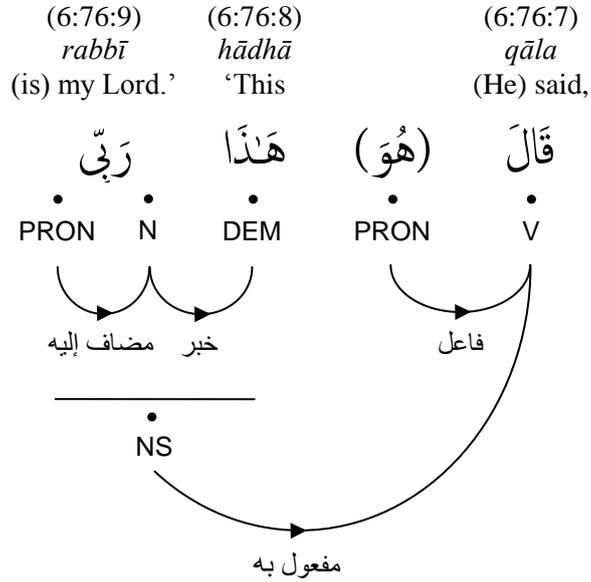

Figure 1.5: Hybrid dependency-constituency graph.

A hybrid representation can be used to formalize this analysis. For example, Salih analyses the phrase 'This is my Lord' as a dependency of the verb 'said'. Although *i'rāb* describes dependencies between morphological segments, this shows that the grammar also describes dependencies between words and phrases. Arabic grammatical theory could be interpreted as either a pure dependency or constituency representation, but a hybrid representation more closely aligns to traditional analysis. Figure 1.5 annotates verse (6:76) of the Quran using the hybrid formalism that will be presented in Chapter 6. The diagram shows a graph with nodes that are either morphological segments with part-of-speech tags (V = Verb, PRON = Pronoun, DEM = Demonstrative, N = Noun) or phrase nodes (NS = Nominal Sentence). Edges are tagged with dependency relations such as object, subject and predicate, shown in Arabic using traditional terminology.

The second research question addressed in this thesis is to determine if a hybrid dependency-constituency representation is better suited to parsing Classical Arabic compared to a pure dependency representation. This question will be answered by annotating the Quran using the hybrid representation and comparing the two approaches to parsing.

### 1.2.3 Can Crowdsourcing be used for Annotating Arabic?

Of potential wider interest beyond Classical Arabic parsing is the use of crowdsourcing to construct the annotated treebank which will be used to train a statistical parser. Statistical parsers require high-quality training data in the form of sentences annotated according to a chosen syntactic representation. A typical annotation methodology involves paid experts who perform offline annotation. However, the alternative of online collaboration has recently emerged as a viable alternative to more conventional approaches for developing tagged corpora (Chamberlain et al., 2009). Online collaboration has been used for a wide variety of linguistic tagging tasks ranging from named-entity resolution of international hotels (Su et al., 2007) to syntactic annotation of Latin and Ancient Greek texts (Bamman et al., 2009).

In this thesis, crowdsourcing will be used to develop the first treebank for Classical Arabic. Following initial automatic tagging, the main task that volunteer annotators are asked to perform is to proofread morphological and syntactic annotation. Annotators verify this against gold standard analyses from Arabic reference works of Quranic grammar. Although the reference material contains equivalent grammatical information, because its content is unstructured prose that is not easily machine readable, a manual cross-checking stage is required.

The third research question to be investigated in this thesis is to determine if a form of crowdsourcing can be used as an annotation methodology for producing high-quality tagging of Classical Arabic. Volunteer crowdsourcing can be cost effective, but consistency and accuracy need to be ensured if the data is to be used for statistical modelling. In the Quranic Arabic Corpus, expert annotators are promoted to a supervisory role, reviewing and discussing the work of others online using an interactive message board forum. In this thesis, the collaborative annotation methodology will be compared to the alternative of crowdsourcing without expert supervision, and evaluated for accuracy.

## 1.3   Original Contributions of the Thesis

### 1.3.1 Theoretical Contributions

The main theoretical contributions that will be presented in the thesis are:

- The first formalism of *i'rāb* and the first morphosyntactic annotation scheme for Classical Arabic. This includes a novel hybrid dependency-constituency representation, with a fine-grained tagset for parts-of-speech and phrases, morphological features and dependency relations.

- The first evaluation of a methodology for online supervised collaboration for Arabic annotation. This methodology combines crowdsourcing with expert supervision to produce highly-quality annotation for Arabic text.

### 1.3.2 Practical Contributions

The main practical contributions to be presented are:

- The first treebank for Classical Arabic. This includes manually-verified morphological annotation for 77.4K words tagged with 783K feature-values together with syntactic tagging for 37.6K words. Supplementary annotation includes named-entity tagging, an ontology of concepts, a word-by-word English translation and a morphological lexicon.

- The first web-based platform for capturing, editing and visualizing Arabic morphosyntactic annotations online. This includes a comprehensive set of supplementary linguistic tools to access and search corpus annotations.

- The first statistical parser for Classical Arabic. In addition, this is also the first dependency-based statistical parser in any language that handles elliptical structures, dropped pronouns and a hybrid representation.

## 1.4   Thesis Outline

This thesis is divided into five parts with 12 chapters, shown in Figure 1.6 below:

---

**Part I: Introduction and Background**
    1      Introduction
    2      Literature Review
    3      Historical Background
**Part II: Modelling Classical Arabic**
    4      Orthographic Representation
    5      Morphological Representation
    6      Syntactic Representation
**Part III: Developing the Quranic Arabic Corpus**
    7      Annotation Methodology
    8      Annotation Platform
**Part IV: Statistical Parsing**
    9      Hybrid Parsing Algorithms
    10    Machine Learning Experiments
**Part V: Further Work and Conclusion**
    11    Uses of the Quranic Arabic Corpus
    12    Contributions and Future Work

---

Figure 1.6: Organization of thesis chapters.

Part I provides relevant background information. Following this introductory chapter, Chapter 2 contains the literature review, discussing Arabic treebanks and annotation methodologies. Recent morphological analyzers and statistical parsers for Arabic are also compared. Relevant historical background on the Arabic linguistic tradition is provided in Chapter 3.

Part II presents a formal model of Classical Arabic, with a representation for orthography (Chapter 4), morphology (Chapter 5) and syntax (Chapter 6). The representation is presented both as a well-defined set-theoretic description and as an annotation scheme.

Part III describes the development of the Quranic Arabic Corpus. Chapter 7 discusses the annotation methodology of supervised collaboration. Chapter 8 describes the web-based software platform used to capture annotations online and the supplementary linguistic tools developed for annotators.

Part IV focuses on statistical parsing. In Chapter 9, two algorithms for hybrid parsing are compared: a multi-step process using graph transformations and a novel one-step algorithm without post-processing. Chapter 10 evaluates the parser using statistical models induced from the treebank by machine learning. A series of experiments consider the effect of using different morphological features for parsing and the results are compared to recent parsing work for Modern Arabic.

Part V concludes the thesis. Chapter 11 describes recent research that has made use of the annotations in the Quranic Arabic Corpus and Chapter 12 summarizes the main contributions and presents recommendations for future research. The last chapter concludes with a discussion of the challenges and limitations of the work as well as its implications for theoretical and computational linguistics.

# 2 Literature Review

## 2.1 Introduction

Arabic is a major world language. Together with Chinese, English, French, Russian and Spanish, it is one of the six official languages of the United Nations. Including its literary form and its various dialects, it is the first language for 280 million native speakers across the Middle East and North Africa (Procházka, 2006). Classical Arabic is the liturgical language of prayer and worship for the world's Muslim population, estimated at between 1.57 billion (Lugo, 2009) and 1.65 billion people (Kettani, 2010), up to a quarter of the world's population.

Arabic has recently become the focus of an increasing number of natural language processing projects (Habash, 2010). This review describes relevant work in four areas: morphology, syntax, parsing and annotation methodologies. The first part of the review describes recent work for Arabic morphology, including an analysis of the limitations of previous morphological work for the Quran. To provide context for the syntactic representation developed for the Quranic Arabic Corpus, the review compares the Penn, Prague and Columbia Arabic treebanks, focusing on the approaches used to formalize Arabic syntax.

Following the description of morphological and syntactic projects, parsing work for Arabic is reviewed, describing how different syntactic representations affect accuracy. Attention is also given to dual dependency-constituency parsing work for German and Swedish, as these methods are relevant to the hybrid parsing work described in Chapter 9. Models for ellipsis are also reviewed, which are often ignored in parsing work but are developed in this thesis. The review of parsing work concludes with a discussion of recent work for Hebrew. This related Semitic language presents similar challenges to statistical parsing, and illustrates recent trends in parsing that are also applicable to Arabic.

Methodologies for other relevant annotation projects beyond Arabic are also reviewed, comparing offline expert annotation to collaborative online annotation and crowdsourcing. Finally, the conclusion summarizes the implications of the reviewed work in relation to the thesis research questions.

## 2.2   Arabic Morphological Analysis

This section of the review discusses different approaches to Arabic computational morphology. Morphological analysis tasks for Arabic include segmentation (the division of compound word-forms into prefixes, stems and suffixes), part-of-speech tagging (assigning a tag to each morphological segment), lemmatization (assigning lemmas to stems) and the identification of the roots and patterns used in inflected Arabic word-forms.

### 2.2.1 The Buckwalter Arabic Morphological Analyzer

The Buckwalter Arabic Morphological Analyzer (BAMA) is a freely available rule-based morphological analyzer, developed to perform initial tagging of Penn Arabic Treebank (Buckwalter, 2002). This previous work is relevant because an analyzer based on BAMA's algorithm will be used in Chapter 7 to perform initial morphological tagging for the Quranic Arabic Corpus.

BAMA's analysis algorithm depends on its lexicon. Version 2.0 of the analyzer contains 78,839 lexical entries representing 40,219 lemmas. This data is organized into segment tables with entries for prefixes, stems and suffixes, and compatibility tables listing permitted combinations of segments. The part-of-speech tagset used in these dictionary files is the same as that used for the Penn Arabic Treebank.

The morphological analyzer processes undiacritized Arabic text, returning several possible analyses for each word. Its analysis algorithm generates all possible segmentations into prefixes, stems and suffixes. For each combination, the segment tables are checked to determine if the analysis is linguistically plausible. The resulting filtered analyses are output with full diacritization and morphological annotation, augmented by features from the lexicon.

BAMA is widely used by the Arabic computational research community for a variety of tasks including diacritic restoration (Ananthakrishnan et al., 2005), automatic speech recognition and machine translation (Soltau et al., 2007) and named entity recognition (Farber et al., 2008). Its lexicon has also been used as one source of data for the Arabic version of Google's online translation service. However, BAMA is limited by producing multiple analyses for each word. To overcome this limitation, BAMA's lexicon has been used as the basis for more sophisticated statistical disambiguation systems, described in the next section.

## 2.2.2 Lexeme and Feature Representations

Habash (2007a) notes that Arabic morphological resources use different, often incompatible, representations to model morphology. Electronic dictionaries and lexicons are based around headwords and lemmas. Stemmers focus on extracting the stems of word-forms and deeper analyzers extract roots and patterns. Habash proposes a lexeme-plus-feature representation to relate these different resources. This work is relevant to Classical Arabic because the Quranic Arabic Corpus uses a similar representation for morphological annotation, as described in Chapter 5.

For morphologically-rich languages such as Arabic, the term lexeme is used to denote an abstract grouping of words that share the same base meaning, but differ through inflection. A lemma, also known as a citation form, is a conventional choice of one word that represents a lexeme. Dictionary entries are usually organized by lemma. For example, in English the set of words 'eat', 'eats', 'ate' and 'eating' form a lexeme, with 'eat' as the lemma.

| Feature | Value | Definition |
|---|---|---|
| Part of Speech | POS:N | Noun |
| | POS:PN | Proper Noun |
| | POS:V | Verb |
| | POS:AJ | Adjective |
| | POS:AV | Adverb |
| | POS:PRO | Pronoun |
| | POS:P *and others* | Preposition |
| Conjunction | w+ | *'and'* |
| | f+ | *'and', 'so'* |
| Preposition | b+ | *'by', 'with'* |
| | k+ | *'like'* |
| | l+ | *'for', 'to'* |
| Verbal Particle | s+ | *'will'* |
| | l+ | *'so as to'* |
| Definite Article | Al+ | *'the'* |
| Verb Aspect | PV | Perfective |
| | IV | Imperfective |
| | CV | Imperative |
| Voice | PASS | Passive |
| Gender | FEM | Feminine |
| | MASC | Masculine |
| Subject | S:PerGenNum | Person = {1, 2, 3} |
| Object | O:PerGenNum | Gender = {M, F} |
| Possessive | P:PerGenNum | Number = {S, D, P} |
| Mood | MOOD:I | Indicative |
| | MOOD:S | Subjunctive |
| | MOOD:J | Jussive |
| Number | SG | Singular |
| | DU | Dual |
| | PL | Plural |
| Case | NOM | Nominative |
| | ACC | Accusative |
| | GEN | Genitive |
| Definiteness | INDEF | Indefinite |
| Possession | POSS | Possessed |

Table 2.1: Features used in ALMORGEANA's morphological representation.

The ALMORGEANA system described by Habash (2007a) uses lexemes and features to provide bidirectional morphological analysis and generation, suitable for a variety of processing tasks, such as machine translation. The system utilizes a lexicon based on dictionary data from BAMA, but applies a different algorithm to perform morphological processing. In ALMORGEANA, the BAMA segment tables are converted to the lexeme-plus-feature representation. Table 2.1 (page 16) lists the converted morphological features. Figure 2.1 below illustrates how these features are used to represent the morphology of the compound Arabic word-form *lilkutubi* (translated as '*for the books*').

[kitAb_1 POS:N PL Al+ l+]

للكتب

'*for the books*'

Figure 2.1: Lexeme-plus-feature representation for an Arabic word.

The lexeme for this surface form is represented by the lemma *kitāb*, displayed using Buckwalter transliteration as kitAb_1. The suffix _1 is part of a numbering scheme used to distinguish word senses with the same name. Four features follow the lemma. POS:N is the part-of-speech tag for nouns, and PL denotes a plural word. Al+ indicates that the word-form has the Arabic *al-* prefix to denote definiteness ('the'), and l+ indicates the *lām* prefixed preposition ('for').

Like the Buckwalter analyzer, ALMORGEANA outputs several possible morphological analyses for each input Arabic word. Habash and Rambow (2005) extend the system to select a statistically most-probable analysis. Using data from the Penn Arabic Treebank converted to the lexeme-plus-feature representation, they build a statistical model to rank possible analyses using support vector machines trained to recognize individual morphological features. Testing against the Penn Treebank, they report high accuracy scores of 99.3% for morphological segmentation at word-level, and 98.1% for part-of-speech tagging over all tokens, using a reduced tagset.

Based on this work, Habash, Rambow and Roth (2009b) describe a toolkit consisting of two Arabic morphological systems, MADA and TOKAN. Like ALMORGEANA, the toolkit utilizes the BAMA lexicon. MADA (Morphological Analysis and Disambiguation for Arabic) is a statistical morphological analyzer that selects the best possible BAMA analysis using weighted predicted features. TOKAN is a flexible Arabic tokenizer that provides morphological segmentation of Arabic words according to a number of possible tokenization schemes. The toolkit has been used for a variety of further work including English-to-Arabic machine translation (Badr et al., 2008) and named entity recognition (Farber et al., 2008; Benajiba et al., 2008).

Compared to the Buckwalter Analyzer, this toolkit is attractive because it produces a single morphological analysis for each Arabic word. The use of a lexeme-plus-feature representation is notable for providing a computational model of Arabic morphology that is flexible enough to support different processing tasks. This representation will be extended to Classical Arabic morphology in Chapter 5.

## 2.2.3 Fine-Grained Morphological Analysis

In contrast to previous work, the SALMA tagger (Standard Arabic Language Morphological Analysis) uses a more fine-grained morphological tagset based on concepts from the Arabic linguistic tradition (Sawalha and Atwell, 2010; Sawalha, Atwell and Abushariah, 2013). This work compares to the annotation presented in this thesis, which is also fine-grained.

The SALMA tagger utilizes a lexicon of inflected surface forms containing 2.7 million vowelized word-root pairs, built by combining 23 Arabic dictionaries. Arabic text is annotated using a set of 22 morphological features that include part-of-speech, gender, number, person, case, mood, definiteness, voice, emphasis, transitivity, variability, roots and verb structure. The tagging algorithm segments words by applying a sequence of regular expressions to produce a list of candidate analyses. Segmented stems are matched to the lexicon to extract possible roots. A pattern database consisting of 2,730 patterns for verbs and 985 for nouns is used to search for appropriate root-pattern pairs. Morphological features are then annotated using the lexicon.

Sawalha et al. (2013) measure the tagger's accuracy by manually annotating a gold-standard dataset of 2,000 words using samples from two corpora. For Classical Arabic, they annotate the morphological analysis of the Quran by Dror et al. (2004), described in the next section. For Modern Arabic they use data from the Corpus of Contemporary Arabic (Al-Sulaiti and Atwell, 2006). For a set of 15 morphological features, they report an estimated accuracy score of 98.53% for tagging Modern Arabic and 90.1% for Classical Arabic.

This work demonstrates that automatic fine-grained morphological analysis of Arabic is possible. The morphological representation in Chapter 5 will also use a fine-grained tagset based on traditional grammar. It differs by using an alternative set of tags with morphological features developed specifically for Classical Arabic and designed to integrate with a syntactic representation.

## 2.2.4 Finite State Morphological Analysis of the Quran

This section describes the use of Finite State Machines (FSMs) to annotate the Arabic morphology of the Quran (Dror et al., 2004). To the best of the author's knowledge, this work is the only other wide-coverage computational analysis of Classical Arabic morphology, before the new work presented in this thesis. However, unlike the Quranic Arabic Corpus, the FSM analysis has not been manually verified by expert annotators. Dror et al. provide several different possible analyses for each word in the Quran, but do not disambiguate these to bring their annotations up to gold-standard level.

Their approach uses finite state computing using FSMs. These are abstract mathematical models of computation that consist of multiple states, together with rules that determine transitions between states. They have been applied to a wide variety of morphologically-rich languages, for which lexicons and morphological rules are developed manually by linguistic experts and encoded as state transition (Roche and Schabes, 1997; Beesley and Karttunen, 2002). The output of FSM systems are typically in a lexeme-plus-feature representation. In the description of their system for Classical Arabic, Dror et al. note that the language of the Quran remains relatively unexplored in contrast to Modern Arabic:

> Except for isolated efforts, little has been done with computer-assisted analysis of the text. Thus, for the present, computer-assisted analysis of the Quran remains an intriguing but unexplored field.

Their FSM analysis utilizes a new morphological lexicon based on the Quranic concordance by Abdalbaqi (1987). The lexicon associates lexemes with roots and patterns, and consists of 2,500 noun-forms, 100,000 possible verb bases and several hundred closed-class words. The verb bases were generated automatically by applying a list of Arabic word patterns to the roots in the Quran. As a result, most of the verbs bases in the lexicon do not occur in the text. To perform morphological analysis, an FSM consisting of approximately 300 hand-written rules for verbs and 50 rules for nouns are used to generate a list of possible analyses for each word in the Quran. In their evaluation, Dror et al. note that they do not perform full morphological disambiguation to select a single analysis for each word. However, by performing manual verification on a 1,250 word sample of the Quran, they estimate that 86% of words have a correct morphological analysis in the list of possible outputs produced by their analyzer.

This work is notable for being the first automatic morphological analysis of the Quranic text. However, their analysis has three limitations. Without manual correction, the annotations cannot be considered to be of gold-standard. Secondly, the Classical Arabic script of the Quran is not used, which makes it difficult to relate their work to other Arabic computational resources. Instead a phonetic transcription into the English alphabet is used as their orthographic representation. Thirdly, they do not publish a well-defined annotation scheme. Although they provide example output for their analyzer, they do not fully describe their tagset or list their set of morphological features. However, this could be inferred by processing their annotations to build up a list of possible tags. These limitations will be addressed in this thesis by providing manually-verified annotation using a well-defined morphosyntactic representation. To address the limitations with their approach to orthography, a new orthographic representation for Classical Arabic script that is convertible to Unicode will be presented in Chapter 4.

## 2.3  Arabic Syntactic Treebanks

Over the last several decades, the development and use of annotated corpora has grown to become a major focus of research for both linguistics and computational natural language processing. Corpora provide the empirical evidence that is used to advance various theories of language (Sampson and McCarthy, 2005). They are also used by computational linguists to engineer state-of-the-art natural language systems and resources such as electronic lexicons (Hajič et al., 2003; Kucera and Francis, 1967) and part-of-speech taggers (Brants, 2000a; Spoustová et al., 2009; Søgaard, 2011). Treebanks are annotated corpora that include morphological and syntactic annotation. This section reviews previous work for developing the three major treebanks for Arabic: The Penn, Prague and Columbia Arabic treebanks.

### 2.3.1 The Penn Arabic Treebank

The Penn English Treebank (Marcus, Santorini and Marcinkiewicz, 1993) was the first large-scale syntactic annotation project in any language, and helped introduce an alternative methodology for parser construction. Parsers that had previously been developed using hand-written grammatical rules were supplemented by parsers using statistical models induced from treebank data (Collins, 1999; Charniak, 2000; Nivre et al., 2007b). Over the last two decades, the Penn Treebank has remained one of the standard datasets for benchmarking English parsing, with state-of-the-art statistical parsers achieving F1-scores of 90-92% against Penn Treebank data.

For Modern Arabic, The Penn Arabic Treebank (Maamouri et al., 2004) is a related project designed to support the development of data-driven morphological analyzers and syntactic parsers. This project is important as it is the first treebank for the Arabic language. It uses the same constituency representation as the English Treebank, with the same tags used to annotate phrase structure. Maamouri et al. (2004) argue that using the English tagset for Arabic makes it easier to train annotators and that existing linguistic tools for English can be reused, simplifying the annotation process.

However, after the initial release of the treebank several constituency parsers previously developed for English were adapted to Arabic. Compared to English, the Arabic Treebank has been found to be more challenging to parse, with parsers achieving lower F1-scores of 74-83%. Recent work has shown that the treebank's choice of constituency representation has affected both parsing accuracy and annotation consistency (Kulick et al., 2006; Green and Manning, 2010). Section 2.4 reviews this parsing work and describes the causes of underperformance.

Figure 2.2 (overleaf) shows an example tree from the Penn Arabic Treebank annotated using constituency syntax. As per the annotation guidelines (Bies and Maamouri, 2003), this tree is shown in bracketed form and annotates a sentence that is a single Arabic verb stem with attached clitics. The word-form has been segmented into four morphemes shown both in Arabic script and Buckwalter Transliteration. In the parse tree, the tags are the same as that used for the Penn English Treebank (S = sentence, VP = verb phrase, PRT = particle, NP-SBJ = noun phrase / subject, NP-OBJ = noun phrase / object). The tree also contains an empty category denoted by an asterisk (*). In Arabic, the subjects of verbs are often dropped pronouns and are implied by the verb's morphological inflection features. In comparison to the work in this thesis, the Penn Arabic Treebank is the only other Arabic resource to annotate elliptical structure.

```
(S wa- و
   (VP (PRT -sa- س
            -tu+$Ahid+uwna- نُشاهِدُون
      (NP-SBJ *)
      (NP-OBJ -hA ها)))
```

وستشاهدونها

'*and you will observe her*'

Figure 2.2: Constituency tree from the Penn Arabic Treebank.

The first version of the Penn Arabic Treebank was annotated over a three year period using a two-stage process. The first stage is morphological annotation, where each sentence is processed using BAMA (described previously in section 2.2.1), to produce a list of possible morphological segmentations with part-of-speech tags, lemmas and morphological features for each word. Following automatic tagging, morphological annotation is manually corrected by paid linguistic experts who select the most suitable analysis from the list of available possibilities. The second stage is syntactic annotation. Bikel's parser is used to generate a constituency tree for each sentence using the reviewed morphological annotation (Bikel, 2004a). The constituency trees are then reviewed and corrected by annotators. Using this two-stage process, the initial release of the treebank contained morphosyntactic annotation for approximately half a million words of Arabic (Maamouri et al., 2004).

For newer versions of the Penn Arabic Treebank, Maamouri et al. (2008) have suggested changes to the annotation scheme to improve parsing accuracy. They note that annotation inconsistencies in the Arabic treebank arise when expert annotators, who are familiar with traditional Arabic grammar and concepts from *i'rāb*, attempt to interpret their analyses using an annotation scheme originally designed for English. They propose a revised set of guidelines that include new tags to better represent the fine-grained distinctions of Arabic syntax. These changes align the tagset more closely to traditional concepts already familiar to annotators, such as the traditional categorization of nominals and particles. This compares to the work presented in this thesis, which uses a tagging scheme based on traditional grammar, but using an alternative hybrid syntactic representation. In contrast, the new guidelines for the Penn Arabic Treebank fall short of suggesting any changes to the syntactic representation, which remains constituency-based, despite the accuracy limitations this imposes on Arabic parsing.

## 2.3.2 The Prague Arabic Treebank

The syntactic representation to be presented in Chapter 6 is a dependency-based hybrid that includes aspects of constituency syntax. This compares to the second major Arabic treebank to be released after the Penn Treebank, the Prague Arabic Treebank (Hajič et al., 2004; Smrž and Hajič, 2006). This treebank uses a pure dependency representation and annotates the same source text as the Penn Treebank – collections of Arabic news articles distributed by the Linguistic Data Consortium (LDC).

The Prague Arabic Treebank shares its grammatical framework with the Prague Czech Treebank (Hajič, Hladká and Pajas, 2001), and focuses on three levels of annotation: morphological, analytical (surface syntax) and tectogrammatical (deep syntax and linguistic meaning). The first version of the treebank, published in 2004, contains morphological annotation for 148,000 words and syntactic annotation for 113,500 words, with tectogrammatical annotation still under development at the time of its publication.

The grammatical framework used for the Prague Treebank is the Functional Generative Description (Sgall, Hajičová and Panevová, 1986; Hajičová and Sgall, 2003). This is a dependency-based representation that emphasizes the difference between form (including word-forms and morphological realizations) and function (such as the syntactic roles of subject, object and predicate). This grammatical description was originally designed for Czech, a language that is morphologically rich, possessing a high degree of free word order. Both of these aspects of Czech are also found in Arabic. The authors of the treebank argue that using a dependency representation has resulted in annotations better suited to Arabic's linguistic constructions, compared to the constituency representation used for the Penn Treebank. Smrž and Hajič (2006) note the similarities between their dependency representation and the Arabic linguistic tradition:

> Not only are the notions of dependency and function central to many modern linguistic theories and 'inherent' to computer science and logic, their connection to the study of the Arabic language and its meaning is interesting too, as the traditional literature on these topics, with some works dating back more than a thousand years, actually involved and developed similar concepts.

Hajič et al. (2004) describe the annotation methodology used to develop the treebank as multi-staged. Initial morphological tagging was performed by a data-driven maximum entropy tagger that was previously developed for Czech (Hajič and Hladká, 1998). This tagger was adapted to Arabic through retraining by using morphological data from the Penn Arabic Treebank. They report a 10.8% error rate for tagging parts-of-speech, but only a 0.8% error rate for segmentation of Arabic words into constituent morphemes.

Following automatic tagging, expert annotators corrected the morphological analysis and manually added syntactic annotation. Once an initial section of the treebank was completed, a syntactic parser was trained on the annotated data in order to automatically parse the remainder of the corpus. The resulting dependency trees were then manually corrected by annotators.

Figure 2.3 (overleaf) shows an example tree from the Prague Arabic Treebank. Individual Arabic words have been morphologically segmented into morphemes, with one morpheme annotated per line. The first line is reserved for the abstract root of the dependency tree. This differs from other dependency treebanks, such as the Columbia Arabic Treebank, in which all nodes including the root node correspond to morphemes (Habash and Roth, 2009c).

The diagram is organized into four columns. Reading from left-to-right, the first column contains the dependency tree. The tree's nodes are morphemes and the tree's edges are labelled with syntactic roles. The syntactic tags shown in the diagram are the same as those found in the Czech Treebank (AuxS = Root Node, AuxY = Adverbial Particle, AuxP = Preposition, Adv = Adverb, Atr = Attribute, Pred = Predicate, Sb = Subject, Obj = Object, Coord = Coordination, AuxK = Punctuation). This approach is similar to the Penn Arabic Treebank, which also does not use traditional Arabic grammar for its syntactic tags, but instead reuses an annotation scheme for another language. The second column shows surface forms, displayed using both Arabic script and a phonetic English transcription. The third column is a gloss for each morphological segment. Finally, the fourth column displays morphological tagging using positional notation. The positions are slots for major and minor parts of speech, mood, voice, person, gender, number, case and state features. Unset values are indicated by dashes (-). For example, the Arabic word for 'the magazine' is tagged as N-----FS1D, denoting a feminine singular noun in the nominative case with definite state.

Since its initial release, the treebank has been extended with morphological annotation for 393,000 words, syntactic annotation for 125,000 words and tectogrammatical annotation for 10,000 words. Data from this extended version of the treebank was used in the CoNLL shared task on multilingual dependency parsing to benchmark the performance of several Arabic statistical parsers (Nivre et al., 2007a). This parsing work is reviewed in section 2.4.3.

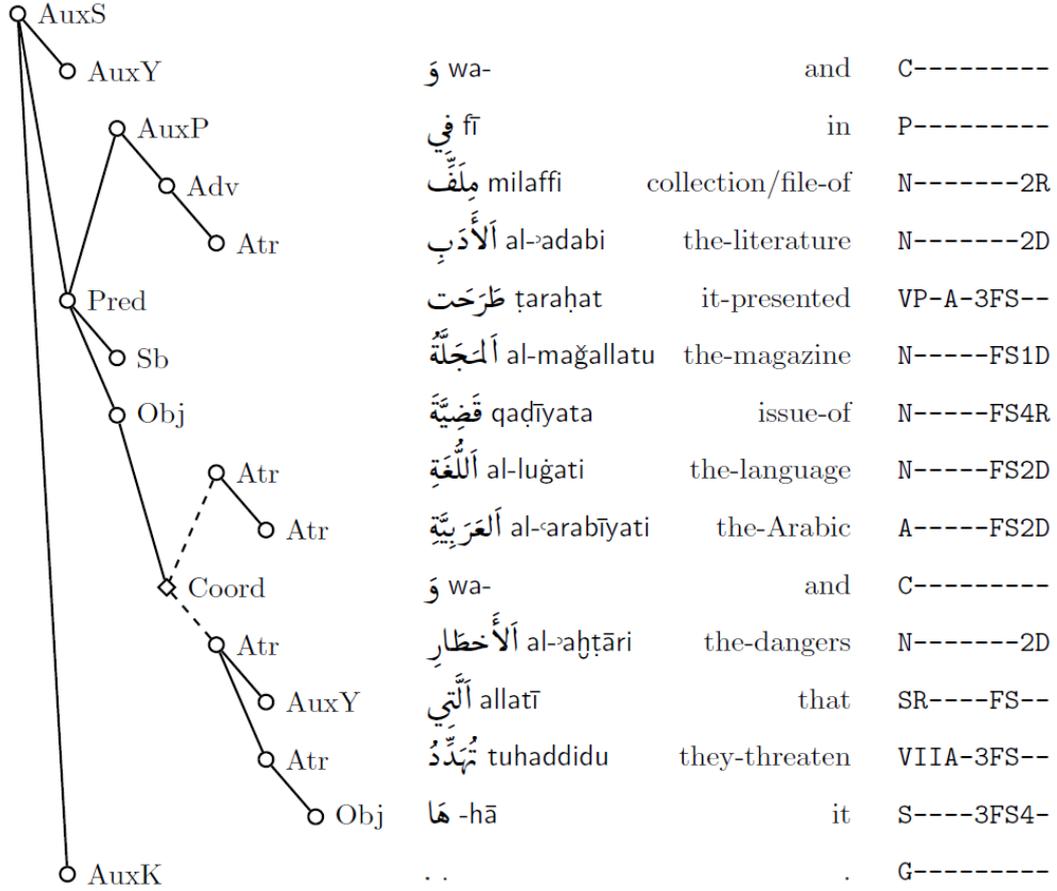

| | | | |
|---|---|---|---|
| AuxS | | | |
| AuxY | وَ wa- | and | C--------- |
| AuxP | فِي fī | in | P--------- |
| Adv | مِلَفِّ milaffi | collection/file-of | N-------2R |
| Atr | أَلأَدَبِ al-ʾadabi | the-literature | N-------2D |
| Pred | طَرَحَت ṭaraḥat | it-presented | VP-A-3FS-- |
| Sb | أَلمَجَلَّةُ al-maǧallatu | the-magazine | N-----FS1D |
| Obj | قَضِيَّةَ qaḍīyata | issue-of | N-----FS4R |
| Atr | أَللُّغَةِ al-luġati | the-language | N-----FS2D |
| Atr | أَلعَرَبِيَّةِ al-ʿarabīyati | the-Arabic | A-----FS2D |
| Coord | وَ wa- | and | C--------- |
| Atr | أَلأَخطَارِ al-ʾaḫṭāri | the-dangers | N-------2D |
| AuxY | أَلَّتِي allatī | that | SR----FS-- |
| Atr | تُهَدِّدُ tuhaddidu | they-threaten | VIIA-3FS-- |
| Obj | هَا -hā | it | S----3FS4- |
| AuxK | . | . | G--------- |

وفي ملف الأدب طرحت المجلة قضية اللغة العربية والأخطار التي تهددها.

'*In the section on literature, the magazine presented the issue of the Arabic language and the dangers that threaten it.*'

Figure 2.3: Dependency tree from the Prague Arabic Treebank.

## 2.3.3 The Columbia Arabic Treebank

The Columbia Arabic Treebank (CATiB) is the third major syntactic treebank for Arabic (Habash, Faraj and Roth, 2009a; Habash and Roth, 2009c). The treebank is designed to facilitate the development of statistical parsers for Modern Arabic. Like the Prague Treebank, the Columbia Treebank is also annotated using dependency grammar. However, the Columbia Treebank contrasts with both the Penn and Prague treebanks by adopting a minimalistic syntactic representation. The methodology for treebank construction focuses on rapid annotation using a smaller number of tags, allowing annotators to correct large amounts of text as quickly as possible. The treebank's tagset has six part-of-speech tags, shown in Table 2.2 below:

| Part-of-Speech Tag | Meaning |
| --- | --- |
| NOM | Nominals (nouns, pronouns, adjectives and adverbs) |
| PROP | Proper nouns |
| VRB | Verbs |
| VRB-PASS | Passive-voice verbs |
| PRT | Particles (including prepositions and conjunctions) |
| PNX | Punctuation |

Table 2.2: Part-of-speech tags in the Columbia Arabic Treebank.

Similarly, the dependency tagset is also minimal with only seven tags (Table 2.3, overleaf). With the exception of the modifier tag (MOD), the dependency relations are based on well-known traditional syntactic roles. These tags are easily understandable by expert annotators familiar with traditional Arabic grammar. The annotation scheme purposely excludes additional relations used for deep tagging, such as the functional tags for time and place in the Penn Treebank.

| Dependency Tag | Meaning |
|---|---|
| SBJ | Subject |
| OBJ | Object |
| TPC | Topic |
| PRD | Predicate |
| IDF | Possessive (*iḍāfa*) |
| TMZ | Specification (*tamyīz*) |
| MOD | Modifier |

Table 2.3: Dependency tags in the Columbia Arabic Treebank.

Habash et al. (2009a) emphasize that basing their scheme on concepts from the Arabic linguistic tradition simplifies the annotation process. This compares to the approach used for the Quranic Arabic Corpus, which also uses a tagset based on traditional grammar, but utilizes a more fine-grained set of tags:

> CATiB uses a linguistic representation and terminology inspired by Arabic's long tradition of syntactic studies. This makes it easier to train annotators without being restricted to hire annotators who have degrees in linguistics. CATiB uses an intuitive dependency representation and relational labels inspired by Arabic grammar such as *tamyīz* (specification) and *iḍāfa* (possessive construction) in addition to universal predicate-argument structure labels such as subject, object and modifier.

The initial version of the treebank provided morphological and syntactic annotation for 200,000 words of Arabic, annotated rapidly over five months. The annotator training period was only two months, compared to between six months to a year for the Penn and Prague Arabic treebanks (Habash and Roth, 2009c).

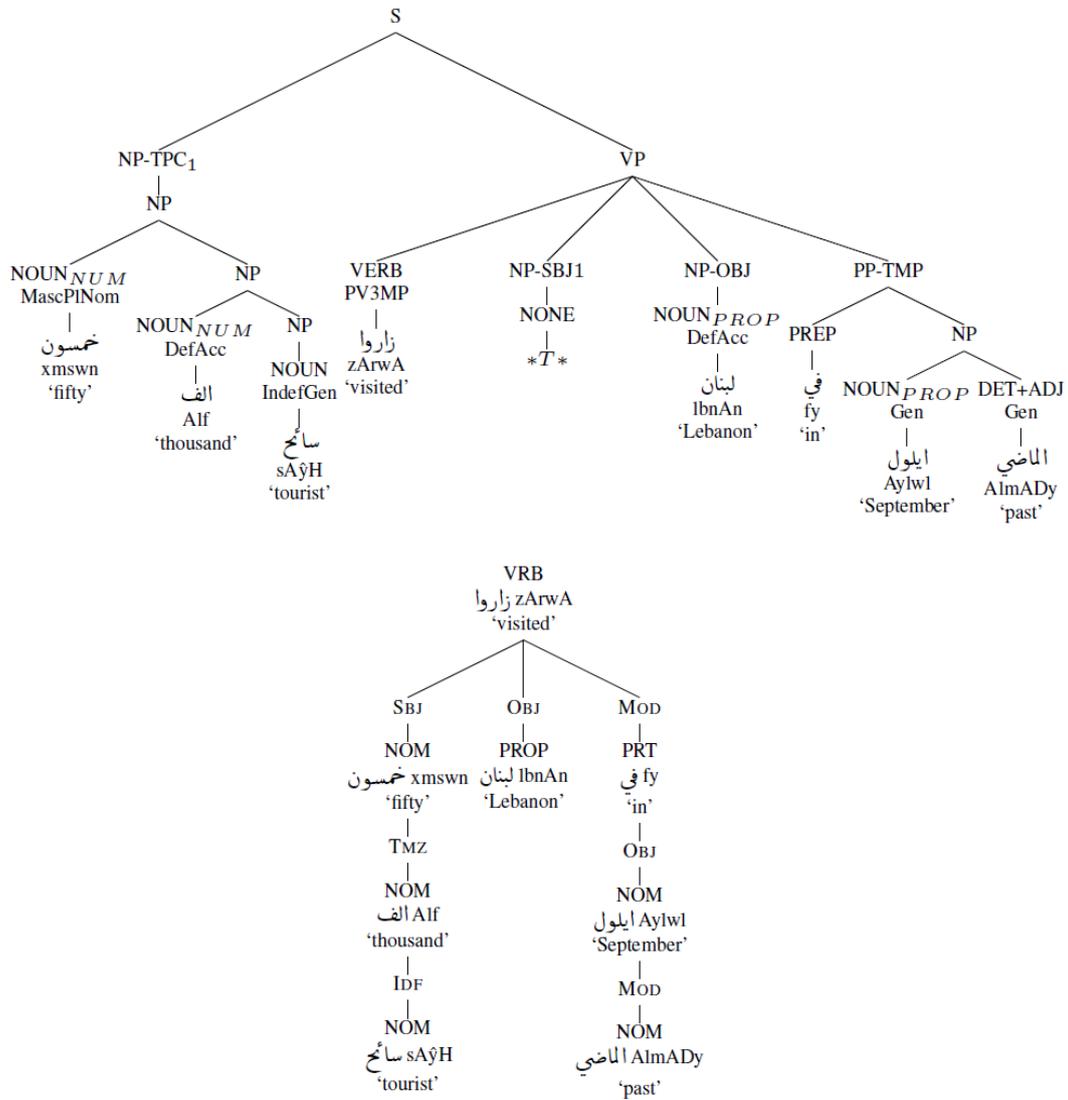

خمسون الف سائح زاروا لبنان في ايلول الماضي

'*50 thousand tourists visited Lebanon last September.*'

Figure 2.4: Constituency tree from the Penn Arabic Treebank (upper tree)
and a dependency tree from the Columbia Arabic Treebank (lower tree).

As with previous treebanks, the annotation methodology proceeds in multiple stages. In the first stage, the text is part-of-speech tagged and morphologically segmented using the MADA+TOKAN toolkit (Habash and Rambow, 2005). The F1 accuracy scores reported for these two morphological processing tasks is 99.7% and 97.7% respectively. The automatically tagged data is corrected by annotators. Following morphological annotation, initial dependency parsing was performed using MaltParser (Nivre et al., 2007) and then manually reviewed. The parser was trained using data from the Penn Arabic Treebank by automatically converting constituency trees into dependency trees. Following completion of the first section of the treebank, the parser's statistical model was improved by retraining using the additional annotated data.

To illustrate the differences in representation between in the Penn Treebank and the Columbia Treebank, Figure 2.4 (page 30) shows the same Arabic sentence annotated using both schemes. The upper tree in the diagram uses Penn Treebank-style constituency annotation. The lower tree is a dependency tree from the Columbia Treebank. Similar to the Prague Treebank, this tree has nodes which are morphological segments and edges labelled with syntactic dependency roles.

Work for developing the Columbia Arabic treebank demonstrates that high-quality morphosyntactic annotation of Arabic is possible using an annotation scheme based on concepts from traditional Arabic grammar. Compared to the Penn Arabic Treebank, Habash et al. (2009c) report higher inter-annotator agreement for morphological and syntactic annotation, as the tagset is based on concepts familiar to annotators. However, due to the focus on rapid annotation, the treebank lacks fine-grained morphological or syntactic annotation. This differs from the work for Classical Arabic presented in this thesis. For example, although ellipsis is commonly used to describe syntactic structure in traditional grammar, the Columbia treebank does not annotate empty categories. In contrast, the Quranic Arabic Corpus provides a fine-grained morphological representation with a richer tagset, as well as being more closely aligned to traditional concepts.

## 2.4   Statistical Parsing Models

### 2.4.1 Classical Arabic Parsing

Despite lower accuracy scores compared to English, Modern Arabic parsing is well established in computational linguistics research. State-of-the-art Modern Arabic parsers utilize data-driven statistical models and have been evaluated on large datasets, for both constituency and dependency representations. In contrast, almost no previous work has been published for parsing Classical Arabic. The few published studies are either descriptions of small experiments, or are discussion papers that outline possible approaches without providing clear descriptions of methodology or results. For example, Shokrollahi-Far et al. (2009) discuss their rule-based constituency parser. Although they outline a parsing experiment using verses of the Quran, they fail to explain their evaluation process in detail and do not report accuracy scores. Similarly, Shatnawi and Belkhouche (2012) describe a small experiment for parsing the Quran using a recursive descent parser. They generate constituency trees for a small 60-word sample of the Quran using hand-written grammatical rules but do not evaluate parsing performance.

Previous work for Classical Arabic parsing has been limited by lack of data. Unlike for Modern Arabic, treebanks for Classical Arabic have not previously been developed, ruling out data-driven approaches to parsing using statistical methods. In contrast, the statistical parser described in this thesis is made possible by learning from a new manually-verified treebank.

### 2.4.2 Arabic Constituency Parsing

For Modern Arabic, using constituency phrase-structure to represent Arabic syntax has resulted in parsing underperformance. For example, Kulick et al. (2006) parse the Penn Arabic Treebank using Bikel's parser (Bikel, 2004b). This is an improved reimplementation of Collins' parser, a well-known model for constituency syntax (Collins, 1999). They report an F1-score of 74% for Arabic, but a much higher score of 88% for a similar sized English dataset. This suggests that parsing using a constituency representation is more suitable for English than for languages with relatively free word order such as Arabic.

In a more recent comparison, Green and Manning (2010) measure the accuracy of three constituency parsers, including their own Stanford parser, against the Penn Arabic Treebank. Their results are not directly comparable to Kulick et al. since they use an alternative metric for measuring accuracy. Instead of Parseval, they use a leaf-ancestor metric, and report scores of 77.5% for Bikel's parser, 80% for the Stanford Parser and 83.1% for the Berkeley parser (Petrov, 2009).

These results fall short of state-of-the-art parsing performance for English. In addition to measuring accuracy, they investigate the causes of poor parsing results for the Penn Arabic Treebank. They conclude that low annotation consistency is a problem. They also note that using a constituency representation for Arabic does not capture important syntactic constructions not found in English:

> It is well-known that constituency parsing models designed for English often do not generalize easily to other languages and treebanks. The Penn Arabic Treebank (ATB) syntactic guidelines (Maamouri et al., 2004) were purposefully borrowed without major modification from English (Marcus et al., 1993). Further, Maamouri and Bies (2004) argued that the English guidelines generalize well to other languages. But Arabic contains a variety of linguistic phenomena unseen in English. The ATB is similar to other treebanks in gross statistical terms, but annotation consistency remains low relative to English. Our results suggest that current parsing models would benefit from better annotation consistency and enriched annotation in certain syntactic configurations.

However, Green and Manning are able to improve parsing performance by supplementing the Penn Arabic Treebank with additional morphosyntactic features. Using this approach, they are able to boost the accuracy of a probabilistic context-free parser from 75.95% to 80.95%, measured using the leaf-ancestor metric. The additional features they add to the treebank are designed to capture linguistic constructions that only occur in Arabic and not English, and are partly based on linguistic considerations from traditional grammar:

> For verbs we add two features. First we mark any node that dominates a verb phrase. This feature has a linguistic justification. Historically, Arabic grammar has identified two sentences types: those that begin with a nominal (الجملة الإسمية), and those that begin with a verb (الجملة الفعلية). But foreign learners are often surprised by the verbless predications that are frequently used in Arabic. Although these are technically nominal, they have become known as 'equational' sentences. [This feature] is especially effective for distinguishing root S nodes of equational sentences. We also mark all nodes that dominate an SVO (subject-verb-object) configuration. In MSA, SVO usually appears in non-matrix clauses.

This thesis will address the limitations that the Penn Treebank's constituency representation has on Arabic parsing performance. For example, the annotation improvements suggested by Green and Manning are implemented in the Quranic Arabic Corpus. The suggested tags for nominal phrases[2] (الجملة الإسمية) and verbal phrases (الجملة الفعلية) are explicitly annotated, as these are among the structures described by traditional Arabic grammar in Chapter 6.

## 2.4.3 Arabic Dependency Parsing

Most recent parsing work for Arabic has focused on dependency grammar, a representation better suited to modelling languages with free word order such as Arabic. The 2007 Conference on Computational Natural Language Learning (CoNLL) featured a shared task that evaluated statistical dependency parsers for several languages (Nivre et al., 2007a). State-of-the-art parsers for Modern Arabic were tested in the shared task using data from the Prague Arabic Treebank developed by Hajič et al. (2004). As input, the parsers were provided with Arabic text with gold-standard morphological annotation, including part-of-speech tags, segmentation and features annotated from the treebank. The same approach is used in this thesis, where gold-standard morphological annotation is also assumed as input for evaluating a new Classical Arabic parser.

---

[2] In Arabic grammar, the concept الجملة الإسمية applies to clauses as well as phrases. The term 'nominal phrase' is used here generally, to refer to nominal syntactic structures.

| Lead Author | Parsing Model | Score |
| --- | --- | --- |
| Nilsson | Ensemble (combination of six models) | 76.52 |
| Nakagawa | Global graph features using Gibbs sampling | 75.08 |
| Hall | MaltParser | 74.75 |
| Sagae | Ensemble (combination of three models) | 74.71 |
| Chen | Unlabelled MaltParser + SVM labelling | 74.65 |

Table 2.4: Top five statistical parsers for Arabic in the CoNLL shared task.

A total of 20 Arabic dependency parsers were evaluated in the shared task. Table 2.4 summarizes the results of the top five parsers, measured using a labelled attachment score (LAS) metric. The best performing parser by Nilsson, described in Hall et al. (2007a), uses an ensemble system that combines the results of six parsing models using MaltParser (Nivre et al., 2007b). However, the top score of 76.52% falls short of the performance of 88.1% reported for English dependency parsing in the same task. This work demonstrates that parsing the Prague Arabic Treebank is more challenging than English dependency parsing.

These results contrast with recent work by Marton et al. (2013), who report improved parsing results for the Columbia Arabic Treebank. Like Hall et al., they also use MaltParser, and report a baseline F1-score of 81% for their Arabic dependency parsing model. They are able to increase parsing accuracy to 84% by introducing a more fine-grained tagset with additional morphological features not included in the Columbia Treebank's original annotation scheme. They conclude that the most useful features for dependency parsing that are missing from the treebank are definiteness, person, number, gender and lemma. This limitation will be shown to be addressed in the Quranic Arabic Corpus, which includes these additional features as part of its fine-grained annotation scheme.

## 2.4.4 Dual Dependency-Constituency Parsing

Within published literature, previous work that most closely resembles the hybrid dependency-constituency parsing algorithm developed in this thesis is the approach by Hall et al. for German (Hall and Nivre, 2008) and for Swedish (Hall, Nivre and Nilsson, 2007b). However, in contrast to the hybrid parser presented in Chapter 9, their combined model outputs two parse trees for an input sentence, providing distinct annotation for dependency and constituency representations. They also describe their approach as hybrid parsing. To avoid confusion, this thesis instead uses the term 'dual parsing' for their model. The term 'hybrid parsing' is reserved for the new algorithms presented in Chapter 9, which output a single graph using a hybrid dependency-constituency representation.

The dual parsing algorithm described by Hall et al. extends MaltParser to output constituency trees by merging the two representations into dependency structures. The merged structures encode additional constituency information on enriched edge labels. The two diagrams overleaf illustrate the merging process for Swedish (Hall et al., 2007b). Figure 2.5 shows a constituency representation with an equivalent dependency representation. In Figure 2.6, the lower tree is a dependency structure with merged edges. Merging is possible if for every word $w$ in a sentence, the sequence of words governed by $w$ in the dependency tree is equal to the set of leaf nodes covered by a non-terminal node $n$ in the constituency tree. In the merged representation, compound edge labels are of the form $X \mid Y$, where $X$ is $w$'s dependency relation, and $Y$ is $n$'s phrase-structure tag if $n$ is not a preterminal, or an asterisk (*) otherwise.

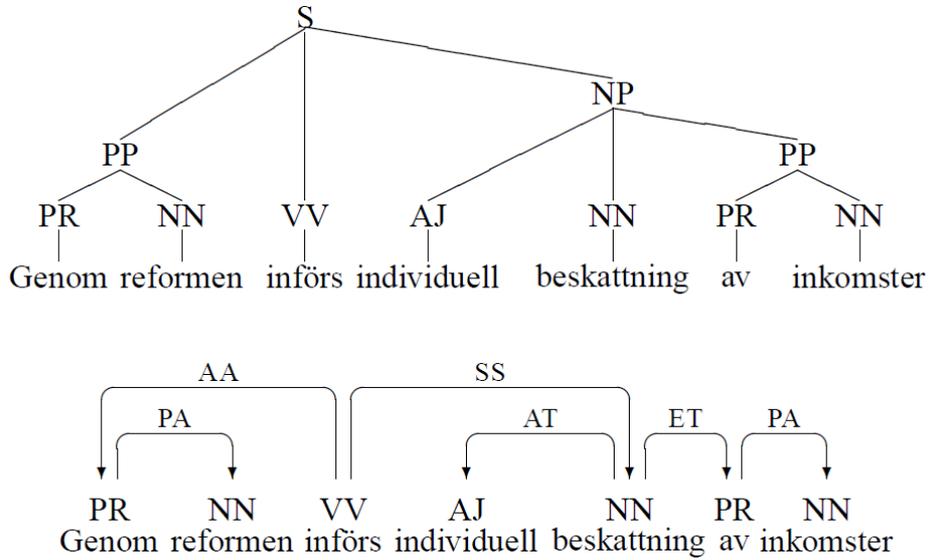

Figure 2.5: Constituency and dependency representations for Swedish.

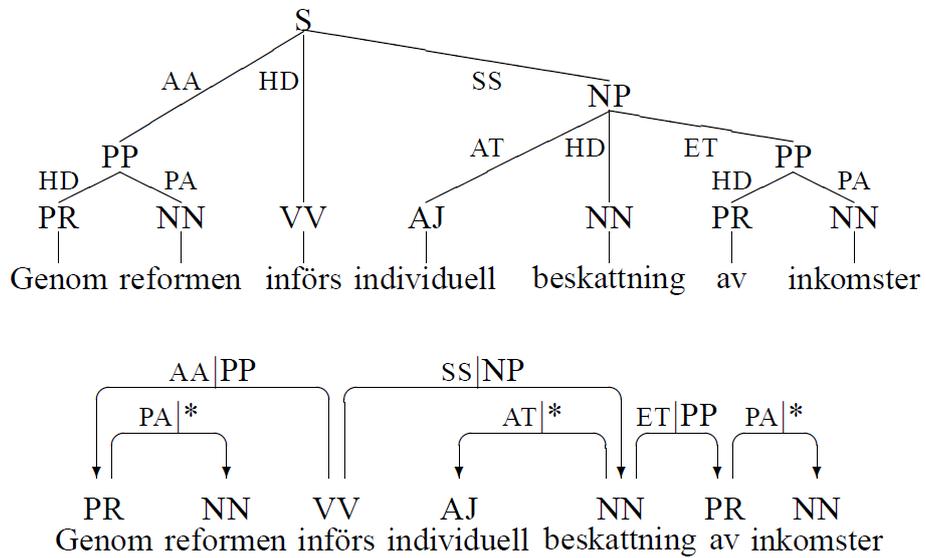

Figure 2.6: Dual dependency-constituency representation for Swedish.

Hall et al. build their statistical model for dual parsing by training MaltParser using data converted to the merged representation. To produce constituency trees, the merged output is post-processed after dependency parsing. An inverse transformation is applied that uses the information encoded on merged edges to restore constituency nodes and phrase-structure tags. For German, Hall and Nivre (2008) measure performance using constituency data from two German treebanks: the TIGER Treebank (Brants and Hansen, 2002) and the TüBa-D/Z Treebank (Hinrichs et al., 2004). Using head-finding rules, dependency data is collected by automatically converting from the constituency representation in the treebanks. They report accuracy close to 90% for dependency parsing, measured using a labelled attachment score. Similarly, for Swedish, Hall et al. (2007b) report results of over 80% using the same metric.

Dual parsing algorithms are relevant to the work in this thesis, which compares a hybrid parser to a multi-step dependency model that uses post-processing. A similar approach to Hall et al. will be used to encode constituency information onto merged edge labels for multi-step hybrid parsing. However, this approach will be adapted to the Classical Arabic syntactic annotation scheme.

## 2.4.5 Parsing Models for Ellipsis

To the best of the author's knowledge, the work in this thesis describes the first dependency-based parsing model in any language for elliptical constructions. In syntactic treebanks, empty categories are used to represent words or phrases that are not written or pronounced in the original text, such as the elliptical annotation in the Penn Treebank for null complementizers and wh-movement. Figure 2.7 overleaf shows an example from the Penn Treebank for the noun phrase 'the man Sam likes'. This constituency tree annotates two empty categories. The node marked 0 is a null complementizer, i.e. 'the man (that) Sam likes'. The second node marked *T*-1 is a co-indexed trace.

Although no previous work exists for dependency parsing with ellipsis, related work has been done for constituency parsing. Gabbard et al. (2006) show that it is possible to fully recover Penn Treebank-style trees for English including function tags and empty categories, by training a cascade of statistical classifiers.

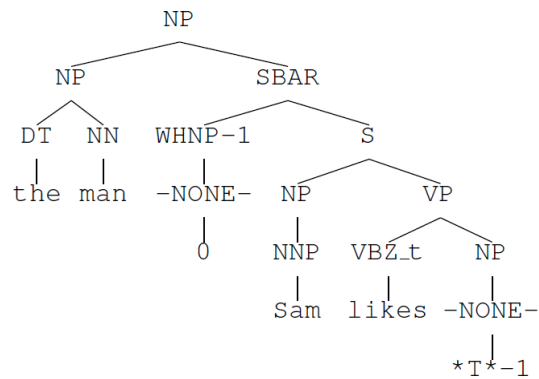

Figure 2.7: Empty categories in a Penn English Treebank constituency tree.

For Arabic constituency representations, Gabbard (2010) extends this approach to recover the empty categories annotated in the Penn Arabic Treebank. In his description of ellipsis restoration, Gabbard notes that both functional tags and elliptical structures are not generally considered in constituency parsing work:

> The syntactic structures produced by the most commonly used parsers are less detailed than those structures found in the treebanks the parsers were trained on. In particular, this is true of Collins (1999), Bikel (2004) and Charniak (2000), which are very commonly used. The parsers do not recover two sorts of information present in all the Penn Treebanks (English, Arabic, Chinese, and historical). The first are annotations on constituents indicating their syntactic or semantic function in the sentence (Gabbard et al., 2006). The second kind of information is tree nodes which do not correspond to overt (written or pronounced) words.

For dependency representations, although various treebanks annotate elliptical structures, these have previously been ignored in parsing work. For example, Rello and Ilisei (2009) develop a Spanish corpus annotated with dropped subject pronouns using dependency grammar. This compares to Classical Arabic, where dropped subject pronouns also frequently occur. However, they use manual annotation for this task, as no dependency or constituency parsers for Spanish exist for these constructions. In related work, Bengoetxea and Gojenola (2010) use MaltParser to parse the Basque Dependency Treebank, which originally included empty categories to represent ellipsis and coordination. However, their work uses a newer version of the treebank in which the empty categories are no longer annotated in order to minimize the number of non-projective edges in the treebank and simplify parsing.

Similarly, previous Arabic dependency treebanks do not annotate ellipsis, a limitation addressed in this thesis. In contrast to the post-processing approached described by Gabbard et al., the dependency-based parser that will be presented for Classical Arabic handles ellipsis in the hybrid representation directly in the parsing process.

## 2.4.6 Hebrew Parsing Models

Hebrew, another Semitic language, faces a similar set of challenges in comparison to parsing Arabic. Both languages have relatively free word order and require morphological disambiguation for syntactic parsing. Similar to recent work for Arabic, parsing work for Hebrew focuses on both constituency and dependency representations. For dependency parsing, Goldberg and Elhadad (2010), apply a pipeline approach by disambiguating morphology and syntax in two separate steps. They report an 84.2% labelled attachment score using gold-standard morphological input, and 76.2% using predicted morphological tagging.

More recent work for Hebrew parsing has focused on joint morphological and syntactic models. In contrast to pipeline approaches, in which the output of a morphological analyzer is given to a syntactic parser, this approach utilizes an integrated statistical model. Tsarfaty (2006) argues that for Semitic languages such as Arabic and Hebrew, morphological disambiguation is dependent on syntactic context, and that combined models lead to improved performance. This is demonstrated by Goldberg and Elhadad (2011), who perform joint parsing using a lattice segmentation model for Hebrew. Using the Berkley parser (Petrov, 2009), they report an F1-score of 77.3% using a pipeline approach, and 79.9% for joint disambiguation.

Similar to Goldberg and Elhadad's evaluation methodology, the Classical Arabic parser developed in this thesis will be evaluated by considering a pipeline approach as a baseline, in which the output of a dependency parser is converted to the hybrid representation. This will be compared to a one-step dependency-constituency parser that uses a joint model for the hybrid representation. However, joint morphological disambiguation for Classical Arabic is beyond the scope of this thesis. Although recent work for Hebrew suggests that joint models outperform pipeline approaches, joint morphological disambiguation has not yet been performed for Arabic, and Arabic statistical parsers are generally evaluated using gold-standard morphological input.

## 2.5   Annotation Methodologies

This section reviews previous work for three annotation methodologies: offline expert annotation, online crowdsourcing, and supervised collaboration – the methodology used to annotate the Quranic Arabic Corpus.

Most annotated corpora are developed by experts who annotate a corpus manually, following an annotation scheme and a set of annotation guidelines. Crowdsourcing is an emerging alternative methodology in which a large number of non-experts repeatedly annotate a corpus. These independent annotations are combined to achieve high reliability, using an aggregate metric such as majority voting or statistical weighting. These methodologies contrast with recent work for supervised collaboration, a third approach to annotation where non-experts produce annotations collaboratively under expert supervision.

## 2.5.1 Expert Annotation

Inter-annotator agreement for corpora annotated by experts is important for consistent and high-quality annotation. However, agreement between annotators can be difficult to achieve, requiring training, clear guidelines, and reconciling different annotator results to produce the final gold-standard annotation. Kilgarriff (1998) investigates the factors that affect inter-annotator agreement for word-sense tagging. He notes that two important reasons for inconsistent results between experts are a poorly-defined annotation scheme and mistakes by annotators due to lack of motivation or misunderstanding the annotation task.

For syntactic annotation, Brants (2000c) analyzes the annotation accuracy of the German NEGRA Treebank. Initial annotation of the treebank was performed quickly by two experts who manually corrected the output of a syntactic parser (Skut et al., 1997; Brants, Skut and Uszkoreit, 1999). Brants reports an initial annotation speed of 50 seconds per sentence for each annotator on average. In contrast, total annotation time was measured at 10 minutes per sentence for the final gold-standard. This included the time spent by two annotators independently reviewing each sentence, performing a comparison of each other's work, and discussing and correcting differences. Initial inter-annotator agreement before discussion was 98.57%. Agreement between the initial versions and the final gold-standard was 98.8%. This work shows that despite comparison and review, disagreement between experts leads to an upper bound on annotation accuracy when measured using inter-annotator agreement.

Even widely used resources such as the Penn English Treebank have limits on data quality. Marcus et al. (1993) report an inter-annotator agreement of 97% for the part-of-speech tagging in the treebank. However, Manning (2011) analyses the quality of annotation by training a part-of-speech tagger and classifies its errors against a sample of sentences from the Penn Treebank (Table 2.5, overleaf).

| Class | Frequency |
|---|---|
| Lexicon gap | 4.5% |
| Unknown word | 4.5% |
| Could plausibly get right | 16.0% |
| Difficult linguistics | 19.5% |
| Underspecified/unclear | 12.0% |
| Inconsistent/no standard | 28.0% |
| Gold standard wrong | 15.5% |

Table 2.5: Errors for automatic part-of-speech tagging for the Penn Treebank.

Manning classifies 12% of errors from the output of the tagger as due to underspecified or unclear part-of-speech tags. These errors resulted from tags being ambiguous or unclear to annotators, such as whether to choose a verbal or noun tag for gerunds. A further 28% of errors are attributed to inconsistent guidelines. Similar to Kilgarriff's work on inter-annotator agreement for word-sense tagging, this work shows that annotation guidelines need to be clear and easily understandable even to expert annotators.

## 2.5.2 Crowdsourcing, Voting and Averaging

In contrast to expert annotation, crowdsourcing is an alternative approach that has proven to be effective for a wide variety of tagging tasks, with accuracy approaching that of expert annotation. Crowdsourcing is attractive because it is cost effective, allowing for large-scale annotation tasks that would otherwise be prohibitively expensive.

Nowak and Rüger (2010) investigate the effectiveness of crowdsourcing for annotating Flickr photos with concept tags. Using 11 expert annotators, they report an inter-annotator agreement of over 90%. Expert annotation was compared to the results of using Amazon Mechanical Turk, an online crowdsourcing marketplace. Using an averaging method based on majority voting, inter-annotator agreement was found to be comparable to expert annotation. Although these results indicate that crowdsourcing is viable, Nowak and Rüger suggest further analysis by annotating larger datasets.

A wider variety of linguistic annotation tasks are considered by Snow et al. (2008). Amazon Mechanical Turk is used for five tagging tasks: affect recognition (100 sentences), word similarity (30 word pairs), recognizing textual entailment (800 sentence pairs), event temporal ordering (462 verb event pairs) and word sense disambiguation (177 sentences). They note that Amazon Mechanical Turk is cost effective. For example, they paid only USD \$2 to collect 7,000 non-expert annotations for the affect recognition task.

To boost annotation accuracy, a statistical model is used to correct for the reliability and biases of individual annotators. Using a multinomial model similar to naïve Bayes, results are combined by assigning annotators who are more than 50% accurate positive votes, annotators whose judgments are pure noise zero votes and anti-correlated annotators negative votes. This statistical weighting increases the accuracy of the annotation tasks by up to 4%, compared to majority voting. Snow et al. report that for most annotation tasks, only a small number of non-experts are required to achieve accurate annotation. For example, for the affect recognition task, the combined results of just four non-experts are required to emulate the quality of expert-level annotation.

In contrast to the small-scale experiments described above, an example of a large-scale corpus developed through crowdsourcing is the Phrase Detectives corpus, containing 1.1 million words annotated with 380,000 anaphoric relations (Chamberlain et al., 2009). In the description of their annotation methodology, they note that crowdsourcing is an attractive alternative to expert annotation:

> The statistical revolution in natural language processing (NLP) has resulted in the first NLP systems and components really usable on a large scale, from part-of-speech (POS) taggers to parsers (Jurafsky and Martin, 2008). But it has also raised the problem of creating the large amounts of annotated linguistic data needed for training and evaluating such systems. This requires trained annotators, which is prohibitively expensive both financially and in terms of person-hours (given the number of trained annotators available) on the scale required.

Their solution is to motivate annotators through entertainment, by casting the annotation task as an online game. Phrase Detectives provides an interactive web-based interface for non-experts to learn how to annotate text and make annotation decisions. Following a training phase, the game runs in two modes. In annotation mode, players locate the closest markable antecedent of an anaphor. In validation mode, players are asked to review previously annotated sentences. Final annotations are selected through majority voting. The effectiveness of this methodology is measured by annotating a section of the corpus using two expert annotators. Inter-annotator agreement between the experts was 94%, compared to 93% between experts and non-experts. This demonstrates that large-scale annotation tasks can be highly reliable using crowdsourcing.

## 2.5.3 Supervised Collaboration

Supervised collaboration is an annotation methodology involving the online collaboration of multiple annotators whose work is reviewed by supervisors acting as editors. This methodology can be considered to be a middle ground between offline expert annotation and crowdsourcing. Supervised collaboration is also cost effective, but ensures reliability through expert supervision.

Perhaps the best example of a fully collaborative resource is Wikipedia, constructed entirely by unpaid volunteers who are motivated by the interest they share in the articles being developed. Recent research has consistently shown that the effectiveness of Wikipedia depends on incremental edits to improve quality, but also crucially on open communication and discussion between editors to resolve issues, and to promote common understanding (Kittur and Kraut, 2010).

Collaborative annotation with inter-annotator discussion has recently been used to develop specialist corpora that require the participation of expert annotators. For example, the Ancient Greek Dependency Treebank (Bamman et al., 2009) is developed by annotators with backgrounds ranging from advanced undergraduate students to recent PhD graduates and professors. The treebank provides syntactic annotation for 200,000 words of Ancient Greek texts, including the works of Hesiod, Homer and Aeschylus. It is unlikely that annotating the treebank could be performed effectively using a crowdsourcing marketplace such as Amazon Mechanical Turk, given the prerequisite knowledge required. Instead, the treebank was annotated using supervised annotation, with different groups of annotators developing different sections of the treebank. Every sentence was annotated by two annotators and the differences were reconciled by an expert with specialist knowledge of the text.

In addition to an initial training period, annotators are actively engaged in new learning and collaboration by means of an online forum in which they can ask questions of each other and of project supervisors. Using this method, average annotator agreement for dependency relations was 80.6% compared to the final gold standard, measured using a labelled attachment score metric.

The complexity of syntactically tagging Ancient Greek is demonstrated by the time and effort required to produce annotations. Average annotation speed was measured at only 124 words per hour. This compares to the Penn English Treebank, where annotator speed has been reported as 1,000 words per hour after four months training (Taylor, Marcus and Santorini, 2003). Bamman et al. argue that a collaborative methodology is more suitable for the creation of a scholarly treebank, given the specialist nature of the annotations. Supervised collaboration allows annotators with different levels of expertise to participate in the annotation process, while ensuring that annotations remain consistent and of a high quality.

## 2.6 Conclusion

This chapter reviewed previous work in four areas: morphological representations, syntactic representations, parsing and annotation methodologies. This section summarizes the implications of the reviewed work in relation to the thesis research questions.

For annotation methodologies, the review contrasted the approaches of expert annotation, crowdsourcing and supervised collaboration. In comparison to expert annotation, crowdsourcing was found to be cost effective for a wide variety annotation tasks, producing annotation of comparable accuracy (Snow et al., 2008; Chamberlain et al. 2009). Supervised collaboration is an alternative approach that is also cost effective but is better suited to tasks requiring expert supervision, such as syntactic annotation of the Ancient Greek Treebank (Bamman et al., 2009). This compares to the Quranic Arabic Corpus, where annotation also requires specialist knowledge. The implication of this work is that supervised collaboration may be an appropriate methodology for annotating Classical Arabic, a research question that will be addressed in Chapter 7.

From the literature on Arabic syntactic representations, a key theme is that although both representations are used, dependency representations are preferred to constituency representations, as Arabic is a language with free word order. The Penn Arabic Treebank (Maamour et al., 2004) is the only treebank that uses a constituency representation. In contrast, the Prague (Smrž and Hajič, 2006) and Columbia (Habash et al., 2009c) treebanks are dependency based, although only the Penn Treebank performs fine-grained syntactic annotation of constructions such as ellipsis. The work reviewed for Arabic parsing (Kulick et al., 2006; Green and Manning, 2010) implies that constituency representations impose limitations on annotation consistency and parsing accuracy. However, both types of representation have resulted in lower performance for Modern Arabic compared to English using similar parsing models.

A second theme that emerged from the review on Arabic morphological and syntactic work is that many projects base their representations on traditional

Arabic grammar. For morphology, there is consensus in the literature that using a fine-grained approach based on traditional concepts leads to improved annotation (Habash, 2007a; Sawalha and Atwell, 2010). For syntax, Smrž and Hajič (2006) note that despite traditional Arabic grammar being over a thousand years old, it is based on similar concepts to modern representations such as dependencies and functional roles. Work on syntactic annotation for the Columbia Arabic Treebank (Habash and Roth, 2009c) has shown that annotators prefer to work with traditional grammar using familiar concepts and terminology. This has resulted in less annotator training and improved inter-annotator agreement and annotation consistency.

The implication of these two themes is that although traditional grammar is often cited as an inspiration for Arabic computational work, there is ongoing debate on how best to represent Arabic syntax using traditional concepts, with opinion in favour of dependency representations. An alternative representation could be a hybrid representation. Work on dual dependency-constituency parsing for German (Hall and Nivre, 2008) and for Swedish (Hall et al., 2007b) has demonstrated the feasibility of merged syntactic representations for statistical parsing. Similarly, work reviewed for Hebrew showed that integrated models can outperform pipeline approaches. For example, Goldberg and Tsarfaty (2008) integrate morphological and syntactic disambiguation and report improved parsing performance for their task.

For Classical Arabic, a thesis research question asks if a dependency-based representation that incorporates aspects of constituency syntax will be suitable for statistical parsing. This thesis will argue that this representation is more closely aligned to historical traditional analyses. The next chapter provides relevant context for this argument, providing background information on the Arabic linguistic tradition.

# 3 Historical Background

## 3.1 Introduction

Together with the Indian, Greek and Chinese languages, Arabic has one of the world's major linguistic traditions. The key developments in Arabic linguistics occurred during the Islamic Golden Age (750-1250), a time of rapid advances in philosophy, science and medicine (Hayes, 1992; Meri and Bacharach, 2006). A large number of grammarians contributed to Arabic linguistics. From 750-1500, the names of over 4,000 grammarians are known (Versteegh, 1997a). Figure 3.1 (overleaf) shows a timeline of historical events relevant to the work in this thesis.[3]

## 3.2 Motivations of the Early Arabic Grammarians

Arabic grammarians were motivated to understand and describe the details of Classical Arabic because it is the language of the Quran. Adherents of the Islamic faith believe that the Quran is the literal word of God, revealed to the Prophet Muhammad over a 23 year period, from 609 to 632, the year of the his death (Lings, 1983; Al-Azami, 2003). The Quran is written in Classical Arabic, largely in a style of rhymed prose known as *saj'* (سجع). Even among non-Islamic scholars of Arabic, the Quran is widely regarded as a masterpiece of literature due to its eloquent and beautiful use of language. For example, Stewart (2000) notes that:

---

[3] A detailed description of the history of the Arabic linguistic tradition is beyond the scope of this chapter. Introductory surveys can be found in Owens (1988), Bohas et al. (1990), Versteegh (1997a), Al-Liheibi (1999) and Jiyad (2010).

**328**: Earliest known Arabic inscription at Namara in the Nabataean alphabet.

**609-632**: Revelation of the Quran to the Prophet Muhammad.

**632**: Death of the Prophet Muhammad. Islam begins to spread rapidly.

**603-688**: Abu al-Aswad al-Du'ali: First Arabic grammarian. Analyzed parts of speech, conjunctions, attributes, exclamations and interrogatives.

**685-705**: Reign of the Caliph Abd al-Malik ibn Marwan. Arabic becomes the lingua franca and sole administrative language of the Islamic empire.

**750**: Islamic empire controls a vast area of land including Southern France, Spain, North Africa, the Middle East, the Indus Valley and Central Asia.

**718-786**: Al-Khalil: Introduces vowel marks into Arabic script (*ḥarakāt*) and the study of prosody (*al-'arūḍ*). First Arabic dictionary (*kitāb al-'ayn*).

**731-822**: Al-Farra: Establishes that grammar is key to understanding the Quran.

**760-796**: Sibawayh: The Book of Grammar (*al-kitāb fī an-naḥw*), a seminal treatise that introduces syntactic governance (*'amal wa 'āmil*).

**830**: Al-Akhfash: Describes rhetorical structures in the Quran.

**826-898**: Al-Mubarrad: Collects a corpus of Classical Arabic prose and poetry.

**892-951**: Al-Zajjaji: Explores the relationship between grammar and logic.

**932-1002**: Ibn Jinni: Detailed work on Arabic phonology and morphology.

**1075-1144**: Al-Zamakhshari: Deep linguistic analysis of the Quran.

**1256-1345**: Abu Hayyan: Concepts from Arabic linguistics are applied to develop functional grammars for Turkic, Ethiopian and Mongolian.

**1308-1359**: Ibn Hisham: Fine-grained classification of parts-of-speech.

**1859**: Publication of Wright's grammar in English for the Arabic language.

**1863**: Lane's lexicon: An Arabic-English lexicon based on traditional sources.

Figure 3.1: Timeline of key developments in Classical Arabic grammar.

> It is widely agreed that the Quran is a beautiful text. Umar ibn al-Khattab, later the second Caliph, vehemently opposed the Prophet's early preaching in Mecca but was so moved upon hearing [the Quran] recited that he converted on the spot. What is it that makes the Quran so beautiful and that renders any translation a pale shadow of the original? Rhyme and rhythm are certainly the most outstanding elements lost in translation. The Quran is a profoundly artistic and indeed poetic text.

Following the rapid spread of Islam, the Quran became the central religious text for a large number of non-Arabs, with Arabic as their lingua franca. By 750, the Umayyad Caliphate had grown to become the largest empire the world had ever seen up to that time, controlling a vast area of land that included Southern France, Spain, North Africa, the Middle East, the Indus Valley, and Central Asia up to the borders of China (Hawting, 2000). However, grammatically correct Arabic was often not spoken among the diverse ethnic groups within Islamic civilization. Solecisms, termed *laḥn* (لحن), became more frequent as Islam spread (Al-Liheibi, 1999). Concerns over incorrect recitation of the Quran motivated early Arabic grammarians to produce detailed work documenting its linguistic rules.

A later motivation was *shu'ūbiyya*. This movement sought to counter the spread of Arabic culture through the Islamic principle of racial equality. Following the conquest of Persia, from the late 8th century a resurgence in Persian identity questioned the dominance of Arabic. Prominent Arabic grammarians responded by detailing the language's unique features (Suleiman, 2003). For example, Al-Zamakhshari (1075-1144) felt motivated to produce deep linguistic analyses of the Quran in response to criticisms of Arabic on cultural grounds.

In comparison to modern linguistics, the aims and motivations of traditional Arabic grammar differed in two respects. Firstly, concerned by ungrammatical language and motivated to preserve the language of the Quran, grammarians were primarily interested in describing Arabic's linguistic rules. Secondly, in common with believers of Islam today, the grammarians considered the Quran's language to be perfect. Driven by their beliefs, they produced detailed analysis of a wide variety of linguistic phenomena, developing a comprehensive theory of grammar.

## 3.3   Analytical Methods in Traditional Grammar

### 3.3.1 Analogical Deduction (*qiyās*) and Causation (*ta'līl*).

Despite their different motivations, the analytical methods used by traditional grammarians are similar to modern empirical methods. For example, they placed importance on using linguistic data in preference to constructed examples. The grammarians were interested in describing the purest form of Arabic and focused on examples from which evidence could be drawn to support various linguistic arguments. Their corpora included the Classical Arabic text of the Quran, collections of pre-Islamic poetry, and the speech of the Bedouin, who were believed to speak a pure form of Arabic having avoided contact with foreigners. An example of this method is the work of Al-Mubarrad (826-898) who collected a corpus of Classical Arabic prose and poetry for linguistic analysis in The Book of the Perfect (*kitāb al-kāmil*).

Based on linguistic data, the two main analytical methods used by traditional grammarians were analogical deduction (*qiyās* – قياس) and causation (*ta'līl* – تعليل). Analogy is a process used in Islamic jurisprudence, where rulings for situations not described in the Quran are derived through deduction. The same principle was used in linguistics. Arabic grammarians described the structure of new sentences in their corpora based on previous analyses using analogy, by comparing them to similar structures from the Quran and related texts.

The principle of causation was also a key analytical method. The grammarians believed the form of language used by native speakers had underlying causes, such as the rules that relate syntactic function to inflectional case endings. For example, for certain sentences, the cause of a noun being in the nominative case would be due to a grammatical rule that states that all nouns which are subjects of verbs are found in the nominative. Similarly, the reason for certain nouns being in the accusative case would be the rule that all nouns which are objects of verbs are found in the accusative (Owens, 1989). Using the data from corpora together with the principles of analogy and causation accelerated the elucidation of Classical Arabic's rules, as various linguistic theories could be efficiently evaluated against accepted grammatically correct texts.

## 3.3.2 The Basran and Kufan Schools

Although traditional grammarians made advances in Arabic linguistics, there was not always consensus in their approaches. Early on in the development of traditional grammar, two competing schools emerged in the Iraqi cities of Basra and Kufa. The Kufans are usually credited with initiating grammatical analysis. For example, although there are several candidates, Abu al-Aswad al-Du'ali (603-688) is often cited as the first Arabic grammarian. He was commissioned by the fourth Caliph, Ali ibn Abi Talib to document the rules of the Arabic language. Jiyad (2010) recounts the following story, often cited in later works of traditional grammar:

> I came to the leader of the believers, Ali ibn Abi Talib. He said, 'I have been thinking about the language of the Arabs and how it has been corrupted through contact with foreigners. I have decided to put something that they (the Arabs) refer to and rely on.' He gave me a note which said: 'Speech is made of nouns, verbs and particles. Nouns are names of things, verbs provide information, and particles complete meaning.' He said to me, 'Follow this approach and add to it what comes to mind.' I wrote chapters on conjunctions, attributes, exclamations and interrogatives. Every time I finished a chapter I showed it to him until I covered what I thought to be enough. He said, 'How beautiful is the approach you have taken!' From there, the concept of grammar came to exist.

The Basran and Kufan schools developed Arabic grammar at the same time, and were often engaged in competitive discussions. Although the Kufans are credited with originating grammar, Basran works have been far more influential to later grammarians (Owens, 1988). In contrast to Kufa, a city that attracted many Bedouin Arabs, Basra had a more mixed population combining Arabic and Persian cultures. The two schools of thought had different approaches to linguistic analysis. The Basran school made stronger use of analogy and restricted their analysis to the pure speech of Arabs. The Kufans had more prescriptive views. They tended to cite anomalous linguistic forms in the analysis of grammatical constructions, and were more interested in different readings of the Quran.

Both schools adopted different terminology for linguistic constructions. Due to the larger influence of the Basran school, their terminology became more standardized and was used in later works. For example, the Arabic linguistic construction of specification is today widely known by the Basran term *tamyīz* instead of the Kufan term *mufassir* (Al-Liheibi, 1999). Kufan terminology is rarely used today, except in comparative work.

### 3.3.3 Al-Khalil and Sibawayh

The grammarian Al-Khalil (718-786) was a founding member of the Basran school. His accomplishments include introducing standardized vowel marks into Arabic script (*ḥarakāt*) and founding the study of Arabic prosody (*al-'arūḍ*). He also produced the first Arabic dictionary (*kitāb al-'ayn*) using citations from the Quran and Classical Arabic poetry. His convention of organizing the lexicon by root then lemma has been adopted by later Arabic dictionaries, including those for Modern Arabic. However, he chose to sort entries using a phonetic listing instead of alphabetically, the method more commonly used today.

Al-Khalil's student Sibawayh (760-796) is widely regarded as the greatest of all Arabic grammarians. He originally arrived in Basra with the intention of studying Islamic law. A well-documented incident tells of Sibawayh learning a phrase that contained an important religious ruling. When asked to recite the phrase back to his tutor, Sibawayh mispronounced the vowelized case-ending of a single word, and his tutor publically corrected him. Aware that this mistake would have never been committed by a native Arabic speaker, Sibawayh, a Persian, felt shamed and embarrassed. He declared, 'I will seek such knowledge, that no-one will be able to accuse me of making mistakes' (Carter, 2004).

Instead of continuing to study law, Sibawayh turned his attention to mastering Arabic grammar. His magnum opus was a 1,000-page sophisticated and detailed treatise known simply as 'The Book' (*al-kitāb*), which to this day remains the authoritative work on Classical Arabic grammar. Sibawayh's *kitāb* is often ranked on par with work of other great historical linguists, such as Panini's *Ashtadhyayi* for Classical Sanskrit (Baalbaki, 2008). Sibawayh envisioned an all-encompassing grammatical system that would account for the phonology, morphology and syntax of Classical Arabic. Carter (2004) notes that:

> Sibawayh is the founder not only of Arabic grammar but also of Arabic linguistics, which are by no means the same thing. Furthermore, as becomes obvious with every page of his *kitāb*, he was also a genius, whose concept of language has a universal validity. When we bear in mind that he was probably not even a native speaker of Arabic, being the son of a Persian convert, his achievement becomes all the more astonishing.

A crucial insight of Sibawayh's analysis is that words in an Arabic sentence govern other words to produce distinctive changes in pronunciation. For example, if certain particles are placed before a verb, they change the verb's grammatical mood and affect its morphological inflection and surface form. This simple idea led to grammatical analysis that focused on analyzing sentence structure by describing the syntactic relationships between words in order to explain morphological inflection. Concepts from Sibawayh's seminal work on syntactic governance will be used for the syntactic representation in Chapter 6.

## 3.4  Further Developments

Sibawayh's grammatical analysis had a lasting influence on the Arabic linguistic tradition, and his *kitāb* introduced concepts that were extended and refined by later grammarians. These included Al-Zajjaji (892-951), who considered the relationship between grammar and logic (Zabarah, 2005; Versteegh, 1995), Abu Hayyan (1256-1345) who applied concepts from Arabic linguistics to develop functional grammars for other languages including Turkic, Ethiopian and Mongolian (Versteegh, 1997b), and Ibn Hisham (1308-1359) who introduced a fine-grained classification for parts-of-speech, focusing on grammatical particles (Gully, 1995). By the time of grammarians such as Ibn Hisham, Arabic linguistic analysis reached a stage of sophistication approaching that of modern theories, with highly detailed descriptions of Arabic's phonology, morphology, syntax and rhetorical structures. Later work by Orientalists introduced the Arabic linguistic tradition to the Western world. Examples include Lane's Arabic-English Lexicon, published in 1859, (Lane, 1992), and Wright's grammar of the Arabic Language in 1863 (Wright, 2007). Both of these works are based on traditional sources, use terminology from traditional Arabic grammar and are highly cited in later work.

   Although the early Arabic grammarians provided detailed analysis of examples from the Quran, more recent work has focused on comprehensive analysis of the entire text. The Quranic Arabic Corpus uses as its primary reference Salih's work *al-i'rāb al-mufaṣṣal likitāb allāh al-murattal* ('A Detailed Grammatical Analysis of the Recited Quran using *i'rāb*'), which collates previous analyses of historical Arabic grammar into a single reference work. This analysis of the Quran's morphology and syntax is over 10,000 pages long, spans 12 volumes, and provides detailed linguistic analysis for each of the 77,429 words in the Quran (Salih, 2007). This detailed work would not have been possible without building on centuries of previous analysis by historical Arabic grammarians.

## 3.5   Conclusion

This chapter provided historical background on the Arabic linguistic tradition, describing the aims, motivations and analytical methods of early Arabic grammarians. The Arabic linguistic tradition is a synthesis of the work of many grammarians, but certain key works have defined the field, introducing standardized terminology and grammatical concepts. Although this thesis will use sources from across this tradition, the syntactic work of Sibawayh stands out as one of the main sources of inspiration for developing the hybrid representation for Classical Arabic syntax. As will be discussed further in Part II, later works that build on this tradition, such as the comprehensive analysis by Salih (2007), will be used as primary references for annotation work.

# Part II: Modelling Classical Arabic

The invention of the alphabet was a singular event in human history, a revolutionary as well as unique gift to human civilization.

*– Frank Moore Cross*

# 4    Orthographic Representation

## 4.1    Introduction

Part II of this thesis is divided into three chapters that describe a formal model of Classical Arabic. The model consists of representations for Classical Arabic's orthography (this chapter), morphology (Chapter 5), and syntax (Chapter 6). The representations are based on concepts from the Arabic linguistic tradition, and are used for two purposes. Firstly, they are used to develop the annotation scheme for the Quranic Arabic Corpus, described in this part of the thesis. Secondly, the representations are used to develop a computational model for Classical Arabic statistical parsing, described in Part IV.

Formal models are representations of systems within a defined mathematical framework. They are descriptions that utilize formal concepts such as set theory, logic, data structures and transformational rules. In formal linguistics, they are used to analyze linguistic structures, such as the grammatical rules that underlie sentence construction. In corpus linguistics, formal representations lead to annotation schemes for annotating corpora. Although the formalization of Classical Arabic in this thesis draws on a large body of work from the Arabic linguistic tradition, adapting these works into a well-defined representation is challenging. In Arabic grammatical theory, linguistic structures are analyzed through prose, in contrast to modern approaches that utilize formal methods. Despite this, similar concepts are used in comparison to modern linguistics, such as morphological segmentation, part-of-speech classification, dependencies and semantic analysis.

The comparison between formal methods and historical analysis in Arabic grammar parallels the development of early Islamic mathematics. For example, Al-Khwarizmi (780-850) (from whose name the term 'algorithm' is derived) put forward solutions to the quadratic equation as part of the development of algebra (Kleiner, 2007; Katz, 1998). Al-Khwarizmi did not use formal notation for his equations, but instead performed mathematics rhetorically, recording his analysis in prose. However, his analysis for solving equations remains relevant today. Although modern mathematical notation for the quadratic appeared around the 16th century (e.g. Viete), the widespread use of formal notation for linguistic structures is more recent, starting with Chomsky (1957). In comparison, the use of formal methods for Classical Arabic can be seen as introducing notation and convention to an existing tradition. The aim of the formal model in Part II of this thesis is to represent the same analyses found in historical works of traditional Arabic grammar. This difference is that unlike the descriptions in prose, formal descriptions allow for further computational work such as parsing.

This chapter focuses on an orthographic representation for Classical Arabic. To relate to other Arabic resources, such as electronic lexicons, this representation must be convertible to Unicode, the computing standard for multilingual text. However, Unicode may not be the best choice as an internal format because the same Classical Arabic word can have multiple representations in Unicode as different combinations of diacritics and letters, or as pre-composed characters. In addition, the Arabic script of the Quran requires special processing to handle complex markings such as prosodic recitation marks not found in Modern Arabic. To address these issues, this chapter describes JQuranTree, a new open source component for the Quran. The component uses a novel character-plus-diacritic representation that has an unambiguous mapping to Classical Arabic, simplifying its processing.

The remainder of this chapter is organized as follows. Section 4.2 provides an overview of Quranic orthography. Section 4.3 describes the formal orthographic representation and section 4.4 describes the computational model, relating this to other approaches such as Buckwalter transliteration. Section 4.5 concludes.

## 4.2 Quranic Orthography

### 4.2.1 The Uthmani Script

Historically, copies of the Quran have been written in almost exactly the same way, with the exception of slight variations in spelling. The two most prominent variations are *warsh* (رواية ورش), used in North Africa, and *ḥafṣ* (رواية حفص), the narration used more widely across the Islamic world (Brockett, 1988). As comparative work is beyond the scope of this thesis, a single copy of the Quran was chosen for annotation. The Quranic Arabic Corpus is based on the *madīnah musḥaf* (مصحف المدينة النبوية), published by the Quran Printing Complex in Madinah. This copy is a *ḥafṣ* narration written in the Uthmani script, named after its calligrapher Uthman Taha. The *madīnah musḥaf* is widely considered to be highly accurate in comparison to traditional sources, and since 1985, the Quran Complex has printed over 200 million copies of the Quran (Mattson, 2012).

Figure 4.1 (overleaf) shows the composition of the Uthmani script for part of verse (6:76). Arabic is written from right-to-left using a connected cursive script that is more complex compared to scripts for languages such as English. In early historical copies of the Quran, letters were written in their base form, similar to (A) in Figure 4.1 (Al-Azami, 2003). This form includes consonants and long vowels. However without pointing, letters are ambiguous, such as the letters *fā'* and *qāf* in their frontal positions. Later copies included points to distinguish letters (B), and diacritics known as *tashkīl* for the precise pronunciation of short vowels (C). The *madīnah musḥaf* also includes pause marks to indicate when readers should start and stop in longer verses, as part of a prosodic mark-up system (D).

Due to the nature of the Quran as a central religious text, the script is designed to be as unambiguous as possible, encoding detailed information about correct pronunciation and recitation. These diacritics will be used in Chapter 7 to guide automatic morphological annotation of the Quran. In contrast, this supplementary data is not available in Modern Arabic, which is almost always written without diacritics, requiring readers to infer vowelization using linguistic knowledge.

| | |
|---|---|
| (A) Base script | رءاكوكبا ڡال �هذارى |
| (B) Pointed script | رءاكوكبا قال هذارى |
| (C) Diacritics | رَءَا كَوْكَبًا قَالَ هَذَارَبِّي |
| (D) Pause marks | رَءَا كَوْكَبًا قَالَ هَذَارَبِّي |

Figure 4.1: Structure of Quranic script for verse (6:76) in the *madīnah muṣḥaf*.

## 4.2.2 The Tanzil Project

Although digital copies of the Arabic text of the Quran have been available since the early 1980s, these were not as accurate as printed copies, often containing typographical errors (Khan and Alginahi, 2013). As recently as 2008, searching for Quranic verses using Google would result in spelling mistakes in the highest ranked search result, such as تَخَيَّرُونَ instead of يَتَخَيَّرُونَ in Figure 4.2:

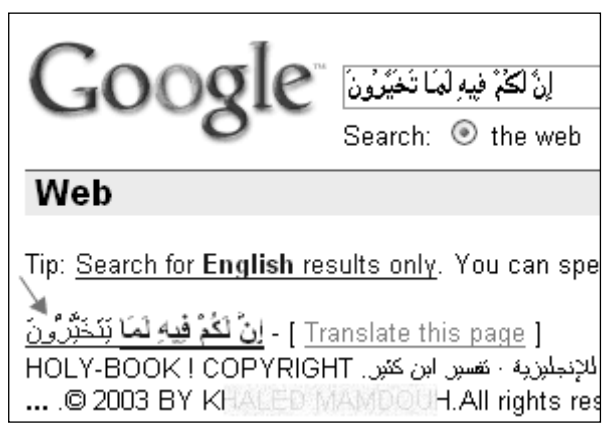

Figure 4.2: Incorrect Google results for verse (68:38), as of January 21, 2008.

In contrast to previous work, such as the morphological analysis by Dror et al. (2004) described in section 2.2.4, JQuranTree uses orthographic data from the Tanzil project (Zarrabi-Zadeh, 2011). Released in 2008, this is the only accurate digital copy of the Quran. To ensure accuracy, this project was developed using a multi-stage approach. In the first stage, previous digital copies of the Quran were compared to produce an initial candidate text. This was followed by automatic verification using a set of morphological rules based on traditional grammar. The final stage was manual verification. Verse checksums were computed manually using all letters and diacritics from the *madīnah muṣḥaf* and then compared to the digital version. The orthographic representation in this chapter is based on the Uthmani *ḥafṣ* data published by Tanzil Project as a Unicode dataset.

## 4.3   Formal Representation

Unicode is a computing standard for representing text that covers most of the world's writing systems and is used as a data format for exchanging multilingual information. Formally, a Unicode string $s$ is a sequence of Unicode characters:

$$s = (c_1, \ldots, c_n) \mid c_i \in U \quad (1 \leq i \leq n)$$

Each Unicode character $c$, from the set of all characters $U$, has an associated numerical code. Different code ranges are reserved for different languages. For Arabic, Unicode characters represent either letters or diacritical marks, with diacritics following letters in multiple permitted permutations. For the Quran, there have been proposals to extend Unicode to allow for more fine-grained representations. For example, Pournader (2010) suggests new characters to represent subtle variations in diacritics such as open *tanwīn* and the combined versions of small *wāw* used in Quranic script. Despite not implementing these extensions, the orthographic Tanzil data represents the Uthmani script with sufficient accuracy for the morphosyntactic annotation work in this thesis.

| Character | Glyph | Character | Glyph |
|-----------|-------|-----------|-------|
| *alif* | ا | *tatwīl* | ـ |
| *bā'* | ب | | |
| *tā'* | ت | | |
| *thā'* | ث | Small high *sīn* | ڛ |
| *jīm* | ج | | |
| *ḥā'* | ح | Small high rounded zero | ْ |
| *khā'* | خ | | |
| *dāl* | د | | |
| *dhāl* | ذ | Small high upright rectangular zero | ٌ |
| *rā'* | ر | | |
| *zayn* | ز | Small high *mīm* (isolated form) | ۭ |
| *sīn* | س | | |
| *shīn* | ش | | |
| *ṣād* | ص | Small low *sīn* | ڛ |
| *ḍād* | ض | | |
| *ṭā'* | ط | Small *wāw* | و |
| *dthā'* | ظ | | |
| *'ayn* | ع | Small *yā'* | ے |
| *ghayn* | غ | | |
| *fā'* | ف | | |
| *qāf* | ق | Small high *nūn* | ن |
| *kāf* | ك | | |
| *lām* | ل | Empty center low stop | ۖ |
| *mīm* | م | | |
| *nūn* | ن | Empty center high stop | ۖ |
| *hā'* | ه | | |
| *wāw* | و | Rounded high stop with filled center | ۬ |
| *yā'* | ي | | |
| *hamza* | ء | | |
| *alif maqṣūra* | ى | Small low *mīm* | ۖ |
| *tā' marbūṭa* | ة | | |

Table 4.1: Base characters in the orthographic representation.

| Diacritic | Position / description | Glyph |
|---|:---:|:---:|
| *fatḥa* | Above | َ |
| *ḍamma* | Above | ُ |
| *kasra* | Below | ِ |
| *fatḥatān* | Double *fatḥa* | ً |
| *ḍammatān* | Double *ḍamma* | ٌ |
| *kasratān* | Double *kasra* | ٍ |
| *shadda* | Above | ّ |
| *sukūn* | Above | ْ |
| *madda* | Above | آ |
| *hamza* above | Above[1] | أ |
| *hamza* below | Below[1] | إ |
| *hamzat waṣl* | Above *alif* | ٱ |
| *alif khanjarīya* | Superscript *alif* | ٰ |

[1] Diacritic *hamza* shown above/below *alif* for illustrative purposes.

Table 4.2: Attached diacritics with their positions relative to base characters.

For orthographic processing, JQuranTree does not use Unicode for two reasons. Firstly, locating a letter by ordinal position requires scanning up to that point in a verse, as diacritic sequences can have variable length, resulting in linear, instead of constant, time complexity. Secondly, characters such as *alif* and *alif khanjarīya* are in fact the same underlying Arabic letter with only a stylistic difference, and should be handled uniformly in tasks such as morphological analysis. Instead, JQuranTree uses a character-plus-diacritic representation. In this representation variations such as *alif* and *alif khanjarīya* map to the same base characters with distinguishing marking features, simplifying text comparisons with diacritics.

The character-plus-diacritic representation uses two sets of glyphs. To define the representation, let $B$ be the set base characters, and $D$ be the set of diacritics. The set of base characters is derived from the Tanzil data and includes the letters and recitation marks used in the Quran (Table 4.1, page 64). The set of diacritics is shown in Table 4.2 (page 65). A string $s$ of Arabic text is then formally defined as a sequence of compound characters, each of which is a base character (from $B$), together with a set of zero or more attached diacritics (a subset of $D$):

$$s = (c_1, \ldots, c_n)$$

$$c_i = (b_i, d_i) \mid b_i \in B \land d_i \subseteq D \ (1 \leq i \leq n)$$

An example of this representation for the third word of verse (70:8) is shown in Figure 4.3. This word is pronounced *al-samā'u* ('the sky'). The diagram shows the word written in Classical Arabic script, followed by its composition into six base characters with diacritics attached to five of these. The lower part of the diagram shows the character-plus-diacritic representation as a list of pairs $(b_i, d_i)$:

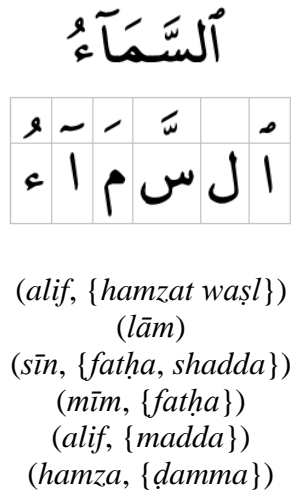

(*alif*, {*hamzat waṣl*})
(*lām*)
(*sīn*, {*fatḥa, shadda*})
(*mīm*, {*fatḥa*})
(*alif*, {*madda*})
(*hamza*, {*ḍamma*})

Figure 4.3: Character-plus-diacritic representation for Arabic script.

## 4.4   Computational Model

### 4.4.1 Java Object Model

JQuranTree uses Object Oriented Programming (OOP) to represent orthography. This is the computational design paradigm used for Java programming. Figure 4.4 shows the classes used to implement the character-plus-diacritic representation.[4]

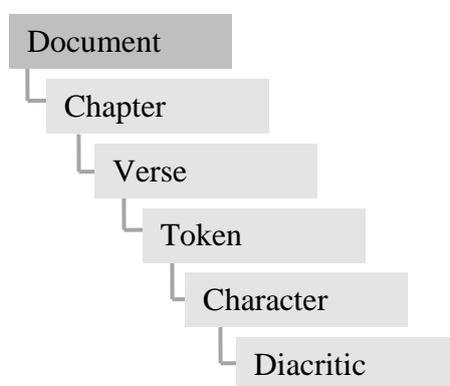

Figure 4.4: Class hierarchy for orthography in JQuranTree.

These Java classes are based on the following definitions:

**Document:** The Quran is modelled as a single text document.

**Chapter:** One of the 114 numbered chapters in the Quran.

**Verse:** One of the numbered verses in a chapter.

**Token:** A whitespace-delimited span of text within a verse.

**Character:** A base character from the set $B$ in Table 4.1 (page 64).

**Diacritic:** A diacritic from the set $D$ in Table 4.2 (page 65).

---

[4] This implementation is freely available online: http://corpus.quran.com/java

In Arabic computational processing, the term 'token' can have multiple meanings depending on the processing task, such as a word or its subdivisions. JQuranTree uses the term token to denote a whitespace-delimited run of text within a Quranic verse. These are often words, although in the Quran multiple words with different stems are occasionally fused as a compound word-form. Morphological segmentation for compound forms is discussed in Chapter 5.

## 4.4.2 Location Notation

The Quran is divided into 114 chapters, with each chapter divided into a sequence of numbered verses. The pair notation (*c*:*v*) is often used in scholarly works to reference verses within the Quran. For example (6:76) refers to verse 76, chapter 6. This thesis extends this notation to tokens using the following definition:

> A **location** uniquely identifies a token as a triple (*c*:*v*:*t*) where *c* is a chapter number, *v* is a verse number, and *t* is a token number.

The Location class in JQuranTree models this concept computationally. In the Quran Arabic Corpus, this notation is used to assign a unique reference number to tokens in the Quran, and appears in morphological and syntactic diagrams online. Location numbers are also used by annotators during online discussion to refer to particular parts of verses and chapters. They will also be used in the syntactic representation in Chapter 6, in which each token is annotated with its location number in the corpus.

## 4.4.3 Internal Representation

Internally, JQuranTree uses a byte-encoded representation for orthographic data that has been optimized for efficient access. This allows the morphological and syntactic algorithms described later in this thesis to rapidly process the Quranic text. As described in section 4.3, given a block of Unicode Arabic text with diacritics, locating a letter by offset requires a linear-time scan, as sequences of diacritics are of variable length. The class hierarchy in JQuranTree allows access to individual Arabic letters. However, for the entire Quran, representing each letter with its own Java object would not be a memory-efficient approach.

Both of these concerns are addressed by using a byte buffer, with a fixed width for each letter including its diacritics. In JQuranTree, character objects are a view on the buffer, and are created on demand and garbage collected. Each character is represented by three bytes. The first byte encodes the character type. The second and third bytes form a vector of bits. Each attached diacritic has a fixed position in the bit vector, and if the bit is set then the diacritic is present. The maximum range of values possible in this encoding scheme would be 256 types of base character, and combinations of 16 diacritic types. In practice, only 44 base character types and 13 diacritic combinations are used in Classical Arabic.

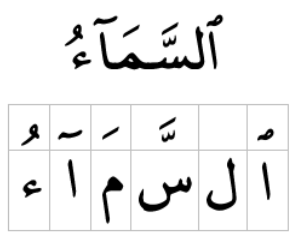

| Character-plus-diacritics | Byte 1 | Byte 2 | Byte 3 |
|---|---|---|---|
| (*alif*, {*hamzat waṣl*}) | 0 | 0 | $2^3$ |
| (*lām*) | 22 | 0 | 0 |
| (*sīn*, {*fatḥa*, *shadda*}) | 11 | $2^0 + 2^6$ | 0 |
| (*mīm*, {*fatḥa*}) | 23 | $2^0$ | 0 |
| (*alif*, {*madda*}) | 0 | 0 | $2^0$ |
| (*hamza*, {*ḍamma*}) | 28 | $2^1$ | 0 |

Figure 4.5: Internal orthographic encoding.

As an example, the upper part of Figure 4.5 (page 69) shows the character-plus-diacritic representation for token (70:8:3). The table in the lower part of the diagram shows the internal encoding. In contrast to Unicode, where multiple byte-encodings are possible, the token's six characters and their attached diacritics are unambiguously represented using the following 24 bytes:

$$(0, 0, 8, 22, 0, 0, 11, 65, 0, 23, 1, 0, 0, 0, 1, 28, 2, 0)$$

The Quran contains 6,236 verses. Representing all orthographic data from the Tanzil project in Unicode would require 1,389,662 bytes (1.33 megabytes). The bit-packed representation used by the orthography model uses 1,242,006 bytes (1.18 megabytes). Dividing this by three, we get 414,002 characters for all verse text including whitespace, as the internal representation has a constant ratio of characters to bytes, regardless of the number of attached diacritics.

## 4.4.4 Unicode Conversion

Converting to and from Unicode is supported by JQuranTree to allow the Uthmani script to be loaded into the orthographic model, and for exporting Arabic text for display on the corpus website. The decoding process is reversible and is tested via the round trip method: a Unicode encoder is used to serialize the orthography model back into Unicode, and tests are run to ensure that the original character data is recovered and no orthographic information is lost.

Unicode decoding (converting from Unicode into the character-plus-diacritic representation) is performed using table lookup.[5] For each Unicode character in the Uthmani script, the orthographic base character and diacritics are determined. Several Unicode characters may be decoded as a single orthographic base character. If table lookup results in a character, then a new base character is formed. Otherwise, if the lookup results in only a diacritic, then that diacritic marker will be combined with the previous base character.

---

[5] http://corpus.quran.com/java/unicode.jsp

Unicode encoding (converting from the character-plus-diacritic representation into Unicode) is more complex than decoding. A given subset of the orthographic model could have multiple representations in Unicode. This is not only because Unicode allows combining marks to be ordered arbitrarily, but also because certain combinations of letters and diacritics (such as *alif* and *hamza*) have an alternative representation as a single pre-composed Unicode character.

The encoding algorithm is shown in Figure 4.6 below. The algorithm's steps ensure that round trip testing is possible. Given Tanzil orthographic data, the original sequence of Unicode characters will be recovered after deserializing then reserializing. The algorithm uses the same conversion table for decoding so that Unicode serialization is perfectly reversible.

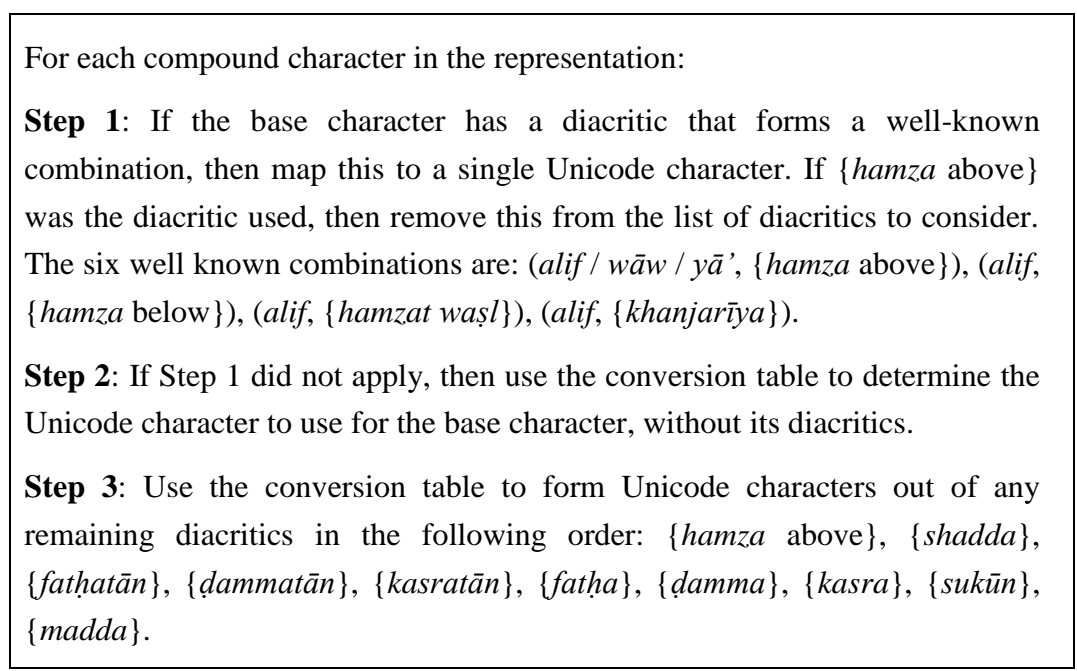

For each compound character in the representation:

**Step 1**: If the base character has a diacritic that forms a well-known combination, then map this to a single Unicode character. If {*hamza* above} was the diacritic used, then remove this from the list of diacritics to consider. The six well known combinations are: (*alif / wāw / yā'*, {*hamza* above}), (*alif*, {*hamza* below}), (*alif*, {*hamzat waṣl*}), (*alif*, {*khanjarīya*}).

**Step 2**: If Step 1 did not apply, then use the conversion table to determine the Unicode character to use for the base character, without its diacritics.

**Step 3**: Use the conversion table to form Unicode characters out of any remaining diacritics in the following order: {*hamza* above}, {*shadda*}, {*fatḥatān*}, {*ḍammatān*}, {*kasratān*}, {*fatḥa*}, {*ḍamma*}, {*kasra*}, {*sukūn*}, {*madda*}.

Figure 4.6: Unicode encoding algorithm.

## 4.4.5 Extended Buckwalter Transliteration

In addition to Unicode conversion, JQuranTree supports converting to and from Buckwalter transliteration. This is an ASCII-based encoding scheme that is fully reversible, so that no information is lost during transliteration. A reversible transliteration scheme can be used for precisely specifying the orthography of Arabic words in computational work. The BAMA system described in section 2.2.1 stores its morphological lexicon in this format, and this data will be used in Chapter 7 for Classical Arabic annotation.

JQuranTree extends Buckwalter's scheme to include additional symbols in the Uthmani script. Four non-Arabic characters in the original scheme (not found in the Quran) are used for dialects and foreign words: P (*peh*), J (*tcheh*), V (*veh*) and G (*gaf*). The combination character (*alif*, {*madda*}), encoded as a vertical bar '|', is also not used in the Tanzil orthographic data. These characters are not implemented by JQuranTree. Similarly, 14 Quranic symbols do not feature in the original scheme. In the extended scheme these are assigned to ASCII punctuation marks, which is unambiguous as modern punctuation does not occur in the Quran.

Table 4.3 (overleaf) shows the additional characters. As an example, token (19:7:6) in the Quran is the proper noun Yahya (يَحْيَىٰ), which would be encoded as yaHoyaY`. The Token class in JQuranTree implements this conversion process. Figure 4.7 shows an example Java program for accessing this implementation:

```java
public class BuckwalterExample {

    public static void main() {

        Token token = Document.getToken(19, 7, 6);

        System.out.println(token.toBuckwalter());

    }
}
```

Figure 4.7: Example JQuranTree program for Buckwalter transliteration.

| Symbol | Encoding |
|--------|----------|
| *Madda* | ^ |
| *hamza* above | # |
| Small high *sīn* | : |
| Small high rounded zero | @ |
| Small high upright rectangular zero | " |
| Small high *mīm* (isolated form) | [ |
| Small low *sīn* | ; |
| Small *wāw* | , |
| Small *yā'* | . |
| Small high *nūn* | ! |
| Empty centre low stop | - |
| Empty centre high stop | + |
| Rounded high stop with filled center | % |
| Small low *mīm* | ] |

Table 4.3: Additional characters in extended Buckwalter transliteration.

## 4.4.6 Orthographic Search

JQuranTree implements the class TokenSearch for orthographic search. This finds all tokens that match an orthographic form specified using extended Buckwalter transliteration and is useful for tasks such as implementing a concordance. Figure 4.8 (overleaf) shows an example Java program that uses this class to find occurrences of the orthographic form *qamar* (the word 'moon') in the Quran. When run, this program will display all exactly matching surface forms (قَمَر).

```
public class TokenSearchExample {
    TokenSearch search
        = new TokenSearch(EncodingType.Buckwalter);
    search.findSubstring("qamar");
    System.out.println(search.getResults());
}
```

Figure 4.8: Orthographic search using Buckwalter transliteration.

Because orthographic search is used to find tokens that match a specific spelling with diacritic markers, this type of search is used to find exact matches regardless of morphological inflection. Online, the corpus website extends this search to provide users with a more flexible search based on matching lemmas, parts-of-speech tags and morphological features (described in section 8.4.2).

## 4.5   Conclusion

The Uthmani script of the Quran has complex orthography and includes additional characters and markings not used in Modern Arabic. These include verse pause marks for specifying detailed pronunciation, and diacritical marks used to indicate inflection as part of Arabic's morphological and syntactic rules.

This chapter described a formal orthographic representation for the Quran, as well as JQuranTree, the representation's realization as a computational system. To represent the Quranic text, orthographic data from the Tanzil project was used (Zarrabi-Zadeh, 2011). This work was required to unambiguously represent the Classical Arabic script of the Quran in a computational system, so that no orthographic information is lost during processing. JQuranTree is made freely available online as an open source project for accessing and searching the original Arabic text of Quran. The orthographic model presented here will be next used for the morphological representation described in the following chapter.

*The Semitic root is one of the great miracles of man's language.*

*– Johannes Lohmann*

# 5 Morphological Representation

## 5.1 Introduction

This chapter describes the formal representation used to develop morphological annotation in the Quranic Arabic Corpus. The representation provides a model for Classical Arabic word structure that is designed to be fine-grained and suitable for statistical parsing. Computationally, the formalism is based on the lexeme-plus-feature representation reviewed in section 2.2.2 (Habash, 2007a) for two reasons. Firstly, analyzing word structure using a lemma and a set of features is an intuitive approach to Arabic morphology that is easily understandable by annotators. Secondly, the feature-value data structures in Habash's representation are directly applicable to machine learning and parsing work.

However, the representation described in this chapter differs in several respects. Following the direction taken by Sawalha and Atwell (2013), a more fine-grained approach is used for Arabic morphology. As described in the literature review, annotating a set of detailed morphological features during treebank construction improves parsing accuracy. Another difference is that Habash's scheme is designed for Modern Arabic. For Classical Arabic, different features and part-of-speech tags are used that more closely align the representation to traditional sources. Finally, an alternative segmentation scheme is used that is better suited to the Quranic text. Inspired by recent computational work for Arabic morphology by Smrž (2007) and Habash (2007a; 2010), both form and function are modelled. Form is modelled by segmenting Arabic words into their constituent morphemes. Function is modelled by associating a set of morphological features with each segment, such as person, gender, number and syntactic inflection features.

The remainder of this chapter is organized as follows. Section 5.2 provides an overview of Classical Arabic morphology and defines key terminology. Section 5.3 provides a formal description of the representation. Sections 5.4, 5.8 and 5.9 describe the part-of-speech tagset, the feature set and the segmentation scheme respectively. Section 5.10 compares formal representations of Classical Arabic morphological structures to traditional analyses and section 5.11 concludes.

## 5.2 Classical Arabic Morphology

### 5.2.1 Traditional Morphological Analysis

Classical Arabic is a morphologically-rich language with complex word structure. In traditional Arabic grammar, morphological analysis is a well-established field of study known as *ṣarf* (صرف), which has been continuously developed from the start of the Arabic linguistic tradition by grammarians. Prominent examples include Sibawayh (760-796), who devoted half of *al-kitāb* to the subject. He described Arabic's inflectional and derivational processes, as well as its root and pattern system (Carter, 2004). Al-Farra (731-822) and Al-Akhfash (d. 830) each wrote linguistic works focused entirely on morphological analysis. Ibn Jinni (932-1002) further developed the field, and was the first Arabic grammarian to explicitly define the difference between morphology and syntax, famously stating:

> Morphology deals with the form of words, while syntax studies words in their different contexts.

By the time of the grammarian Ibn Mas'ud (ca. 1250-1300), morphological analysis for Classical Arabic was highly developed, and on par with modern linguistic work. His treatise *marāḥ al-arwāḥ* contained detailed descriptions of verb and noun patterns, providing phonological and semantic context for Arabic's rich morphology, building on a large body of previous work (Akesson, 2011).

Concepts from Classical Arabic morphology are also applicable to Modern Arabic, as both forms of the language share a common morphological system. However, there are distinctions between the two. For example, in spoken Modern Arabic inflection is simplified and case endings are generally omitted, whereas Classical Arabic is fully vocalized. Similarly, Classical Arabic has a richer set of particles that are used as concatenative prefixes, such as the *hamza* of equalization (همزة التسوية), requiring a different set of segmentation rules.

## 5.2.2 Roots and Patterns

A distinguishing feature of Arabic, and other Semitic languages such as Hebrew, is nonconcatenative morphology (Habash, 2007; Boudelaa and Marslen-Wilson, 2001). Most Arabic words can be structured as the combination of two abstract morphemes: a lexical root and a template pattern. This approach is termed nonconcatenative because the root's letters are not always found consecutively in derived words. The use of roots and patterns was an early development in the Arabic linguistic tradition (Muhammad, 2007; Versteegh, 1997b). For Modern Arabic, this has remained the standard approach in morphological analysis (Mace, 2007; Wightwick and Gaafar, 2008). For example, both Classical and Modern Arabic dictionaries are organized by root. For the purposes of computational work in this thesis, the following definitions will be used for Classical Arabic:

A **root** (*jithr* – جذر) is a sequence of three or four consonants (known as radicals) that is used to derive a group of related words. These sequences are known as triliteral and quadriliteral roots respectively.

A **pattern** (*wazn* – وزن) is a template consisting of consonants and vowels together with placeholders for a root's radicals.

**Derivation** (*ishtiqāq* – اشتقاق) is the morphological process in which a root in combination with a pattern generates a derived word.

The nonconcatenative system for word generation in Arabic is well developed. Several hundred patterns in combination with thousands of roots allows for a large number of possible derived words, although in practice the number of roots is limited. For Classical Arabic, Lane's Lexicon lists 3,775 roots based on traditional sources (Lane, 1992). A more comprehensive Classical Arabic dictionary is *lisān al-'arab* (لسان العرب) by Ibn Manzur (1233-1312). Hegazi and El-Sharkawi (1985) estimate that the lexicon contains 6,350 triliteral roots and 2,500 quadriliteral roots, although only 1,200 of these are still used in Modern Arabic. For Modern Arabic as a whole, Ryding (2005) estimates that between 5,000 and 6,500 roots are currently in use.

In both varieties of Arabic, roots are used to form words with related meanings. For this reason, a root is said to generate a semantic field (Badawi and Haleem, 2008). The canonical example used to illustrate this is the root *ka ta ba* (ك ت ب), used in both Classical and Modern Arabic. This root generates the verb 'write' (*kataba* – كتب) and the nouns 'writing' (*kitābah* – كتابة), 'writer' (*kātib* – كاتب), 'book' (*kitāb* – كتاب) and 'desk / office' (*maktab* – مكتب). In traditional analysis, the patterns used to derive these words are specified using the placeholder letters *fā' 'ayn lām* (ف ع ل). For example, the pattern for *kātib* (كاتب) is *fā'il* (فاعل), a form I active participle. In the Quranic Arabic Corpus, root tagging is the basis for further annotation including derived and inflectional morphological forms.

## 5.2.3 Inflection and Concatenation

In Arabic, derived words can undergo two changes before appearing in their final surface form, due to semantic and syntactic context:

> **Inflection** (*taṣrīf* – تصريف) is the morphological process in which the form of a word is modified by grammatical attributes or syntactic function.

> **Concatenation** is the morphological process in which the form of a word is modified by attaching prefixes and suffixes. A **stem** is the part the word to which prefixes and suffixes are attached.

In the process of inflection, words are modified by grammatical attributes. For example, the masculine form for teacher, *mu'alim* (معلم) becomes *mu'alimah* (معلمة) in the feminine. Relevant to parsing work, words are also inflected for syntactic function through case endings. In morphological concatenation, words are further modified by attaching prefixes and suffixes. Unlike in English, where the syntactic unit is primarily the word, in Arabic, stems, prefixes and suffixes are units for syntactic analysis, requiring decomposition as a prerequisite for parsing:

**Segmentation** is the reverse process of concatenation.

**Morphological segments** are the concatenative morphemes that result from segmentation. These are stems, prefixes and suffixes.

To illustrate these concepts, Figure 5.1 below shows token (14:22:30) from the Quran. This compound word بِمُصْرِخِكُم (translated as 'with your helper') exhibits rich morphology. Its surface form (*bimus'rikhikum*) is a concatenation of a prefixed preposition (*bi*), a stem (a form IV active participle – *mus'rikh*) and a suffixed pronoun (*kum*). The stem's surface form is related to its syntactic function. Due to the prefixed preposition, the stem is inflected for the genitive case (*mus'rikhi*). Figure 5.2 (overleaf) shows how the word is composed through a combination of derivation, inflection and concatenation.

(14:22:30)

*bimus'rikhikum*

'with your helper'

بِمُصْرِخِكُم

Figure 5.1: A compound Classical Arabic word-form in verse (14:22).

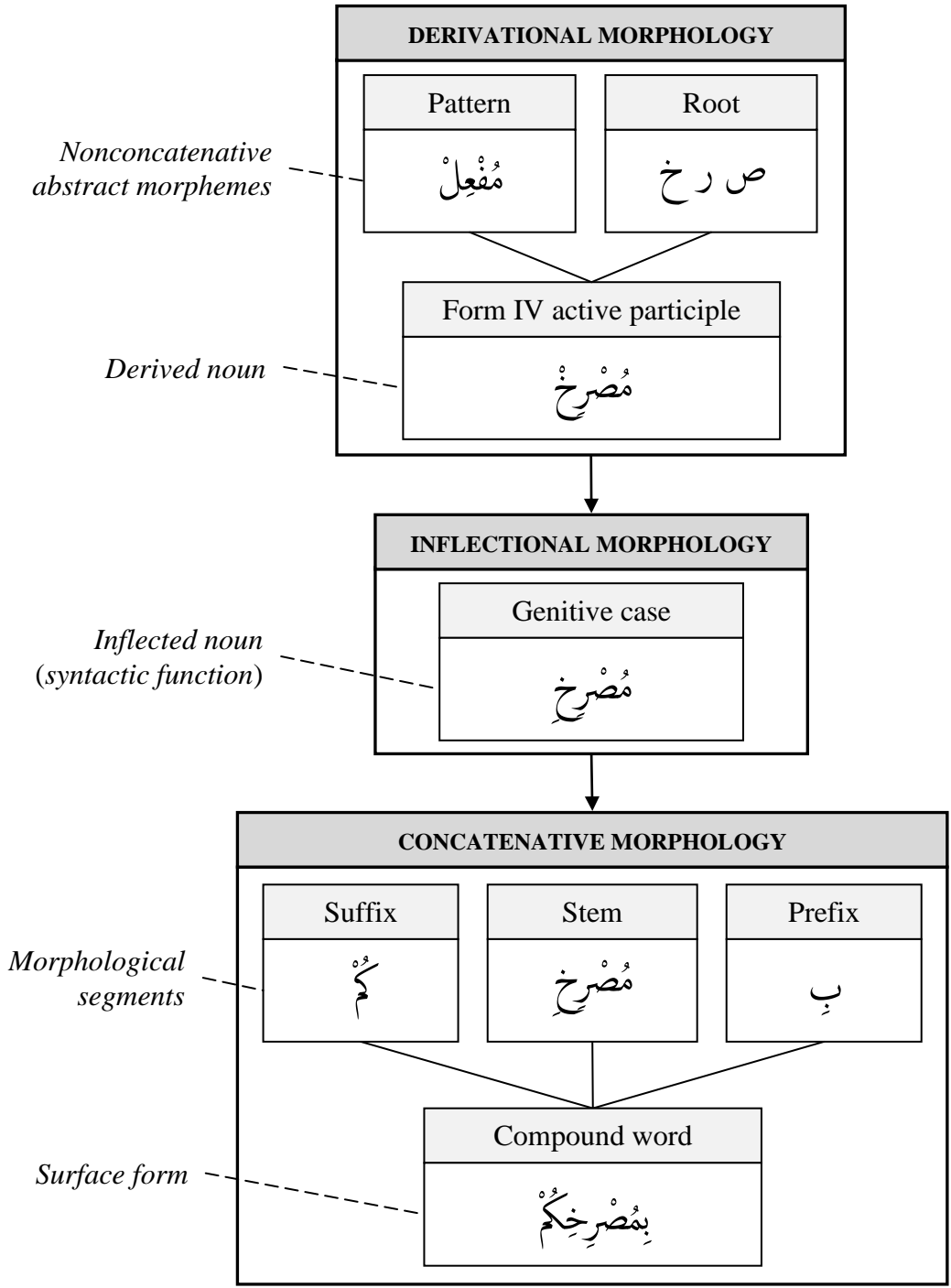

Figure 5.2: Derivational and inflectional morphology with form and function.

## 5.2.4 Lemmas

In Arabic, a root gives rise to a group of derived words with related meanings. Each of these derived words gives rise to a secondary group of words that differ only by inflection. In Arabic lexicographic analysis, this inflection group is known as a lexeme:

> A **lexeme** is a group of words with the same derivational morphology that differ only by inflection.

> A **lemma** (also known as a citation form) is a conventional choice of one word that represents a lexeme.

Both Modern and Classical Arabic dictionary entries are organized by root then lemma, but stop short of enumerating inflected or concatenated forms due to the large number of inflection patterns.

## 5.3   Formal Representation

## 5.3.1 Segmentation

This section formalizes Classical Arabic morphological structures by extending Habash's lexeme-plus-feature representation for Modern Arabic (Habash, 2007a). This is based on the concept of using a lemma and a set of feature-value pairs. In contrast to Habash's work, the representation here supports multiple stems. This is due to the frequent occurrence of contractions in Classical Arabic script, such as the fused word-form 'about-what' (*'amma* – عَمَّ) consisting of the particles 'about' (*'an* – عَن) and 'what' (*mā* – مَا), each with a distinct stem and syntactic function. For this reason, the lemma and features are attached to individual morphological segments, instead of the word-level attachment in Habash's scheme. As a consequence, each segment in a Classical Arabic word has its own part-of-speech.

The first part of the formalization describes segmentation. A token was defined in Chapter 4 as a whitespace-delimited span of text. This is either a single stem or a compound word-form constructed by concatenating multiple segments. Using the orthographic representation from section 4.3, a token $w$ is a sequence of base characters with attached diacritics:

$$w = (c_1, \ldots, c_n)$$

$$c_i = (b_i, d_i) \mid b_i \in B \land d_i \subseteq D \ (1 \leq i \leq n)$$

Morphologically, a token is partitioned into a sequence of $m$ segments. Let each segment $s_i$ ($1 \leq i \leq m$) span base characters in the token from positions $S(i)$ to $E(i)$. The following constraints are used to ensure that the partition covers the entire token continuously:

$$w = (s_1, \ldots, s_m)$$

$$S(1) = 1 \land \ E(m) = n$$

$$S(i+1) = E(i) + 1 \ \ (1 \leq i < m)$$

$$E(i) \geq S(i) \ \ (1 \leq i \leq m)$$

This definition of segmentation applies to all segments except those of zero-length. These are abbreviated suffixed pronouns represented by a diacritic, such as (3:35:5) *rabbi* (رَبِّ) – 'my lord'. This special case is described in section 5.9.

## 5.3.2 Feature-Value Pairs

The representation associates a set of feature-value pairs with each morphological segment in a token $w = (s_1, \ldots, s_m)$. Formally, a feature is a function that maps a segment to a feature-value:

$$f_j(s_i) \in F_j \;\; (1 \leq i \leq m, \, 1 \leq j \leq M)$$

Here $M$ is the number of features in the representation and $F_j$ is the set of possible values for feature $f_j$. In the annotation scheme, the term 'feature' is used in a functional sense. These include segment type (stem, prefix or suffix), root, lemma and grammatical features such as person, gender and number.

## 5.3.3 Feature Notation

The Quranic Arabic Corpus uses a formal notation for morphological annotation, written as a sequence of tags in square brackets. Morphologically annotated data is stored in the corpus database using this format. Each tag either starts a new segment, or describes a feature-value pair associated with the previous segment. For example, the compound word-form *bimus'rikhikum* (بِمُصۡرِخِكُم) in Figure 5.1 (page 79) is tagged as:

[bi+ POS:N ACT PCPL (IV) LEM:muSorix ROOT:Srx M GEN PRON:2MP]

In this example, the symbol bi+ is the prefixed preposition *bi*. POS:N is a noun (a stem) followed by derivation features (active participle, form IV). The next two features are the stem's lemma and root specified using Buckwalter transliteration, followed by inflection features for masculine and the genitive case. The symbol PRON:2MP is a suffixed second person masculine plural pronoun. These tags correspond to the morphological analysis in Figure 5.2 (page 80). This notation is designed to be machine-readable but is also purposefully verbose so that annotators do not have to frequently consult annotation guidelines to look up the meaning of tags. The remainder of this chapter describes the part-of-speech tags and morphological features for Classical Arabic in more detail.

## 5.4 Parts of Speech

### 5.4.1 The Part-of-Speech Hierarchy in Arabic Grammar

In traditional Arabic grammar, parts of speech are organized into a hierarchy consisting of three main classes that are divided into subclasses (Owens, 1989). The main classes are nominals (*ism* – اسم), verbs (*fi'il* – فعل) and particles (*ḥarf* – حرف). This classification was introduced at the beginning of the Arabic linguistic tradition. For example, Sibawayh's *kitāb* opens by establishing that the topic of his book is speech (*kalām*) and that speech is divided into three main categories. He divides the class of nouns into subclasses including explicit nouns and pronouns, and organizes the class of particles by their syntactic function (Carter, 2004; Baalbaki, 2008). Later grammarians refined these subdivisions, such as Ibn Hisham who developed a detailed classification of particles according to syntactic and semantic usage (Gully, 1995).

   However, a frequent simplification for certain computational tasks is that Arabic has only three parts of speech. In contrast to the work in this thesis, several Arabic computational systems have previously relied on only the three main classes. Examples of underrepresentation includes parsing work by Mehdi (1985) and Shokrollahi-Far at al. (2009), verbal representations by Islam et al. (2010) and stemming work for information retrieval by Moukdad (2006). As noted by Attia (2008), the simplification that Arabic has only three parts of speech arises by only considering the main classes and not their subdivisions:

> It is quite surprising to see many morphological analyzers today influenced by the misconception that Arabic parts of speech are exclusively nouns, verbs and particles. The Xerox Arabic morphological analyzer is a good example of this limitation (Beesley, 2001). In Xerox morphology, words are classified strictly into verbs, nouns and particles; no other categorical description is used.

In deeper computational analysis, such as the work presented in this thesis, part-of-speech tagsets are more fine-grained. Other examples of rich tagsets for Arabic include the Penn Arabic Treebank tagset by Buckwalter (2002), the Prague Arabic Dependency Treebank tagset by Hajič et al. (2004), and the theory-neutral tagset by Sawalha and Atwell (2010). Modern Arabic computational work often cites traditional grammar as a source of inspiration. For example, the tagger developed by Khoja (2001) uses a tagset based on traditional sources:

> Since the grammar of Arabic has been standardized for centuries, [the tagset] is derived from this grammatical tradition rather than from an Indo-European based tagset. Arabic grammarians traditionally analyze all Arabic words into three main parts-of-speech. These parts-of-speech are further subcategorized into more detailed parts-of-speech which collectively cover the whole of the Arabic language.

## 5.4.2 Part-of-Speech Analysis in *al-i'rāb al-mufaṣṣal*

For Classical Arabic part-of-speech tagging, the Quranic Arabic Corpus uses as its primary reference *al-i'rāb al-mufaṣṣal likitāb allāh al-murattal* ('A Detailed Grammatical Analysis of the Recited Quran using *i'rāb*') (Salih, 2007). Because this work builds on multiple sources, it provides morphological and syntactic analysis for the entire Quran. Salih provides more detail in comparison to related works such as Darwish (1996), who instead provides more concise grammatical analysis alongside exegetic commentary.

Developing a part-of-speech tagset using Salih as a reference is complicated by several factors. Firstly, he does not list or define his grammatical terminology, assuming the reader has expertise with traditional grammar and is familiar with its conventions. At over 10,000 pages of prose, the reference work is also lengthy, using alternative terminology in different places. Finally, the text is not available in an easily machine-readable form, making automatic extraction of its analyses unviable. Consequently, deriving a complete listing of grammatical terminology in Salih's work is only possible by reviewing the complete text.

The part-of-speech tagset presented here is based on a manual review of Salih's analysis. During this review, the key terms for parts-of-speech, morphological features and syntactic constructions were documented and compared to Darwish's terminology. The two works were found to use essentially the same standardized terms. However, although both works primarily use Basran terminology, Salih also uses Kufan. For example, he often uses the Kufan term *na't* (نعت) alongside the Basran *ṣifa* (صفة) for adjectives (Carter, 2000). An example of Salih's analysis for verse (77:21) is shown in Figure 5.3 below. This provides morphological analysis with segmentation and part-of-speech tagging, together with a description of syntactic structure: [6]

﴿فَجَعَلْنَٰهُ فِى قَرَارٖ مَّكِينٍ ٢١﴾

فجعلناه : الفاء عاطفة . جعل : فعل ماضٍ مبني على السكون لاتصاله بنا . و «نا» ضمير متصل مبني على السكون في محل رفع فاعل والهاء ضمير متصل مبني على الضم في محل نصب مفعول به .

في قرار مكين : جار ومجرور متعلق بجعلناه وهو في مقام المفعول الثاني لأن المعنى فصيرناه في موضع أو مكان حصين أو منيع . مكين : صفة - نعت - لقرار مجرورة مثلها بالكسرة .

Figure 5.3: Salih's grammatical analysis for verse (77:21).

---

[6] Salih (2007). Volume 12, page 297.

In his morphological analysis, Salih's divides the first word in the verse (فجعلناه) into four segments: a prefixed conjunctive particle (الفاء عاطفة), a verb (فعل ماض), a suffixed subject pronoun (ضمير متصل في محل رفع فاعل) and a suffixed object pronoun (ضمير متصل في محل نصب مفعول به). The second and third words in the verse are described as a prepositional phrase (جار ومجرور). This concise analysis assumes that the reader is sufficiently familiar with traditional grammar to understand that these two words are a preposition and a noun respectively. Finally, the last word of the verse is tagged as an adjective (صفة – نعت).

## 5.4.3 Part-of-Speech Tags for Classical Arabic

The complete part-of-speech tagset adapted from Salih's analysis contains 44 tags (Table 5.1, overleaf). In the table, tags have been organized into a hierarchy with three levels. The first level (column one) consists of the three main parts-of-speech from traditional grammar: the nominals (*ism* – اسم), verbs (*fi'il* – فعل) and particles (*ḥarf* – حرف). The second level (column two) is an intermediate category. The third level in the tagset consists of the fine-grained parts-of-speech used to tag morphological segments (columns three to five). Only part-of-speech tags from this level are stored in the corpus database. The other two levels are abstract groups that are used to describe morphology and parts-of-speech in general terms.

The last two columns in Table 5.1 provide descriptions using both English and Arabic terminology. For Arabic, Salih's most commonly used term is listed for each part-of-speech. For English, equivalent terminology for nominal tags was derived by comparing three Classical Arabic reference grammars and selecting the most suitable translation based on Salih's usage of each term (Wright, 2007; Haywood and Nahmad, 1990; Fischer and Rodgers, 2002). For particles, terminology from Gully (1995) was adapted by comparing to the dictionary of Quranic usage by Badawi and Haleem (2008).

Figure 5.4 (page 89) shows example morphological segmentation and part-of-speech tagging for verses (1:1-7) of the Quran. The next three sections describe the part-of-speech tagset for Classical Arabic in more detail.

| Class | Subclass | Tag | Description | Arabic Term |
|---|---|---|---|---|
| Nominals | Nouns | N | Noun | اسم |
| | | PN | Proper noun | اسم علم |
| | Derived nominals | ADJ | Adjective | صفة |
| | | IMPN | Imperative verbal noun | اسم فعل أمر |
| | Pronouns | PRON | Personal pronoun | ضمير |
| | | DEM | Demonstrative pronoun | اسم اشارة |
| | | REL | Relative pronoun | اسم موصول |
| | Adverbs | T | Time adverb | ظرف زمان |
| | | LOC | Location adverb | ظرف مكان |
| Verbs | Verbs | V | Verb | فعل |
| Particles | Prepositions | P | Preposition | حرف جر |
| | *lām* prefixes | EMPH | Emphatic *lām* prefix | لام التوكيد |
| | | IMPV | Imperative *lām* prefix | لام الامر |
| | | PRP | Purpose *lām* prefix | لام التعليل |
| | Conjunctions | CONJ | Coordinating conjunction | حرف عطف |
| | | SUB | Subordinating conjunction | حرف مصدري |
| | Other particles | ACC | Accusative particle | حرف نصب |
| | | AMD | Amendment particle | حرف استدراك |
| | | ANS | Answer particle | حرف جواب |
| | | AVR | Aversion particle | حرف ردع |
| | | CAUS | Particle of cause | حرف سببية |
| | | CERT | Particle of certainty | حرف تحقيق |
| | | CIRC | Circumstantial particle | حرف حال |
| | | COM | Comitative particle | واو المعية |
| | | COND | Conditional particle | حرف شرط |
| | | EQ | Equalization particle | حرف تسوية |
| | | EXH | Exhortation particle | حرف تحضيض |
| | | EXL | Explanation particle | حرف تفصيل |
| | | EXP | Exceptive particle | أداة استثناء |
| | | FUT | Future particle | حرف استقبال |
| | | INC | Inceptive particle | حرف ابتداء |
| | | INT | Particle of interpretation | حرف تفسير |
| | | INTG | Interrogative particle | حرف استفهام |
| | | NEG | Negative particle | حرف نفي |
| | | PREV | Preventive particle | حرف كاف |
| | | PRO | Prohibition particle | حرف نهي |
| | | REM | Resumption particle | حرف استئنافية |
| | | RES | Restriction particle | أداة حصر |
| | | RET | Retraction particle | حرف اضراب |
| | | RSLT | Result particle | حرف واقع في جواب الشرط |
| | | SUP | Supplemental particle | حرف زائد |
| | | SUR | Surprise particle | حرف فجاءة |
| | | VOC | Vocative particle | حرف نداء |
| | Quranic initials | INL | Disconnected letters | حروف مقطعة |

Table 5.1: Part-of-speech tags for Classical Arabic.

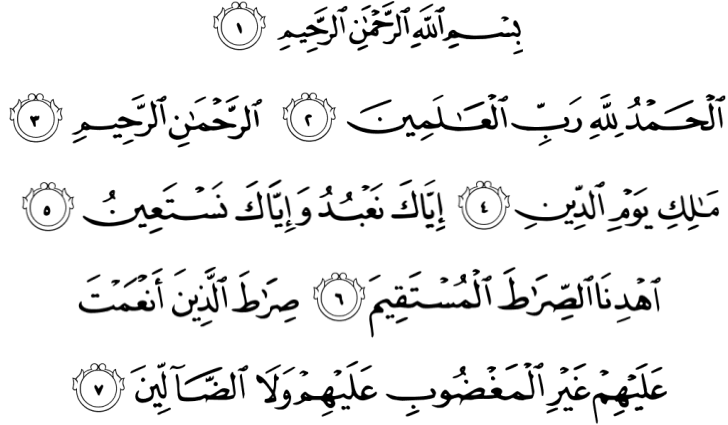

(1:1) *bi*/P *s'mi*/N *allāhi*/PN *al*/DET *raḥmāni*/ADJ *al*/DET *raḥīmi*/ADJ

(1:2) *al*/DET *ḥamdu*/N *li*/P *llāhi*/PN *rabbi*/N *al*/DET *'ālamīna*/N

(1:3) *al*/DET *raḥmāni*/ADJ *al*/DET *raḥīmi*/ADJ

(1:4) *māliki*/N *yawmi*/N *al*/DET *dīni*/N

(1:5) *iyyāka*/PRON *na'budu*/V *wa*/CONJ *iyyāka*/PRON *nasta'īnu*/V

(1:6) *ih'di*/V *nā*/PRON *al*/DET *ṣirāta*/N *al*/DET *mus'taqīma*/ADJ

(1:7) *ṣirāta*/N *alladhīna*/REL *an'am*/V *ta*/PRON *'alay*/P *him*/PRON *ghayri*/N
*al*/DET *maghḍūbi*/N *'alay*/P *him*/PRON *wa*/CONJ *lā*/NEG *al*/DET *ḍālīna*/N

Figure 5.4: Uthmani script and part-of-speech tagging for verses (1:1-7).

## 5.5   Nominals

The term *ism* (اسم) in Arabic linguistics is an autohyponym, used by traditional
grammarians to refer to one of the three main parts-of-speech, as well as one of its
subclasses. These two cases are distinguished in Arabic computational work by using the term 'nominal' for the general class, and the term 'noun' for the specific subclass (Diab, 2007; Smrž, 2007; Habash and Roth, 2009c). In the Quranic corpus, nine tags are used for nominals: POS:N and POS:PN for nouns and proper nouns, POS:PRON, POS:DEM and POS:REL for personal, demonstrative and relative pronouns, POS:ADJ for adjectives, POS:LOC and POS:T for adverbs of place and time, and POS:IMPN for the imperative verbal noun.

## 5.5.1 Nouns

In Arabic grammar, words are classified as nouns (POS:N) primarily according to syntactic criteria (Owens, 1989). For example, Al-Zajjaji (892-951) defined a noun as a word occurring as the subject or object of a verb. Ibn Jinni (932-1002) included the more specific criteria that nouns are words placed into the genitive case by prepositions (حرف جر). Remarkably similar criteria are used in modern linguistics to define nouns. For example, Loos et al. (2004) propose the universal definition that nouns are words acting as the subjects or objects of verbs, or as the objects of prepositions or postpositions.

## 5.5.2 Proper Nouns

Classical Arabic script makes no orthographic distinction between nouns and proper nouns (اسم علم), unlike English where capitalization is used. However, most proper nouns (tagged as POS:PN) have the grammatical property that they are definite without having to carry the *al-* determiner prefix. Many proper nouns in the Quran are of a foreign or ancient origin. Morphologically, these fall outside the root and pattern system and are subject to restricted inflection rules. For example, the name Aaron (*harūn* – هارون) is a diptote (ممنوع من الصرف) and has same inflected case-ending for both the genitive and accusative case. Although Salih flags diptotes, he does not generally indicate which nominals are proper nouns. A prominent exception to this is the name Allah (الله), which is referred to as *lafth al-jalālah* (لفظ الجلالة), literally 'the majestic name'.

### 5.5.3 Personal Pronouns

In traditional grammar, personal pronouns (POS:PRON), are classified into two types. Suffixed pronouns are known as *ḍamīr muttaṣil* (ضمير متصل). These require segmentation for annotation, described further in section 5.9. The second type are separate words known as *ḍamīr munfaṣil* (ضمير منفصل), forming a small closed class of inflected forms (Table 5.2). In Arabic, personal pronouns include forms not found in English, such as the second person dual *antumā* ('you two'). To simplify the segmentation process, members of the lexeme *iyyā* (إيّا), such as the third person masculine singular form *iyyāhu* (إيّاهُ), are also tagged as POS:PRON and annotated as a single word. These are known traditionally as *ḍamīr naṣb munfaṣil* (ضمير نصب منفصل), and are syntactically used as objects.

| Person | | Singular | Dual | Plural |
|---|---|---|---|---|
| First | | أَنا | (none) | نحن |
| Second | Masculine | أَنتَ | أَنتُما | أَنتُم |
| | Feminine | أَنتِ | | أَنتُنّ |
| Third | Masculine | هو | هُما | هُم |
| | Feminine | هي | | هُنّ |

Table 5.2: Independent personal pronouns.

### 5.5.4 Demonstrative Pronouns

Demonstrative pronouns are known as *ism ishāra* (اسم اشارة) and are tagged as POS:DEM. Traditional grammarians distinguish between demonstratives used for objects that are near (*ism ishāra lilqarib* – اسم اشارة للقريب) and far (*ism ishāra lilba'id* – اسم اشارة للبعيد). The same distinction is found in other languages such as English. The main inflection forms are shown in Table 5.3 (overleaf).

| Type | Number | Gloss | Gender | Form | |
|------|--------|-------|--------|------|------|
| Near | Singular | this | Masculine | هذا | *hādhā* |
| | | | Feminine | هذه | *hādhihi* |
| | Dual | these (two) | Masculine | هذان | *hādhāni* |
| | | | Feminine | هتان | *hātāni* |
| | Plural | these (all) | All | هؤلاء | *hā'ulā'i* |
| Far | Singular | that | Masculine | ذلك | *dhālika* |
| | | | Feminine | تلك | *tilka* |
| | Dual | those (two) | Masculine | ذانك | *dhānika* |
| | | | Feminine[1] | تانك | *tānika* |
| | Plural | those (all) | All | أولئك | *ulā'ika* |

[1] This inflected form is not used in the Quran.

Table 5.3: Main inflection forms for demonstrative pronouns.

## 5.5.5 Relative Pronouns

Relative pronouns (POS:REL) are known as *ism mawṣūl* (اسم موصول) in Arabic. Syntactically, these connect a relative clause to its main clause. Certain words such as inflected forms of *alladhī* (الذي) are easily tagged as relative pronouns as this is their main part-of-speech. Other relative pronouns include *man* (من) and *mā* (ما). However, because these two words frequently occur in more than one grammatical category, syntactic context is required to choose the correct part-of-speech tag. For example, the word *mā* ('what') is tagged as POS:REL in verse (109:2): *lā aʿbudu mā taʿbudūna* (لَا أَعْبُدُ مَا تَعْبُدُونَ) – 'I do not worship what you worship.' In contrast, *mā* ('what') is tagged as an interrogative (POS:INTG) in verse (99:3): *waqāla al-insānu mā lahā* (وَقَالَ الْإِنسَانُ مَا لَهَا) – 'And man says, "What is [wrong] with it?"'.

## 5.5.6 Adjectives

Adjectives (*ṣifa* – صفة) are tagged as POS:ADJ and are closely related to nouns (POS:N). Without context, it can be difficult to distinguish the two as both occur with similar morphological features. For example, both can carry the prefix *al-* ('the'). For this reason, adjectives are tagged according to syntactic criteria. In Classical Arabic, an adjective appears after the noun it describes, and is subject to a set of grammatical agreement rules. An example is the two-word verse (101:11) which consists of a noun followed by an adjective. Both words are indefinite and in the nominative case: *nā'run ḥāmiyatun* (نَارٌ حَامِيَةٌ) – 'a blazing fire'.

## 5.5.7 Adverbs

The term 'adverb' is used to describe a variety of grammatical categories in part-of-speech tagsets for English, with different classifications used for different tagged corpora (Atwell, 2008; Nancarrow, 2011). For part-of-speech tagging in the Quranic Arabic Corpus, the term is specifically used for the adverbs of place (POS:LOC) – *dharf makān* (ظرف مكان) and the adverbs of time (POS:T) – *dharf zamān* (ظرف زمان). These usually appear in adverbial expressions in the accusative case. For example, *warā'a* ('behind') is tagged as POS:LOC in verse (84:10): *wa-ammā man ūtiya kitābahu warā'a dhahrihi* (وَأَمَّا مَنْ أُوتِيَ كِتَابَهُ وَرَاءَ ظَهْرِهِ) – 'But as for he who is given his record behind his back'. Similarly, *aḥqāban* ('ages') appears in the accusative case and is tagged as POS:T in verse (78:23): *lābithīna fīhā aḥqāban* (لَّابِثِينَ فِيهَا أَحْقَابًا) – 'In which they will remain for ages'.

## 5.5.8 Imperative Verbal Nouns

Salih uses the grammatical term *ism fi'il 'amr* (اسم فعل أمر) in only a few places in the Quran. In the Quranic Arabic Corpus, these words are tagged as imperative verbal nouns (POS:IMPN). For example, this tag is used for the word *misāsa* (مِسَاسَ) in verse (20:97). In this context, the word appears as a nominal, yet has an imperative meaning: *lā misāsa* (لَا مِسَاسَ) – 'do not touch'.

## 5.6   Verbs

Verbs are one of the three main parts-of-speech in traditional Arabic grammar, and are known as *fi'il* (فعل). Historically, grammarians classified words as verbs primarily using semantic and morphological criteria. For example, Al-Zajjaji defined a verb semantically as a word that represents past, present and future actions. Ibn Hisham defined a verb morphologically as a word derived from a root using a well-known verbal pattern (Owens, 1989). In the Quranic Arabic Corpus, verbs are annotated using the POS:V tag. Morphological features are used to subclassify verbs according to their template pattern, inflection attributes and syntactic group. For example, verbs in the group known as *kāna wa akhwātuhā* (كان واخواتها) are tagged as POS:V together with a feature marker. In contrast, nominals derived from verbs, such as participles, are tagged as either POS:N or POS:ADJ according to their syntactic usage.

## 5.7   Particles

In traditional Arabic grammar, a word is classified as a particle, *ḥarf* (حرف), if it is neither a nominal (اسم) nor a verb (فعل). In contrast to previous tagged Arabic corpora, the Quranic Arabic Corpus provides deep annotation of particles using 34 tags. In the tagset hierarchy, particles are subclassified into Quranic initials (POS:INL), prepositions (POS:P), conjunctions (POS:CONJ and POS:SUB), prefixed *lām* particles (three additional tags), and other particles (27 tags).

### 5.7.1 Quranic Initials

Quranic initials, *ḥuruf muqaṭṭa'ah* (حروف مقطعة), are sequences of disconnected letters, such as *alif lām mīm* (م ل أ), that appear at the start of several chapters in the Quran. Their interpretation has no firm consensus in Quranic exegesis, and in Islam their meaning is generally considered to be a divine secret (Shahid, 2000). As their grammatical function is not specified, they are tagged as a separate part-of-speech (POS:INL).

## 5.7.2 Prepositions

Prepositions (POS:P) are known as *ḥarf jar* (حرف جر). They precede nominals, placing them into the genitive case. Independent prepositions include '*alā* (على) and *fī* (في), usually translated as 'on' and 'in' respectively. POS:P is also used to tag vowelized prepositional prefixes, including *bā'* (ب), *kāf* (ك), *tā'* (ت), *wāw* (و), and one of the senses of *lām* (ل). In contrast to Modern Arabic which has a reduced set of prefixes, *tā'* and *wāw* occur in Classical Arabic as particles as oath. For example *tāllah* ('by Allah') in verse (37:56): *qāla tāllahi in kidtta laturdīni* (قَالَ تَاللَّـهِ إِن كِدتَّ لَتُرْدِينِ) – 'He will say, "By Allah, you almost ruined me."'.

## 5.7.3 Prefixed *lām* Particles

The prefix *lām* (ل) has four uses including its use as preposition. POS:EMPH is used for the emphatic prefix (لام التوكيد), such as (4:66:23) *lakāna* (لَّكَانَ) – 'surely it would have been'. POS:IMPV is used for the imperative prefix (لام الامر) which precedes imperfect verbs placing them into the jussive mood, such as (106:3:1): *falya'budū* (فَلْيَعْبُدُوا) – 'so let them worship'. The prefix *lām* also occurs as a particle of purpose (لام التعليل) tagged as POS:PRP. In this construction, the particle introduces a subordinate clause and places the following verb into the subjunctive mood, such as (72:17:1) *linaftinahum* (لِّنَفْتِنَهُمْ) – 'that we might test them'.

## 5.7.4 Coordinating and Subordinating Conjunctions

In traditional grammar, coordinating conjunctions (حرف عطف) are particles that connect two words or phrases, and are tagged as POS:CONJ. The prefixed particle *wāw* (و) used in its conjunctive sense ('and') is the most common coordinating conjunction. Independent coordinating conjunctions include *thumma* (ثُمَّ) 'then', as well as *aw* (أَوْ) and *am* (أَمْ), usually translated as 'or'. Subordinating conjunctions are tagged as POS:SUB. In Classical Arabic, the most common subordinating conjunction (حرف مصدري) is one sense of the particle *an* (أَن), usually translated as 'that'. Syntactically, particles tagged as POS:SUB introduce subordinate clauses.

## 5.7.5 Other Particles

In addition to the part-of-speech tags described in the preceding sections, a further 27 tags are used for other particles (the fourth subclass in Table 5.1, page 88). Some of these particles appear only in Classical and not Modern Arabic such as the prefixed *hamza* of equalization (همزة التسوية), tagged as POS:EQ. Historically, grammarians such as Ibn Hisham provided detailed analysis of Arabic particles (Gully, 1995). Based on traditional sources, the Quranic Arabic Corpus tagset is used to classify particles according to both syntactic and semantic criteria.

Syntactically, traditional Arabic grammar describes the rules that determine the way in which particles modify the inflection of surrounding words. An example is the vocative particles (حرف نداء), tagged as POS:VOC. These precede nouns and place them into the nominative or accusative case according to syntactic context and the nature of the individuals being addressed. Similarly, exceptive particles (أداة استثناء) tagged as POS:EXP place nouns into the accusative case depending on contextual negation and ellipsis (Ansari, 2000; Jones, 2005). Another example of the syntactic classification of particles is the frequently occurring accusative particles (*ḥarf naṣb*), tagged as POS:ACC. In traditional Arabic grammar, a group of accusative particles known as *inna wa akhwātuhā* (ان واخواتها) are considered to be verb-like (حرف مشبه بالفعل), as they appear in syntactic constructions similar to verbs. Like the verb *kāna* (كان), these particles take a subject and a predicate. However, they differ from verbs syntactically by placing their subjects (اسم ان) into the accusative case, and their objects (خبر ان) into the nominative case.

Other particles are classified on semantic grounds. These include the negative particles (حرف نفي) tagged as POS:NEG, prohibition particles (حرف نهي) tagged as POS:PRO and interrogative particles (حرف استفهام) tagged as POS:INTG. The tag POS:SUP is used for supplemental particles (حرف زائد), which occur infrequently in the Quran. Grammarians consider these particles to supplement an existing sentence. Although they do not generally add extra meaning, they often make a sentence sound better when recited aloud, improving a verse's prosodic balance (Wohaibi, 2001).

## 5.8   Morphological Features

In addition to part-of-speech tagging, morphological segments are annotated with multiple feature-value pairs encoded as a sequence of feature tags. Table 5.4 (overleaf) summarizes the feature tags used in the corpus.

### 5.8.1 Prefixes

During morphological segmentation, word-forms are segmented into prefixes, stems and suffixes. Prefix features are annotated using the notation X:C+ where X is the prefixed particle and C is its part-of-speech tag. For example, f:CONJ+ is used for words with the particle *fā'* (ف) prefixed as a coordinating conjunction (الفاء عاطفة). The notation X+ is used for prefixes that belong to only a single part-of-speech, such as the prefix feature Al+ for the determiner *al* (لام التعريف).

### 5.8.2 Suffixes

Two suffix features are annotated using the notation +X. The first is the vocative suffix +VOC. This is only used with the word *allāh* to produce the vocative word-form *allāhumma* (اللَّـهُمَّ) that occurs several times in the Quran. The second suffix tag is +n:EMPH, used to denote an emphatic suffixed letter *nūn* (نون التوكيد). The compound PRON: tag is used for suffixed pronouns (ضمير متصل) in combination with person, gender and number features. For example, PRON:3MS represents a suffixed pronoun inflected for the third person masculine singular.

### 5.8.3 Classification Features

In addition to the part-of-speech tag (formally considered a feature) a further three features are used to classify words. ROOT: and LEM: indicate roots and lemmas, specified using Buckwalter transliteration. For example LEM:kitaAb for the lemma *kitāb* (كتاب). The SP: feature is used to group words with a special syntactic function in traditional grammar. It is used for *kāna wa akhwātuhā* (كان واخواتها), *kāda wa akhwātuhā* (كاد واخواتها) and *inna wa akhwātuhā* (ان واخواتها).

| Type | Category | Tag | Description |
|---|---|---|---|
| Prefixes | Letter *alif* as a prefixed particle | A:INTG+ | Interrogative *alif* (همزة استفهام) |
| | | A:EQ+ | Equalization *alif* (همزة التسوية) |
| | Letter *wāw* as a prefixed particle | w:CONJ+ | Conjunction *wāw* (الواو عاطفة) |
| | | w:REM+ | Resumption *wāw* (الواو استئنافية) |
| | | w:CIRC+ | Circumstantial *wāw* (حرف حال) |
| | | w:SUP+ | Supplemental *wāw* (الواو زائدة) |
| | | w:P+ | Preposition *wāw* (حرف جر) |
| | | w:COM+ | Comitative *wāw* (واو المعية) |
| | Letter *fā'* as a prefixed particle | f:CONJ+ | Conjunction *fā'* (الفاء عاطفة) |
| | | f:REM+ | Resumption *fā'* (الفاء استئنافية) |
| | | f:SUP+ | Supplemental *fā'* (الفاء زائدة) |
| | | f:RSLT+ | Result *fā'* (الفاء واقعة في جواب الشرط) |
| | | f:CAUS+ | Cause *fā'* (الفاء سببية) |
| | Letter *lām* as a prefixed particle | l:P+ | Preposition *lām* (حرف جر) |
| | | l:EMPH+ | Emphasis *lām* (لام التوكيد) |
| | | l:PRP+ | Purpose *lām* (لام الأمر) |
| | | l:IMPV+ | Imperative *lām* (لام التعليل) |
| | Other prefixes | Al+ | Determiner *al* (لام التعريف) |
| | | bi+ | Preposition *bā'* (حرف جر) |
| | | ka+ | Preposition *kāf* (حرف جر) |
| | | ta+ | Preposition *tā'* (حرف جر) |
| | | sa+ | Future particle *sīn* (حرف استقبال) |
| | | ya+ | Vocative particle *yā'* (أداة نداء) |
| | | ha+ | Vocative particle *hā'* (أداة نداء) |
| Core Features | Classification features | POS | Part-of-speech |
| | | LEM: | Lemma |
| | | ROOT: | Root (جذر) |
| | | SP: | Special group (e.g. كان واخواتها) |
| | Verbal features | Form | I to XII (وزن) |
| | | Aspect | Perfect, imperfect or imperative |
| | | Mood | Indicative, subjunctive or jussive |
| | | Voice | Active (معلوم) or passive (مجهول) |
| | Nominal features | Derivation | Participle or verbal noun |
| | | State | Definite (معرفة) or indefinite (نكرة) |
| | | Case | Nominative, accusative or genitive |
| | Phi features | Person | First, second or third (الاسناد) |
| | | Gender | Masculine or feminine (الجنس) |
| | | Number | Singular, dual or plural (العدد) |
| Suffixes | Suffix features | +VOC | Vocative suffix (used for اللَّهُمَّ) |
| | | +n:EMPH | Emphasis *nūn* (نون التوكيد) |
| | | PRON: | Pronoun suffix (ضمير متصل) |

Table 5.4: Morphological feature tags for Classical Arabic.

## 5.8.4 Phi Features

The phi-features for Classical Arabic are person, gender and number, and are annotated using a compound tag. For example, 3MS represents third person masculine singular. The values for the person feature are first person (المتكلّم), second person (المخاطَب) and third person (الغائب). Gender (الجنس) is a complex topic in Arabic and words may have different values for semantic, morphemic and grammatical gender. In the corpus, grammatical gender is tagged, as this is the most useful type of gender for syntactic annotation.

## 5.8.5 Verbal Features

The features aspect, mood, voice and form apply to verbs and their derivatives: active and passive participles and verbal nouns. In Arabic grammar, aspect is closely related to but distinct from tense. The aspects tags are PERF for perfect (فعل ماض), IMPF for imperfect (فعل مضارع) and IMPV for imperative (فعل أمر). The mood tags are IND for indicative (مرفوع), SUBJ for subjunctive (منصوب) and JUS for jussive (مجزوم). Voice is tagged as either ACT for active (مبني للمعلوم) or PASS for passive (مبني للمجهول). Verb forms are tagged using roman numerals (I to IX), a convention introduced in Western works describing traditional Arabic grammar (Haywood and Nahmad, 1990; Wright, 2007).

## 5.8.6 Nominal Features

In Arabic, nominals may be in a definite (معرفة) or indefinite (نكرة) state. These are tagged using the features DEF and INDEF respectively. Nominals derived from verbs are tagged using a derivation feature. The possible values are ACT PCPL for the active participle (اسم فاعل), PASS PCPL for the passive participle (اسم مفعول) and VN for verbal nouns (مصدر). In various linguistic constructions, nominals with these derivation tags function similarly to verbs. Syntactically, nominals are also found in one of three cases: NOM for the nominative case (مرفوع), ACC for the accusative case (منصوب) and GEN for the genitive case (مجرور).

## 5.9 Segmentation Rules

A segmenter is a computational component that divides words into segments. The segmenter developed for the Quranic Arabic Corpus splits words using annotated morphological features. For example, a word tagged as w:CONJ+ POS:N will be divided into the prefixed letter *wāw* followed by the remaining letters as a stem. Segmentation for the Quran is challenging due to the Uthmani script's complex orthography with multiple possible forms for prefixes and suffixes as well as the presence of zero-length morphological segments. Table 5.5 below summarizes the morphological segmentation rules used in the corpus:

| Type | Feature | Segmentation | Example |
|------|---------|--------------|---------|
| Prefixes | w:CONJ+, … | Single letter particles | (5:15:22) وَكَتَبٌ |
| | *alif* prefixes | Single letter *alif* | (21:36:9) أَهَذَا |
| | | Single letter *hamza* | (56:59:1) ءَأَنتُمْ |
| | ya+ or ha+ | Single letter vocative | (20:94:2) يَبْنَؤُمَّ |
| | | Two letter vocative | (20:36:5) يَمُوسَىٰ |
| | Al+ | Two letter determiner | (2:2) ٱلْكِتَٰبُ |
| | | Single letter after *lām* | (16:69:18) لِّلنَّاسِ |
| | | Elided letter *alif* | (26:176:3) لَٔيْكَةَ |
| Stems | POS: | Single stem | (67:1:3) بِيَدِهِ |
| | | Two stems | (15:32:5) أَلَّا |
| Suffixes | +VOC | Single letter suffix | (10:10:4) ٱللَّهُمَّ |
| | +n:EMPH | Emphatic letter *nūn* | (3:188:2) تَحْسَبَنَّ |
| | | Emphatic letter *alif* | (12:32:17) وَلَيَكُونًا |
| | Verb subjects | Subject pronoun | (1:7:3) أَنْعَمْتَ |
| | | Subject with object | (18:76:3) سَأَلْتُكَ |
| | PRON: | Elided (zero-length) | (3:35:5) رَبِّ |
| | | Single object | (38:20:2) مُلْكَهُ |
| | | Two objects | (8:43:2) يُرِيكَهُمُ |
| | | Two objects and subject | (33:37:31) زَوَّجْنَٰكَهَا |

Table 5.5: Morphological segmentation rules for Classical Arabic.

Most of the rules for segmenting prefixes relate to a single letter segment. For example, the features w:CONJ+ and f:REM+ represent segments consisting of the letters *wāw* (و) and *fā'* (ف) respectively. Other rules depend on orthographic and morphological context, such as the prefix feature Al+ used to tag determiners. In Arabic, the determiner is the letter *lām* (لام التعليل). However, this takes a different surface form according to the presence of a preceding *lām* particle. In the Uthmani script, this is written in three different ways forming a segment either one or two letters long, such as (2:2) ٱلْكِتَٰبُ, (16:69:18) لِلنَّاسِ or (26:176:3) لِـَٔيْكَةِ.

Stems are constructed after the segmenter processes prefixes and suffixes. The remaining letters in a word either form one or two stems. Double stems occur in Classical Arabic as compound contractions, such as (15:32:5) – *allā* (أَلَّا) 'that-not'. In the context of its verse, this word is tagged as POS:SUB POS:NEG, a subordinating conjunction *an* (أَنْ) 'that' and a negative particle *lā* (لا) 'not'. There are a limited number of two stem combinations, and the segmenter builds these by using a lookup table of concatenated surface forms.

The rules for suffixes apply to vocative and emphatic particles, and pronouns. In Classical Arabic, suffixed pronouns occur in several forms as they inflect for person, gender and number. The segmenter builds two types of pronoun segments. The first type are subject pronouns. These are attached to verbs and their surface form depends on the phi-features as well as the verb's aspect. For example, the second person masculine singular verb *an'amta* (أَنْعَمْتَ) in verse (1:7) is divided into a verb stem and the suffixed letter *tā'* (ت). In his grammatical analysis for this verse, Salih refers to the letter *tā'* as an attached pronoun in the syntactic role of a nominative subject (التاء ضمير متصل في محل رفع فاعل).[7] The second type of suffixed pronoun segments are object pronouns. In Classical Arabic, these also inflect for phi-features but can be abbreviated, such as (3:35:5) – *rabbi* (رَبِّ) 'my Lord'. In this example, the letter *yā'* (ي) has been omitted (الياء محذوفة) from the possessive pronoun, but is indicated by the presence of a diacritic *kasra*.[8]

---

[7] Salih (2007). Volume 1, page 10.
[8] Ibid. Volume 2, page 42.

## 5.10    Morphological Structures

This section compares the formal representation to Salih's traditional analysis for two short verses of the Quran. The first verse (4:68) shows example tagging for a noun, an adjective and a verb with concatenative morphology. The second verse (74:42) illustrates how correctly annotating the inflectional case of nominals requires understanding both morphological and syntactic context.

### 5.10.1   Prefix and Suffix Concatenation

Figure 5.5 below shows morphological annotation for verse (4:68) in the Quranic Arabic Corpus, using the part-of-speech and feature tags described in this chapter. Morphologically, the first word (4:68:1) consists of five segments: two prefixes, a stem and two suffixes. Salih describes the first segment as a prefixed conjunction, annotated in the corpus as w:CONJ+ (الواو حرف عطف) followed by an emphatic prefix l:EMPH+ (اللام حرف توكيد), and a perfect verb stem (فعل ماض). He describes two suffixes: a subject pronoun 'we' (نا» ضمير متصل فى محل رفع فاعل) and an object pronoun 'them' (هم» ضمير الغائبين مبني على السكون في محل نصب مفعول به أول).[9]

Figure 5.5: Morphological annotation for verse (4:68).

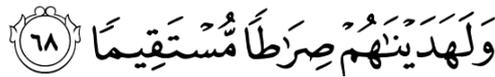

*walahadaynāhum ṣirāṭan musˈṭaqīman*

'And we would have guided them to a straight path.'

(4:68:1) [w:CONJ+ l:EMPH+ POS:V PERF LEM:hadaY ROOT:hdy 1P PRON:3MP]

(4:68:2) [POS:N LEM:Sira`T ROOT:SrT M INDEF ACC]

(4:68:3) [POS:ADJ ACT PCPL (X) LEM:m~usotaqiym ROOT:qwm M INDEF ACC]

---

[9] Salih (2007). Volume 2, page 317.

The second word in the verse (*ṣirāṭan*) is described as a second object inflected for the accusative case (مفعول به ثانٍ منصوب بالفتحة). This is tagged as a noun in the corpus. He analyses the last word in the verse (*mus'taqīman*) as an adjective. Since this describes the preceding noun, it is also in the accusative case. Additional features annotated in Figure 5.5 include root, lemma, derivation and phi-features. For example, (4:68:3) is annotated as POS:ADJ ACT PCPL (X), indicating a form X active participle adjective. These features are not present in Salih's analysis but are included in the Quranic Arabic Corpus as part of its fine-grained annotation.

## 5.10.2  Diptote Inflectional Case

Annotation for verse (74:42) is shown in Figure 5.6 below. This verse consists of four Arabic words, translated as 'What put you in Saqar?' The proper noun 'Saqar' is one of the Classical Arabic names for Hell, and is morphologically ambiguous. This word is a diptote with the same surface case ending (a diacritical *fatḥa*) for both the accusative and genitive cases. Correctly annotating the proper noun's case requires determining its syntactic role in the verse.

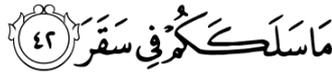

*mā salakakum fī saqara*

'What put you in Saqar?'

(74:42:1) [POS:INTG LEM:maA]

(74:42:2) [POS:V PERF LEM:salaka ROOT:slk 3MS PRON:2MP]

(74:42:3) [POS:P LEM:fiY]

(74:42:4) [POS:PN LEM:saqar GEN]

Figure 5.6: Morphological annotation for verse (74:42).

In Salih's analysis, the first word is specified an interrogative particle acting as a subject (اسم استفهام مبني على السكون في محل رفع مبتدأ). This is tagged as POS:INTG in the corpus. The second word is a perfect verb (POS:V) which Salih indicates is inflected for third person masculine by specifying the form of its dropped subject pronoun (فعل ماض مبني علي الفتح والفاعل ضمير مستتر فيه جوازاً تقديره هو). Salih also describes a suffixed pronoun attached to the verb composed of two Arabic letters (الكاف ضمير متصل – ضمير المخاطبين – مبني على الضم في محل نصب مفعول به). He indicates that the second letter *mīm* is a plural marker (والميم علامة جمع الذكور اي ما ادخلكم). In contrast, the Quranic Arabic Corpus uses a simplified representation where these two letters consist of a single morphological segment tagged as PRON:2MP. The relationship between morphological surface form and syntactic function for the word 'Saqar' is made clear in Salih's analysis of the last two words of the verse. These are described as a prepositional phrase (جار ومجرور). For this reason, the word is in the genitive and not the accusative case, indicated by the diacritical *fatḥa* (علامة جر الاسم الفتحة بدلاً من الكسرة لانه منوع من الصرف للتأنيث والمعرفة).[10]

## 5.11   Conclusion

Classical Arabic has a complex morphological system that includes derivational, inflectional and concatenative morphological processes. This chapter discussed the morphological representation used in the Quranic Arabic Corpus, defining key terminology in Arabic computational morphology, as well as providing a formal description of segmentation structures. The annotation scheme was also described, consisting of a fine-grained part-of-speech tagset and a lexeme-plus-feature representation that is closely aligned to traditional sources. The work in this chapter demonstrated the relationship between morphological form and syntactic function. It was also shown that Arabic words require segmentation into multiple morphemes, as these are the basic syntactic unit in traditional Arabic grammar. Morphological segments and feature-values will be used in the next chapter to develop the hybrid syntactic representation.

---

[10] Salih (2007). Volume 12, page 250.

Language is a process of free creation; its laws and principles are fixed, but the manner in which the principles are used is free and infinitely varied.

*– Noam Chomsky*

# 6    Syntactic Representation

## 6.1   Introduction

The Quranic Treebank is the syntactic layer in the Quranic Arabic Corpus. This chapter describes its hybrid representation that in contrast to previous formal work for Arabic, combines aspects of both dependency and constituency syntax. Computationally, the resulting structures are more complex in comparison to previous Arabic treebanks. However, annotators developing the Quranic Treebank have found this approach to be intuitive and closely aligned to traditional sources. The hybrid representation is inspired by two traditional concepts. The first is syntactic position (*maḥal* – محل), such as the subject and predicate in nominal sentences. Due to substitution, positions can be filled not only by words but also by phrases and sentences, leading to phrase-structure. The second concept is governance (*'amal* – عمل), realized as a lexical element's inflectional change due to a governing element (*'āmil* – عامل). Elements related through governance form dependency relations, such as a verb governing its subject in the nominative case.

The remainder of this chapter is organized as follows. Section 6.2 describes Classical Arabic syntax. Section 6.3 reviews previous work that relates traditional Arabic grammar to constituency and dependency theories, and compares this to a hybrid representation. Section 6.4 provides a formal definition of the representation using directed labelled graphs. Sections 6.5 and 6.6 describe the dependency relations and phrase-structure tags used in the Quranic Treebank. Section 6.7 compares the annotation scheme to traditional analysis for example syntactic structures and section 6.8 concludes.

## 6.2 Classical Arabic Syntax

This section describes the traditional concepts of position (محل), governance (عمل) and ellipsis (حذف), using examples from Salih's *al-i'rāb al-mufaṣṣal* (Salih, 2007).

### 6.2.1 Syntactic Position

In traditional *i'rāb*, words and phrases are found in different syntactic positions known as *maḥal* (محل).[11] Figure 6.1 below shows a nominal and a verbal sentence with each position occupied by a single word. From right-to-left, the nominal sentence in verse (112:2) has a subject (مبتدأ) and a predicate (خبر) position. The main positions for verbal sentences such as (29:44) are the verb (فعل), its subject (فاعل) and for transitive verbs, an object (مفعول به).

| VERBAL SENTENCE (29:44) | | | | NOMINAL SENTENCE (112:2) | |
|---|---|---|---|---|---|
| Object | Subject | Verb | | Predicate | Subject |
| *alsamāwāti* | *allahu* | *khalaqa* | | *alṣamadu* | *allahu* |
| ٱلسَّمَٰوَٰتِ | ٱللَّهُ | خَلَقَ | | ٱلصَّمَدُ | ٱللَّهُ |
| Acc. | Nom. | Ind. | | Nom. | Nom. |

Figure 6.1: Nominal and verbal positions in verses (112:2) and (29:44).

When a single word occupies a position, it will inflect as nominative (مرفوع) if a subject or predicate, or as accusative (منصوب) if an object. Similarly, verbs if ungoverned, conjugate as indicative (مرفوع). Different named positions are used for other sentence types. For example, a position termed the subject representative (نائب فاعل) is used to describe a verb's subject in passive constructions.

---

[11] Alternative terms include *makān* (مكان) and *mawqi'* (موقع) (Versteegh, 1978).

## 6.2.2 Dependencies

In modern linguistic theory, a dependency is a binary relation that relates two lexical elements such as words or morphemes. A dependency is asymmetrical, distinguishing a dependent lexical element from its head (Mel'čuk, 1988). Similar to modern theory, the concept of governance (*'amal* – عمل) in traditional grammar explains the syntactic effect of one element on another using a binary relation between a governing element (*'āmil* – عامل) and its dependent (*ma'mūl* – معمول) (Versteegh, 1997b).

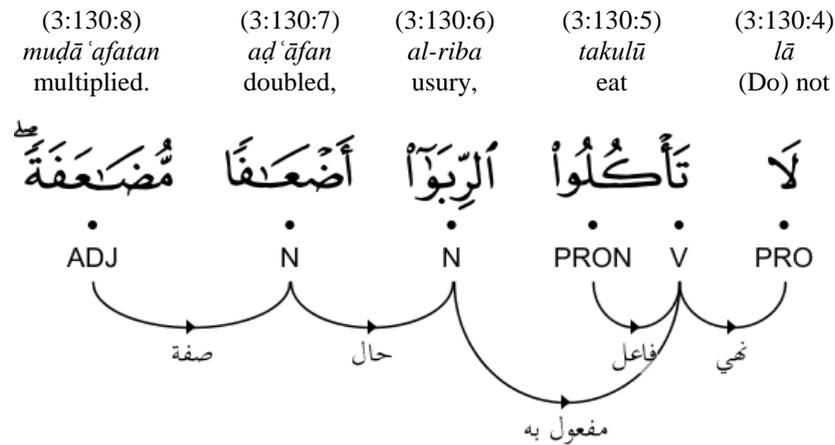

Figure 6.2: Dependency relations in verse (3:130).

    Although the Quranic Treebank uses a hybrid representation, when single words occupy positions, sentences are annotated using pure dependencies. Figure 6.2 shows an example of this with edges pointing towards heads. Reading from right-to-left, the prohibitive particle (حرف نهي) governs the verb in the jussive mood. The verb governs its suffixed pronoun as a subject (فاعل) and the first noun as an object (مفعول به) placing it into the accusative. The second noun depends on the first as a circumstantial accusative (حال). The last dependency relates the second noun to a dependent adjective (صفة), also in the accusative case.

The main focus of syntax in Arabic grammar is explaining inflection using binary dependencies. This is evident in the etymology of the term *i'rāb*. This originally meant Arabic's system of declension, but was later used to describe syntactic theory as a whole (Samsareva, 1998). Since *i'rāb* attempts to account for all reasons of inflection, a rich set of binary relations are utilized by grammarians, with each pair of related elements uniquely named for each relation type. For example, in an adjectival relation, the element being described is *mawṣūf* (موصوف) and the adjective is *ṣifa* (صفة). Similarly, in apposition structures, the head is *mubdal minhu* (مبدل منه) and the dependent is termed *badal* (بدل).

## 6.2.3 Phrase Structure

Arabic grammatical theory does not only utilize dependency relations. Phrase-structure is used to analyze syntactic constructions such as embedded sentences in direct speech. An example from the Quran is the verb *qāla* (قال) 'to say', shown in Figure 6.3 (read from right-to-left). In his analysis for this verse, Salih describes the embedded sentence 'We are the helpers of Allah' as occupying the position of an accusative object (الجملة في محل نصب مفعول به «مقول القول»).[12]

| VERBAL SENTENCE | | |
|---|---|---|
| *'We are the helpers of Allah'* | *the disciples* | *said* |
| Object | Subject | Verb |
| *naḥnu anṣāru allahi*<br>نَحْنُ أَنصَارُ ٱللَّهِ | *al-ḥawāriyūna*<br>ٱلْحَوَارِيُّونَ | *qāla*<br>قَالَ |
| Embedded sentence | Nominative | Indicative |

Figure 6.3: An embedded sentence as a direct object in verse (3:52).

---

[12] Salih (2007). Volume 2, page 64.

Another example use of phrase structure is in the analysis of conjunctions. In contrast to most versions of dependency grammar, Arabic uses dependencies between phrases to describe sentences that include coordination. An example can be found in verse (8:40): *ni'ma al-mawlā wani'ma al-naṣīru* (نِعْمَ الْمَوْلَىٰ وَنِعْمَ النَّصِيرُ) – 'Excellent is the protector and excellent is the helper'. Salih analyzes this structure syntactically as two sentences directly related through a conjunctive dependency (الجملة الفعلية «نعم النصير» معطوفة بالواو على «نعم المولى» وتعرب إعرابها)[13].

Phrase structure also occurs in the analysis of prepositions. In traditional Arabic grammar, prepositional phrases are known as *jār wa majrūr* (جار ومجرور). In contrast to coordination, which is analyzed as a relation between two phrases, prepositions occur in constructions with a prepositional phrase attached to a word. For example, in (7:85): *dhālikum khayrun lakum* (ذَٰلِكُمْ خَيْرٌ لَّكُمْ) – 'That is better for you'. This sentence is analyzed traditionally as a demonstrative pronoun in the subject position (اسم اشارة في محل رفع مبتدأ) with its predicate in the nominative case (خبر المبتدأ مرفوع بالضمة). In his analysis, Salih describes the prepositional phrase as attached (*muta'alliq* – متعلق) to the nominative predicate (جار ومجرور متعلق بالخبر)[14].

## 6.2.4 Ellipsis (*ḥadhf*) and Reconstruction (*taqdīr*)

Elliptical constructions are considered to be part of Classical Arabic's eloquent style and succinctness (Al-Liheibi, 1999). In traditional grammar, the term *ḥadhf* (حذف) denotes the omission of words from a sentence, and *taqdīr* (تقدير) refers to the process of reconstructing them. To closely align to traditional sources, three types of elliptical structure are annotated in the Quranic Treebank that depend on either morphological, syntactic or semantic context.

The first type of ellipsis is related to the morphological form of verbs. Classical Arabic is a pro-drop language and certain verbs imply a pronoun subject which may be dropped from the sentence. The form of the dropped pronoun depends on the verb's phi-features (Fischer and Rodgers, 2002). Traditional analysis restores these dropped pronouns, known as *ḍamīr mustatir* (ضمير مستتر). This is because the grammar requires certain obligatory positions in a sentence to be filled, while other positions are optional. In verbal sentences the subject position must be filled. An example of this is the verbal phrase *lam yalid* (لَمْ يَلِدْ) in verse (112:3). In traditional analysis, the subject in this phrase is a dropped pronoun in third person masculine singular form (الفاعل ضمير مستتر تقديره هو).

---

[13] Salih (2007). Volume 4, page 202.
[14] Ibid. Volume 4, page 28.

Similar to dropped subject pronouns, syntactic ellipsis arises in order to satisfy other constraints. For example, in certain structures prepositional phrases that follow nouns are attached to a reconstructed adjective. In contrast, semantic ellipsis involves an omitted word that is reconstructed based on the sentence's meaning and its situational context. In all three types of ellipsis, omitted words are restored through *taqdīr* and assigned a syntactic role. Section 6.7.2 provides further examples of ellipsis in the treebank.

## 6.3   The Representation Problem

As demonstrated by the examples of traditional analysis in the previous sections, Arabic grammatical theory makes use of dependency relations between words, as well as between phrases. Ellipsis and reconstruction are also utilized to describe sentence structure. In this thesis, a central research question asks if a hybrid representation can be used to model Classical Arabic syntactic structures. This chapter addresses this research question by showing that a hybrid representation for Arabic closely aligns to traditional grammatical concepts. Before describing the hybrid approach, the limitations of two previous approaches are discussed: the constituency interpretation by Carter (1973) and the dependency interpretation by Owens (1984).[15] Both of these interpretations attempt to relate historical analyses that use traditional concepts to modern syntactic theory.

---

[15] The author would like to thank Jonathan Owens and Michael Carter who kindly reviewed this chapter. Although reconciling their different viewpoints has been a source of inspiration, this thesis presents a new hybrid representation as an alternative to both interpretations.

## 6.3.1 Constituency Representations

Carter (1973) suggests that there is a strong similarly between the work of the early grammarian Sibawayh (760-796) and the modern notion of using immediate constituency analysis to construct phrase-structure trees. Sibawayh was highly influential to later grammatical thought and introduced the traditional concepts of *'amal* (عمل) and *'āmil* (عامل) that have been used since the inception of the Arabic linguistic tradition. Carter's interpretation differs from other linguists such as Owens and Versteegh because he does not consider these concepts to refer to governance. His argument is based on noting that Sibawayh uses 'syntactic equivalence' whereby a group of words is replaced by an equivalent element having the same syntactic function. As a specific example, he cites Sibawayh's analysis of the sentence *iḍrib 'ayyu man ra'ayta 'afḍalu* (اَضْرِب أَيُّ مَنْ رَأَيْتَ أَفْضَلُ) – 'Strike whichever of those you consider best'. This sentence has a verb with an embedded relative clause. A possible constituency structure that could parallel Sibawayh's analysis is shown in Figure 6.4 below:

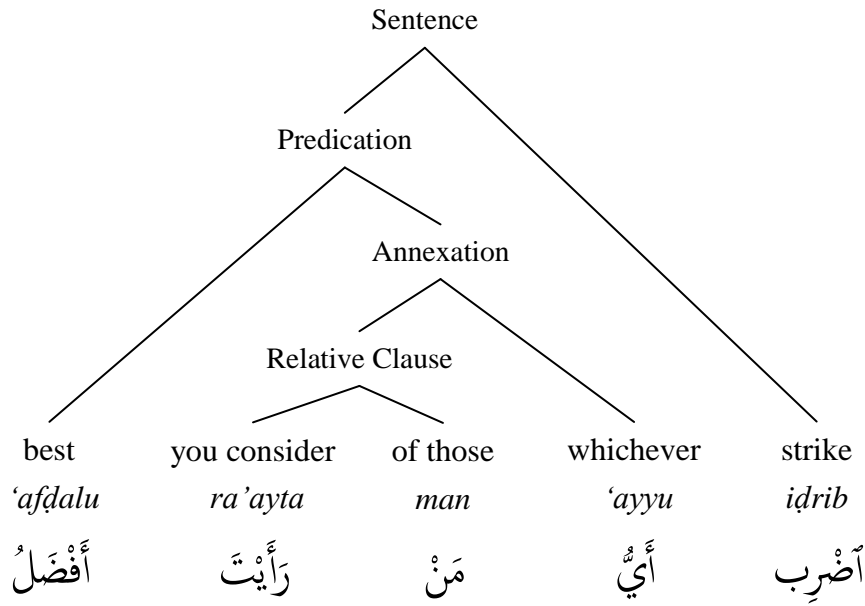

Figure 6.4: Carter's constituency representation based on Sibawayh's analysis.

Carter argues that Sibawayh did not introduce governance into traditional grammatical theory. In his interpretation, the concepts of *'amal* (عمل) and *'āmil* (عامل) instead form a binary constituent. This parallels modern constituency theory in which elements recursively form larger structures through substitution:

> The first systematic work of Arabic grammar, the Book of Sibawayh, presents a type of structuralist analysis unknown to the West until the 20th century. Each function is normally realized as a binary unit containing one active 'operator' (the speaker himself or an element of his utterance) and one passive component operated on (not 'governed') by the active member of the unit. Because every utterance is reduced to binary units, Sibawayh's method is remarkably similar to immediate constituent analysis.

He correctly notes that in common with modern linguistics, Sibawayh uses substitution to determine the syntactic position of words and phrases in a sentence. This technique is often used by traditional grammarians in syntactic analysis. However, a deeper analysis would find that substitution is nearly always used to describe syntax by replacing a larger structure by a single word instead of other intermediate structures (Versteegh, 1997a; Owens, 1998; Salih, 2007).

Carter's view that Sibawayh's grammar is similar to a constituency theory is not closely aligned to traditional thought and has several limitations. For example, in a constituency representation, a more complex construction than the sentence in Figure 6.4 will form a larger binary tree with many more intermediate nodes. It is difficult to see how all intermediate nodes in this representation would correspond to the traditional concept of syntactic position. A wider issue is that Carter views dependency structure as incompatible with Sibawayh's grammar. However, the view in this thesis is that the traditional notion of *'amal* (عمل) corresponds to the governance, or dependency, of two elements in a sentence. These elements may be either words or complete phrases, depending on the type of relation used in the dependency structure.

## 6.3.2 Dependency Representations

In contrast to Carter's constituency representation, the majority consensus in modern literature is that Sibawayh's work and that of later Arabic grammarians is based on dependency. An example of this includes Kruijff (2006; 2002) who puts forward the view that Arabic grammatical theory is based on concepts that form the core of modern dependency grammar. He argues that *'āmil* and *ma'mūl* are equivalent to the notions of heads and dependents in modern grammar, and notes that dependencies in Arabic are used to explain syntactic function.

Versteegh (1997a; 1997b) also considers Arabic grammar to be dependency based. He concludes that grammarians formulated two of the principles used in modern theory to define well-formed pure dependency structures – the existence of exactly one root element in a sentence and the constraint that all elements except the root must have exactly one head:

> The status of declension is thus directly connected with the important principle of *'amal*, governance. The relationship between governor (*'āmil*) and declension (*i'rāb*) is formulated by the Arabic grammarians in terms that suggest a dependency between two constituents. Just like Western dependency-type grammars the Arabic grammarians explicitly specify that within each syntactic structure all elements, except one, depend on another element, but never directly on more than one. One of the strictest rules in Arabic syntactic theory is precisely that there can never be more than one governor (*'āmil*) for a governed element, although one governor may govern more than one element at the same time.

As with Carter's constituency representation, Owens (1984) also draws on the work of Sibawayh, although in contrast he argues for an alternative dependency-based representation. He cites the example sentence *lan yaḍriba al-rajulu ghulāma zaydin* (لَنْ يَضْرِبَ الرَّجُلُ غُلَامَ زَيدٍ) – 'The man won't hit Zayd's son'.

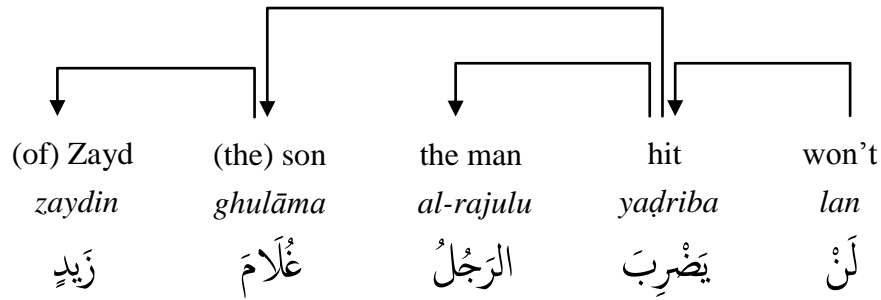

Figure 6.5: Owens' dependency representation based on traditional grammar.

Owen interprets this sentence as a dependency structure (Figure 6.5 above) and argues that the dependencies between words are primarily a consequence of the traditional concept of governance. The edges in his diagram point from heads towards the words that depend on them. Reading from right-to-left, the negative particle *lan* (لَنْ) governs the verb *yaḍriba* (يَضْرِبَ), placing it into the subjunctive mood. The verb *yaḍriba* governs the noun *al-rajulu* (الرَجُلُ) as its subject, placing it into the nominative case, and governs the noun *ghulāma* (غُلَامَ) as its object placing it into the accusative. Finally, the last noun *zaydin* (زَيْدٍ) is in the genitive due to a possessive relation with the previous noun *ghulāma* (غُلَامَ).

Although this example is a pure dependency structure, the main limitation of the analysis by Owens (1984) is that only a few sample sentences are considered. However, he comes close to suggesting a hybrid representation for more complex sentences by observing that dependencies occur between syntactic positions:

> Three of the key principles of Arabic grammatical theory are structure, class and dependency. Items occur in classes at positions of structure and are bound together syntactically in terms of dependency relations. The Arabic notation of dependency is very similar to the modern Western conception. The examination of one structure does not prove that Arabic and modern dependency grammar are based on the same principles, though it does create a strong *prima facie* case.

This thesis takes the next step of noting that positions can also hold complete phrases, allowing for dependencies between items that need not be words. Owens also describes other differences to modern dependency grammar. For example, in most modern theories, verbs are the root of a sentence (Tesnière, 1959; Hays, 1964; Robinson, 1970; Hudson, 1984). However, the previous example showed that particles can be the root of verbal sentences as they govern verbs. Another difference to modern grammars is that Arabic includes binary relations between words that are not always based on governance per se. For example, modifiers known as *tawābi'* (توابع), which include adjectives and words in apposition, are not generally thought of as participating in *'amal*. However, although these words are not governed, they are still dependent on other head words in the sentence.

In his review of Owens and Carter's interpretations, Itkonen (1991) concludes that it is more accurate to say that traditional Arabic grammar combines both representations. His viewpoint is adopted in this thesis, which argues that Arabic grammar is primarily dependency-based while also incorporating constituency:

> It is perfectly right to say that in addition to its preponderant dependency aspect, Arab syntax also has a constituency aspect. This is evident from the role that substitution plays in it. It is explicitly recognized that [positions] can be filled by units of varying size and category-membership. Thus both the dependency view and the constituency view are present in Arab syntax (though not to an equal extent).

## 6.3.3 Hybrid Representation

This section introduces a new hybrid representation for Arabic by building on Itkonen's insight that its grammar combines both constituency and dependency syntax. Section 6.2 provided several examples of traditional syntactic analysis by Salih (2007). Based on these examples, it is possible to deduce a list of concepts that a syntactic formalism for Classical Arabic should account for. The following are necessary but not sufficient for close alignment to traditional grammar:

(A) Prepositions cannot be governed and have no head words in a sentence.

(B) An embedded sentence is a syntactic element that has an explicit relation to a head word in its enclosing sentence.

(C) Coordinating conjunctions are particles that introduce a direct relation between two elements that are words, phrases or sentences.

(D) If there is no relevant governing element within a sentence, inflection is explained using ellipsis.

Itkonen concludes that concepts (A), (B) and (D) are found in the work of Sibawayh and other early Arabic grammarians, but he stops short of providing a formal hybrid representation. To the best of the author's knowledge, (C) has not previously been noted in modern research describing Arabic grammatical theory.

In the remainder of this section, these four concepts are discussed within a hybrid dependency-constituency representation, using diagrams with dependents pointing to heads. This is the convention used in the Quranic Treebank. These diagrams can be used to visualize the reasons for inflection. For example, nouns are found in the nominative if they are a verb's subject (فاعل), the accusative if they are an object (مفعول به) and the genitive (مجرور) if they are governed by a preposition. Figure 6.6 below visualizes these word-to-word dependencies, shown from right-to-left respectively:

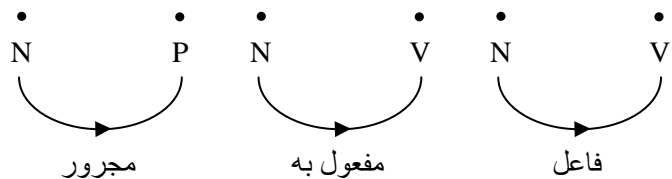

Figure 6.6: Subject, object and prepositional dependencies.

The first concept (A) required for close alignment to traditional grammar is the rule that a preposition cannot be governed. This rule stands in contrast to modern dependency theory, which attempts to assign a head to every word in a sentence except for a root word. However, Nivre (2005) notes that prepositional phrases are challenging to modern dependency theory and are handled differently in its various versions:

> There are also many constructions that have a relatively unclear status. This group includes constructions that involve grammatical function words, such as articles, complementizers and auxiliary verbs, but also structures involving prepositional phrases. For these constructions, there is no general consensus in dependency grammar as to whether they should be analyzed as head-dependent relations at all and, if so, what should be regarded as the head and what should be regarded as the dependent.

In a pure dependency representation, the syntax of the following structure is problematic: (2:71) *ji'ta bilḥaqqi* (جِئْتَ بِالْحَقِّ) – 'You came with the truth'. This consists of a verb, a preposition and a noun. This could be analyzed with the preposition depending on either the verb or the noun. In Arabic grammar, no such dependencies exist. A preposition governs the noun that follows it and heads a prepositional phrase that is attached to another word in the sentence. Visually, this *muta'alliq* (متعلق) dependency is shown in Figure 6.7. In the treebank's hybrid representation, horizontal bars are used to indicate phrase structure:

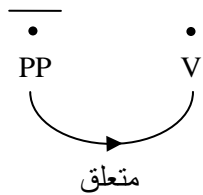

Figure 6.7: Dependency between a prepositional phrase and a verb.

A similar representation can be used for embedded sentences in rule (B). In the example from 6.2.3, an embedded sentence occurred as direct speech: (3:52) *qāla al- ḥawāriyūna naḥnu anṣāru allahi* (قَالَ ٱلۡحَوَارِيُّونَ نَحۡنُ أَنصَارُ ٱللَّهِ) – 'The disciples said, "We are the helpers of Allah."'. In a pure dependency analysis, the head word of the embedded sentence would have a dependency on the verb 'said'. In Arabic grammar, the embedded nominal sentence (NS) in this verse is a complete syntactic unit that is the object (مفعول به) of the verb:

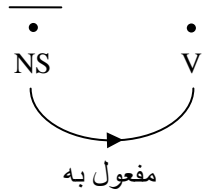

The rule (C) for coordination also differs from pure dependency grammar. For example, Nivre (2005) describes an analysis of coordination as a relation between the first phrase and the conjunction, and a relation between the conjunction and the second phrase. In traditional grammar, a conjunction introduces a single direct dependency between elements, such as in verse (8:40): *ni'ma al-mawlā wani'ma al-naṣīru* (نِعۡمَ ٱلۡمَوۡلَىٰ وَنِعۡمَ ٱلنَّصِيرُ) – 'Excellent is the protector and excellent is the helper'. Figure 6.8 illustrates the traditional analysis for this verse as a conjunctive dependency (معطوف) between two verbal sentences (VS). A conjunctive particle in traditional grammar (POS:CONJ) has no direct relation with other words in a sentence and occupies no syntactic position (لا محل له من الإعراب).

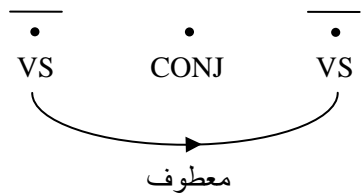

Figure 6.8: Conjunctive dependency between two verbal sentences.

There are several examples of similar analyses in modern linguistics. Although as noted by Nivre (2005), this use of dependencies occurs only in hybrid theories:

> Another way in which theories may depart from a pure dependency analysis is to allow a restricted form of constituency analysis, so that dependencies can hold between strings of words rather than single words. This possibility is exploited, to different degrees, in the frameworks of Hellwig (1986; 2003), Mel'čuk (1988) and Hudson (1990), notably in connection with coordination.

To be closely aligned to traditional analyses, a syntactic representation must also account for ellipsis in rule (D). Traditionally, the dependency relations in elliptical structures have head or dependent elements that are reconstructed words. For example, the start of the Quran opens with verse (1:1) *bis'mi allahi al-raḥmani al-raḥimi* (بِسْمِ ٱللَّهِ ٱلرَّحْمَٰنِ ٱلرَّحِيمِ) – 'In the name of Allah, the most beneficent, the most merciful'. The prepositional phrase *bis'mi* ('in the name of') is said to be attached (متعلق) to a reconstructed verb in traditional grammar, viz. '(I begin) in the name of Allah' (Al-Liheibi, 1999). Figure 6.9 below illustrates this elliptical dependency graphically, using an asterisk (*) to denote the reconstructed verb as an empty category:

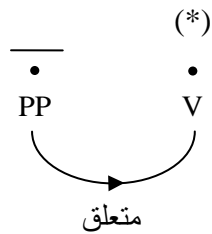

Figure 6.9: Dependency of a prepositional phrase on a reconstructed verb.

## 6.4   Formal Representation

This section provides a formal description of the hybrid representation used in the Quranic Treebank and outlined in the previous section. The formalization is based on directed labelled graphs. These are triples $G = (V, E, L)$ where $V$ is a set of vertices (also known as nodes), $E$ is a set of edges connecting vertices and $L$ is a set of edge labels. In dependency grammar for languages such as English, vertices are words, edges are dependencies and edge labels denote syntactic function. In the Quranic Treebank's hybrid dependency graphs, nodes do not only represent words. Instead, four types of node are used:

1. Morphological segments: terminal nodes resulting from segmentation.
2. Empty categories: terminal nodes used to annotate reconstructed words.
3. Phrases: non-terminal nodes with an associated phrase tag.
4. Referenced words: words referenced from other graphs in the treebank.

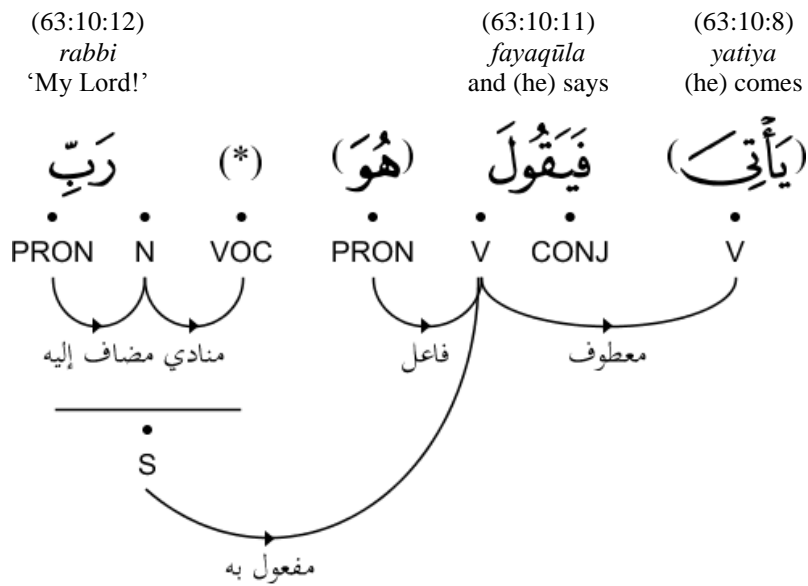

Figure 6.10: Hybrid dependency-constituency graph for verse (63:10).

The two types of terminal nodes are morphological segments and reconstructed words. As described in Chapter 5, morphological segments are the basic syntactic unit in Arabic. In Figure 6.10 (page 120), a dropped pronoun is shown in brackets as the subject of the verb (63:10:11).[16] In the treebank, empty categories are indicated by an asterisk (*), such as the implied omitted vocative at (63:10:12), and phrases are indicated by horizontal bars. In contrast to pure constituency analysis, in the hybrid representation phrases do not have other phrases as explicit immediate constituents. Instead, a phrase is a continuous span of terminal nodes, so that phrases are only implicitly nested. The fourth type of node is referenced words. In the treebank, a word in one verse may have a syntactic relation to a word in another verse. Similarly, long verses are split into multiple dependency graphs. Reference nodes are used to relate words across graphs. Visually, these are shown in brackets, such as in (63:10:8) in Figure 6.10.

Formally, hybrid graphs use the morphological representation described in section 5.3. Let $(s_1, ..., s_n)$ be an input sentence that has been morphologically segmented, and let $R$ denote the set of dependency relations. A hybrid dependency graph is defined as a triple $G = (V, E, L)$ where $E \subseteq V \times V$ are the graph's edges and $L : E \rightarrow R$ are the edge labels. The vertices $V$ are morphological segments, phrases, elliptical nodes or referenced words:

$$V = \{s_1, ..., s_n\} \cup P \cup H \cup W$$

Here $P \subseteq \mathbb{P}$, where $p_{ij} = (s_i, s_j)$ denotes the phrase that spans the segments from $s_i$ to $s_j$ inclusively, and $\mathbb{P}$ is the set of all such possible phrases. Similarly $H \subseteq \mathbb{H}$ and $W \subseteq \mathbb{W}$ where $\mathbb{H}$ and $\mathbb{W}$ are the set of all possible elliptical and referenced words respectively. In the representation, each phrase node $p_{ij}$ has a phrase tag and each edge is labelled with a dependency relation. The dependency relations and phrase tags are defined in the following sections.

---

[16] Salih (2007). Volume 12, page 29.

## 6.5   Dependency Relations

The remainder of this chapter describes the treebank's annotation scheme. The dependency tagset was developed using a similar methodology to the POS tagset described in Chapter 5. Traditional analyses from two reference works were compared: Salih (2007) and Darwish (1996), with Salih as the primary reference. The extracted dependencies are listed in Table 6.1 (overleaf). This tagset consists of 45 relations, with six tags used for nominal dependencies, eight tags for verbal dependencies, six tags for phrasal dependencies, four for adverbial dependencies and the remaining 21 tags for particle-related dependencies.

### 6.5.1 Nominal Dependencies

Figure 6.11 illustrates dependencies used to annotate a possessive construction (مضاف إليه) and apposition (بدل), as well as a dependency for predicate-subject structure (خبر). Other nominal dependencies in the tagset include the adjective (صفة) and the compound dependency (مركب) used for multiword numeric expressions, as in (74:30) *tis'ata 'ashara* (تِسْعَةَ عَشَرَ), which literally means, 'nine (and) ten' for nineteen. Another nominal dependency is specification (تمييز) used for degree or extent, as in (69:32) 'its length is seventy cubits' (ذَرْعُهَا سَبْعُونَ ذِرَاعًا).

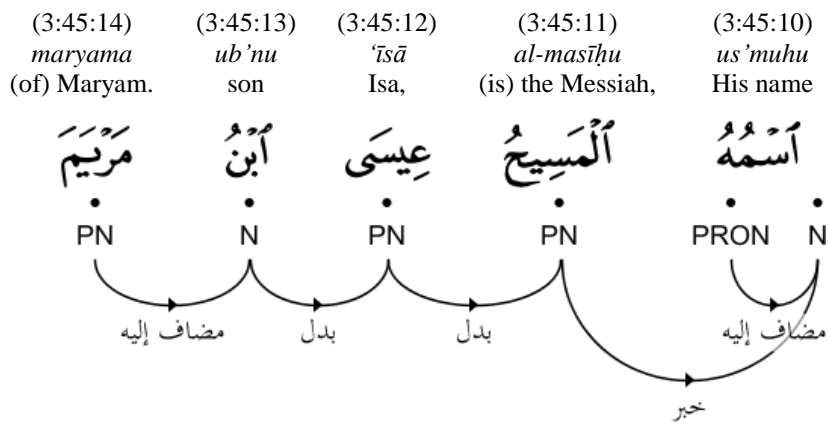

Figure 6.11: Nominal dependencies in verse (3:45).

| Category | Tag | Description | Arabic Term |
|---|---|---|---|
| Nominal Dependencies | adj | Adjective | صفة |
| | poss | Possessive construction | مضاف إليه |
| | pred | Predicate of a subject | مبتدأ وخبر |
| | app | Apposition | بدل |
| | spec | Specification | تمييز |
| | cpnd | Compound | مركب |
| Verbal Dependencies | subj | Subject of a verb | فاعل |
| | pass | Passive verb subject representative | نائب فاعل |
| | obj | Object of a verb | مفعول به |
| | subjx | Subject of a special verb or particle | اسم كان |
| | predx | Predicate of a special verb or particle | خبر كان |
| | impv | Imperative | أمر |
| | imrs | Imperative result | جواب أمر |
| | pro | Prohibition | نهي |
| Phrases and Clauses | gen | Prepositional phrase construction | جار ومجرور |
| | link | PP or adverbial attachment | متعلق |
| | conj | Coordinating conjunction | معطوف |
| | sub | Subordinate clause | صلة |
| | cond | Condition (protasis) | شرط |
| | rslt | Result (apodosis) | جواب شرط |
| Adverbial Dependencies | circ | Circumstantial accusative | حال |
| | cog | Cognate accusative | مفعول مطلق |
| | prp | Accusative of purpose | المفعول لأجله |
| | com | Comitative object | المفعول معه |
| Particle Dependencies | emph | Emphasis | توكيد |
| | intg | Interrogation | استفهام |
| | neg | Negation | نفي |
| | fut | Future clause | استقبال |
| | voc | Vocative | منادى |
| | exp | Exceptive | مستثنى |
| | res | Restriction | حصر |
| | avr | Aversion | ردع |
| | cert | Certainty | تحقيق |
| | ret | Retraction | اضراب |
| | prev | Preventive | كاف |
| | ans | Answer | جواب |
| | inc | Inceptive | ابتداء |
| | sur | Surprise | فجاءة |
| | sup | Supplemental | زائد |
| | exh | Exhortation | تحضيض |
| | exl | Explanation | تفصيل |
| | eq | Equalization | تسوية |
| | caus | Cause | سببية |
| | amd | Amendment | استدراك |
| | int | Interpretation | تفسير |

Table 6.1: Dependency relations for Classical Arabic based on Salih (2007) and Darwish (1996).

## 6.5.2 Verbal Dependencies

Dependencies involving verbs include the subject (فاعل), object (مفعول به) and subject representative (نائب فاعل) for passive verbs. Certain verbs (كان واخواتها) form dependencies other than subject and object (اسم وخبر كان). Another dependency is the prohibitive construction (نهي) in which a prohibitive particle governs a verb placing it into the jussive mood, shown in (5:87:5) in Figure 6.12 below.

## 6.5.3 Phrasal Dependencies

Figure 6.12 also illustrates two phrasal dependencies in the tagset. In the graph, (5:87:10) is a prefixed preposition and a pronoun. These are a prepositional phrase (جار ومجرور) attached (متعلق) to a verb. This graph also has a verbal sentence (VS) as a subordinate clause (صلة) introduced by a relative pronoun (اسم موصول). Other phrasal dependencies in the tagset include coordination (عطف) and conditional sentences relating a protasis clause (شرط) to an apodosis clause (جواب شرط).

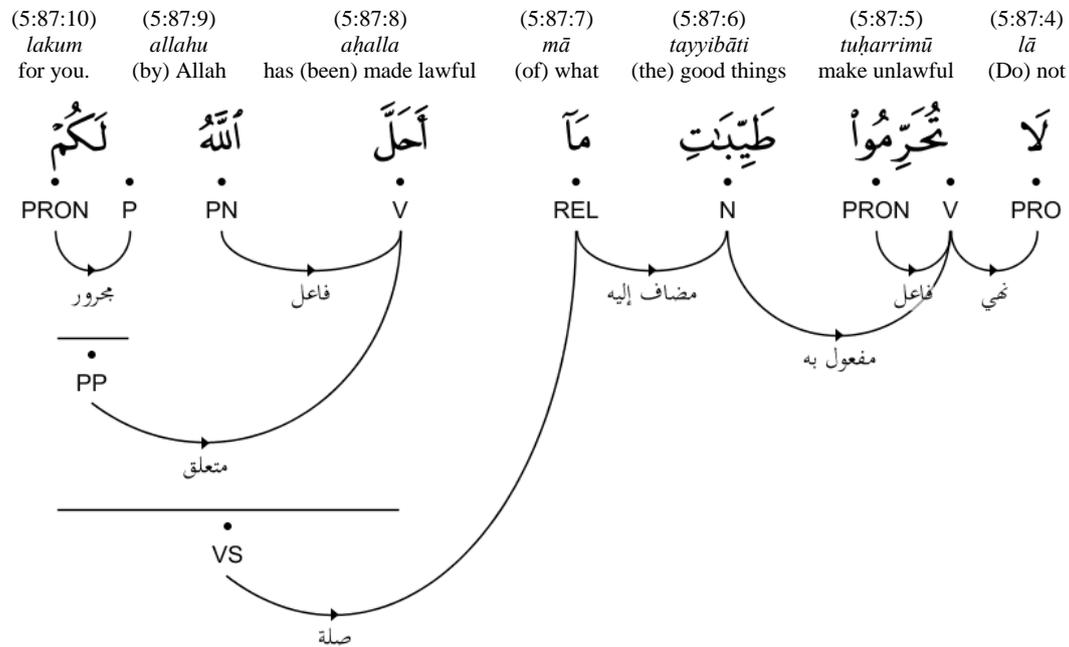

Figure 6.12: Verbal and phrasal dependencies in verse (5:87).

## 6.5.4 Particle Dependencies

Similar to the part-of-speech tagset for particles described in section 5.7, the tagset for particle dependencies is also fine-grained. These relations generally correspond to the equivalent POS tags. For example, the reconstructed vocative in Figure 6.10 (page 120) governs the noun placing it into the accusative case through a vocative dependency (منادي). Other particle dependencies are described in examples of traditional analyses in section 6.7.

## 6.5.5 Adverbial Dependencies

Figure 6.13 illustrates two of the adverbial relations in the tagset that place nouns into the accusative case. The noun at (3:13:16) is a circumstantial accusative (حال), a syntactic role that describes the circumstances of an event or concept. In contrast, (3:13:17) is a cognate accusative (مفعول مطلق). These add emphasis by using a verbal noun derived from the main verb that governs it. In most uses of the cognate accusative, both the accusative and the verb will resonate phonetically as they share the same triliteral root.

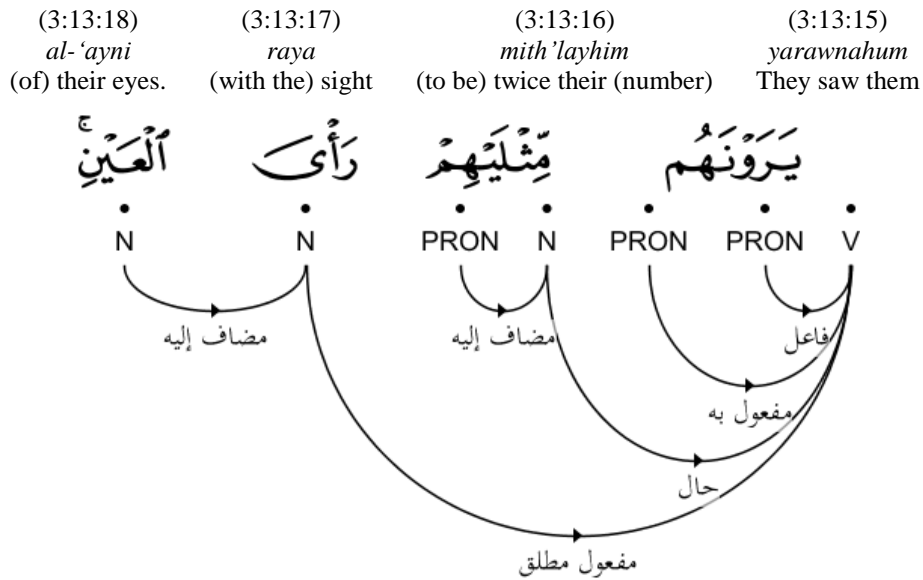

Figure 6.13: Accusative adverbial dependencies in verse (3:13).

## 6.6   Phrase Structure Tags

Phrase structure is used when embedded phrases occupy syntactic positions (محل), although the number of phrase types is restricted. Table 6.2 lists the six phrase tags used in the Quranic Treebank. The PP tag is used for prepositional phrases, and the S tag is used for general sentences when other tags do not apply. The more specific NS, VS, CS and SC tags are described in the following sections.

| Tag | Description | Arabic Term |
|-----|-------------|-------------|
| S | Sentence | جملة |
| NS | Nominal sentence | جملة اسمية |
| VS | Verbal sentence | جملة فعلية |
| CS | Conditional sentence | جملة شرطية |
| PP | Prepositional phrase | جار ومجرور |
| SC | Subordinate clause | تأويل مصدر |

Table 6.2: Phrase-structure tags for Classical Arabic.

## 6.6.1 Nominal and Verbal Sentences

In Arabic grammatical theory, the main distinction between nominal and verbal sentences is that the former starts with a verb and the latter with a noun. However, these criteria are known to inadequately represent more complex cases (Owens 1998; Gully, 1995). For example, the first word of a verbal sentence may be a particle as in (2:78): 'They do not know the book' (لَا يَعْلَمُونَ ٱلْكِتَٰبَ). Similarly, nominal sentences also need not start with a noun, such as in (3:86) 'Indeed the Messenger is truthful' (أَنَّ ٱلرَّسُولَ حَقٌّ). In the Quranic Treebank, a more precise functional definition is used: sentences are tagged as nominal sentences (NS) if they contain the syntactic roles of subject and predicate (مبتدأ وخبر), and are tagged as verbal sentences (VS) if they contain a verb (فعل) with a subject role (فاعل). These tags are based on Classical Arabic's sentence classification rules.

## 6.6.2 Conditional and Subordinate Clauses

Figure 6.14 below shows an example dependency graph with a CS tag used to annotate an embedded conditional sentence (جملة شرطية). These are headed by conditional particles (POS:COND) or an adverbs of time (POS:T). Similarly, embedded subordinate clauses are tagged as SC (تأويل مصدر), such as the object clause headed by 'that' (an – أَن) in verse (2:75) 'Do you hope that they will believe you?' (أَفَتَطْمَعُونَ أَن يُؤْمِنُوا لَكُمْ).

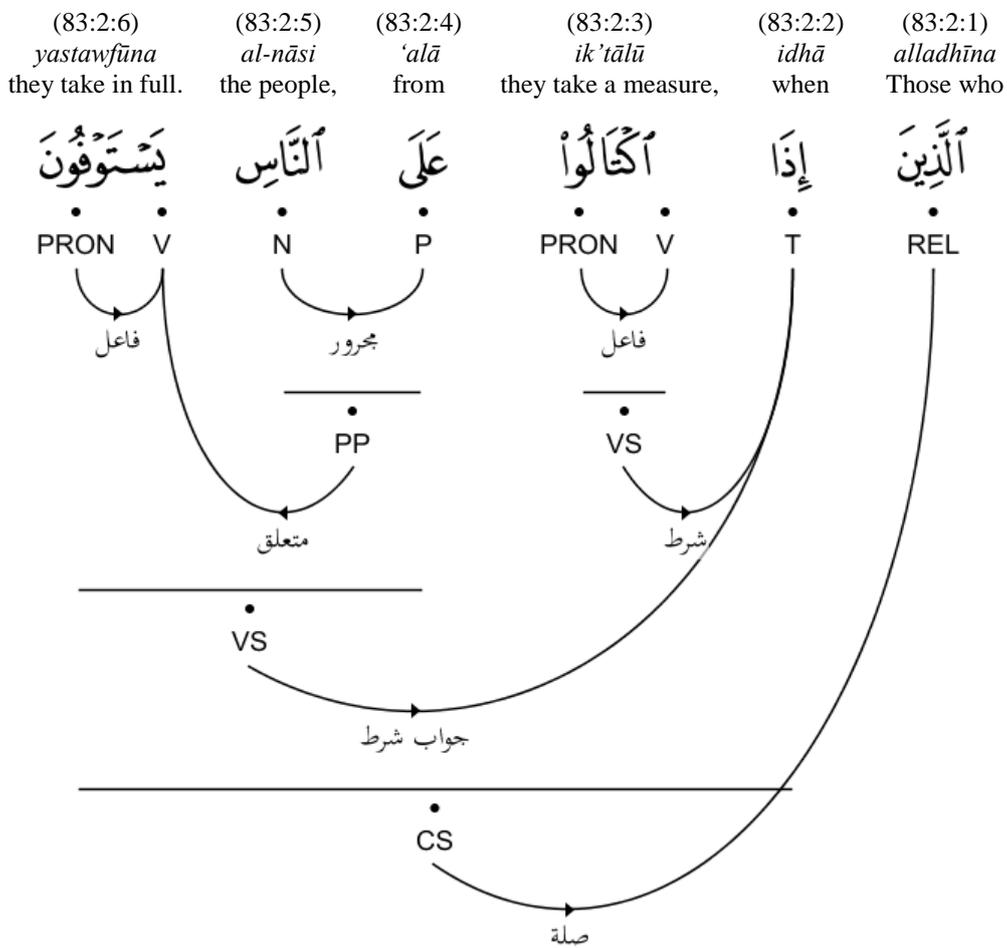

Figure 6.14: Embedded conditional clause in verse (83:2).

## 6.7   Syntactic Structures

The syntactic representation described in this chapter serves two purposes in this thesis, described in parts III and IV respectively. Firstly, Salih (2007) is used as a primary reference work to develop the Quranic Treebank, resulting in a collection of Classical Arabic sentences annotated as hybrid dependency graphs. Secondly, the treebank is used to induce a statistical model for Classical Arabic parsing. This section compares the representation to Salih's traditional analysis for several syntactic structures that highlight the challenges to parsing compared to simpler pure dependency or constituency representations. These include hybrid structures with non-projective dependencies (6.7.1), ellipsis (6.7.2) and disconnected nodes (6.7.3). Examples of syntactic ambiguity in traditional Arabic grammar are also described, including the different syntactic roles for the accusative (6.7.4) and prepositional phrase attachment (6.7.5).

### 6.7.1 Non-Projective Dependencies

In dependency grammar, non-projective edges occur in structures where the dependency relation that connects a pair of words crosses other edges in the graph. Non-projectivity can be formally defined. Let $G = (V, E, L)$ be a pure dependency structure with vertices $(v_1, ..., v_n)$. The graph is non-projective if a pair of edges $(v_a, v_b)$ and $(v_c, v_d)$ exist with ordered vertices such that $a < c < b$ and $d > b$. For the hybrid representation, a graph is non-projective if one of its pure dependency substructures is non-projective, but also if edges that connect phrases and words cross.

Figure 6.14 (overleaf) illustrates two types of non-projectivity based on Salih's analysis of verse (2:127). The non-projectivity is a consequence of four related dependencies. The first dependency in this analysis is the verb (2:127:2) governing the following proper noun as a subject, placing it into the nominative case (فاعل مرفوع بالضمة). Secondly, the verb governs (2:127:4) as an object, placing it into the accusative (مفعول به منصوب بالفتحة). Similarly, the preposition (2:127:5) governs the noun at (2:127:6) placing it into the genitive. These two words form a prepositional phrase attached to the verb at (2:127:2) (جار ومجرور متعلق بيرفع). In the fourth dependency, the proper noun 'Ishmael' at (2:127:7) has a conjunctive dependency on the previous proper noun in the subject position at (2:127:3). Although in a discontinuous position, it is nominative due to agreement with the subject (معطوف على «ابراهيم» وهو مرفوع مثله بالضمة وهو ايضاً ممنوع من الصرف). Because of its unusual position, the coordinating dependency crosses both the prepositional phrase edge and the pure dependency edge between the verb and its object.

Although discontinuity is found in most languages, Nivre (2009) estimates that for some languages, 25% of sentences are non-projective. The example above shows that discontinuity also occurs in Classical Arabic due to governance, as elements may become separated because of flexible word order. This is different compared to non-projectivity in other dependency treebanks due to the inclusion of phrase-structure. However, compared to the traditional analysis written in prose, non-projectivity is easier to identify computationally in the hybrid representation as it is a formal property of directed graphs.

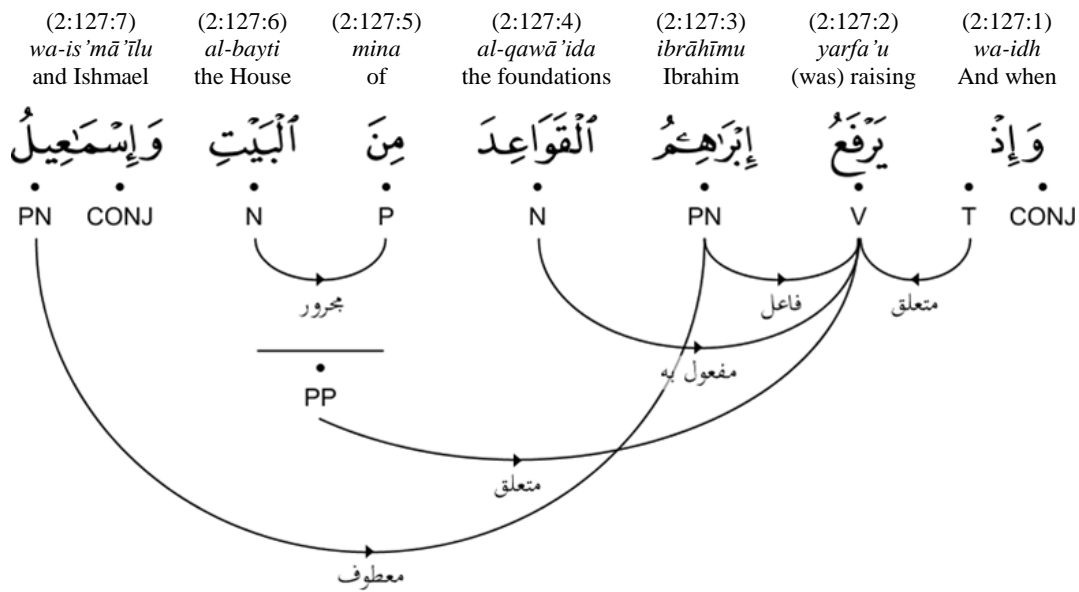

Figure 6.15: Non-projective dependencies in verse (2:127).

## 6.7.2 Ellipsis

In contrast to previous dependency-based treebanks for Arabic (Habash and Roth, 2009c; Hajič et al., 2004) the Quranic treebank annotates ellipsis. This is inspired by Arabic grammar, which often reconstructs hypothesized omitted words to describe sentence structure. As outlined in section 6.2.4, the three types of ellipsis in the treebank depend on either morphological, syntactic or semantic context.

In the first type of ellipsis, dropped subject pronouns are reconstructed based on verb morphology. Because Arabic is a pro-drop language, these frequently occur in the treebank. Figure 6.16 below shows an example dropped pronoun that has been annotated for two reasons. Firstly, in traditional Arabic grammar the subject position of a sentence must be filled. In Salih's analysis for this verse, the verb's subject is a reconstructed pronoun (الفاعل ضمير مستتر فيه جوازاً تقديره هو). Secondly, the pronoun explains why the noun at (3:199:14) is inflected for the accusative case. This word has the role of a circumstantial accusative (حال) with a dependency on the dropped pronoun (حال من فاعل «يؤمن» منصوب بالياء لأنه جمع مذكر سالم).[17]

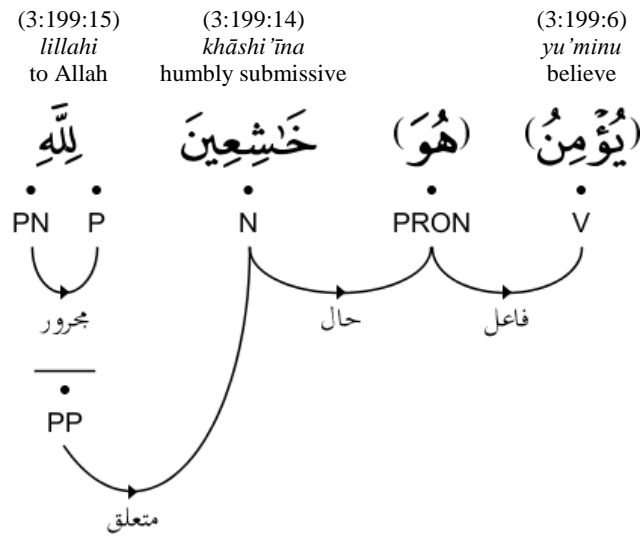

Figure 6.16: Dropped subject pronoun in verse (3:199).

In the second type of ellipsis, reconstructed words satisfy syntactic constraints. For example, in nominal sentences the predicate position must be filled. This occurs in Figure 6.17 shown below. The token at (7:186:4) is segmented into two particles. The first is a result particle *fā'* (ف) that marks the start of an apodosis clause (الفاء رابطة لجواب الشرط). The second segment is a negative particle *lā* that acts syntactically as the particle *inna* (إنّ) (لا نافية للجنس تعمل عمل), governing the following noun as its accusative subject (اسمها مبني على الفتح في محل نصب). Because the nominal sentence does not have a predicate, the prepositional phrase is attached to a reconstructed noun at this position (له جار ومجرور متعلق بالخبر المحذوف). In his grammatical analysis for this verse, Salih states that the form of the reconstructed noun should be analogous to 'there is' (كائن) shown in brackets in the word-by-word translation at (7:184:4) (والتقدير لا هادي كائن له).[18]

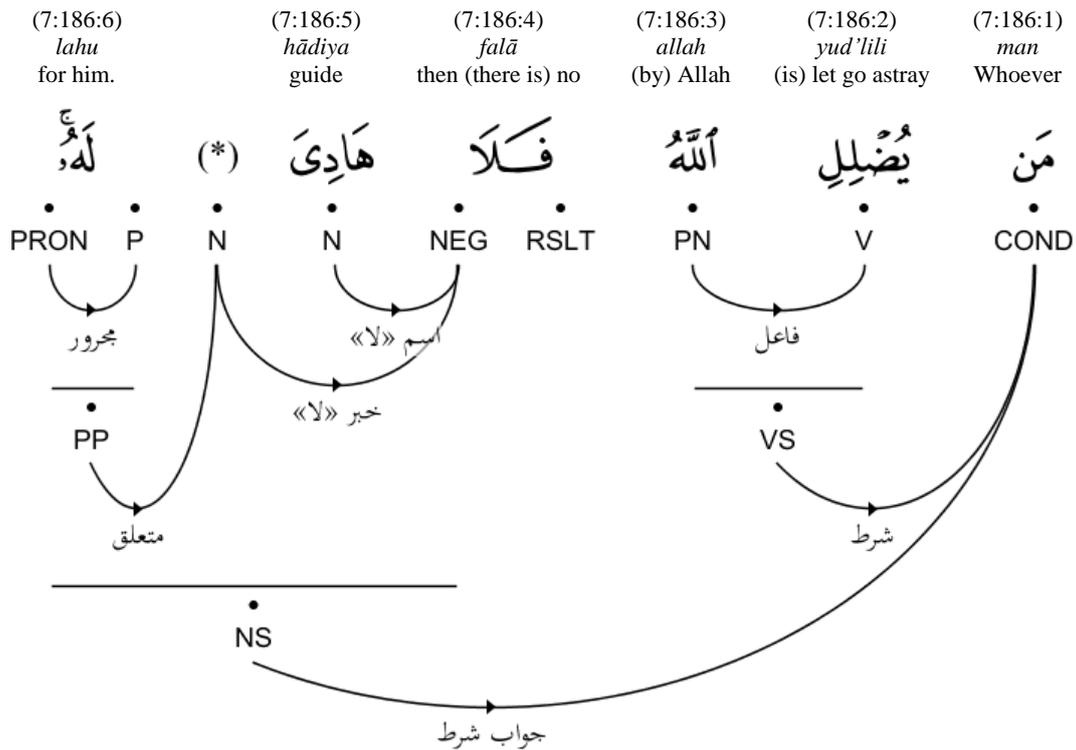

Figure 6.17: Syntactic ellipsis in verse (7:186).

---

[18] Salih (2007). Volume 4, page 140.

The third type of ellipsis involves words that are reconstructed due to semantic context to explain inflection. For example, in Figure 6.18, the first three nouns in the verse are in the nominative case. Using on the context of preceding verses, an elliptical pronoun is annotated based on implied meaning, viz. '(They are) deaf, dumb and blind'. The reconstructed pronoun is also the head of the pronoun at (2:18:4) in a conjunctive dependency. Salih's complete analysis for the verse is given below, demonstrating that the hybrid representation is closely aligned.[19]

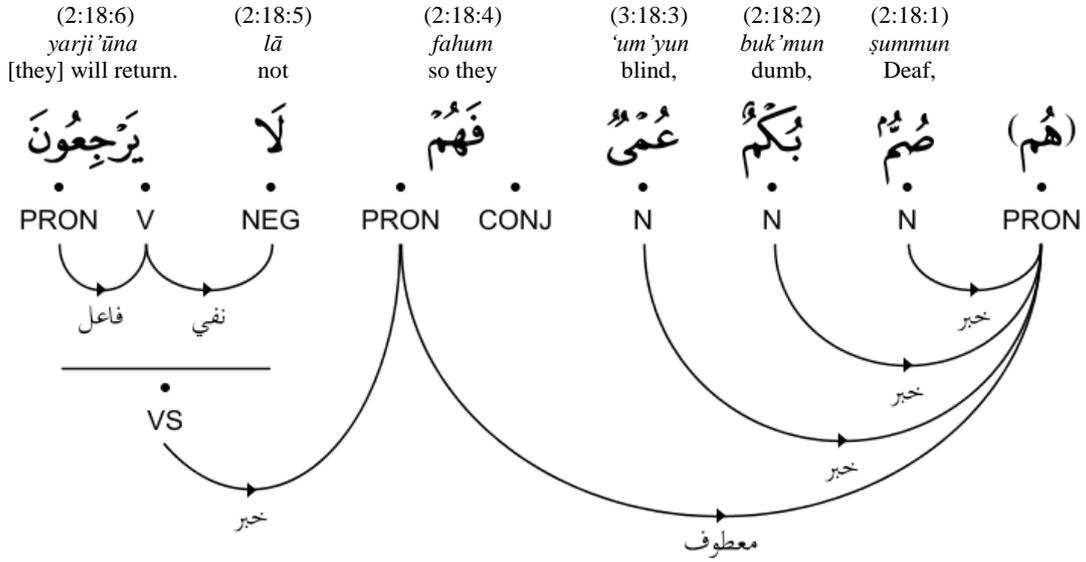

صم : جبر لمبتدأ محذوف تقديره : «هم» . مرفوع بالضمة . بكٌم عميٌ : خبران آخران للمبتدأ مرفوعان بالضمة . الفاء : حرف عطف . هم : ضمير رفع منفصل معطوف على المبتدأ المحذوف «هم» في محل رفع مبتدأ . لا : نافية . يرجعون : فعل مضارع مرفوع بثبوت النون لأنه من الافعال الخمسة . والواو : ضمير متصل في محل رفع فاعل . والجملة الفعلية «لا يرجعون» في محل خبر المبتدأ «هم» .

Figure 6.18: Reconstructed pronoun with Salih's analysis for verse (2:18).

---

### 6.7.3 Coordination and Connectivity

As described in section 6.2.3, the Quranic Treebank annotates coordination differently to previous Arabic dependency and constituency treebanks. In traditional grammar, a coordinating conjunction is neither the head nor the dependent of other words in a sentence. Instead, the two elements on either side of the conjunction are linked through a direct dependency, and are said to be *ma'ṭūf* (معطوف), or connected to one another. This dependency is used to link pairs of words or pairs of phrases, which are found in the same type of syntactic position.

Figure 6.19 below shows the dependency graph for verse (8:40), based on traditional analysis. Salih analyzes the coordination structure in this verse as two sentences directly related through a dependency introduced by a conjunctive particle (الجملة الفعلية «نعم النصير» معطوفة بالواو على «نعم المولى» وتعرب إعرابها). The dependency graph shows the conjunctive particle *wāw* (و) (tagged as POS:CONJ) as disconnected from the rest of the graph because it has no direct syntactic role.[20]

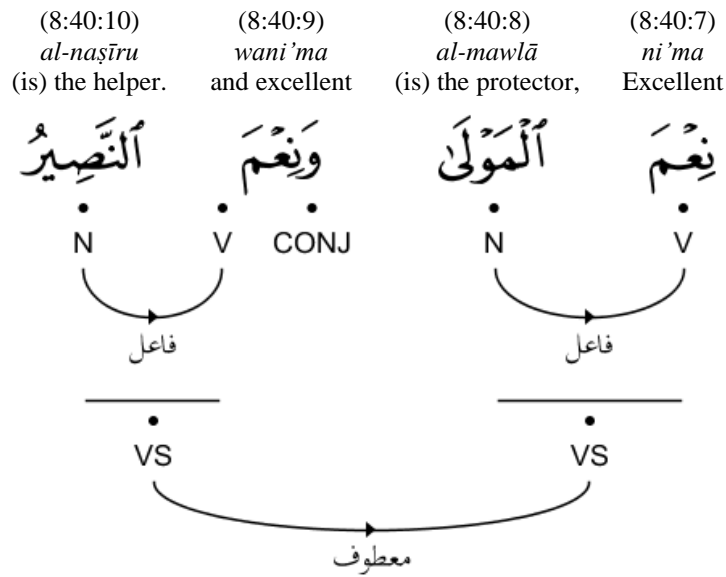

Figure 6.19: Coordinating conjunction as a disconnected node in verse (8:40).

[20] Salih (2007). Volume 4, page 202.

In the Prague and Columbia Arabic dependency treebanks, graphs are fully connected and conjunctions are heads or dependents of other words (Hajič et al., 2004; Habash and Roth, 2009c). Connected dependency graphs that are directed and acyclic are formally dependency trees. In contrast, the disconnected hybrid graphs in the Quranic Treebank will require special processing in the statistical parsing work described in Chapter 9, as previous algorithms have assumed fully connected structures for pure-dependency parsing.

In addition to coordinating conjunctions, other particles in the treebank can also cause graphs to become disconnected. For example, in conditional sentences the result particle is not connected to the rest of the graph, such as the prefixed particle *fā'* at (7:186:4) in Figure 6.17 (page 131). Traditionally, these particles play no role in governance and do not form dependencies (لا محل له من الإعراب). Although the dependency graphs in the Quranic treebank could be made fully connected by adding additional edges, this is intentionally not done so that the syntactic representation remains closely aligned to Arabic grammatical theory.

## 6.7.4 The Accusative Case

Arabic grammar aims to explain all reasons for inflection. However, syntactic role labelling for nominals in the accusative is an example of a parsing task that can be ambiguous. Traditionally, the syntactic roles of the accusative are known as the *manṣūbāt* (منصوبات). The Quranic Treebank uses a fine-grained set of syntactic roles consisting of 45 dependency tags. In 16 of these roles, nominals can occur in the accusative case, listed in Table 6.3 (overleaf). The first role is the most frequent use of the accusative – a nominal used as a verb's object (مفعول به). The next tag is the circumstantial accusative (حال), which has a more semantic usage. This role describes the circumstance or condition of a concept or action. Circumstantial accusatives are also suggested by their morphology. They are generally participles derived from verbs and unlike adjectives which are subject to agreement rules, they are always indefinite. When describing a noun, the noun will always be in the definite state (Rafai, 1998).

| Accusative Role | Dependency Tag | Arabic Term |
|---|---|---|
| Object of a verb | obj | مفعول به |
| Circumstantial accusative | circ | حال |
| Emphasis | emph | توكيد |
| Purpose | prp | المفعول لأجله |
| Specification | spec | تمييز |
| Cognate accusative | cog | مفعول مطلق |
| Time or location adverbial attachment | link | متعلق |
| Vocative | voc | منادى |
| Exceptive | exp | مستثنى |
| Comitative object | com | المفعول معه |
| Subject of the particle *inna* | subjx | اسم ان |
| Predicate of the verb *kāna* | predx | خبر كان |
| Compound | cpnd | مركب |
| Adjective of another accusative word | adj | صفة |
| Apposition to another accusative word | app | بدل |
| Conjunction to another accusative word | conj | معطوف |

Table 6.3: Accusative syntactic roles in Classical Arabic.

Figure 6.20 (overleaf) shows the same word 'a messenger' (*rasūlan* – رَسُولًا) tagged differently as a circumstantial accusative and a direct object in two verses. The upper dependency graph represents the traditional analysis for verse (4:79). Salih analyzes the word *rasūlan* in this verse as a circumstantial accusative as it describes a condition (رسولاً حال مؤكد لعاملها في اللفظ والمعنى منصوبة بالفتحة).[21] This usage is reflected in the word-by-word translation above the graph, viz. 'And we have sent you (as) a messenger'. This contrasts with the use of the same word in (73:15) as a direct object (رسولاً مفعول به منصوب وعلامة نصب الفتحة).[22] Both these usages differ from the word at (73:15:5), which is also in the accusative case but is syntactically in the role of an adjective (شاهداً صفة لرسولاً منصوبة بالفتحة).

---

[21] Salih (2007). Volume 2, page 332.
[22] Ibid. Volume 12, page 225.

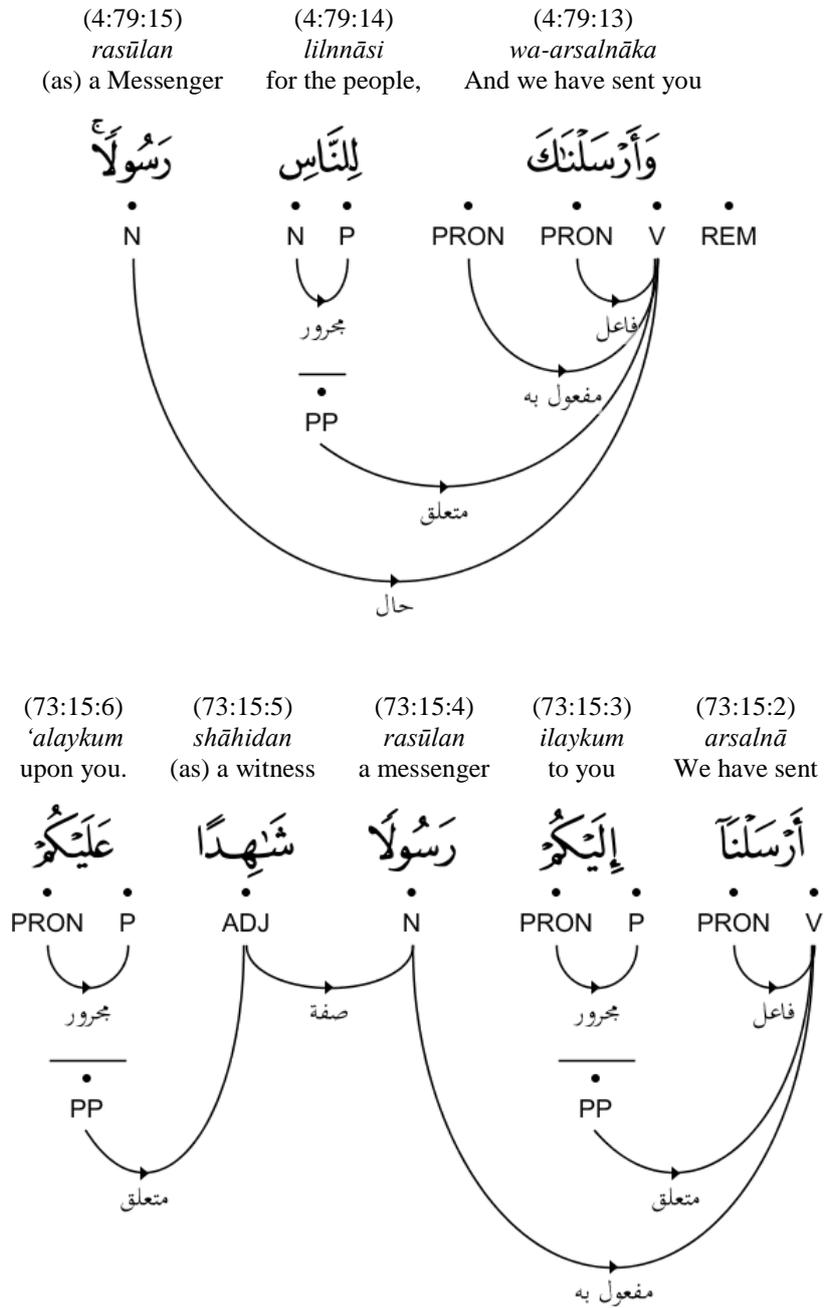

Figure 6.20: The word *rasūlan* ('a messenger') as a circumstantial accusative in verse (4:79) and as a direct object in verse (73:15).

## 6.7.5 Prepositional Phrase Attachment

As a final example, this section describes a hybrid approach for prepositional phrase attachment. This is challenging from a parsing perspective as it involves the interaction of several components in the grammar, including phrase-structure, dependencies, ellipsis and ambiguity resolution. In Classical Arabic, prepositional phrases (جار ومجرور) are generally attached (متعلّق) to verbs, nouns or adjectives. This section focuses on the elliptical form of attachment, in which a prepositional phrase depends on a reconstructed word. In this construction, a prepositional phrase does not directly occupy a position in a sentence but is instead attached to a hypothesized word (محذوف) which fills a position.

In traditional analysis, elliptical PP-attachment occurs because prepositional phrases cannot fill positions that require either words or complete sentences. For example, a sentence consisting of a noun and a prepositional phrase, such as in verse (1:2) 'All praise (be) to Allah' (الحمد لله), is traditionally analyzed as an elliptical construction, with the preposition attached to a reconstructed predicate (متعلّق بخبر محذوف). In elliptical attachment, prepositional phrases are known as *shibh jumla* (شبه جملة), literally a 'quasi-sentence'.[23] The most frequently occurring reconstructed empty categories used with PP-attachment are:

1. A predicate: (1:2) الحمد [مختصّ] لله
2. An adjective: (37:5) بكأس [حالة كونها] من معين
3. A circumstantial accusative: (76:2) إنا خلقنا الإنسان [حال كونه] من نطفة أمشاج
4. A subordinate clause: (2:21) والذين [هم كائنون] من قبله

For statistical parsing, distinguishing these cases requires ambiguity resolution as a prepositional phrase may attach to one of several words in a sentence, or attach to several types of reconstructed words in elliptical constructions.

---

[23] The term *shibh jumla* is also used for phrases headed by locative or temporal adverbs. As with prepositional phrases, these are also attached (متعلّق) and are subject to similar ambiguities.

Figure 6.21 below shows a simple case of non-elliptical PP-attachment. In this verse, the nominal sentence has both its predicate and subject positions occupied, with the prepositional phrase attached to the predicate (جار ومجرور متعلق بالخبر).[24] Two examples of elliptical attachment are given in Figure 6.22 (overleaf). The dependency graph for verse (4:141) in the upper part of the diagram shows both non-elliptical attachment as well as elliptical attachment to a reconstructed circumstantial accusative (جار ومجرور متعلق بحال لأنه صفة مقدمة عليه).[25] The analysis of verse (4:98) is more complex as it depends on a preposition being partitive (بياني). Salih provides two analyses, suggesting attachment to a reconstructed circumstantial accusative (متعلق بحال محذوفة لأن «من» حرف جر بياني) or to an adjective (أو متعلق بصفة لأن «المستضعفين» غير معرفة فيها «أل» لانها اسم جنس).[26] The second analysis is used in the Quranic Treebank after cross-referencing with Darwish (1996).

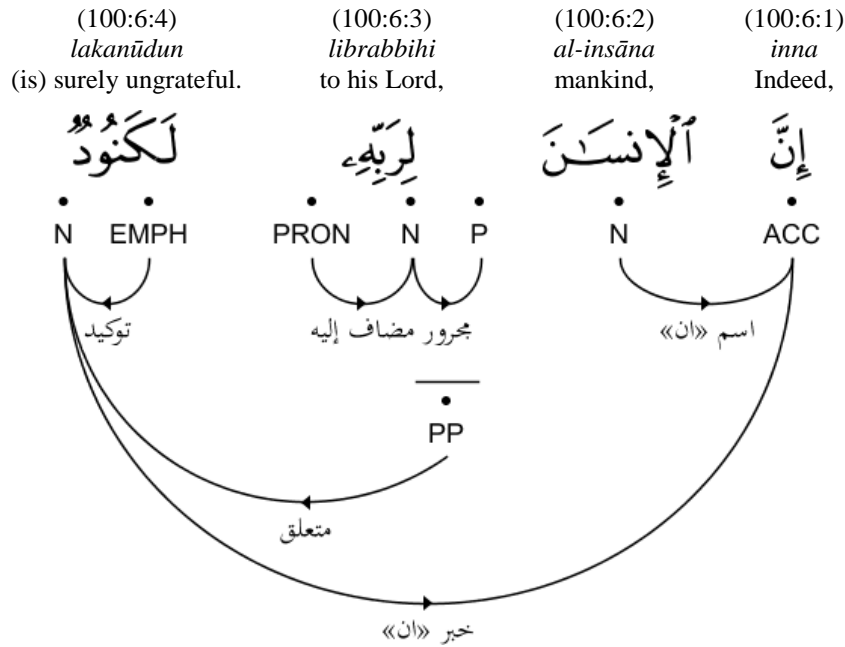

Figure 6.21: Prepositional phrase attachment in verse (100:6:1).

---

Figure 6.22: Elliptical PP-attachment to a reconstructed circumstantial accusative in verse (4:141) and to a reconstructed adjective in verse (4:98).

## 6.8   Conclusion

Part II of this thesis consisted of three chapters that together provided a formal representation for Classical Arabic's orthography, morphology and syntax. This chapter described the syntactic representation used in the Quranic Treebank. Although no previous work exists that formalizes Classical Arabic syntax using graph-theoretic concepts, two previous interpretations of Arabic grammatical theory were compared: the constituency interpretation by Carter (1973) and the dependency interpretation by Owens (1984). Both these were found to be of limited scope. In contrast to previous approaches to Arabic annotation, which has involved adapting Arabic grammatical theory to fit other theories of syntax, the Quranic Treebank adopts a different approach. A new syntactic formalism was constructed based on a hybrid dependency-constituency representation. This was shown to be closely aligned to traditional sources and able to represent a wide variety of linguistic constructions using fine-grained dependencies. It was also interestingly shown that some of these dependencies, such as the circumstantial accusative, are closer to semantic roles than syntactic ones.

In this chapter, the hybrid representation was given a formal definition using directed labelled graphs, and the tagset for dependency relations and phrase nodes were described and illustrated by examples from the treebank. Traditional grammatical analysis was compared to a formal approach for several syntactic structures that present challenges to statistical parsing. This included hybrid dependency-constituency structures, non-projective dependencies, ellipsis, and disconnected nodes in coordination and conditional sentences. Part IV of the thesis will describe how these constructions are handled in statistical parsing work. The representation of Classical Arabic's morphology and syntax also provides a formal basis for annotating the Quran using gold-standard traditional sources to develop the Quranic Arabic Corpus. This annotation methodology is described in the next part of the thesis.

Part III:

Developing the Quranic Arabic Corpus

Alone we can do so little; together we can do so much.

*– Helen Keller*

---

# 7   Annotation Methodology

## 7.1   Introduction

The previous part of the thesis described the representation and annotation scheme used in the Quranic Arabic Corpus. Part III of the thesis consists of two chapters that describe the development of the corpus using this scheme. This chapter focuses on annotation methodology. Chapter 8 describes the custom web-based software architecture used to store and access annotations online.

As described in section 2.5 of the literature review, developing a fine-grained annotated corpus using paid linguistic experts can be prohibitively expensive. Recent work has suggested that crowdsourcing may be more cost effective, by aggregating the results of smaller paid tasks. Examples include concept annotation by Nowak and Rüger (2010) and linguistic tagging using Amazon Mechanical Turk by Snow et al. (2008). However, using motivated volunteers for annotation can be more effective than paid crowdsourcing. For example, Chamberlain et al. (2009) cast their annotation task as an interactive game and successfully develop a one million word anaphoric corpus using unpaid volunteers.

In contrast, choosing the Quran as a dataset for Classical Arabic annotation allows access to a large number of potential volunteers willing to participate in the annotation effort, motivated by their interest in the Quran as a central religious text. Due to the importance of the Quran to the Islamic faith, there is a strong interest to understand the text in its original Classical Arabic form. Morphological and syntactic annotation can aid the understanding process, and a proportion of those who make use of annotations may become annotators.

The two main challenges in this process are attracting participants and ensuring data quality. Data quality is addressed by using supervised collaboration. To apply this methodology to the Quran, sentences are first annotated automatically and then improved by volunteers who compare against traditional works that contain gold-standard analyses. A small group of volunteers who are promoted to expert status supervise and review the work of others to ensure high-quality annotation. However, attracting participants online requires a user-friendly website with additional relevant content. If only a small fraction of visitors become annotators and a smaller fraction of those become supervisors, attracting a large number of visitors is essential. The Quranic Arabic Corpus website focuses on freely available linguistic data, providing part-of-speech tagging and morphological annotation for the complete Quran, and syntactic annotation for 50% of the text. Supplementary linguistic information designed to attract users includes parallel translations of the Quran into English, verse-aligned audio recitations, a searchable Quranic dictionary, a concordance and grammatical reference works. The annotation task is subtly incorporated into the website by encouraging visitors to suggest corrections to the existing linguistic tagging as they make use of it.

As of 2013, the corpus website (http://corpus.quran.com) is frequently cited online as Quranic reference work, and is reported by Google Analytics to have been used by over two million visitors in the past 12 months. It has grown rapidly because it is the first educational resource for Classical Arabic and Quranic research backed by a linguistic treebank. However, supporting collaborative annotators and a large number of general users requires a scalable platform that can efficiently organize linguistic data. The custom software architecture designed for this purpose is described in the next chapter.

The remainder of this chapter focuses on annotation methodology and is organized as follows. Section 7.2 provides an overview of the methodology. Sections 7.3 and 7.4 describe the initial stages of the annotation process, including automatic annotation and offline manual correction respectively. Section 7.5 describes online supervised collaborative annotation and includes a comparison to a small-scale paid crowdsourcing experiment. Finally, section 7.6 concludes.

## 7.2   Methodology Overview

### 7.2.1 Annotation Stages

This chapter describes a new methodology for linguistic annotation of a corpus: online supervised collaboration using a multi-stage approach. The different stages are automatic annotation, offline correction, and online volunteer proofreading. Figure 7.1 below provides an overview of the annotation process.

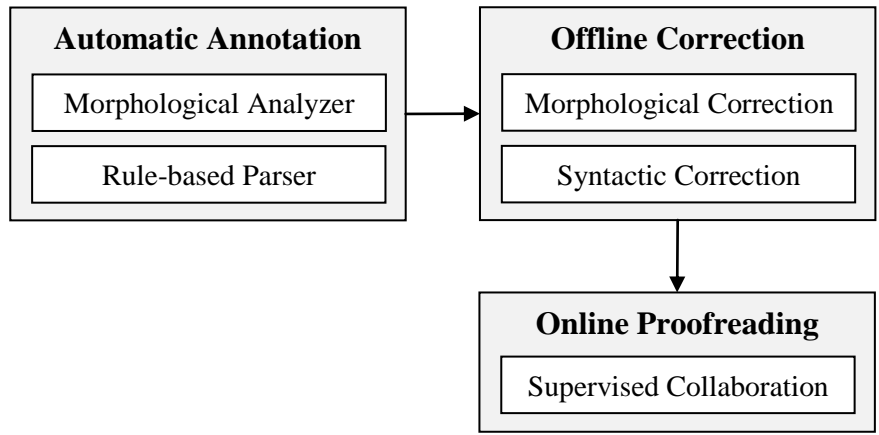

Figure 7.1: Multi-stage annotation process.

The initial stage of automatic annotation uses a rule-based dependency parser. This component is distinct from the statistical parser described in Part IV of the thesis, which was developed separately after the treebank was constructed. The second phase is offline manual correction by experts. In contrast to previous tagged Arabic corpora, in the final stage the corpus is made freely available online for correction by volunteers under expert supervision. To closely align annotation to traditional sources, collaborators are encouraged to compare their analyses to Quranic reference works. For syntactic annotation, Salih (2007) and Darwish (1996) are the primary references.

## 7.2.2 Corpus Size

The Quranic Arabic Corpus is not as large as other tagged Arabic corpora when using word count as a measure of size. The initial release of the corpus annotates the *ḥafṣ* narration of the Quran, consisting of 78K words. In comparison, the first releases of the Prague, Columbia and Penn Arabic Treebanks annotated 113K, 200K and 460K words respectively (Smrž and Hajič, 2006; Habash and Roth, 20009c; Maamouri et al., 2004). However, these treebanks were constructed by paid experts. In contrast, the Quranic corpus is primarily annotated by volunteers.

Another measure of size that may be more applicable to fine-grained annotation is feature-value count. As the corpus provides deep morphological and syntactic annotation, annotators are asked to review a substantial amount of linguistic information per word. The morphological layer in the corpus consists of 128,223 segments. Together with segment type, the feature set in Table 5.4 (page 98) has 42 features, each with multiple possible values. This gives the potential for 5.4 million ($42 \times 128,223$) items of morphological information. In practice, not all features are applicable to every segment. Despite its smaller word count, the corpus annotates 783K feature-values, at an average of 6.1 values per segment.

The syntactic layer covers 37,578 words (~ 49% of the full Quranic text). The total size of the dependency graphs in the treebank is 50,955 terminal nodes formed from morphological segments, including 3,775 empty categories. This node count excludes the determiner Al+ prefix which is not considered to be a terminal segment during syntactic annotation. In addition, the dependency graphs contain a total of 9,847 phrase nodes and 38,642 edges.

## 7.3   Automatic Annotation

The Quranic Arabic Corpus uses orthographic data from the Tanzil Project, described in Chapter 4 (Zarrabi-Zadeh, 2011). In the automatic annotation stage, this text was morphologically and syntactically tagged using new computational components developed for Classical Arabic.

The morphological component is an analyzer derived from the Buckwalter Arabic Morphological Analyzer (BAMA), described in section 2.2.1 (Buckwalter, 2002). BAMA was chosen for two reasons. Firstly, it is freely available and in the public domain. Because the Quranic Arabic Corpus is an open source dataset, any tools used to produce annotation should also be open source (or with less restrictive licenses) to avoid copyright restrictions on the resulting data. Secondly, the analyzer is widely used in the Arabic computational linguistics research community. The Penn Arabic Treebank was initially tagged using BAMA (Maamouri et al., 2004), and the Prague and Columbia Arabic treebanks were tagged using analyzers based on the BAMA lexicon (Smrž and Hajič, 2006; Habash and Roth, 2009c).

However, adapting BAMA is computationally challenging as it is designed for Modern Standard Arabic (MSA). The adapted analyzer used for the Quranic Arabic Corpus is written in Java. It was initially developed by porting version 2.0 of BAMA's source code from the Perl programming language. The analyzer was extended in four ways to make it more suitable for Classical Arabic:

1. Adapting the tagset to align with the tags developed for Classical Arabic.
2. Normalizing text to handle spelling differences.
3. Filtering and ranking results to select a single morphological analysis.
4. Adding additional morphological features such as roots.

The first extension was adapting the part-of-speech tagset. BAMA uses the Penn Arabic Treebank tagset. In contrast, the Quranic Arabic Corpus uses a tagset based on traditional grammar (Table 5.1, page 88). For the majority of words such as verbs, nouns, pronouns and adjectives, the conversion of tags was a one-to-one process. However, the Quranic tagset is more fine-grained. For example, particles are annotated using a set of 27 tags. Quranic Arabic also requires some genre-specific tags such as Quranic initials, used to annotate sequences of disconnected letters. For these fine-grained tags, full automatic conversion was not possible and manual disambiguation was required.

The second extension was text normalization. Running an unmodified analyzer against the Quran produces low accuracy for part-of-speech tagging, because the spelling of the Quran differs from Modern Arabic. Most of the differences involve orthographic variation of the Arabic *hamza* and the *alif khanjarīya* (a diacritic used for the long vowel *ā*). BAMA was extended to account for these differences.

The third extension improves the analysis algorithm using filtering and ranking. BAMA uses its own detailed lexicon of Arabic to identify possible choices for segmentation and tagging for each word. However, the unmodified BAMA algorithm operates on one word at a time to produce multiple candidate analysis. Because the algorithm accepts a single word as input, it does not make use of context. Filtering is used to remove ungrammatical analyses using a small number of hand-written linguistic rules that refer to the context of surrounding words. For example, a genitive noun (مجرور) following a perfect verb (فعل ماض) is very likely to be an incorrect analysis, as nouns are placed into the genitive case either by prepositions (حرف جر) or by following another noun. In addition to incorrect case tagging, another improvement was made to account for BAMA's lexicon, which contains a large number of adjectives incorrectly classified as nouns (Attia, 2008). For certain words, it is often difficult to distinguish between nouns (اسم) and adjectives (صفة) as both occur with similar surface forms. Contextual syntactic rules were used to correct this, as adjectives follow the nouns they describe.

After filtering incorrect results using context, ranking is used to select a single morphological analysis. When used for Modern Arabic, BAMA tags undiacritized text and produces multiple possible morphological analyses for each input word with added diacritics. However, the Quranic text comes with the advantage that it is fully diacritized unlike most other Arabic texts. In the modified analyzer, the different diacritized analyses are ranked in terms of their edit-distance from the Quranic diacritization, with the closer matches ranked higher. The BAMA analysis with the highest rank is then chosen as the unique part-of-speech for that word.

Finally, BAMA was extended to include additional morphological features. For example, it was possible to automatically annotate roots by importing these from the open source Zekr Quran browser (http://zekr.org). This contains an accurate verified root list for the Quran, used to support the software's search feature.

Following automatic morphological annotation, the tagging was manually corrected. Using manually corrected morphological data as input, a rule-based dependency parser was used to produce initial syntactic annotation. The rule-based parser shares the same transition system as the statistical parser described in Part IV. The difference is that the rule-based parser uses a hand-written classifier using traditional Arabic grammar rules, instead of using a statistical model derived from the gold-standard annotations. Due to the similarly between these two transition systems, the rule-based parser is described alongside the statistical parsing work in Chapter 9.

## 7.4   Offline Correction

After applying the automatic annotation algorithm to the corpus, two annotators manually verified the results in turn, with the second annotator reviewing the text after the initial set of corrections made by the first annotator. This process was followed twice, once for morphological and once for syntactic correction. Given the similarities between these two processes, the section focuses on morphology.

A custom Java annotation tool was used for offline morphological correction (Figure 7.2, overleaf). The depth of morphological analysis planned for the corpus exceeded that provided by BAMA. Although the analyzer produced most of the planned features, certain key parts of the morphological analysis could only be produced manually. This included missing verb voice (active or passive), the energetic mood for verbs, the interrogative *alif* prefix, identifying participles, verb forms, and disambiguating *lām* prefixes. Although each of these features had to be added by hand, most do not occur very often, and the analyzer nearly always correctly identified the remaining set of features.

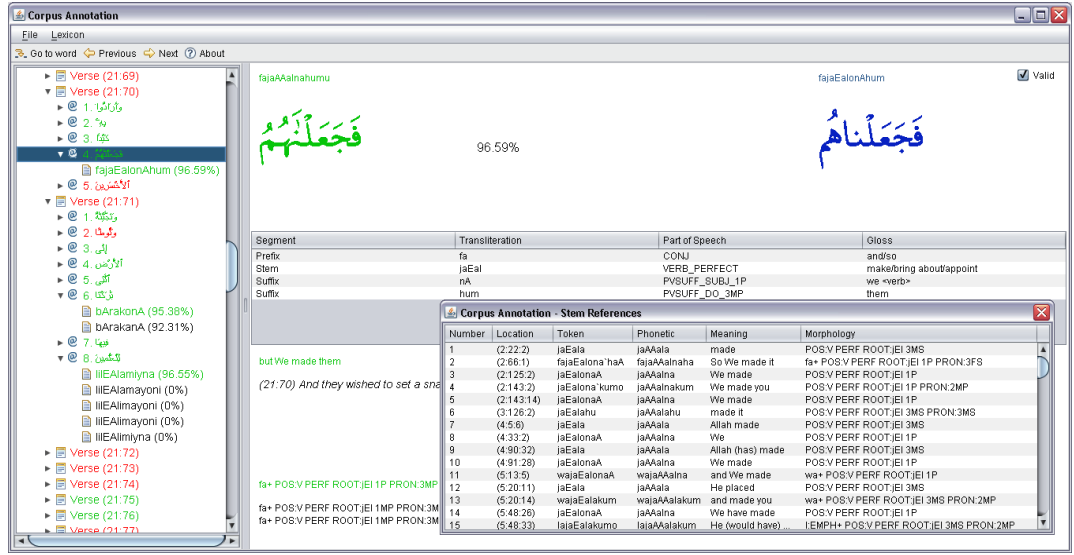

Figure 7.2: Custom Java application used for offline morphological correction.

A useful measure for estimating the accuracy of offline annotation is the number of words that required revision at each stage of correction. The automatic algorithm outlined in the previous section produced an analysis for 67,516 out of 77,430 words (87% unchecked recall). Complete coverage was not possible due to out-of-vocabulary errors in the BAMA lexicon. The rate of out-of-vocabulary errors was lower than expected given the differences between Modern and Classical Arabic. One explanation for this is that the BAMA lexicon contains many Classical Arabic words as traditional dictionaries are one source of its lexical data. Although previous work has shown this to impact the performance of the analyzer for Modern Arabic (Attia, 2008), this was in fact a benefit for annotating the Classical Arabic text of the Quran.

Following automatic analysis, the morphological annotation was reviewed in stages by two annotators. A paid native speaker of Arabic reviewed each word in the Quran working full-time over a three-month period. At this stage, corrections were made to 21,550 words (28%). This included the 9,914 words not analyzed by the automatic algorithm (13% of all words), as well as 11,636 corrections to existing analyses (15% of all words). This allows the performance of automatic morphological annotation to be measured as 72% (recall), 83% (precision) and 77% (F-measure). Recall and accuracy are identical in this case since every word received only one analysis (or no analysis). A second annotator (a trained Arabic linguist) then reviewed the morphological annotations again, including the first annotator's corrections, and made changes to 1,014 words (1.3% of all words). Table 7.1 below summarizes each stage of the process. The automatic algorithm correctly analyzed approximately 3/4 of all words. Without using BAMA, it is likely to have taken a single annotator far more than three months to manually tag all words in the corpus.

| Annotation stage | Words revised | % of Quran |
|---|---|---|
| Automatic algorithm | 67,516 | 87.19 |
| Annotator #1 | 21,550 | 27.83 |
| Annotator #2 | 1014 | 1.3 |

Table 7.1: Number of modifications during morphological annotation.

| Component | Precision | Recall | F-Measure |
|---|---|---|---|
| Morphological analyzer | 72 | 83 | 77 |
| Rule-based parser | 91 | 68 | 78 |

Table 7.2: Estimated accuracy scores for automatic annotation.

The process for syntactic annotation followed a similar methodology to the morphological annotation process. The main difference between these two tasks was that the morphological task required correcting more in-depth information due to higher automatic recall (Table 7.2). In contrast, for automatic syntactic annotation the lower recall but high precision implied that more time was spent manually adding missing edges in dependency graphs.

## 7.5   Supervised Collaborative Annotation

### 7.5.1 Role-Based Collaboration

The final stage of annotation is online supervised collaboration through the corpus website. This has similarities to the Wikipedia model, in which articles are improved through incremental edits (Kittur and Kraut, 2010). For the Quranic corpus, a message board is used to gather suggested corrections.[27] There are three different collaborative roles: contributors, editors and supervisors. New users who have recently registered will be general contributors who have read-only access to the annotations but can post suggested corrections online. Editors are project organizers, and have both read and write access to the linguistic database. When a suggestion is a genuine correction, the corpus annotations are updated.

Online annotation progressed initially with multiple volunteer contributors providing suggestions, but with only two editors reviewing these and making edits (phase A). During a later second stage, the supervisor role was introduced by promoting a small number of contributors to this status (phase B). Supervisors retain their read-only access to annotations, but are differentiated by their ability to veto incorrect suggestions made by other contributors. These trusted experts are chosen if they consistently provide high-quality corrections and have suitable academic credentials. Introducing a supervisory role increased the accuracy of suggestions considered for edits in phase B by 22%. This is due to supervisors filtering out incorrect comments from non-experts, allowing editors to focus on considering suggestions that are more likely to be genuine corrections.

Collaborators participate using free text entry as opposed to restricted multiple-choice responses. This more natural form of expression promotes communication between annotators. Messages are organized into threads that discuss correct tagging for individual words. For example, a common case is a thread in which a contributor suggests a correction that is reviewed by a supervisor:

---

[27] http://corpus.quran.com/messageboard.jsp

> 20th April, 2010
>
> FS: Is this not a LOC - accusative location adverb as opposed to a noun?
>
> AR: Yes, it is indeed *zarf makaan mansoob*.
>
> FS: Thank you.

In the following related example, a contributor participates in order to highlight incorrect tagging as well as to clarify their own understanding of Arabic grammar:

> 24th April, 2010
>
> TH: I am a beginner grammar student. I thought this word is 2nd person masculine singular. Please help me understand.
>
> AR: You are right. The verb is indeed 2nd person masculine singular. This needs to be corrected.

As well as confirming corrections and providing useful educational feedback to contributors, supervisors veto incorrect suggestions made by non-experts:

> 31st March, 2010
>
> FS: Could we also add in addition to this being a noun that it is *hal*?
>
> RZ: For a noun to be *hal* it must be *mansoob* but here noun is *marfoo'*, so it is not *hal*. Vol 3, page 45. Thanks.
>
> FS: Sure. We can leave it as *khabar* of *inna*.

In the above example, the supervisor vetoes a suggestion for syntactic tagging. As justification, the supervisor provides a reference to Salih's analysis (volume 3 page 45). As shown by these examples, the dual nature of the message board involves common understanding to incrementally improve the accuracy of a shared resource, but is also an open forum for researchers to engage with subject experts.

## 7.5.2 Resolving Disagreement

The public threads archived on the Quranic message board are an interesting case study in collaborative annotation. The interactions most often involve mutual understanding between collaborators and supervisors, but also contain cases of disagreement. Consensus is usually achieved by following a resolution procedure. The most common method for resolution is to refer to the annotation guidelines. If these require enhancing, annotators are challenged to each cite references to justify their analyses. If both annotators provide justifications for differing analyses, the analysis from primary reference texts is adopted as definitive. After a difficult linguistic construction is encountered for the first time and agreement is reached, the annotation guidelines are improved.

An interesting case of disagreement that highlights this process is the gender of angels in the Quran. The historical context for this is a pre-Islamic belief that angels were the daughters of God (Al-Mubarakpuri, 2003), whereas the Quran states that God has no offspring. It is also generally accepted in Islam that angels are not feminine, as indicated by verse (43:19) which refers to pre-Islamic beliefs:

وَجَعَلُوا الْمَلَٰئِكَةَ الَّذِينَ هُمْ عِبَٰدُ الرَّحْمَٰنِ إِنَٰثًا ۚ أَشَهِدُوا خَلْقَهُمْ ۚ سَتُكْتَبُ شَهَٰدَتُهُمْ وَيُسْـَٔلُونَ ﴿١٩﴾

**(43:19)** *And they made the angels, the servants of the Most Merciful, females. Did they witness their creation? Their testimony will be recorded, and they will be questioned.*

Gender in Classical Arabic is an intricate issue, as highlighted by the following example. In traditional exegesis, the noun *mu'aqqibātun* in verse (13:11) refers to angels. In Arabic, gender may refer to semantic, morphemic or grammatical gender. A word can have different values for these three attributes, as gender can differ across meaning, form and syntactic function. The Quranic Arabic Corpus tags grammatical gender. The noun *mu'aqqibātun* (معقبات) has a feminine-sounding morphemic ending, but acts as grammatically masculine. This noun was initially incorrectly tagged as feminine by the morphological analyzer. At the time of the online discussion below, the guidelines did not clarify which type of gender should be tagged. The thread begins with an annotator challenging the incorrect automatic tagging of feminine by comparing to the semantically masculine, but morphemically feminine-sounding 'Caliph' (*khalifa*):

> 17th November, 2009
>
> MN: The word 'angels' does not go with feminine, since the Quran states that only disbelievers describe angels as feminine. Can't *mu'aqqibātun* be considered masculine like *khalifatan*?

A second annotator suggests that grammatical gender should be tagged, but unfortunately provides an incorrect analysis of grammatically feminine:

> KD: The full grammatical analysis for this word is feminine plural, active participle from *'aqqaba*, form II of *'aqiba*. This word is a grammatical feminine. This does not mean that angels are feminine.
>
> MN: How can one accept a grammatical analysis for this word as feminine plural?
>
> KD: Can you please cite a reference for your own grammatical analysis?

A third annotator contributes to the thread using the website's concordance tool, which provides easy access to tagging for previous related words:

> AB: I took a corpus linguistics approach and looked at the concordance lines for the 54 occurrences of *malaekah*. Of these, 32 occurrences used pronouns to refer to the angels in the same verse, and that 21 used masculine and 11 used feminine pronouns. One verse (47:27) used both masculine and feminine pronouns. So, in reality angels are not female (based on 43:19 and

> other verses). But grammatically the majority of the time they are referred to as males and sometimes as females.
>
> KD: It is interesting that both feminine and masculine pronouns are used, purely in the sense of grammatical gender.

For this thread and for related examples, consensus between annotators is reached through discussion. In this particular verse, the word *mu'aqqibātun* although feminine in form, is masculine in meaning as well as in grammatical function. The thread concludes with the next stage of the resolution procedure. The analysis is confirmed by the original collaborator who verifies against a primary reference, in this case a Classical Arabic dictionary (Lane, 1992):

> MN: I got this information from the Lane's Lexicon entry for this word: While feminine in form, grammatically this is masculine. This is a double plural, and so is masculine in the same way.
>
> KD: It looks like your reference from Lane's Lexicon sums this up. This reference does suggest that we change this word to masculine.

Following this discussion, the annotation guidelines were enhanced to specify that grammatical gender is being tagged, as opposed to morphemic or semantic gender.[28] This resolution process and annotation methodology contrasts with recent collaborative efforts that use an aggregation statistic to filter out the noisy judgments of non-experts. For a sensitive corpus such as the Quran, Islam's central religious text, inter-annotator discussion is crucial for accurate results when the number of non-experts outweighs more experienced contributors. Experts proofreading annotations typically cite references and take time to pursue and justify their analyses through discussion with other collaborators.

---

[28] http://corpus.quran.com/documentation/gender.jsp

### 7.5.3 Online Annotation Accuracy

The accuracy of corpus annotations that do not have accompanying reference works to verify against are usually measured via inter-annotator agreement using a metric such as the κ-statistic (Carletta, 1996). For the Quranic corpus it is possible to use alternative methods, as data is verified using gold standard works of Quranic grammar. Indirect evidence for the accuracy of the annotations can be found by contrasting website usage to message board activity. Figure 7.3 shows this activity captured weekly, over a year from June 2009 to May 2010. The inverse trends indicate that although more people continue to make use of the online annotated resource over time, the number of suggested corrections has decreased, since errors are becoming harder to find as accuracy improves.

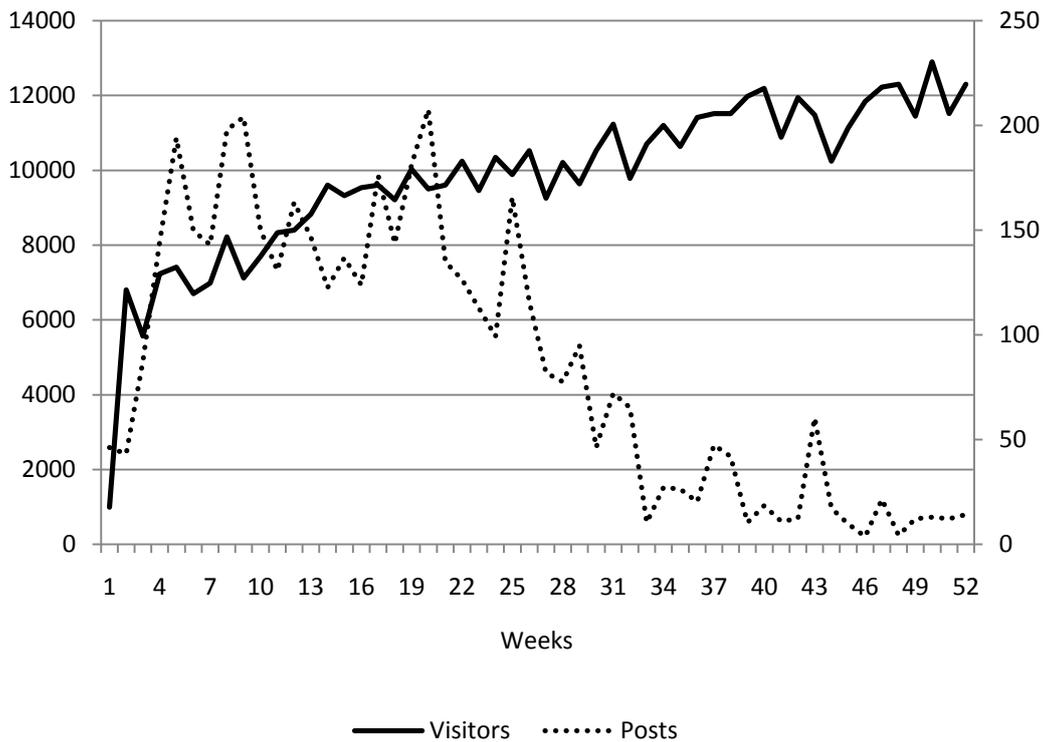

Figure 7.3: Website visitors and message board posts per week over a year.

A more precise measurement of accuracy can be obtained through random sampling. This section focuses on evaluating the accuracy of morphological annotation in the corpus, which is considered to be a stable part of the tagging effort and near completion. As of 2013, The Quranic Treebank provides syntactic dependency graphs for 50% of the Quranic text but is still in progress, while morphological annotation covers 100% of the Quran, and has been proofread online. This section also does not consider the accuracy of ancillary annotation in the corpus (such as phonetic transcription or translation) as their accuracies have no bearing on the core issue of statistical parsing by machine learning.

To measure the accuracy of morphological annotation by random sampling, from the 77,430 words in the Quran, three random non-overlapping samples were collected, with each 1,000 words in size. The words in each sample need not be in sequence or be from the same verses or chapters. The annotations in the corpus for each of these samples were compared to the traditional analyses in reference works of Quranic grammar. Typically, accuracy does not vary significantly across each of these samples, so that they can be averaged to give an estimated accuracy measure for the corpus as a whole. Table 7.3 shows the number of suggestions during the first two 3-monthly periods of online annotation, for the three samples; for the whole Corpus, there were 1,801 suggestions by 3 months, and a further 1,728 suggestions by 6 months. As can be seen, the number of suggestions during these time periods is overall evenly distributed among the samples, which demonstrates that they are representative of the annotation effort.

| Online Time | Suggestions | | |
|---|---|---|---|
| | Sample A | Sample B | Sample C |
| 3 months | 21 | 26 | 23 |
| 6 months | 19 | 24 | 19 |

Table 7.3: Suggestions per random sample.

The accuracy for the morphological annotation of a single word is measured according to strict criteria. A typical word in the Quran will receive multiple tags and features for different items of linguistic information such as segmentation, part-of-speech, gender, person, number, and grammatical case. A word is considered to be accurately annotated only if all of the features have the correct expected values. Table 7.4 summarizes the accuracy of morphological analysis, measured by using the same random samples at 5 different stages of annotation. Each stage of annotation builds on the previous stage by reviewing the existing annotations and making further corrections. Supervisors were introduced after three months of online proofreading by website collaborators. Accuracy is measured at each of these stages, as well as at 6 months and at 12 months into the annotation effort.

| Online Time | Stage | Accuracy |
|---|---|---|
| - | Automatic annotation | 77.2% |
| - | Initial offline correction | 89.6% |
| 3 months | Online proofreading without supervisors | 92.5% |
| 6 months | Online proofreading with supervisors | 96.9% |
| 12 months | - | 98.7% |

Table 7.4: Accuracy of morphological annotation.

The effect of introducing a supervisory role 3 months into the project can be seen from the accuracy measurements in Table 7.4. During the first three months of annotation (without supervisors) accuracy improved by 2.9%. For the next 3 months with supervisors, accuracy improved by a further 4.4%. It is also relevant to consider the quality of message board suggestions. For the first three months of online annotation (without a supervisory role), 1,331 out of 1,801 suggestions resulted in valid corrections to annotations (74%). For the following three months of annotation (with a supervisory role) out of a total of 1,728 suggestions, 401 of these were vetoed by supervisors, and out of the remaining 1,327 suggestions, 1,271 resulted in corrections to the corpus annotations (96%) by editors. Introducing a supervisory role later in the project boosted the quality of suggestions considered by editors by 22%, due to supervisors filtering out inaccurate suggestions made by less experienced contributors. This increase in the quality of suggestions allows editors to focus on considering genuine corrections and comparing only these to grammatical reference works.

## 7.5.4 Unsupervised Crowdsourcing Comparison

In order to compare the methodology of supervised collaboration to unsupervised crowdsourcing, a simple experiment was conducted using Amazon Mechanical Turk (AMT), an online job marketplace where workers are matched with requesters offering tasks. These AMT tasks are known as HITS (Human Intelligence Tasks), and are often presented in a multiple choice format, or make use of restricted text entry. Although recent work has shown high accuracy in using AMT for simple annotation tasks (Su et al., 2007; Snow et al., 2008), it is not clear how well AMT would perform for deep linguistic annotation.

In the AMT experiment, a 500-word part-of-speech tagged section of the Quranic text was put online for correction by Mechanical Turk workers, and was reviewed independently by 6 contributors. To simplify the experiment, only part-of-speech tags were considered instead of the full set of morphological features. This allowed the AMT experiment to run as a simple multiple-choice task. Unlike with the Quranic corpus, AMT workers are paid a small fee for each completed task. These workers are not necessarily Arabic specialists or volunteers interested in the Quran, but can be anyone with the required skills wanting to earn money for participation.

To ensure a baseline level of competency, the experiment required successful completion of an online screening test, which asked 5 challenging multiple-choice questions about Arabic grammar. Only those AMT workers passing the screening test participated in the annotation experiment. The initial data given to AMT was a reduced form of the part-of-speech tagset used to seed the online Quranic Arabic Corpus (stage 2 in Table 7.4, at 89.6% accuracy). This allows for a more accurate comparison between online supervised collaboration and AMT crowdsourcing. The AMT workers were invited to review this tagging and provide corrections. After this review, the final accuracy of the 500-word sample averaged at 91.2% (an increase of 1.6%). This compares with the 92.5% accuracy in Table 7.4 at stage 3, for initial online collaboration in the Quranic corpus without supervisors. This would suggest that involving expert supervisors in the collaborative process, as well as encouraging discussion and communication leads to higher accuracy for a deeply annotated resource such as the Quranic corpus. The current estimated accuracy of morphological annotation in the corpus is measured at 98.7%, using the approach of supervised collaboration.

## 7.6   Conclusion

This thesis asks if a variation of crowdsourcing can be used to accurately annotate Arabic. This chapter addressed this question by providing a description of a multi-stage collaborative effort for Arabic morphological and syntactic annotation. The different stages include automatic rule-based tagging, initial manual verification and supervised collaborative proofreading. The corpus website has approximately 100 unpaid volunteer annotators each suggesting corrections to existing linguistic tagging. To ensure a high-quality resource, 12 expert annotators have been promoted to a supervisory role, allowing them to review or veto suggestions made by other collaborators. This approach was shown to produce superior and needed quality compared to previous crowdsourcing methods that lack supervision. Given the special characteristics of this task, it was decided not to use an existing wiki platform to host the forum used for inter-annotator discussion. Instead the search and feedback mechanisms were developed as part of a custom annotation platform. This platform is described in the next chapter.

# 8   Annotation Platform

## 8.1   Introduction

The central research questions for Classical Arabic in this thesis ask if a hybrid
representation is suitable for statistical parsing and if crowdsourcing is suitable for
annotation. These questions relate to the construction of the Quranic Treebank and
parsing experiments using the annotated syntactic data. A suite of computational
components have been developed to answer these research questions. The custom
linguistic software used for the Quranic Arabic Corpus is implemented using Java,
and consists of 75K lines of programming code, developed over an 18-month
period. These components collectively form a new software system, known as
Linguistic Analysis Multimodal Platform (LAMP). This platform integrates
multimodal data, including deep tagging, interlinear translation, multiple speech
recordings, visualization and collaborative analysis. Annotations are made freely
available online through an accessible cross-referenced web interface.

   This chapter describes the implementation of the annotation platform. Section
8.2 outlines the modular design used for the platform's architecture and provides a
description of the linguistic database and computational components. Section 8.3
describes the website and the associated set of tools used to access annotations.
Section 8.4 provides an overview of supplementary resources made available to
annotators, including grammatical reference material, a morphological search tool
and an ontology of Quranic concepts. Finally, section 8.5 concludes.

## 8.2   Platform Architecture

### 8.2.1 Modular Design

LAMP is a linguistic annotation platform developed using the Java programming language. Java was selected as the implementation language because it is object-oriented and encourages a modular design using distinct components that interlink through the use of interfaces and abstraction. Figure 8.1 (overleaf) shows an architecture diagram that summarizes the interaction of the main components.

LAMP is implemented as a three-tier architecture together with supplementary offline tools. Each tier is organized as a set of related components. The three tiers are: a data access tier for accessing the linguistic database, a service tier consisting of computational linguistic components, and an online presentation tier. The website presents data using servlets and Java Server Pages (JSP), hosted using an Apache Tomcat web server (Brittain and Darwin, 2009). The underlying data is stored in a MySQL database, which includes the treebank, message board threads and supplementary data. The website contains a mix of static HTML pages and dynamic content. In the dynamic pages, computational components in the service tier generate concise summaries and graphical visualizations of annotations from tags stored in the database as users browse the site. This real-time design allows changes to annotated tags to be reflected in the treebank's dependency graphs and displayed on the website instantly without offline rendering. As of 2013, the website has several thousand users accessing dynamic content concurrently during peak hours. To manage this data load, Tomcat and MySQL were chosen to host the platform as they are open source, web-based and highly scalable.

In addition to these online components, LAMP contains components used for offline processing tasks. These include the rule-based morphological analyzer and dependency parser used for initial automatic annotation, as well as a manual annotation tool used for making updates and corrections to the corpus based on volunteer suggestions. The structure of the database and the design of these computational components are described in the following sections.

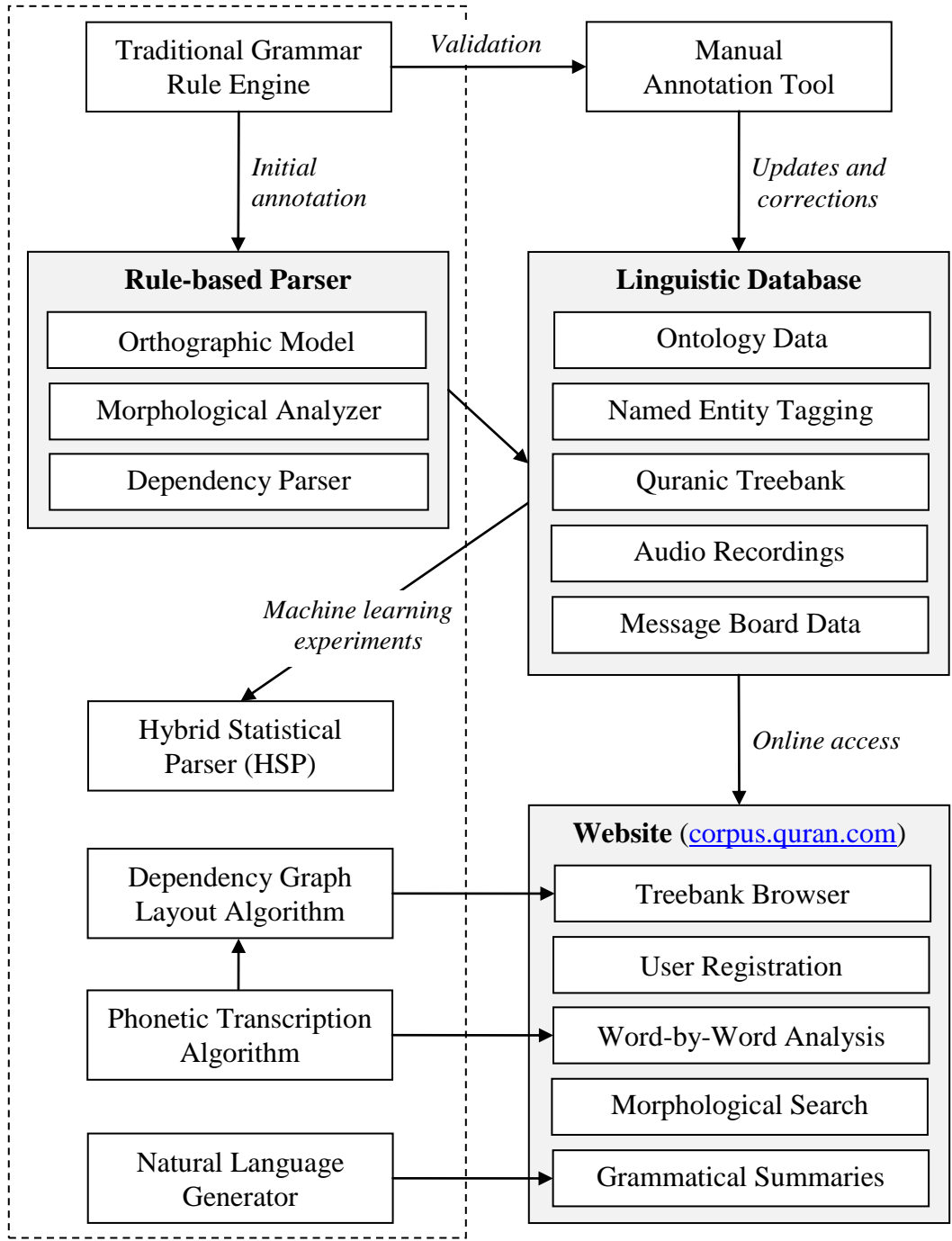

Figure 8.1: LAMP architecture diagram.

## 8.2.2 Linguistic Database

The linguistic database is implemented as a set of related MySQL tables that store morphological, syntactic and supplementary data. The morphological table uses the feature notation described in section 5.3.3, with the analysis for one token stored per row. Table 8.1 below shows an example of this. In the syntactic table, a row represents a node in a dependency graph. A common scheme for encoding dependency treebanks is the CoNLL-X format (Nivre et al., 2007a). Syntactic data is stored in an extension of this format to encode phrases and ellipsis. Table 8.2 shows example rows that correspond to Figure 1.5 (page 8). The extended format adds two new columns: Type indicates the different types of nodes, and Extent defines a phrase by specifying start and end terminal nodes. Head nodes and dependency labels are shown in separate columns.

| Token | Morphology |
|-------|------------|
| (6:76:7) | POS:V PERF LEM:qaAla ROOT:qwl 3MS |
| (6:76:8) | POS:DEM LEM:ha`*aA MS |
| (6:76:9) | POS:N LEM:rab~ ROOT:rbb M NOM PRON:1S |

Table 8.1: Extract from the morphological annotation table.

| Node | Type | Extent | Form | Tag | Head | Dep |
|------|------|--------|------|-----|------|-----|
| 1 | T | _ | qaAla | V | _ | _ |
| 2 | E | _ | Huwa | PRON | 1 | subj |
| 3 | T | _ | ha`*aA | DEM | _ | _ |
| 4 | T | _ | rab~i | N | 3 | pred |
| 5 | T | _ | Y | PRON | 4 | poss |
| 6 | P | 3-5 | _ | NS | 1 | obj |

Table 8.2: Hybrid dependency graph in extended CoNLL-X format.

In addition to the columns shown in Tables 8.1 and 8.2, all tables are indexed by verse number. This allows the website to load all relevant information for a verse in a single data access request. The database is also used to store ancillary information, including English translations of the Quran, named entity-tagging, ontology data, and seven audio recitations. Because the Quran is based on an oral tradition, the recitations reflect different readings of the text, each with subtle differences in prosodic stress.

The tables that are used to support the message board feature of the website store user registration information and discussion threads. The message thread tables are indexed by chapter, verse and token number, as inter-annotator discussion usually focuses on the tagging of individual words in the corpus.

## 8.2.3 Computational Linguistic Components

Figure 8.1 showed the interaction between the platform's website components, the database, and several computational linguistic components. The largest of these computational components in terms of number of lines of programming code is the Traditional Grammar Rule Engine. This is a set of approximately 1,000 linguistic constraints written as Java rules (20K lines of code), which were manually extracted from several grammatical reference works based on the Arabic linguistic tradition (Fischer and Rodgers, 2002; Haywood and Nahmad, 1990; Muhammad, 2007; Rafai, 1998; Wright, 2007).

The linguistic constraints in the rule engine are used offline for three purposes. As described in section 7.3, constraints improve morphological analysis in the initial annotation stage by providing part-of-speech disambiguation using the context of surrounding words. Secondly, rules drive parsing actions in the dependency parser used for initial syntactic annotation. The rule engine is also used to validate the annotation decisions made during manual proofreading. An example of this would be an annotator reviewing a sentence and forgetting to include a dropped pronoun for a verb with no obvious subject. The annotator is alerted to this mistake by a linguistic rule based on traditional Arabic grammar which specifies that all verbs must have a subject (فاعل), with the exception of special verbs known as *kāna wa akhwātuhā* (كان واخواتها) which have different syntactic roles. Another example is a rule which specifies that any words marked as the objects (مفعول به) must be in the accusative case (منصوب), and not in the nominative (مرفوع) or genitive (مجرور). During treebank construction, validation errors are displayed in the annotation tool alongside dependency graphs. This allows annotators to make further amendments before saving their analyses to the database. Annotators can also choose to override the validation rules and force their analyses to be saved. This occurs in special cases that are exceptions to normal sentence structure such as ellipsis. This validation feature helps ensure that annotations remain consistent and of high-quality by reducing the occurrence of obvious mistakes made during manual annotation.

The three main computational components used online in the service tier are a graph layout algorithm, a phonetic transcription algorithm and a natural language generator. Visualization is performed using a custom graph layout algorithm. Because the hybrid dependency-constituency graphs are a new form of syntactic representation, it was not possible to reuse an existing visualizer. Instead a new component was developed based on a two-phase 'measure and arrange' layout algorithm. The visualizer uses a phonetic transcription subcomponent that accepts an Arabic word as input and produces a phonetic transcription in English. These transcriptions are shown in dependency graphs and also in word-by-word morphological analysis web pages.

The final component uses natural language generation (NLG). This simplifies the annotation process by generating concise descriptions of morphological and syntactic tagging in both Arabic and English. Although machine readable, the linguistic tags stored in the database are not easily understood by annotators who are more familiar with standard terminology. The generator reproduces the descriptions from traditional Arabic grammar using a sequence of concatenated templates filled by annotated features. The algorithms for syntactic visualization, phonetic transcription and natural language generation are described in appendices A, B and C respectively (pp. 252 – 258).

## 8.3   The Quranic Arabic Corpus Website

### 8.3.1 User Interface Design

Although a central feature of LAMP is collaborative annotation, the website is presented as an educational study resource to maximize use of the annotated data. They key design principles of the website are usability and ease-of-use. These are essential when online volunteers may not have the motivation or time to follow a non-intuitive annotation process. To encourage volunteer collaborators to assist with annotation, suggesting corrections online is designed to be a subtle and non-intrusive process. Instead of directing users straight to annotation tasks, the website primarily focuses on accessing key information, organized ergonomically. Using statistics provided by Google Analytics, the website's navigation menu has been amended over time so that the most popular sections appear first (left of Figure 8.2, overleaf). This reduces the amount of time users spend searching for relevant information. The menu also lists supplementary resources which are not part of the collaborative effort, but serve to make it a more attractive and useful resource generally, and help to attract and motivate volunteer collaborators.

The use of an accessible website to verify annotations contrasts with more conventional approaches to annotating Arabic corpora. Four recently developed Arabic treebanks (Maamouri et al., 2004; Smrž and Hajič, 2006; Habash and Roth, 2009c; Al-Saif and Markert, 2010) use a small number of paid annotators. Quality is ensured by providing a well-documented set of guidelines, by following a training process, and by having different annotators make multiple passes of the same text. In a collaborative setting, annotation guidelines still apply and are displayed on the corpus website, but training and quality control need to be handled more carefully. When constructing the Quranic Arabic Corpus, it was found that making the annotation process as intuitive as possible led to greater accuracy and consistency, more rapid annotation, and attracted a larger number of expert linguists and Quranic scholars, who are willing to spend more time volunteering contributions.

Figure 8.2: The Quranic Arabic Corpus website.

## 8.3.2 Morphological Annotation

The website provides a drill-down interface (Böhm and Daub, 2008) which is used to 'zoom' into morphological annotations, summarizing linguistic tagging at different levels of detail (Figure 8.3, overleaf). This type of interface is not usually applied to tagged corpora, but is especially useful for a rich, layered dataset such as the Quranic Arabic Corpus. For each verse in the Quran, the original Arabic script (Figure 8.3A) is displayed online alongside seven parallel translations into English. Clicking on the Arabic script displays the website's most used feature, the interlinear format (Figure 8.3B) (Bow et al., 2003; Pietersma, 2002). This shows a running word-by-word summary of annotation for each verse alongside an algorithmically generated phonetic transcription and a word-aligned interlinear translation into English.

Figure 8.3:
Drill-down
interface.

Color-coding is used to highlight morphological segmentation of the Arabic script, with corresponding grammatical summaries displayed in both Arabic and English. Annotators can view further detail for an individual word by clicking through to the analysis web page, where the natural language generation component in the service tier is used to present a more detailed grammatical summary (Figure 8.3C). The analysis page allows collaborators to review all relevant tags for each word in the corpus using a textual summary that describes morphological segmentation, part-of-speech tagging, and syntactic dependency analysis in English and Arabic. Figure 8.4 below shows an extract of the morphological analysis page for token (21:70:4) of the Quran:

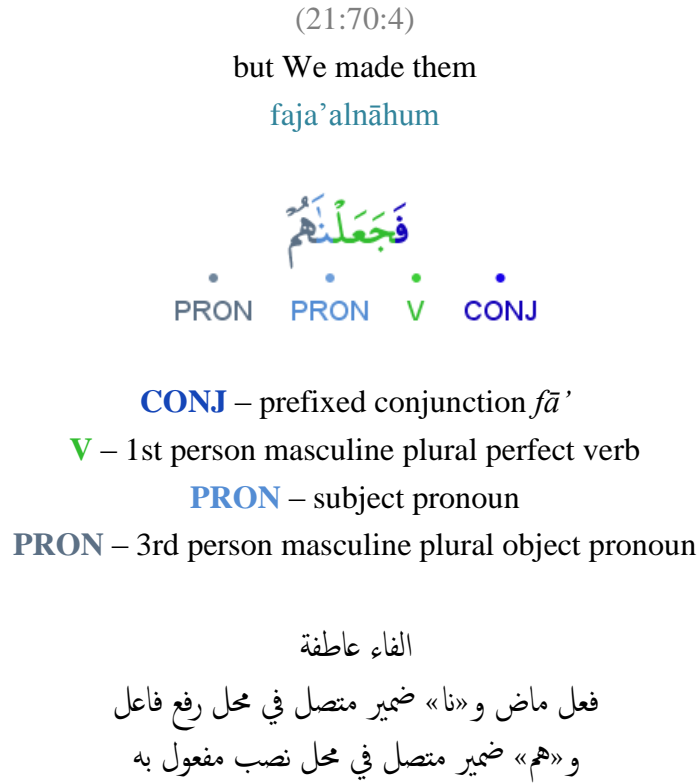

Figure 8.4: Morphological annotation with generated summaries.

The use of natural language generation is a useful addition to the website. For example, a typical Quranic word such as *faja'alnāhum* (فَجَعَلْنَاهُمُ), translated as 'but we made them', has a detailed grammatical description generated automatically using the tags stored in the linguistic database:

> The fourth word of verse (21:70) is divided into 4 morphological segments. A resumption particle, verb, subject pronoun and object pronoun. The connective particle *fā'* is usually translated as 'then' or 'so' and is used to indicate a sequence of events. The perfect verb (فعل ماض) is first person plural. The verb's triliteral root is *jīm 'ayn lām* (ج ع ل). The suffix (نا) is an attached subject pronoun. The attached object pronoun is third person masculine plural.

Based on observing inter-annotator discussion, the majority of collaborators usually prefer to proofread morphological and syntactic analysis in this textual format, instead of reviewing lists of abbreviated tags, features and syntactic relations. The benefit of this approach is that since the grammatical information is equivalent, the underlying tags in the database are indirectly reviewed in parallel. At the same time, a textual format is more easily comparable to the linguistic analyses in gold standard reference works of canonical Quranic grammar. Using the annotation methodology described in the previous chapter, collaborators are invited to review and suggest corrections to this information online. An 'add message' button allows collaborators to start a new discussion thread, with comments for a specific word shown alongside annotations:

> You can add a message if this information could be improved or requires discussion.

To simplify the proofreading process, the analysis page includes a 'See Also' section that provides a set of contextual hyperlinks that are used by annotators to access related resources and tools (Figure 8.3C, page 169). This usability feature allows online collaborators to spend more time making key linguistic decisions. Quick and easy 'one click' access to relevant information provides the ability to see the choices and decisions made previously by other collaborators for related words in the corpus. This compares with other annotation projects for tagging Arabic offline that require annotators to spend time searching through guidelines and other documentation, often without direct access to the work of others who may be working in isolation on the same annotated text.

The contextual hyperlinks in the 'See Also' section are generated dynamically according to the type of word under analysis, depending on part-of-speech, syntactic role and morphology. For example, for the previously discussed Arabic word *faja'alnāhum* in verse (21:70), hyperlinks provide access to the relevant section in the annotation guidelines for verbs, subject and objects. Additional contextual links provide a graphical visualization of syntax using dependency graphs, as well as further links to other online grammatical analyses for the verse at related Arabic grammar and Quran websites.

### 8.3.3 Syntactic Treebank

The syntactic annotation task involves proofreading dependency tagging. In contrast to other syntactically annotated Arabic corpora, the Quranic corpus does not show only bracketed structures or flat lists of relations. To simplify collaboration, a visualization of hybrid dependency-constituency graphs described in Chapter 6 is dynamically generated in the service tier, based on the annotations in the database.[29] The online visualization is backed by the formal syntactic representation, and shows dependency relations, a phonetic transcription and an interlinear translation into English. The treebank can be browsed one verse at a time and is also searchable.

---

[29] http://corpus.quran.com/treebank.jsp

### 8.3.4 Discussion Forum

As described in Chapter 7, the website's message board is used as an online forum to promote open discussion between annotators and users of the corpus, who are typically Arabic students or Quranic researchers. Although the Quranic Arabic Corpus is a useful annotated resource as suggested by user feedback, organizing online collaborative analysis of Quranic Arabic is particularly challenging. Nearly all annotators are in agreement over the most important grammatical features for each word, such as part-of-speech and grammatical case. However, encouraging a large number of volunteers to contribute to annotation through linguistic discussion can lead to differences of opinion that are often hard, if not impossible, to resolve definitively for a small proportion of words in the corpus.

Despite not being one of the key linguistic tagging tasks, most inter-annotator disagreement revolves around the most appropriate interlinear Arabic-to-English translation and the subtly different uses of gender in Quranic Arabic. To ensure that online discussion remains relevant, editors acting as forum administrators close off-topic threads and archive resolved discussions that contain suggestions that have been implemented. As of September 2013, the message board contains 1,512 active threads, with an additional 5,229 archived messages.

## 8.4   Supplementary Resources

### 8.4.1 Reference Material

The following sections describe the supplementary resources made available to annotators. The first of these resources is relevant reference material used to support the annotation tasks. For annotating the Classical Arabic language of the Quran, it is possible to use a collection of certain key reference works as a form of gold standard to measure accuracy and to cross-check and verify analyses. The primary reference for syntax is the analysis by Salih (2007). However, this work does not cover several morphological features which are tagged using online collaboration. For verifying the annotation of derived Arabic verb forms and roots, as well as for grammatical gender, Lane's Lexicon (Lane, 1992) and Wright's reference grammar (Wright, 2007) are used. Both of these are widely considered to be highly authoritative reference works on classical Arabic grammar and for the Quran in particular. Additional Quranic dictionaries used to verify roots, lemmas and verb derivation forms include Omar (2005), Nadwi (2006), and Siddiqui (2008).

Producing a machine-readable annotated resource backed by these existing gold standard analyses is not simply a matter of scanning in the material and applying automatic character recognition. The Quranic Arabic Corpus is designed to be an open source resource, and any material used must be free of copyright. Even if this was not a concern, character recognition for printed Arabic texts such as Salih's *al-i'rāb al-mufaṣṣal* is presently challenging (Amara and Bouslama, 2005). A further obstacle to automatic extraction is that the grammatical analyses in these reference works are not encoded as a series of easily machine-readable tags or tables. Instead the syntactic dependencies and morphological analyses are described in free text, often using detailed technical linguistic language. The approach followed on the website is to use traditional works as references to guide the annotation process, instead of attempting to use them as automatic datasets.

## 8.4.2 Dictionary and Morphological Search

Two other popular resources provided alongside corpus annotations are the Quranic dictionary and morphological search. The online morphological search tool acts as an extended concordance, allowing annotators to search by part-of-speech, stem, lemma, root and other annotated morphological features.[30] This allows collaborators to compare against previous annotations by quickly finding related words. For example, the surface form ذهب in Arabic has two readings, as either the noun 'gold' or the verb 'to go'. By searching using POS tag and root, the occurrences of the correct reading can be easily found in the corpus.

---

[30] http://corpus.quran.com/searchhelp.jsp

Similar to the morphological search tool, the Quranic dictionary uses the annotated morphological data, but presents this information in a format more suited to browsing. The dictionary organizes words first by root then further by lemma, with contextual translations into English. Natural language generation is used to automatically generate summaries for each root. For example, for the root *bā hamza sīn*, the dictionary lists occurrences of word-forms as hyperlinks after generating the following summary information:

> The triliteral root *bā hamza sīn* (ب أ س) occurs 73 times in the Quran, in six derived forms:
>
> 40 times as the form I verb *bi'sa* (بِئْسَ)
> Twice as the form VIII verb *tabta'-is* (تَبْتَئِسْ)
> 25 times as the noun *ba's* (بَأْس)
> Four times as the noun *ba'sā* (بَأْسَاء)
> Once as the adjective *ba'īs* (بَئِيس)
> Once as the active participle *bā'is* (بَائِس)

The concordance lines shown in search results and in the Quranic dictionary are more sophisticated in comparison to previous Arabic corpora. Instead of showing surrounding context using a fixed number of words, a set of contextual rules select a dynamic window size based on a word's syntactic role in a sentence. This uses phrase structure and headword information from the treebank to bound the window and provide readable entries similar to printed Quranic concordances:

Noun – *nasab* (نَسَب)

| | |
|---|---|
| (23:101:6) 'will be relationship' | فَإِذَا نُفِخَ فِي الصُّورِ فَلَا **أَنْسَابَ** بَيْنَهُمْ يَوْمَئِذٍ |
| (25:54:8) 'blood relationship' | وَهُوَ الَّذِي خَلَقَ مِنَ الْمَاءِ بَشَرًا فَجَعَلَهُ **نَسَبًا** وَصِهْرًا |
| (37:158:5) 'a relationship' | وَجَعَلُوا بَيْنَهُ وَبَيْنَ الْجِنَّةِ **نَسَبًا** |

### 8.4.3 Ontology of Concepts

Although not presented as a research result of this thesis, a further resource on the website is an ontology. The motivation for this resource is to encourage users to engage and participate in the annotation effort. To link with the grammatical annotation, concepts were chosen for inclusion in the ontology if they are proper nouns, or if they represent well-defined concepts such as the names of animals, locations and religious entities. The ontology is based on the knowledge contained in traditional sources, including the hadith of the prophet Muhammad and Quranic exegesis by Ibn Kathir (Al-Mubarakpuri, 2003). An overview diagram on the website shows a visual representation of the ontology (Figure 8.5 below). This graph is a network of 300 linked concepts with 350 relations, and supports drill-down into individual concepts and verses. As well as listing the major concepts in the Quran, the ontology also defines a set of core semantic relations between these concepts. An example of this is the set membership relation 'instance' in which one concept is defined to be an instance or individual member of another group.

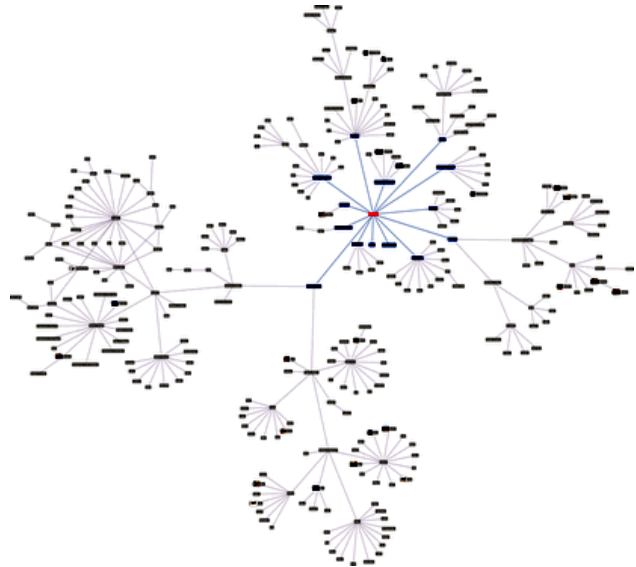

Figure 8.5: Concept nodes in the Quranic ontology.

For example, the relation 'Satan is a jinn' in the ontology would represent the knowledge contained in the Quran that the individual known as Satan belongs to the set of sentient creations named the jinn. Other concepts in the ontology are grouped into logical categories, according to the properties that they share. For example, the 'Sun', 'Earth' and 'Moon' are classified under 'Astronomical Body'.

In the morphological word-by-word view, a small number of pronouns (approximately 100) are hyperlinked to concepts in the ontology in order to resolve certain key anaphoric references.[31] For example, verse (97:1) states 'we revealed it' (إِنَّا أَنزَلْنَاهُ). It is known through traditional Quranic exegesis that this verse refers to Allah revealing the Quran. The analysis online shows this as:

> (97:1:2) *anzalnāhu*
>
> PRON – subject pronoun → Allah
>
> PRON – 3rd person masculine singular object pronoun → Quran

Website users can navigate the concept map online, which shows each concept, its definition and location in the visual map, subcategories, related concepts and predicate logic relations with subclasses and instances. There is also a topic index that supports clicking on a concept in the list to see a summary of that topic and a list of all occurrences of the concept in the Quran with concordance lines.

## 8.4.4 Published Datasets

Morphological annotation is made available as a free download, encoded as UTF-8 plain text file and published under the open source GNU public license. The format of data file is similar to the feature notation described in section 5.3.3, but includes additional tags to make segment types explicit. Each line of the file

---

[31] A more comprehensive tagging of Quranic anaphoric pronouns has since been provided by Sharaf and Atwell (2012b).

corresponds to one morphological segment. For example, the analysis shown in Table 8.1 (page 164) is published using the following segmentation:

| | |
|---|---|
| (6:76:7:1) qaAla | STEM\|POS:V\|PERF\|LEM:qaAla\|ROOT:qwl\|3MS |
| (6:76:8:1) ha`*aA | STEM\|POS:DEM\|LEM:ha`*aA\|MS |
| (6:76:9:1) rab~i | STEM\|POS:N\|LEM:rab~\|ROOT:rbb\|M\|NOM |
| (6:76:9:2) Y | PRON SUFFIX\|PRON:1S |

The syntactic annotation in the Quranic Treebank is not available as a download file as this is annotation effort still in progress. The treebank data is expected to be published once it covers the entire Quran. However, it is assessable for browsing through the website as a set of visual dependency graphs.

## 8.4.5 Mailing List

During the initial phases of treebank design and online annotation, a mailing list was made available to annotators (January 2010 to August 2011).[32] This enabled the annotation guidelines and related tagging questions to be discussed before publishing the guidelines on the website. Several hundred subscribers registered with the mailing list, including active annotators, interested Arabic linguists and Quranic scholars. As the annotation guidelines have since been finalized, the mailing list has been archived and annotator discussion has moved to the message board to discuss the linguistic tagging of individual words in the corpus.

The mailing list was useful for engaging with subject experts. An example of this is the translation of Arabic grammatical terms into English. Although most terms have equivalent translations that can be found in previous literature, certain genre-specific terms applicable to the Quran required discussion. This included the part-of-speech tag known as حرف زائد. Following discussion online, it was decided to translate this as a 'supplementary particle'. For a sensitive text such as the Quran, it was felt that translating this as a 'redundant' or 'extra' particle might imply that parts of the Quranic text were superfluous.

---

[32] http://www.mail-archive.com/comp-quran@comp.leeds.ac.uk

## 8.5   Conclusion

This chapter described the architecture, design principles and features of an online Linguistic Analysis Multimodal Platform (LAMP). Although this platform has been applied to the Classical Arabic language of the Quran, the annotation model and software architecture may be of interest to other related corpus linguistics projects. The platform has been implemented in the Java programming language, organized into a three-tier architecture. The different tiers are a data access layer, a service layer and an online presentation layer. A set of computational linguistic components were described for offline tasks such as manual annotation and for online tasks such as generating dynamic content.

The key design principles of the website were usability and ease-of-use. Based on this design, the platform aims to make the annotation process subtle yet intuitive. A simple user interface leads to greater accuracy, improved consistency and more rapid annotation. The addition of useful supplementary resources has made the website more useful and has attracted a large number of interested linguists and Quranic scholars. The platform has enabled volunteer annotators to contribute time and effort to proofreading the corpus. This has fulfilled one of the main aims of the website, to bring the morphological and syntactic annotations up to gold-standard level.

# Part IV: Statistical Parsing

It should be mentioned that the reader should not
expect to read an algorithm as he reads a novel; such
an attempt would make it difficult to understand...
An algorithm must be seen to be believed.

*– Donald Knuth*

# 9 Hybrid Parsing Algorithms

## 9.1 Introduction

Part IV of this thesis consists of two chapters that describe a Hybrid Statistical
Parser (HSP) designed for hybrid dependency-constituency syntax. The parser's
name was chosen because it can be applied to general hybrid grammars, as its
algorithms do not specifically relate to Classical Arabic. This chapter describes
the design of parsing algorithms. In Chapter 10, these algorithms are combined
with a statistical model to drive parsing actions.

   The motivation for this parsing work comes from the intuition that early Arabic
grammarians had a deep understanding of the structure their language, and that a
hybrid representation is a good model for their conceptualization of sentence
structure. From a linguistic perspective, although traditional Arabic grammar is
primarily dependency-based, it utilizes a restricted form of constituency syntax
(Itkonen, 1991). In contrast, pure constituency models for Arabic parsing have not
generalized well, leading to parsing underperformance (Kulick, Gabbard and
Marcus, 2006; Green and Manning, 2010).

   This thesis asks if a hybrid representation is suitable for statistically parsing
Classical Arabic. This is addressed by using the Quranic Treebank to construct
parsing models. However, fully parsing the treebank is challenging. To restrict the
scope of the problem, gold-standard morphological annotation is assumed as input
to the parser. The parsing models generate phrase structure, dependency relations,

and elliptical structures including empty categories and dropped pronouns. This problem is an extension of the 2007 CoNLL shared task for pure dependency parsing, in which gold-standard morphological annotation was used to benchmark parsing models for several morphologically-rich languages (Nivre et al., 2007a).

The parsing models in this chapter are inspired by two recent approaches. The first is the dual dependency-constituency work by Hall et al. (2007b; 2008). This is a combined model trained on separate dependency and constituency treebanks that is able to output both representations simultaneously. One insight of this work is that merged dependency structures can encode constituency information using enriched edges. The second source of inspiration is the evaluation methodology for joint morphological and syntactic disambiguation for Hebrew (Goldberg and Elhadad, 2011). This work demonstrates that joint disambiguation outperforms a pipeline approach for their task by evaluating both against the same dataset.

The Classical Arabic parser is evaluated using a similar methodology to recent joint work for Hebrew, by comparing two models. The first is a pipeline process that converts the output of a dependency parser to the hybrid representation using enriched edges similar to Hall et al. The second model uses a novel one-step algorithm that is able to construct the hybrid representation directly without post-processing. In the evaluation in the next chapter, it is shown that the pipeline approach achieves an F1-score of 87.47%, compared to an improved F1-score of 89.03% for the integrated model. These accuracy scores are close to state-of-the-art performance for other languages such as English, demonstrating that hybrid statistical parsing is achievable for Classical Arabic.

This chapter focuses on hybrid parsing algorithms, and is organized as follows. Section 9.2 provides relevant background information and gives an overview of transition parsing systems. Section 9.3 presents a description of hybrid graphs by combining the formalizations of morphology and syntax from Chapters 5 and 6. Sections 9.4 and 9.5 describe two parsing algorithms for the pipeline approach and the integrated approach respectively. Finally, section 9.6 concludes.

## 9.2   Transition Parsing Systems

### 9.2.1 Background

Two main approaches to statistical parsing are deterministic and non-deterministic methods. Parsing work using non-deterministic methods generally uses dynamic programming algorithms, such as chart parsing or global optimization. Examples include parsing models by Collins (1999), Charniak (2000) and Bikel (2004) for constituency syntax and McDonald et al. (2006) for dependency grammar. These parsers perform an exhaustive search over all possible parse trees (or dependency graphs) for a sentence. Trees are ranked using a statistical probability measure induced from a treebank, with the most likely tree selected as the final result.

   In contrast, deterministic parsing algorithms do not search through a space of possible parsing solutions. Instead, they operate incrementally by building a result tree one step at a time. These algorithms make a series of local decisions on how best to construct the tree using a statistical model. In this methodology, parsing becomes a classification problem. The parser needs to decide at each step of the incremental process which action to perform next in order to continue building its result tree, guided by contextual information. In addition to their efficiency, deterministic parsers are interesting as models of human parsing. Because they operate incrementally, these parsers relate to work in cognitive modelling, where psycholinguistic evidence has suggested that human parsing is predominantly incremental (Brants and Crocker, 2000b).

   Deterministic parsers are also widely used in computer science to parse formal languages with well-defined grammars. A common example is the use of shift-reduce parsing to compile programming languages, in which a sequence of tokens is read one at a time using look-ahead for context (Knuth, 1965; Wirth, 1996). Variations of shift-reduce parsers have also been successfully used for natural language. For example, initial syntactic annotation in the Penn English Treebank was performed using a deterministic parser based on an extension of a shift-reduce algorithm, driven by hand-written grammatical rules (Hindle, 1983).

More recent work for deterministic natural language parsing has used statistical methods. These have been especially successful for dependency parsing work. As noted by Nivre and Nilsson (2003):

> It can be argued that in order to bring out the full potential of dependency grammar as a framework for natural language parsing, we also need to explore alternative parsing algorithms. Here we investigate deterministic algorithms for dependency parsing. In the past, deterministic approaches to parsing have often been motivated by psycholinguistic concerns, as in the famous Parsifal system (Marcus, 1980). However, deterministic parsing also has the more direct advantage of providing efficient disambiguation. If the disambiguation can be performed with high accuracy and robustness, deterministic parsing becomes an interesting alternative to more traditional algorithms for natural language parsing.

Although using greedy algorithms, the best deterministic parsing models have performance scores only slightly lower than non-deterministic parsers. However, these parsers are attractive because they are relatively easy to implement. In addition, compared to exhaustive search, they typically have improved run-time complexity for larger sentences. Many deterministic parsers are classifier-based and run in linear time, such as the constituency parser by Sagae and Lavie (2005), and the dependency parser by Nivre et al. (2007b). Similar to previous work for constructing the Penn Treebank by Marcus et al., both these parsers are based on variations of a shift-reduce algorithm. However, in contrast to using hand-written rules, these examples use support vector machines (SVMs) to drive parser actions. The algorithms described in this chapter are also deterministic and classifier-based. Although the machine learning experiments described in next chapter also use SVMs, the algorithms presented in this thesis work with a new syntactic representation for hybrid dependency-constituency structures.

Formally, deterministic parsers are state transition systems. These are abstract machines that consist of a set of states and transitions between states. For parsing, the complete state of the system at a point in the algorithm includes the parser's internal state as well as the state of the partially constructed result tree. Together, these represent the configuration of the system. Adapting the notation used by McDonald and Nivre (2007), a transition parsing system is defined as:

1. A set $C$ of parser configurations. Each element of $C$ represents a partially built parse tree (or dependency graph).

2. A set $T$ of state transitions between configurations. Each element of $T$ is a function $t : C \rightarrow C$.

3. For every input sentence $x$:

    (a) a unique initial configuration $c_x$
    (b) a set $C_x$ of terminal configurations

To parse a sentence $x$, the parser follows a transition sequence. Formally, this is a sequence of configurations $C_{x,m} = (c_x, c_1, \dots c_m)$ such that $c_m \in C_x$ is a terminal configuration, and such that each configuration follows using a state transition:

$$\forall\, c_i \; \exists\, t \in T : c_i = t\,(c_{i-1}) \quad (1 < i \leq m)$$

The following sections describe transition systems for shift-reduce constituency and dependency parsing. These are relevant to hybrid parsing, which combines these together with state transitions for elliptical structures. In this chapter, only formal specifications of transition systems are provided. The statistical models for choosing specific transition sequences are described in Chapter 10.

## 9.2.2 Transition Constituency Parsing

This section describes a shift-reduce parser for constituency representations. This system constructs a parse tree using bottom-up processing. To define the system, let $x$ be a sentence that has been divided into a sequence of syntactic units:

$$x = (w_1, \dots w_n)$$

For English, the units $w_i$ would be POS-tagged words. For morphologically-rich languages such as Arabic, the syntactic units are morphological segments. The configuration of the parser is the combined state of two data structures: a queue $Q$ and a stack $S$. The queue contains only syntactic units, whereas the stack contains either units or partially constructed sub-trees. In its initial configuration, all units are placed onto the queue, with the stack empty:

$$Q = (w_1, \dots w_n) \text{ and } S = \emptyset$$

During shift-reduce parsing, two transitions are possible, a shift operation or a reduce operation. To define these, let $Q = (q_1, \dots, q_A)$ and $S = (s_1, \dots, s_B)$ be the state of the parser before a transition, and $Q'$ and $S'$ be the next configuration state. The two transitions are:

1. A shift transition $\Pi$. This operation reads the next item from the queue and moves it to the top of the stack: $Q' = (q_2, \dots, q_A)$ and $S' = (q_1, s_1, \dots, s_B)$.

2. A reduce transition $\Lambda(n, r)$. This removes $n$ elements from the top of stack. They are replaced with a new constituency tree with a root element $r$ having the removed elements as its child nodes: $Q' = Q$ and $S' = (r, s_{n+1}, \dots, s_B)$.

In its final configuration, the parser terminates when the queue is empty and the stack contains a single element:

$$Q = \emptyset \text{ and } S = (p)$$

The resulting parse tree $p$ is the final output of the parser. As an example of this process, Figure 9.1 below shows a constituency tree (right of the figure) together with the corresponding transition sequence (left of the figure). In the initial state, the queue contains words with their POS tags.

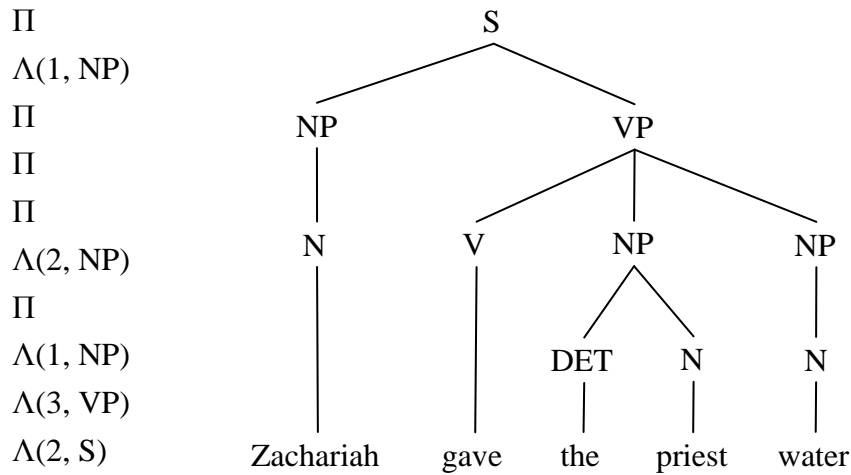

Figure 9.1: Example constituency transition sequence.

Similar transition systems to the one outlined in this section have been used in combination with statistical models to guide parser actions. For example, Sagae and Lavie (2005) describe a shift-reduce parser for the constituency representation in Penn English Treebank. Using tree binarization transformations, they report high accuracy scores of 87.54% precision and 87.61% recall.

### 9.2.3 Transition Dependency Parsing

There are several ways in which transition parsing can be extended to dependency structures by adding extra transitions. Nivre et al. (2007b) classify the two main approaches as 'arc-standard' and 'arc-eager'. The difference between the two is that arc-standard builds its dependency graph using bottom-up processing. This section describes a system that is similar to arc-standard but that more easily generalizes to hybrid parsing. This parser also uses a queue $Q$ and a stack $S$, but includes a dependency graph as part of its state. As with the previous parser, in the initial configuration, all units are placed onto the queue with the stack empty:

$$Q = (w_1, \ldots w_n) \text{ and } S = \emptyset$$

The initial dependency graph consists of the units $w_i$ as nodes, with no edges on the graph. As before, let $Q = (q_1, \ldots, q_A)$ and $S = (s_1, \ldots, s_B)$ be the state of the parser before a transition, and $Q'$ and $S'$ be the next configuration state. In the pure dependency transition system, the four transitions are:

1. A shift transition $\Pi$. This operation reads the next item from the queue and moves it to the top of the stack: $Q' = (q_2, \ldots, q_A)$ and $S' = (q_1, s_1, \ldots, s_B)$.

2. A reduce transition $\Lambda(n)$. This operation removes the $n^{\text{th}}$ element from the stack: $Q' = Q$ and $S' = S \setminus (s_n)$. Only the reductions $n = 1$ and $n = 2$ are used.

3. A left transition $\Phi(r)$. This adds an edge to the graph with $s_1$ as the head node, $s_2$ as the dependent node and $r$ as the edge label.

4. A right transition $\Psi(r)$. This adds an edge to the graph with $s_2$ as the head node and $s_1$ as the dependent node and $r$ as the edge label.

The left and right transitions Φ and Ψ each add a dependency edge to the graph using the top two elements of the stack. These transitions are parameterized by a parameter $r \in R$, where $R$ is the set of all dependency edge labels. By applying these transitions together with shift and reduce, the dependency parser terminates when both the queue and stack are empty:

$$Q = \emptyset \text{ and } S = \emptyset$$

The transition set $T = \{ \Pi, \Lambda, \Phi, \Psi \}$ differs from the arc-standard transitions described by Nivre et al. (2007b) in two ways. Firstly, in the dependency graphs in the Quranic Treebank, labelled edges point from dependents towards heads. The parser in the arc-standard algorithm assumes that dependency graphs are constructed using the opposite convention, so that the left and right operations are reversed. Secondly, arc-standard does not have an explicit reduce transition, but instead uses combined operations. The reduction operation is made explicit here because it is used for phrase structure and subgraphs in the hybrid parser described in section 9.5. However, the combined operations are equivalent to the left and right transitions Φ and Ψ if these are followed by $\Lambda(2)$ and $\Lambda(1)$ reductions:

$$\Phi'(r) \equiv \Phi(r) \, \Lambda(2)$$

$$\Psi'(r) \equiv \Psi(r) \, \Lambda(1)$$

Variations of the arc-standard algorithm have been used for a several transition dependency systems such as the parsers by Yamada and Matsumoto (2003) and Nivre et al. (2007b). The latter also describe a dependency parser that uses the alternative arc-eager algorithm. This differs from arc-standard by including an explicit reduction transition $\Lambda(1)$. In addition, it uses different left and right edge transitions to combine bottom-up and top-down processing which may be more suitable for certain languages. Specifically, edges are added to the dependency graph as soon as the head and dependent nodes are known, even if the dependent node has not been fully parsed with respect to its own dependents. Although they do not consider a wide-coverage study using different algorithms, Nivre et al. report that arc-eager has improved accuracy for Chinese. For Classical Arabic, this thesis uses arc-standard as it is more easily adapted to hybrid parsing.

### 9.2.4 Dependency Parsing Example

This section illustrates the dependency system outlined in the previous section by parsing an example English sentence. This example has been chosen to highlight the use of the $\Lambda(1)$ and $\Lambda(2)$ reduction transitions. In Figure 9.2, an example pure dependency graph has been annotated using a scheme in which dependent nodes point towards heads. The words in the sentence have been labelled $w_1$ to $w_5$, and dependency edges have been labelled using the relation set $R = \{\ subj,\ obj,\ det\ \}$. The dependency graph represents the end state of the parser (the desired terminal configuration):

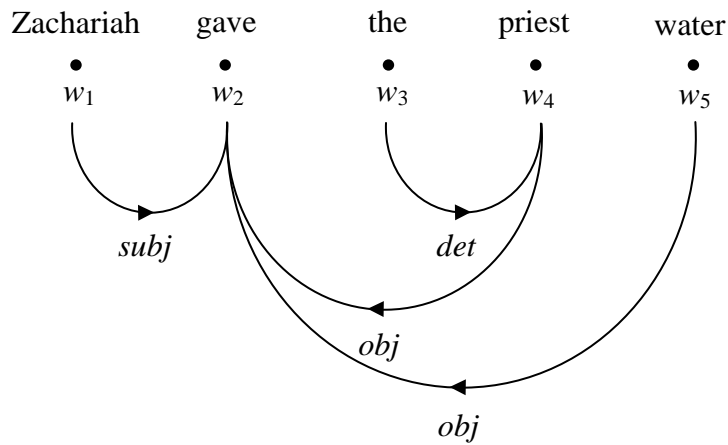

Figure 9.2: Example English dependency graph.

| Action | Stack | Queue | Dependency Graph |
|--------|-------|-------|------------------|
| $\Pi$ | $(w_1)$ | $(w_2, w_3, w_4, w_5)$ | |
| $\Pi$ | $(w_2, w_1)$ | $(w_3, w_4, w_5)$ | |
| $\Pi$ | $(w_3, w_2, w_1)$ | $(w_4, w_5)$ | |
| $\Pi$ | $(w_4, w_3, w_2, w_1)$ | $(w_5)$ | |
| $\Phi(det)$ | $(w_4, w_3, w_2, w_1)$ | $(w_5)$ | |
| $\Lambda(2)$ | $(w_4, w_2, w_1)$ | $(w_5)$ | |
| $\Psi(obj)$ | $(w_4, w_2, w_1)$ | $(w_5)$ | |
| $\Lambda(1)$ | $(w_2, w_1)$ | $(w_5)$ | |
| $\Pi$ | $(w_5, w_2, w_1)$ | $\emptyset$ | |
| $\Psi(obj)$ | $(w_5, w_2, w_1)$ | $\emptyset$ | |
| $\Lambda(1)$ | $(w_2, w_1)$ | $\emptyset$ | |
| $\Phi(subj)$ | $(w_2, w_1)$ | $\emptyset$ | |
| $\Lambda(1)$ | $(w_1)$ | $\emptyset$ | |
| $\Lambda(1)$ | $\emptyset$ | $\emptyset$ | |

Figure 9.3: Example dependency transition sequence.

In the parser's initial configuration, the stack is empty and the queue contains the words in the sentence: $Q = (w_1, ..., w_5)$ with $w_1$ at the top of the queue. Similarly, the dependency graph initially contains the five nodes $w_1, ..., w_5$ with no edges. Formally, the initial dependency graph is disconnected.

Figure 9.3 (page 191) shows a transition sequence for parsing the sentence that takes the system to a terminal configuration state. The first column shows each transition in the sequence, and the second and third columns show the state of the stack and the queue after a transition. In the diagram, transitions are grouped into five sections. Each of these sections has a dependency graph that shows the state of the graph after the transition in the first row of that section. In the first section (rows one to four), four shift operations are executed by the parser. This moves four syntactic units from the queue onto the stack, leaving the graph unchanged. The next sequence is $\Phi(det) \Lambda(2)$. As defined on page 188, this makes the top of the stack a head node using a *det* dependency relation, followed by removing the second item from the stack. Similarly, $\Psi(obj) \Lambda(1) \Pi$ makes the second item on the stack a head node, pops the top of the stack then shifts an element from the queue. The transition sequence continues until the queue and stack are both empty ($Q = \emptyset$ and $S = \emptyset$) and the dependency graph has been fully constructed.

## 9.3   Hybrid Representation

Before describing hybrid parsing algorithms, this section combines the formal representations of morphology and syntax from Chapters 5 and 6, and introduces additional notation that is relevant to parsing work. Using the definition from section 9.2.2, a sentence $x$ is divided into a sequence of syntactic units:

$$x = (w_1, ... w_n)$$

Because the remainder of this chapter focus on examples of Classical Arabic parsing, the syntactic units $w_i$ will be morphological segments.

### 9.3.1 Pure Dependency Graphs

Pure dependency graphs are defined within the context of an annotation scheme for morphological features (section 5.3.2) and dependency relations (section 6.4) These are:

- A set feature functions $F = \{f1, \dots f_m\}$. These associate feature-values with each morphological segment: $f_j(w_i) \in F_j$ ($1 \leq i \leq n$, $1 \leq j \leq m$).

- A set $R$ of dependency relations used to label graph edges.

A pure dependency graph is then defined as the tuple $G = (V, E, L)$, where:

1. $V = \{w_1, \dots, w_n\}$ are the vertices formed from morphological segments.

2. $E \subseteq V \times V$ are the graph's edges.

3. $L : E \rightarrow R$ are the edge labels.

### 9.3.2 Hybrid Dependency-Constituency Graphs

In the syntactic representation described in Chapter 6, nodes in hybrid dependency graphs are of four types: morphological segments, empty categories, phrases and reference nodes. The latter were used to relate words between different graphs. Although distinct for annotation, for the purposes of parsing, reference nodes can be assumed to be the same as other morphological segments.

In addition to feature functions and dependency relations, hybrid graphs also use a set of phrase tags $Z$ for constituency structure, listed in Table 6.2 (page 126). Untagged phrase nodes form a set whose elements are continuous spans over the morphological segments in the sentence. The set of all possible phrase nodes can be formalized as a set of ordered pairs that mark the start and end of each phrase:

$$\mathbb{P} = \{ (w_i, w_j) : 1 \leq j \leq i \leq n \}$$

Similarly, let $\mathbb{H}$ be the set of possible empty categories. In general, a hybrid graph has vertices which are possibly a subset of phrase nodes ($P \subseteq \mathbb{P}$) and empty categories ($H \subseteq \mathbb{H}$). A hybrid graph is then defined as the tuple $G = (V, E, L_1, L_2)$, where:

1. $V = \{w_1, ..., w_n\} \cup P \cup H$ are the vertices.

2. $E \subseteq V \times V$ are the graph's edges.

3. $L_1 : E \rightarrow R$ are the edge labels.

4. $L_2 : P \rightarrow Z$ are the phrase labels.

## 9.4   Algorithm I: Multi-Step Hybrid Parsing

This section describes a pipeline approach to hybrid parsing, which uses graph transformations to covert hybrid graphs to pure dependency graphs, without loss of information. Similar to the methodology by Hall et al. (2007b; 2008) for dual parsing, this is possible by encoding constituency information onto enriched edges in pure dependency graphs. Hybrid parsing is then dependency parsing (using the transition system in section 9.2.1) followed by post-processing. The complete process, including training from a pure dependency version of the treebank, is described in Chapter 10.

This section focuses on specifying transformations used to encode constituency information. By comparing the formalizations in sections 9.3.1 and 9.3.2, it can be seen that the differences between pure and hybrid graphs are phrase structure and ellipsis. Two graph transformations are used to account for these differences. A requirement of these transformations is that they are reversible, as hybrid-to-dependency is used for training, and dependency-to-hybrid is used for parsing.

## 9.4.1 Phrase Structure Conversion

Phrase structure conversion involves replacing a phrase node together with the edge to its head or (dependent node) by a new edge connecting to the head node in the subgraph spanned by the phrase. Figure 9.4 (overleaf) illustrates this process for a graph for verse (19:62) in the Quranic Treebank. In this example, the phrase is a dependent of a morphological segment (an accusative particle).

There are three main scenarios for phrase structure conversion. Consider a node $p = (w_i, w_j)$ in the hybrid graph spanning the morphological segments from $w_i$ to $w_j$ inclusively. The conversion for the phrase node $p$ is based on the observation that the phrase covers a subgraph with root $\omega(p)$. In the example in Figure 9.4, the subgraph root is a pronoun suffix. The conversion rules are:

1. If $p$ is a dependent node with edge $e$, head $h$ and dependency relation $r$, then $e$ and $p$ are removed and a new edge $e'$ is added with dependent $\omega(p)$, head $h$, and enriched dependency label $+r$.

2. If $p$ is a head node with edge $e$, dependent $d$ and dependency relation $r$, then $e$ and $p$ are removed and a new edge $e'$ is added with head $\omega(p)$, dependent $d$, and enriched dependency label $r+$.

3. If two phrases are connected by a dependency edge, then the two phrase nodes and the edge are removed. A new edge is added with the enriched dependency label $+r+$ connecting the roots of the two respective subgraphs.

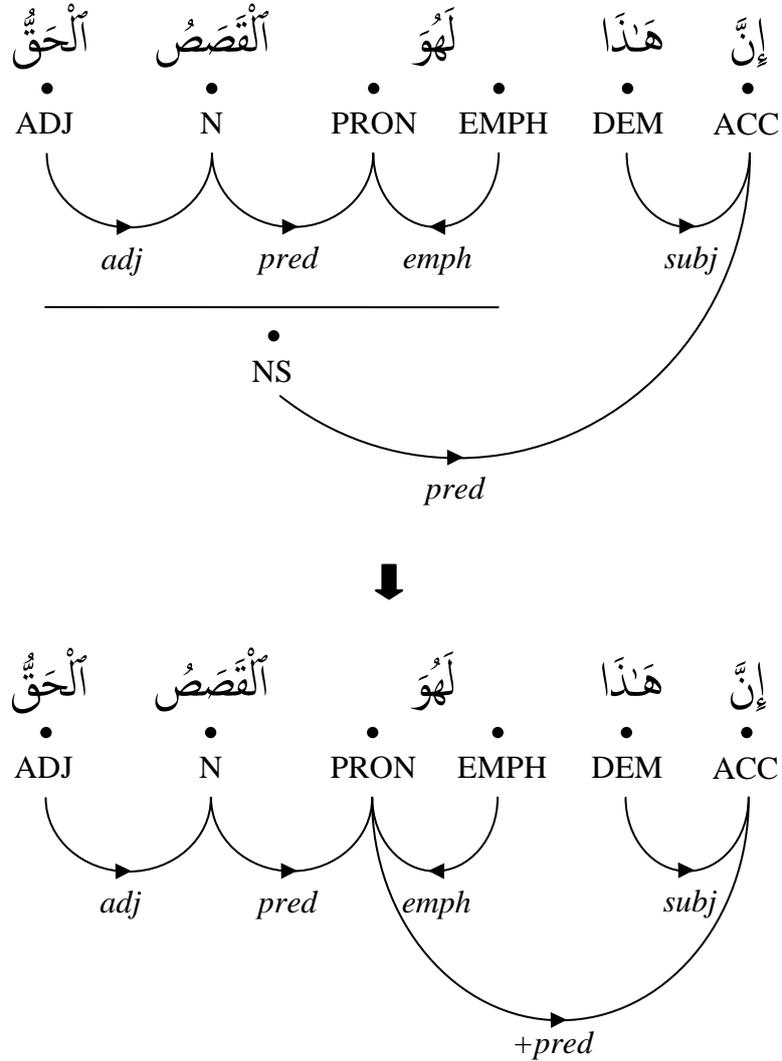

Figure 9.4: Conversion of phrase structure in verse (19:62).

In the inverse process, using the information encoded on enriched edges, +*r* and *r*+ denote expanding an edge's dependent or head into a subgraph respectively, and +*r*+ indicates that both head and dependent nodes should be expanded to produce an edge between a pair of phrases. Phrase tags are reconstructed using a small number of labelling rules, based on traditional Arabic grammar.

The conversion process outlined in this section uses a function $\omega(p)$ that maps a phrase node to the morphological segment that is the root of the subgraph spanned by the phrase. To define this function, let $V$ and $E$ be the vertices and edges of a hybrid graph respectively. A subgraph then has vertices $\bar{V}$ and $\bar{E}$ where

$$\bar{V} \subseteq V \;\text{ and }\; \bar{E} \subseteq E$$

Let $\bar{V} = \{w_a, \ldots w_b\}$ so that the phrase node spans the morphological segments from $w_a$ to $w_b$ inclusively. Let $\delta(x)$ be the function that maps each node $x \in \bar{V}$ in the subgraph to its head node, or $\delta(x) = \emptyset$ if $x$ is headless. If the phrase node covers a pure dependency subgraph, there exists a root node $w_h$ such that:

$$\omega(p) = w_h \text{ where } a \leq h \leq b \text{ and } \delta(w_h) = \emptyset$$

In the scenario where the phrase covers other phrases, the graph transformation process is performed recursively, so that phrases covering pure dependency graphs are converted first in a bottom-up process.

## 9.4.2 Conversion of Ellipsis

Conversion for ellipsis follows a similar process to phrase-structure conversion by building enriched edges in pure dependency graphs. Section 6.2.4 described the different types of ellipsis in the treebank as depending on morphological, syntactic and semantic context. The morphological form of ellipsis involves verbs with dropped subject pronouns. In the conversion process, these are simply removed from dependency graphs, as they can be easily recovered through the verb's morphological features (Figure 9.5, overleaf). To keep the transformation rules simple, only the most common additional case of elliptical structure is considered. Consequently, conversion does not account for all forms of ellipsis.

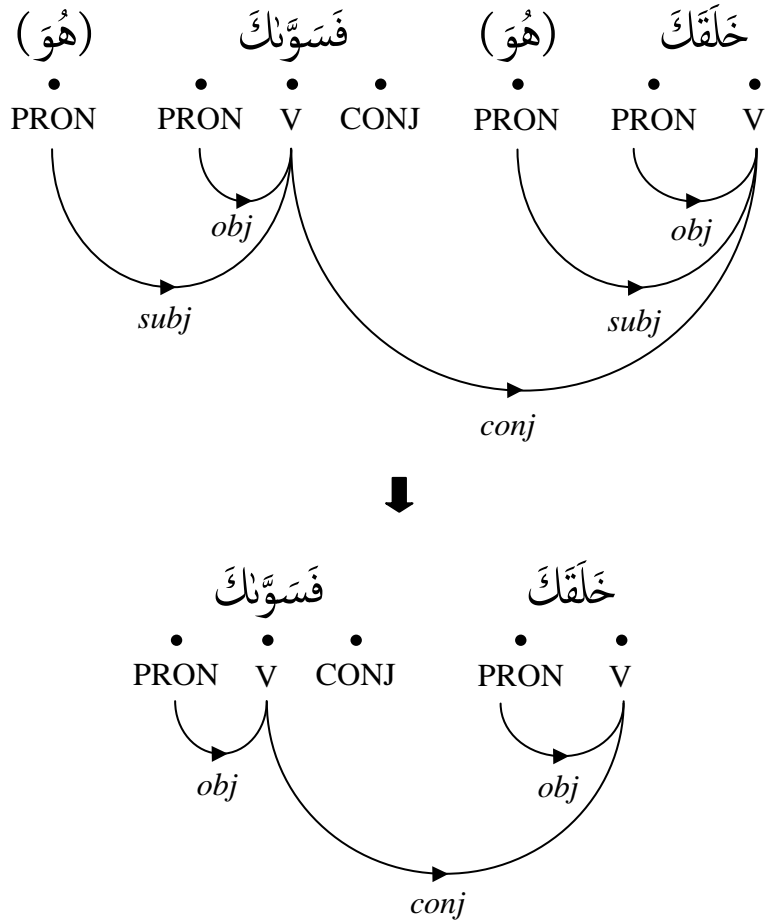

Figure 9.5: Conversion of dropped subject pronouns in verse (82:7).

The second conversion scenario occurs when two nodes are connected via an empty category. In this structure, if node $a$ depends on an empty category $e$ with part-of-speech tag $pos$ and relation $r_1$, and $e$ depends on $b$ with relation $r_2$, then the node $e$ is removed together with the two edges. A new edge is added to the graph with dependent $a$, head $b$ and enriched edge label $r_1 \mid pos \mid r_2$. Figure 9.6 (overleaf) shows an example of this conversion.

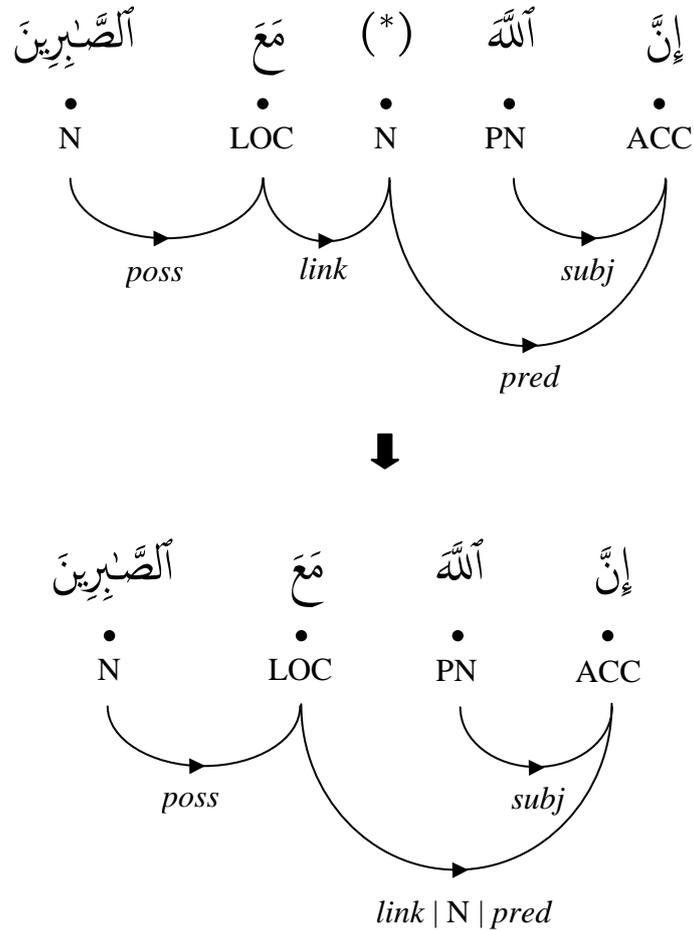

Figure 9.6: Conversion of syntactic ellipsis in verse (2:153).

As will be discussed in the evaluation section in Chapter 10, the performance of the multi-step approach to hybrid parsing is affected by the coverage of the conversion process. However, the small set of rules outlined in this section for phrases and ellipsis allow nearly all edges to be recovered. Using a small sample of the treebank, it was estimated that at most 5% of edges were not recovered in the hybrid graphs through this process.

### 9.4.3 Multi-Step Parsing Example

This section provides a complete example of multi-step hybrid parsing. Figure 9.7 shows a graph that is the desired terminal configuration of the parser after pure dependency parsing, but before post-processing. This corresponds to the graph for verse (4:141) from the treebank, shown in Figure 6.22 (page 139). In the graph below, the transformations described in the previous sections have been applied. The sentence is interesting as it contains a prepositional phrase attached to an empty category. When converted to pure dependency, the graph has an enriched edge that encodes a double transformation (+*link* | N | *circ*).

To parse this sentence, the initial configuration will be a disconnected graph consisting of all morphological segments from the graph in the treebank as terminal nodes, excluding the empty category. Figure 9.8 (overleaf) shows a transition sequence for this sentence. Similar to the previous example, actions in the diagram have been grouped into sections. The state of the dependency graph is shown after the first transition in each group. For brevity, the state of the queue is not shown.

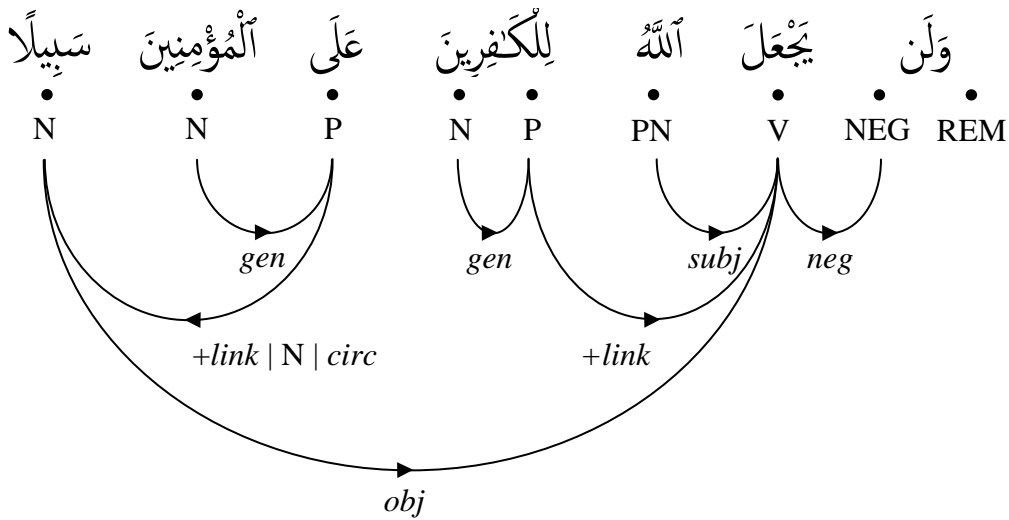

Figure 9.7: Converted graph encoding ellipsis and phrase structure.

| Action | Stack | Dependency Graph |
|--------|-------|------------------|

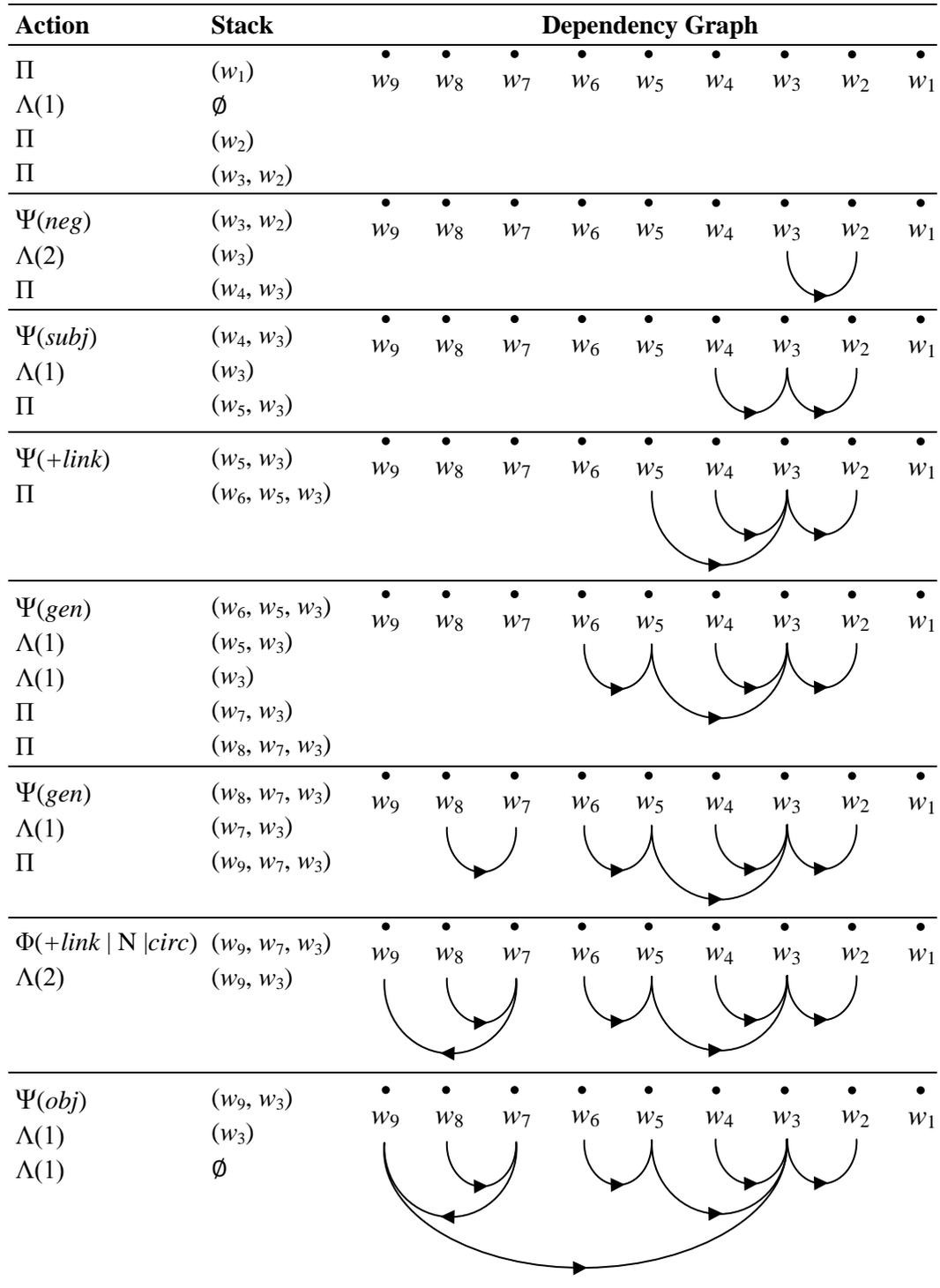

Figure 9.8: Transition sequence for multi-step dependency parsing.

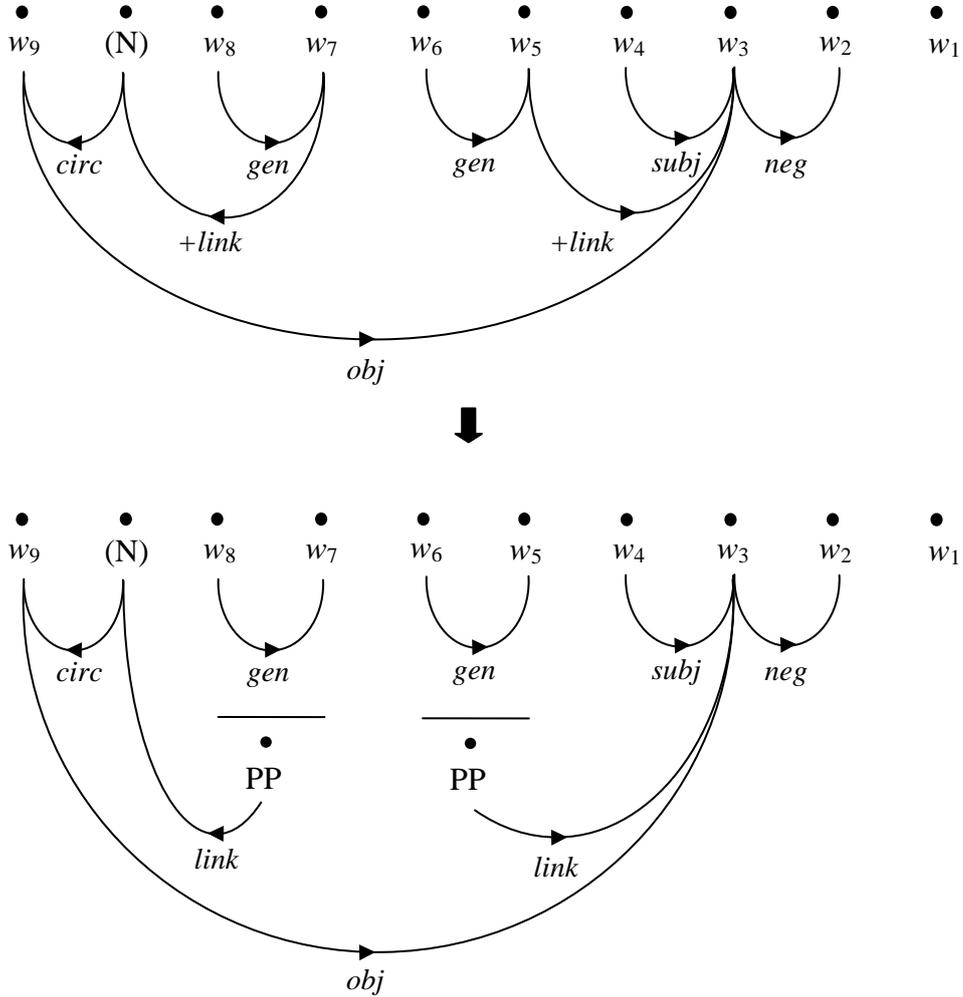

Figure 9.9: Post-processing transformations for verse (4:141).

After pure dependency parsing, the next phase is to transform the graph into a hybrid representation by decoding the enriched edges. Figure 9.9 above illustrates this process. In the first transformation, the edge labelled *+link* | N | *circ* has been converted into an empty category and two edges (upper graph). Finally, the *+link* edges are converted to phrase structure to complete the multi-step parsing process (lower graph).

## 9.5   Algorithm II: Integrated Hybrid Parsing

The multi-step algorithm outlined in the previous section uses pure dependency parsing followed by post-processing to apply graph transformations to build hybrid structures. This section describes an alternative algorithm that parses the hybrid representation directly without post-processing. This algorithm extends the pure dependency transition system described in section 9.2.3 by adding new state transitions for phrases and elliptical structures.

### 9.5.1 Extended Transition Set

To define the extended transition set, let $Q = (q_1, \ldots, q_A)$ and $S = (s_1, \ldots, s_B)$ be the state before a transition, and $Q'$ and $S'$ be the next state. The hybrid parser includes the four transitions { $\Pi$, $\Lambda$, $\Phi$, $\Psi$ } as well as new transitions { $\Theta$, $\Gamma$, $\Omega$ }. The last of these operations use spanning functions $\phi_1(w)$ and $\phi_2(w)$ defined later in this section. The three new transitions are:

1. A transition $\Theta(p)$ for empty categories. This adds an elliptical node $e$ to the graph after $s_1$ with POS tag $p$. The elliptical node $e$ is then pushed onto the stack: $S' = (e, s_1, \ldots, s_B)$.

2. A subject transition $\Gamma$. This is only applicable if $s_1$ is a verb. A dropped pronoun $e$ is inserted after $s_1$, and a *subj* edge is added with $s_1$ as the head node, and $e$ as the dependent node. The elliptical node $e$ is pushed onto the stack: $S' = (e, s_1, \ldots, s_B)$.

3. A subgraph transition $\Omega$. This adds a phrase node $p$ spanning the terminal nodes from $\phi_1(s_1)$ to $\phi_2(s_1)$ (the start and end of the subgraph with root $s_1$). The phrase node $p$ is then pushed onto the stack: $S' = (p, s_1, \ldots, s_B)$.

## 9.5.2 Elliptical Transitions

The transition $\Theta$ is used to add a new elliptical node to the dependency graph. To specify its location, let $V$ and $V'$ be the nodes before and after the transition. Using the notation from section 9.3.2, in general nodes will be terminals or phrases:

$$V = \{w_1, ..., w_n\} \cup P \cup H$$

In dependency diagrams, the morphological segments and empty categories are arranged as a sequence $(v_1, ..., v_k)$ with $k \geq n$ such that

$$v_i \in \{w_1, ..., w_n\} \quad \text{or} \quad v_i \in H \quad (1 \leq i \leq k)$$

Following the transition $\Theta$, a new node $e$ is added to the graph after $s_1$. To define its position, note that $s_1$ is an existing morphological segment on the graph:

$$\exists\, j : s_1 = v_j \text{ where } v_j \in \{w_1, ..., w_n\} \text{ and } 1 \leq j \leq n$$

After the transition, $V' = V \cup \{e\}$ and the new sequence of nodes will be:

$$(v_1, ..., v_j\,,\, e\,,\, ..., v_k)$$

Similar to the empty category operation $\Theta$ described above, $\Gamma$ inserts a dropped pronoun at the same position after $s_1$. However, it does this as a combined operation $\Gamma \equiv \Theta(pron)\,\Phi(subj)$. In addition, the operation takes into consideration the verb's morphology to produce the inflected pronoun's correct surface form.

### 9.5.3 Subgraphs and Phrase Structure

The third extended transition $\Omega(p)$ adds a phrase node with tag $p$ to the hybrid graph. This operation considers the root $s_1$ at the top of the stack and the subgraph rooted by that node. The new phrase node then spans the nodes from $\phi_1(s_1)$ to $\phi_2(s_1)$ inclusively, where these functions denote the start and the end of the subgraph respectively. These can be formally defined using the notation from section 9.4.1. Let $(w_a, \ldots w_b)$ be the sequence of ordered terminal nodes in the dependency subgraph with nodes $\bar{V}$ rooted by $s_1$ such that $\delta(s_1) = \emptyset$. Then

$$\phi_1(s_1) = w_a \text{ and } \phi_2(s_1) = w_b$$

As an example of this operation, Figure 9.10 shows an example graph before and after a $\Omega(p)$ transition with nodes numbered from right-to-left. In this example, if the node $w_2$ is at the top of the stack ($s_1 = w_2$) then $\phi_1(s_1) = w_1$ and $\phi_2(s_1) = w_4$. After the transition, a phrase node $p$ has been added to the graph spanning the nodes $w_1$ to $w_4$ inclusively:

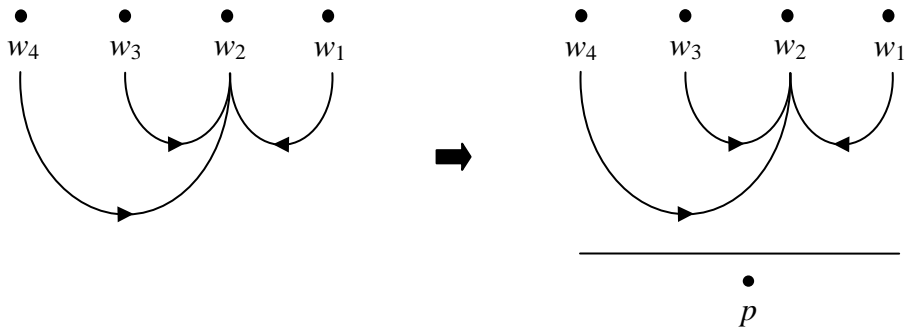

Figure 9.10: Phrase structure transition using a rooted subgraph.

## 9.5.4 Integrated Parsing Example

This section provides an example of integrated hybrid parsing using the graph for verse (7:186) from Figure 6.17 (page 131). For convenience, this graph has been reproduced in Figure 9.11 below using dependency edges labelled with English tag names and using numbered nodes. An elliptical noun denoted by an asterisk (*) is shown in the diagram between nodes $w_6$ and $w_7$. This diagram shows the desired terminal state of the parser. This verse was chosen because it illustrates a nested prepositional phrase (PP) within a nominal sentence (NS). It also includes a dependency edge between the prepositional phrase and an empty category head node (the elliptical noun). In this example, the extended operators $\Theta$ and $\Omega$ are used to construct elliptical dependencies and nested phrase structure respectively.

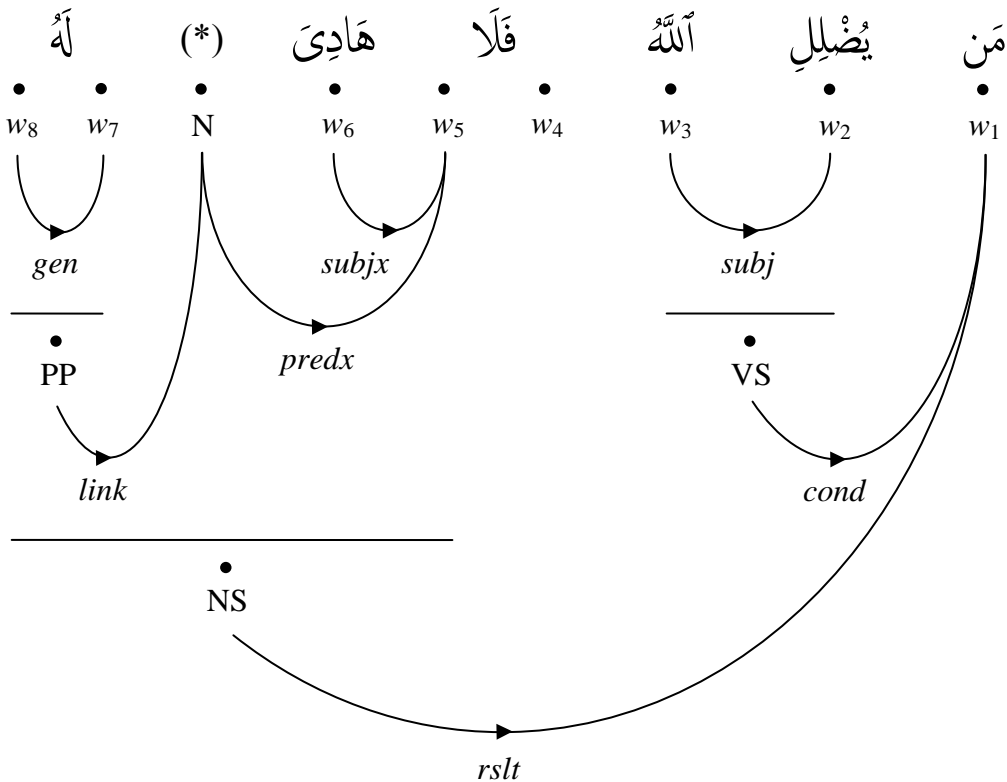

Figure 9.11: Hybrid dependency-constituency graph for verse (7:186).

| Action | Stack | Dependency Graph |
|---|---|---|
| Π Π Π | $(w_3, w_2, w_1)$ | |
| Ψ(*subj*) | $(w_3, w_2, w_1)$ | |
| Λ(1) | $(w_2, w_1)$ | |
| Ω(VS) | $(VS, w_2, w_1)$ | |
| Λ(2) | $(VS, w_1)$ | |
| Ψ(*cond*) | $(VS, w_1)$ | |
| Λ(1) | $(w_1)$ | |
| Π Λ(1) | $(w_1)$ | |
| Π Π | $(w_6, w_5, w_1)$ | |
| Ψ(*subjx*) | $(w_6, w_5, w_1)$ | |
| Θ(N) | $(*, w_6, w_5, w_1)$ | |
| Λ(2) | $(*, w_5, w_1)$ | |
| Ψ(*predx*) | $(*, w_5, w_1)$ | |
| Π | $(w_7, *, w_5, w_1)$ | |
| Π | $(w_8, w_7, *, w_5, w_1)$ | |
| Ψ(*gen*) | $(w_8, w_7, *, w_5, w_1)$ | |
| Λ(1) | $(w_7, *, w_5, w_1)$ | |

Figure 9.12: Hybrid transition sequence (first part).

| Action | Stack | Dependency Graph |
|---|---|---|
| $\Omega$(PP) | (PP, $w_7$, *, $w_5$, $w_1$) | |
| $\Lambda$(2) | (PP, *, $w_5$, $w_1$) | |
| $\Psi$(*link*) | (PP, *, $w_5$, $w_1$) | |
| $\Lambda$(1) | (*, $w_5$, $w_1$) | |
| $\Lambda$(1) | ($w_5$, $w_1$) | |
| $\Omega$(NS) | (NS, $w_5$, $w_1$) | |
| $\Lambda$(2) | (NS, $w_1$) | |
| $\Psi$(*rslt*) | (NS, $w_1$) | |
| $\Lambda$(1) | ($w_1$) | |
| $\Lambda$(1) | $\emptyset$ | |

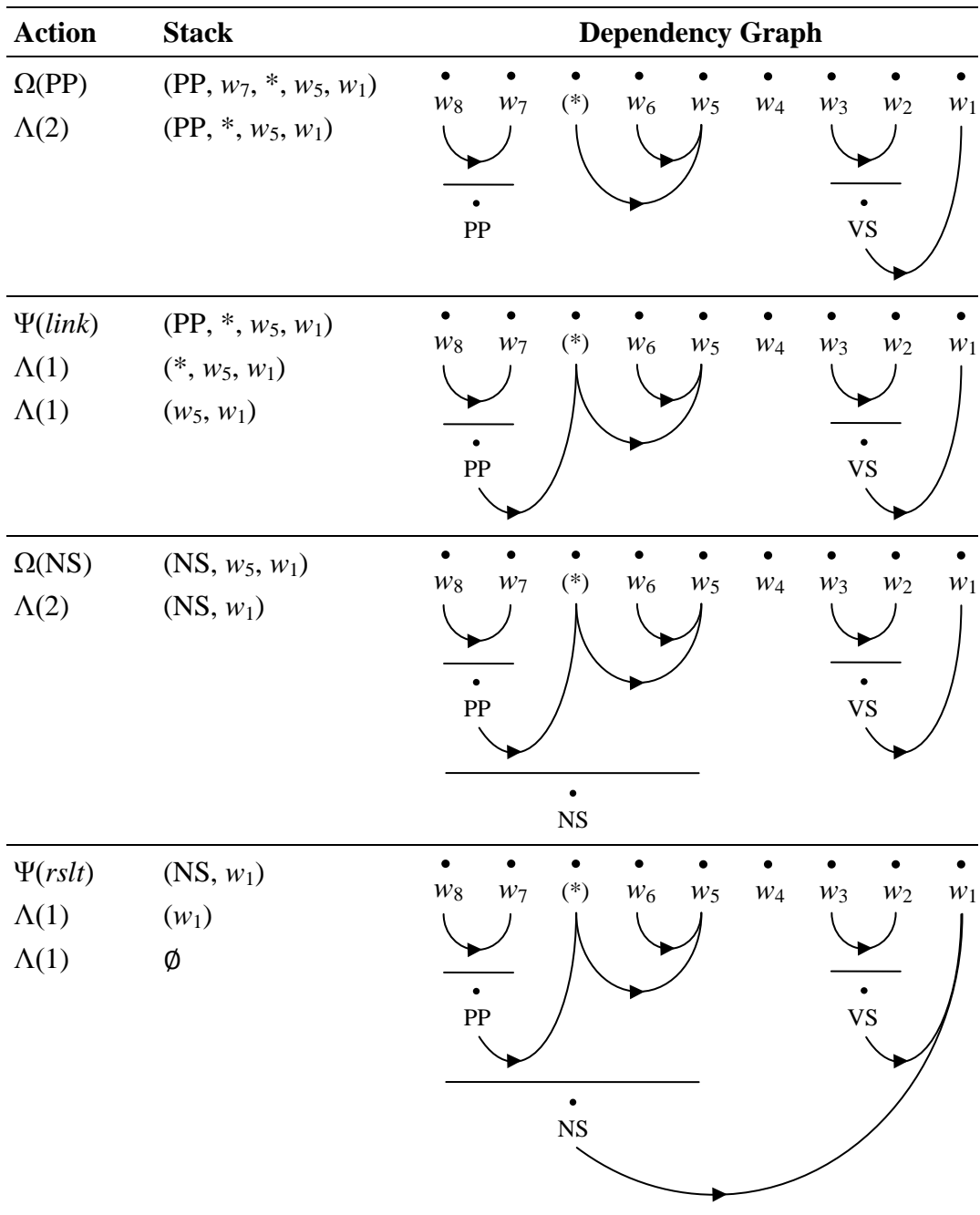

Figure 9.13: Hybrid transition sequence (second part).

The transition sequence for the verse is shown in Figures 9.12 and 9.13 (pp. 207 – 208). In the first section of Figure 9.12, three nodes are moved from the queue to the stack. In the next section, the sequence $\Psi(subj)\ \Lambda(1)$ forms a subject edge and discards $w_3$ as this node has no dependents. In the third section, a phrase structure operation is executed. The sequence $\Omega(VS)\ \Lambda(2)$ builds a verbal sentence (VS) spanning the subgraph headed by $w_2$, the top element of the stack. As shown previously, the reduction $\Lambda(2)$ is used in pure dependency parsing after a left transition $\Phi$. In this hybrid parsing example, $\Lambda(2)$ is useful because an edge should be formed between the first and third elements of the stack. After a $\Omega$ operation, it is possible to use $\Lambda(2)$ to discard the head of the subgraph spanned by the phrase, which would now be at the second element of the stack. After these operations, at the end of the fourth section of the diagram, the transition $\Theta(N)$ builds an elliptical node. This is inserted after $w_6$ as this is the top of the stack.

In the first section of Figure 9.13, a prepositional phrase is constructed using the sequence $\Omega(PP)\ \Lambda(2)$. Similar to the construction of the VS phrase, the head node of the subgraph spanned by the prepositional phrase ($w_2$ in this case) is also discarded from the stack. At this point, the PP node is at the top of the stack, followed by the elliptical node. In the second section, a dependency edge is formed between the phrase node and the elliptical node using a right $\Psi$ operation, as these two nodes are at the top of the stack.

In the third section of the diagram, nested phrase structure is constructed. At the configuration point just before the start of the section, $w_5$ is at the top of the stack. The operation $\Omega(NS)$ constructs an NS phrase node spanning the subgraph headed by $w_5$. The subgraph contains phrase structure itself, spanning the terminal nodes $w_5$ to $w_8$ inclusively. Finally, in the last section of the diagram, the action $\Psi(rslt)$ is executed. This forms a right pointing dependency edge between NS and $w_1$ (the top two nodes of the stack). This completes the dependency graph. The sequence $\Lambda(1)\ \Lambda(1)$ is then used to clear the stack. As both the queue and stack are empty at this point, the parser terminates. By following the transition sequence outlined above, the parser constructed the hybrid graph directly, without requiring further post-processing steps.

## 9.6   Conclusion

This chapter presented formal specifications of transition parsing systems. For hybrid parsing, two algorithms were described: a multi-step process that uses pure dependency parsing followed by post-processing, and a one-step integrated parser that constructs hybrid structures directly using novel state transitions. These systems were compared to the specifications in previous parsing work for pure consistency and pure dependency transition systems.

However, the systems described in this chapter are intentionally underspecified, as there are several ways in which transition sequences can be constructed. One approach is to use a set of hand-written rules to drive parsing actions. For example, Marcus et al. (1993) used a deterministic parser based on a transition system to perform initial automatic annotation of the Penn English Treebank. In comparison, during initial annotation of the Quranic Treebank, the one-step integrated algorithm was driven by hand-written rules based on traditional grammar. An alternative approach to deterministic parsing is to use a statistical model to build transition sequences. This approach is described in the next chapter, which applies machine learning to induce models for parsing actions from the gold-standard annotations in the treebank.

We are all agreed that your theory is crazy. The
question that divides us is whether it is crazy
enough to have a chance of being correct.

*– Niels Bohr*

# 10    Machine Learning Experiments

## 10.1    Introduction

This chapter describes HSP, a new system for hybrid statistical parsing. In
machine learning experiments, the parser is evaluated by dividing the treebank
into training and evaluation datasets. During the training phase, statistical models
for classifying state transitions are constructed for the two algorithms specified in
the previous chapter. During evaluation, the parsing algorithms are tested against
previously unseen sentences. In these experiments, it is not immediately obvious
which of the two algorithms results in higher accuracy. The integrated approach is
simpler because there are no conversion steps, and the parser is trained using the
full hybrid representation. However, although in both cases the same features are
available during training, the two approaches lead to different machine learning
problems. In the multi-step experiment, the parser has to learn more complex edge
labels. In contrast, there are fewer classification classes during one-step parsing as
phrase structure and ellipsis are integrated directly into the parsing process.

   This chapter is organized as follows. Section 10.2 describes the implementation
of HSP as a set of Java modules. Sections 10.3 and 10.4 describe the classification
training problem and the methodology used for machine learning. Section 10.5
describes the experiments and feature sets. Section 10.6 defines a new evaluation
metric for measuring hybrid parsing performance. Section 10.7 presents the
results. The effect of using different feature sets are discussed, and the results are
compared to Modern Arabic parsing work. Finally, section 10.8 concludes.

## 10.2 Parser Implementation

In comparison to previous work, the computational system most similar to HSP is MaltParser (Nivre et al., 2007b). This is an open source pure dependency parser written in Java that uses a shift-reduce transition system, trained using machine learning. Instead of adapting MaltParser, HSP was developed using a new Java codebase. This decision was made to allow for a more flexible architecture that would be easily extensible to hybrid parsing. In comparison to MaltParser, which is a command-line system, HSP includes a graphical user interface, created to help debug the parser. Figure 10.1 below shows a screen from the interface which allows viewing graphs from the treebank, as well as the ability to 'step through' and watch the effect of individual parsing actions in a transition sequence. An example hybrid transition sequence is shown on the left of the screen:

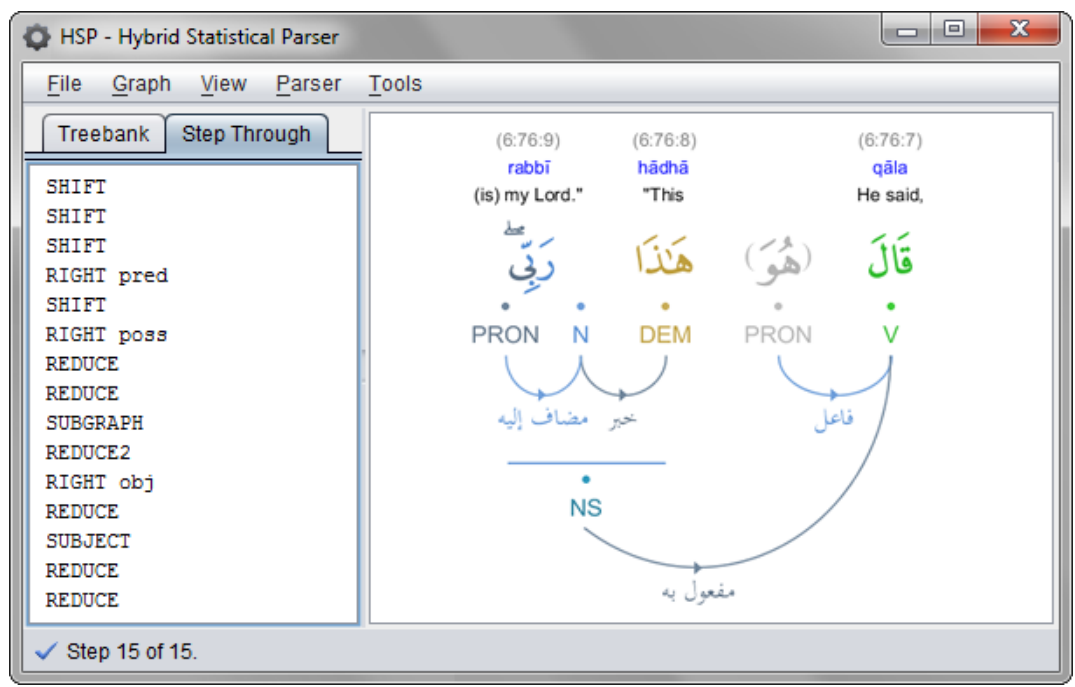

Figure 10.1: Java user interface for HSP with a hybrid transition sequence.

Figure 10.2 below shows the main Java components used to implement HSP. In this diagram, components have been organized into two sections for the multi-step parser (left components) and the integrated parser (right components). During training, an oracle reads from the treebank to construct a statistical model for each algorithm. LIBSVM is used for machine learning (described in section 10.4). During parsing, these models guide parsing actions to build a transition sequence for a given input sentence. The multi-step parser includes additional components for hybrid conversion. This design is described further in the following sections.

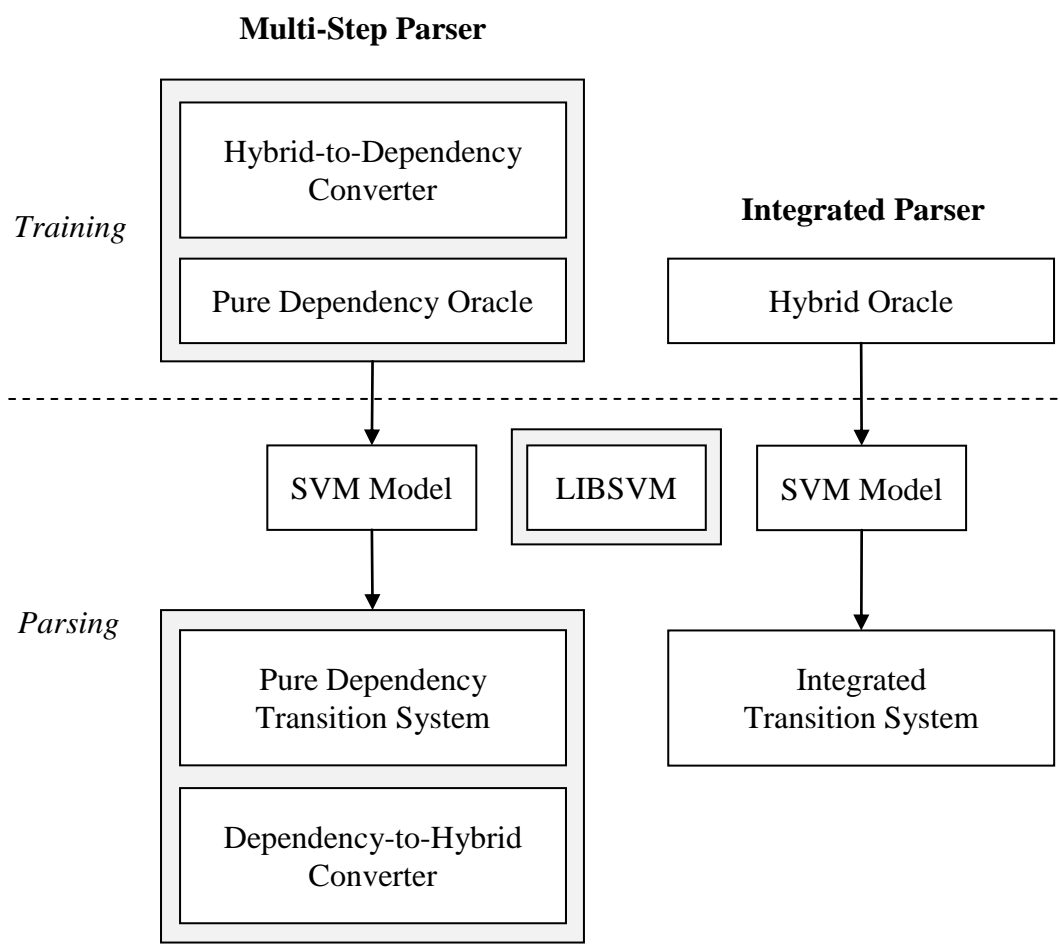

Figure 10.2: Architecture diagram for HSP showing component interaction.

## 10.3    Model Construction

### 10.3.1   Discriminative Probabilistic Models

In computational linguistics, probabilistic models for structured prediction can be categorized into discriminative and generative approaches. Given input data $x$ classified using labels $y$, discriminative models specify a conditional probability distribution $P(y \mid x)$. These contrast with generative models that specify a joint probability distribution $P(x, y)$. Examples of the former approach include logistic regression methods, neural networks, support vector machines and conditional random fields. Examples of the latter include probabilistic context-free grammars, hidden Markov models and naive Bayes classifiers.

Interestingly, the best dependency parsers use discriminative models (Nivre et al., 2007b; McDonald et al., 2006), whereas in contrast the best constituency parsers primarily use generative models (Charniak, 2000; Collins, 1999). Because the hybrid parser extends a dependency transition system, a discriminative model is used. For dependency parsing, the two main discriminative models both solve sequencing problems. McDonald et al. (2006) propose a two stage approach that first identifies dependencies using a deterministic parser then labels dependencies using a sequence labeller. In contrast, Nivre et al. (2007b) perform joint labelling.

The discriminative model used for HSP is a history-based model. In this approach, the transition sequence for an input sentence $x$ represents a sequence of decisions $d_1, \dots d_n$ used to construct the expected hybrid graph. In contrast to MaltParser, for integrated hybrid parsing, these decisions include building phrase structure and ellipsis as well as pure dependency structures. However, similar to the training methodology by Nivre et al., the conditional probability $P(y \mid x)$ can be expressed using the chain rule based on the history of previous decisions:

$$P(y|x) = \prod_{i=1}^{n} P(d_i|d_1 \dots d_{i-1})$$

To turn this into a pure classification problem, these conditional probabilities are estimated using a feature model. The transition systems in Chapter 9 are directly amenable to this type of estimation. For each configuration, the next transition is predicted using a feature vector associated with the first few nodes at the top of the queue and stack. Feature selection for hybrid parsing is discussed in section 10.5.

## 10.3.2  The Oracle

To construct the parsing models, an oracle is used during training (Kay, 2000). This is a computational component that reads each graph from the test part of the treebank, and constructs an expected transition sequence. The oracle is a perfect guide to predicting actions for supervised learning. This is because the expected transition sequence can be used to associate an input feature vector with each transition in the training phase.

| Action | Contextual Rule |
|---|---|
| $\Phi$ or $\Psi$ | $s_1$ and $s_2$ form a left or right edge |
| $\Lambda(2)$ | $s_2$ has all its edges accounted for |
| $\Omega$ | $s_1$ and $s_2$ are adjacent and form a phrase, unless $s_1$ has no dependents |
| $\Omega$ | $s_1$ is the root of a subgraph that is spanned by a phrase |
| $\Gamma$ | $Q = \emptyset$ and $s1$ requires a dropped subject pronoun |
| $\Lambda(1)$ | $s_1$ has all its edges accounted for |
| $\Lambda(1)$ | $Q \neq \emptyset$ |
| $\Lambda(2)$ | $s_1$ and $s_3$ form an edge |
| $\Theta$ | An empty category exists after $s_1$ |
| $\Lambda(1)$ | Default action if no other rules apply |

Table 10.1: Contextual rules used by the hybrid oracle.

In contrast to pure dependency training, for hybrid training, the oracle has to produce a more complex transition sequence. The algorithm used for HSP is incremental. The oracle maps two graphs: the expected graph and a working graph initially containing only terminals. The working graph is constructed using operations from the transition set until it matches the expected graph, and the resulting transitions are recorded. The hybrid oracle is driven by rules that use the current state of the queue and stack as context to select the next transition using the expected graph. These contextual rules are listed in Table 10.1 (page 215), in order of precedence. The table uses the notation from Chapter 9, where $Q$ denotes the queue, and $s_1$, $s_2$ and $s_3$ are the top three elements of the stack.

## 10.4   Machine Learning

### 10.4.1   Support Vector Machines

In principle, different classifiers could be used for hybrid parsing, such as logistic regression or decision trees. Inspired by previous work, HSP uses support vector machines as its algorithms are primarily dependency-based. The use of SVMs for dependency parsing was introduced by Yamada and Matsumoto (2003). Of more relevance to hybrid parsing, they are also used by Hall et al. for dual dependency-constituency parsing work (2007b; 2008). The Java version of LIBSVM was integrated into HSP for classification and training (Chang and Lin, 2011).

SVMs are binary classifiers that solve a linear separation problem by mapping training data points to a higher-dimensional feature space (Vapnik, 2000). Given $n$ points $(x, y)$, where $x$ is a feature vector and $y = \pm 1$, a hyperplane $w \cdot x + b = 0$ is constructed that separates points by a maximum margin (Figure 10.3, overleaf). The hyperplane is found by solving a quadratic programming problem:

$$\min_{w,b} \quad \frac{1}{2} \|w\| = C \sum_{i=1}^{n} \xi_i \quad \text{such that} \quad y_i(w \cdot x_i + b) \geq 1 - \xi_i$$

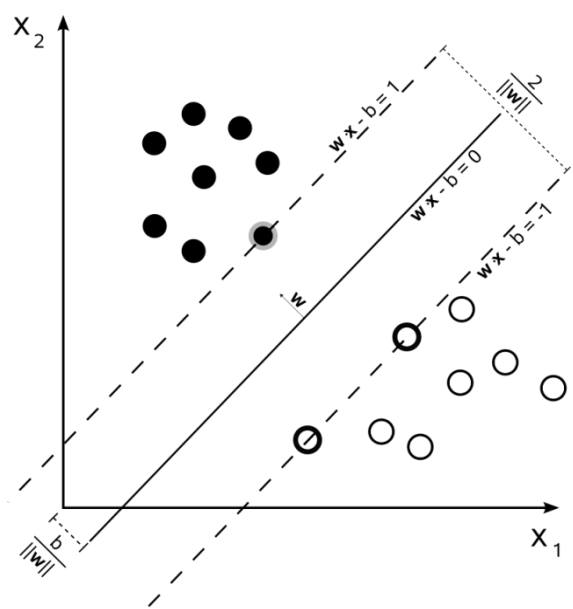

Figure 10.3: Maximum margins in SVM classification.

In this minimization problem, the terms $\xi_i$ are non-negative slack variables used to introduce a soft margin. This is required when no hyperplane exists that exactly separates the training data into two sets. The constant $C$ is used to define a penalty function. This is a free parameter that requires configuration during training.

## 10.4.2  Feature Binarization

HSP has a finite number of possible transitions. For hybrid parsing, these are the seven transition types used in combination with their parameters (POS tags and edge labels). To construct a numerical classification problem, the desired output transitions are represented by integers. HSP also applies binarization of input features in the training data so that a single symbolic feature is represented using many binary predicates. Binarized SVMs have been shown to exhibit improved classification for many learning tasks (Carrizosa et al., 2010).

### 10.4.3   Kernel Selection and Parameters

LIBSVM uses numerical approximation to find an optimal hyperplane, and requires several parameters to be specified during training. These include the parameter $C$ defined in section 10.4.1, and a parameter $\varepsilon$ used to terminate the approximation algorithm. In addition, LIBSVM offers a choice of different kernels to map features. Varying these parameters can result in a large number of machine learning runs. To simplify this process, HSP uses the same parameters that Hall and Nivre (2008) use for their dual parsing work for the German TIGER and TüBa-D/Z treebanks: $C = 0.5$ is used for penalty and $\varepsilon = 1$ for termination. The parameters $\gamma = 0.2$ and $r = 0$ are also used with the same quadratic kernel:

$$K(x_i, x_j) = (\gamma x_i^T x_j + r)^2$$

For machine learning problems, the kernel trick is a standard approach used to map data to a higher dimensional feature space where the hyperplane separation problem is more easily solved. For hybrid parsing with data from the Quranic Treebank, a quadratic kernel was found to give good results.

### 10.4.4   Reducing Learning Time

Learning time for SVMs depends on the size of each feature vector, as well as the number of points in the training set. Running against the Quranic Treebank, the experiments outlined in this chapter took 20 minutes per run, including model construction time using a specific morphological feature set, and evaluation time using 10-fold cross-validation.[33] To reduce learning time, HSP partitions training sets by using $POS(s_1)$, the part-of-speech at the top of the stack. One statistical classifier is then trained for each part-of-speech. This substantially reduced the training phase down from an original run time of several hours per experiment.

---

[33] Experiments were performed on a dual core PC running at 2.66 GHz with 4 GB of memory.

# 10.5    Experiments

The statistical parsing experiments described in this chapter use version 0.5 of the Quranic Treebank, containing 37,578 word-forms (~ 49% of the Quranic text), divided into 47,220 morphological segments. The experiments are organized into different runs that measure the effect of several factors on the performance of the parser. These include the choice of parsing algorithm (multi-step or integrated parsing), and the effect of using different feature sets for prediction.

## 10.5.1   Parsing Algorithms

HSP is designed to output both pure and hybrid dependency graphs. In the first set of experiments, HSP is used as a pure dependency transition parser and the hybrid representation is recovered through post-processing. In this process, the following steps are performed:

1.  The training data is converted to pure dependency by encoding additional information using enriched edge labels.

2.  In the learning phase, HSP is restricted to using only the four transitions that are required for pure dependency parsing: $T = \{ \Pi, \Lambda, \Phi, \Psi \}$.

3.  The parser's output is pure dependency. The hybrid representation is recovered by reversing the transformations in step 1.

For the conversion, the rules described in section 9.4 are applied to the treebank before training the pure dependency model. The size of the unconverted dataset is 50,955 terminal nodes, including 3,775 empty categories. The dependency graphs in the treebank contain 9,847 phrase nodes and 38,642 edges. After conversion, all phrase nodes and empty categories were removed, resulting in 47,220 terminals and a total of 34,849 edges. The number of edges dropped due to collapsing edges between empty categories, as per the conversion rules.

The second set of experiments uses the integrated parser. HSP is trained using the treebank's full hybrid representation without post-processing. In these experiments, the transition set is extended to include the three transitions required for hybrid parsing: $T = \{ \Pi, \Lambda, \Phi, \Psi, \Theta, \Gamma, \Omega \}$.

## 10.5.2 Graph Features

In the experiments, different combinations of features are used. At a specific point in parsing, a feature vector is constructed using features taken from the top three nodes on the stack ($s_1$, $s_2$ and $s_3$) and the top node on the queue ($q_1$). Two different types of features are used: static features and dynamic features. The former are morphological features, which do not change depending on the location of nodes in the graph. In contrast, dynamic features depend on the configuration of the dependency graph at a specific configuration point during parsing. The three graph features are Deprel, IsRoot and Edge. Each of these is a parameterized binary predicate. They are defined as follows:

- Deprel($w$, $r$) is parameterized using a relation $r \in R$ from the relation set. The binary predicate is set to true if the node $w$ has a dependent with that relation. For example, Deprel($s_1$, *subj*) is true the node at the top of the stack has an existing subject dependency that was previously parsed.

- IsRoot($w$) is set to true if the node $w$ is the root of a previously constructed well-formed subgraph. This feature is useful for building phrase structure.

- Edge($w_1$, $w_2$) is set to true if $w_1$ and $w_2$ form a previously parsed edge. Either $w_1$ or $w_2$ may be the head node.

As discussed in section 10.3, in a discriminative history model, features can be used to estimate the conditional probabilities in a decision sequence. The graph features used by HSP were each intentionally selected to represent part of the history of the previously constructed hybrid graph.

## 10.5.3  Morphological Features

After initial work using a subset of the data, it was decided to use five different sets of morphological features for the parsing experiments. These were grouped together to reduce the number of runs. The features are derived from the morphological feature set used for the Quranic Treebank, described in Chapter 5. The members of each group are shown in Table 10.2 below. All feature sets are used in combination with the same graph features described in the previous section.

| Features | Pos | Morph6 | Morph9 | Lemma | Phi |
|----------|-----|--------|--------|-------|-----|
| POS | Y | Y | Y | Y | Y |
| Phrase | Y | Y | Y | Y | Y |
| Voice | - | Y | Y | Y | Y |
| Mood | - | Y | Y | Y | Y |
| Case | - | Y | Y | Y | Y |
| State | - | Y | Y | Y | Y |
| PronType | - | - | Y | Y | Y |
| SegType | - | - | Y | Y | Y |
| Copula | - | - | Y | Y | Y |
| Lemma | - | - | - | Y | Y |
| Person | - | - | - | - | Y |
| Gender | - | - | - | - | Y |
| Number | - | - | - | - | Y |

Table 10.2: Morphological feature sets for parsing Classical Arabic.

The morphological feature sets are:

- **Pos:** This baseline feature set includes the part-of-speech and phrase tags for the selected nodes. In machine learning experiments, only using POS tags tests the accuracy of parsing Classical Arabic without using additional morphological information.

- **Morph6:** This set adds the core morphological features that might help with parsing, based on domain knowledge of traditional Arabic grammar: voice, mood, case and state. For example, case is known to be an important feature related to syntactic structure (Habash et al., 2007b).

- **Morph9:** Adds a further three morphological features. PronType marks a pronoun clitic as either an object pronoun or subject pronoun. As described in Chapter 5, due to Classical Arabic's rich morphology, these different types of clitics are common, and they form either subject or object dependency relations when attached to verbs. The feature SegType indicates if a morphological segment is a prefix, stem or suffix. The copula feature is used for a subset of copular verbs known as *kāna wa akhwātuhā* (كان واخواتها). Although assigned the same part-of-speech tag as normal verbs, in hybrid graphs these words form subject and predicate relations instead of subject and object.

- **Lemma:** To test the effect of lexicalization on the parser, this feature set adds lemmas. After initial experimentation, it was decided not to include Classical Arabic roots as this feature is possibly too general to be of use for parsing.

- **Phi:** This feature set includes the so-called phi-features of person, gender and number. For parsing Classical Arabic, these features may be relevant as they are used in traditional grammar to describe agreement rules.

## 10.6    Evaluation Metrics and Methodology

Two standard metrics for evaluating the performance of parsers are LAS (labelled attachment score) for pure dependency parsing, and Parseval for constituency parsing. LAS is a single measure, whereas Parseval defines three measures: precision, recall, and F1-score, where F1-score is the harmonic mean of precision and recall. For hybrid parsing, this section combines both LAS and Parseval into a single new metric termed ELAS (extended labelled attachment score). Before introducing ELAS, the two existing metrics are first defined in set-theoretic terms. It is then shown how these metrics can be combined.

### 10.6.1  Labelled Attachment Score and Parseval

In the CoNLL shared task on multilingual dependency parsing (Nivre et al., 2007a), LAS was used an official accuracy metric. Let $(w_1, ..., w_n)$ be an input sentence that has been morphologically segmented, $G = (V, E, L)$ be an expected graph from the reference data, and $G' = (V', E', L')$ be the corresponding pure dependency graph output by the parser. Let $H(w)$ be the expected head of the segment $w \in \{w_1, ..., w_n\}$, or $\phi$ if $w$ is headless. Similarly, if $H(w) \neq \phi$, let $l(w) \in L$ denote the expected label of the edge $e \in E$ from $w$ to $H(w)$. The LAS metric for the dependency parse pair $(G, G')$ is then defined as the cardinality ratio:

$$\frac{\left|\{w : H(w) \neq \phi \wedge H(w) = H'(w) \wedge l(w) = l'(w)\}\right|}{\left|\{w : H(w) \neq \phi\}\right|}$$

For a pure dependency graph, this is the fraction of segments that are assigned the correct head node and dependency label. This segment-based definition does not easily generalize to hybrid parsing since hybrid graphs can contain edges between phrase nodes. Therefore, this section provides a second definition of LAS by shifting focus from segments to edges.

For a well-formed pure dependency graph, the number of segments with heads is the same as the number of edges. Consider the edge equivalence relation $e \equiv e'$ defined to be true if and only if $e$ and $e'$ both connect $w$ to $H(w)$ and if $l(e) = l(e')$. This results in the following edge-based definition:

$$\text{LAS} = \frac{\left|\left\{e' \in E' : \exists e \in E \ \left(e \equiv e'\right)\right\}\right|}{\left|E\right|}$$

For constituency phrase structure, the Parseval metric (Black et al., 1991) can also be defined using a similar equivalence relation. Let $C$ denote the set of constituency labels. Given a sentence $(w_1, ..., w_n)$, let $p_{ij} = (w_i, w_j)$ be the phrase that spans the segments from $w_i$ to $w_j$ inclusively with label $c(p) \in C$. Let $P$ denote the set of non-terminal phrases in a parse tree from the reference data, and $P'$ be the corresponding set of phrases output by a pure constituency parser. A phrase $p' \in P'$ is considered to be correct if there exists an equivalent phrase $p \in P$ with the same label that spans the same terminal nodes. The phrase equivalence relation is:

$$p \equiv p' \Leftrightarrow \exists i, j : p = p_{ij} \land p' = p'_{ij} \land c(p) = c(p').$$

For the constituency pair $(P, P')$, Parseval precision and recall are defined as:

$$\text{Precision} = \frac{\left|\left\{p' \in P' : \exists p \in P \ \left(p \equiv p'\right)\right\}\right|}{\left|P'\right|}$$

$$\text{Recall} = \frac{\left|\left\{p' \in P' : \exists p \in P \ \left(p \equiv p'\right)\right\}\right|}{\left|P\right|}$$

## 10.6.2   Extended Labelled Attachment Score

For hybrid parsing, an edge in a parsed graph $G' = (V', E', L')$ is considered to be correct if it has an equivalent edge in the reference graph $G = (V, E, L)$. Two edges are equivalent if they have the same edge label, and connect equivalent vertices. A vertex $v \in V$ may represent a morphological segment, a phrase node or an empty category. Consider the vertex equivalence relation $v \equiv v'$ defined to be true when $v$ and $v'$ are both the same segment. For two vertices that are phrases ($v = p \wedge v' = p'$), the same phrase equivalence relation $p \equiv p'$ can be used from the Parseval metric. For ellipsis, two vertices are equivalent if they have the same POS tag and surface form. For two edges, $e$ from $v$ to $H(v)$, and $e'$ from $v'$ to $H'(v')$, let the edge equivalence relation be defined as:

$$e \equiv e' \Leftrightarrow v \equiv v' \wedge H(v) \equiv H'(v') \wedge l(e) = l(e').$$

For hybrid parsing, the ELAS precision and recall scores are then defined as:

$$\text{Precision} = \frac{\left| \{ e' \in E' : \exists e \in E \ (e \equiv e') \} \right|}{|E'|}$$

$$\text{Recall} = \frac{\left| \{ e' \in E' : \exists e \in E \ (e \equiv e') \} \right|}{|E|}$$

For pure dependency graphs, ELAS recall is the same as LAS. For an edge between phrases, a Parseval-like measure is used for the two phrase nodes.[34]

---

[34] ELAS imposes a strict metric for measuring partially correct hybrid analyses involving phrase structure. For example, because ELAS is defined over edges, two partially correct phrases without a corresponding correct edge connecting them would receive no credit using this metric.

### 10.6.3  Cross-Validation

In contrast to the methodology for evaluating previous parsers, ELAS is used as the evaluation metric for measuring the performance of HSP in both the integrated and multi-step parsing experiments. In addition to using a hybrid metric, the evaluation methodology also accounts for the size of the treebank. In previous work for benchmarking state-of-the-art parsing systems for English, parsers are generally trained using standard sections of the Penn English Treebank, and then evaluated using different standard sections. For a smaller treebank such as the Quranic Treebank a different approach is required.

To reduce sample bias, cross-validation is used. In this process, each round of cross-validation involves partitioning the treebank into different training and evaluation sets. Using 10-fold cross-validation, the experiments were repeated 10 times. In each fold, a different 10% portion of the data is used for evaluating the model, with the remaining 90% of the data used for training. F1-scores are then calculated by aggregating the total number of true positives and false positives across the ten folds. Forman and Scholz (2009) report that this method is more effective than other aggregation methods for cross-validation.

## 10.7    Parsing Results

### 10.7.1  Multi-Step and Integrated Parsing

This section discusses parsing results. Table 10.3 (overleaf) shows the results for the two parsing approaches. Using the best performing feature set, HSP achieves an F1-score of 87.47% for the multi-step approach, and 89.03% for the integrated approach. This high performance may not only be due to the treebank being annotated with rich morphological features or the choice of algorithms. The Quranic text contains many examples of syntactic and stylistic repetition (Salih, 2007). Repetition leads to an easier machine learning problem, as fewer non-standard cases are encountered during training.

| **Features** | **Multi-step Parser** | | | **Integrated Parser** | | | **F1-Diff** |
|:---:|:---:|:---:|:---:|:---:|:---:|:---:|:---:|
| | Precn | Recall | F1-score | Precn | Recall | F1-score | |
| POS | 76.73 | 74.38 | 75.54 | 78.28 | 75.01 | 76.61 | +1.07 |
| Morph6 | 82.52 | 79.74 | 81.10 | 84.62 | 80.64 | 82.58 | +1.48 |
| Morph9 | 86.98 | 85.32 | 86.14 | 89.42 | 86.35 | 87.86 | +1.72 |
| **Lemma** | **88.42** | **86.54** | **87.47** | **90.98** | **87.16** | **89.03** | **+1.56** |
| Phi | 88.23 | 86.35 | 87.28 | 90.87 | 87.02 | 88.90 | +1.62 |

Table 10.3: Accuracy scores for hybrid parsing using different feature sets.

These results should also be compared to the rule-based parsing approach used for initial syntactic annotation of the treebank. In Table 7.2 (page 150) this component was estimated to have an F1-score of 78%, with 91% precision and 68% recall. It is interesting to note that although the hand-written parser had comparable precision to the statistical parser (91% compared to 90.98%), its recall was far worse (68% compared to 87.16%). This demonstrates that although hand-written rules may be accurate, a large number of rules that cover increasingly smaller number of cases are required to produce sufficient coverage for overall accurate parsing. In contrast, a statistical model can more easily learn from the many cases available in the treebank.

## 10.7.2  Effect of Different Feature Sets

For statistical parsing, the five feature sets in Table 10.3 give different results. It is surprising that the Pos feature set alone is already a good baseline. Using no morphological features and only part-of-speech tags, this feature set produces scores of 75.54% and 76.61% for the two approaches respectively. One explanation for this is the fact that the treebank uses a detailed part-of-speech tagset, with 44 tags. For example, many of the particle tags that are based on traditional Arabic grammar are used for words with specific syntactic functions.

However, all five feature sets use the same graph features defined in section 10.5. In a further experiment without using these graph features to estimate the probability of decision histories, accuracy for the baseline Pos feature set dropped to only 21.64%. This is because the graph features provide constraints on possible dependencies. For example, the Deprel features stop additional edges being formed where these would not make sense based on examples in the training data, such as multiple subjects for the same verb.

The next set Morph6 adds voice, mood, case and state. The improvement over the Pos feature set is 5.56% for the multi-step approach and 5.97% for the integrated approach. This is consistent with recent work for parsing Modern Standard Arabic. Marton et al. (2010) use a similar set of morphological features to improve parsing accuracy for the Columbia Treebank (Habash and Roth, 2009c). The next set Morph9 further improves performance by adding segment and copula features.

### 10.7.3  Comparison with Modern Arabic

The work described in this chapter contrasts with recent work for parsing Modern Arabic using both constituency and dependency representations. For example, for Arabic constituency parsing, Kulick et al. (2006) discuss parsing the Penn Arabic Treebank using phrase structure grammar. One conclusion that can be drawn from their results is that parsing using a constituency representation leads to lower accuracy for Arabic in comparison to English. They report a Parseval F1-score of 74% for version 1 of the Penn Arabic treebank, and 88% for English using a similar sized corpus, trained using Bikel's parser (Bikel, 2004b).

In contrast, this work is more similar to dependency parsing work for Modern Arabic. In traditional Arabic grammar, the basic unit of analysis is the morphological segment and compound word-forms are segmented into independent grammatical units. This agrees with other recent treebanking efforts for Modern Arabic using dependency representations such as the Columbia Arabic Treebank (Habash and Roth, 2009c), the Prague Arabic Dependency

Treebank (Smrž et al., 2008), and the Penn Arabic Treebank (Maamouri et al., 2004). However, in contrast to these other Arabic dependency treebanks that define their own segmentation schemes, morphological annotation in the Quranic Treebank closely follows segmentation rules from *i'rāb*, and as a consequence is more fine-grained. In addition to part-of-speech, the grammar describes multiple features at morpheme-level, including person, gender, number, verb mood, noun case and state. The fine-grained annotation scheme may be one contributing factor to improved performance, in addition to the use of a hybrid representation.

In comparison to parsing Modern Arabic, the best feature set is Lemma, which boosts performance by a further 1.33% and 1.17% respectively over Morph9. However, the feature set Phi that adds person, gender and number, surprisingly degrades performance by 0.19% and 0.13% for the two approaches. This differs from recent work for parsing the Columbia Arabic Treebank (Marton, Habash and Rambow, 2013), where the phi-features have been shown to be helpful. It can be concluded that adding these features may not be statistically significant for parsing the Quranic Treebank using 10-fold cross-validation, or that this last feature set possibly includes too many features for the SVM models, given the relatively smaller size of the current version of the treebank.

## 10.7.4  Effect of the Conversion Process

The results in section 10.7.1 show that the integrated parser outperforms the multi-step parser for all of the five feature sets. However, it is interesting that the absolute difference between the two F1-scores consistently lies in the narrow band $1.4 \pm 0.32$. This suggests that the two parsers have similar sensitivities to feature selection.

Another factor affecting the performance of the multi-step parser is the accuracy of the conversion process from the hybrid representation to pure dependency, and then back to hybrid. One example of complexity that is not handled in the conversion process is the combination of nested phrase structure and non-projective dependencies. In the treebank, phrase nodes are used to model constituency structure. In a pure dependency representation, the grammatical relationship between a pair of phrases is implicit in the edge that connects the head words of the two phrases. In the traditional Arabic grammar of the Quran, phrase-level relations such as conjunction and apposition are made explicit in syntactic analysis. Since the grammatical rules that determine these phrase structures allow recursion, the Quranic Treebank includes hybrid graphs that contain multiple levels of nested consistency structure, occasionally with non-projective dependencies. However, the rule-based conversion algorithm outlined in section 9.4 correctly recovers 94.81% of edges. Although it might have been possible to improve the accuracy of the conversion process, this would have required a larger set of more complex rules for uncommon structures, such as the few cases of non-projective edges in the treebank, or for semantic ellipsis.

To measure the effect of the conversion process, a further experiment was performed. All graphs that did not have a perfect reversible conversion to pure dependency were excluded from the treebank (~ 8% of all graphs). The 10-fold cross-validation tests were then repeated using the best performing configuration for both approaches, the Lemma feature set. On this subset of the data, the multi-step parser achieved an F1-score of up to 88.89% (89.33 precision, 88.45 recall), and the integrated parser's F1-score was up to 90.24% (91.48 precision, 89.03 recall). The difference between the two F1-scores was +1.35, which lies in the same narrow band of $1.4 \pm 0.32$.

These results suggest that the absence of a conversion process is not the largest contributing factor to integrated parser's improved performance. Although additional investigation into optimizing the multi-step parsing algorithm could be further pursued, this may have diminishing returns. In contrast, the integrated approach is not only simpler as there is no conversion, but is also better suited to the hybrid representation in the treebank.

## 10.8    Conclusion

This chapter presented the first results for statistically parsing Classical Arabic. In this evaluation, the Quranic Treebank was parsed using HSP, a new hybrid statistical parser developed specifically for this task. This chapter also defined a new extended labelled attachment score (ELAS) for measuring the performance of hybrid dependency-constituency parsers. Two parsing algorithms were compared using different sets of rich morphological features. Out of the two approaches, the integrated shift-reduce algorithm is able to parse hybrid syntactic representations using a one-step process.

This work showed that accurate statistical parsing results for Classical Arabic are achievable using a hybrid syntactic representation. Based on the performance metrics, it can be concluded that the novel integrated algorithm is not only more elegant, but that encoding information this way improves performance, resulting in a 1.6% ELAS absolute increase over the multi-step baseline for the integrated approach. Although not directly comparable due to different training and evaluation datasets, these parsing results contrast with recent work for Modern Arabic, which suggests an improvement over pure constituency models. In comparison to the feature sets recently used for Modern Arabic dependency parsing, the same improvements were gained, with the interesting exception of the use of the Classical Arabic phi-features.

The problem presented in this chapter is an extension of the 2007 CoNLL shared task for pure dependency parsing, in which gold-standard morphological annotation was used as input (Nivre et al., 2007a). Morphological disambiguation is an important component of the hybrid parsing architecture. One factor not considered in the experiments is the effect of using predicted morphological input. However, Marton et al. (2010) show that for Modern Arabic at least, parsing using predicted instead of gold morphological input gives similar results across multiple feature sets. This additional extension to the parsing task, together with joint morphosyntactic disambiguation is described as further work in Chapter 12.

Part V: Further Work and Conclusion

# 11     Uses of the Quranic Arabic Corpus

## 11.1    Introduction

Part V of this thesis consists of two chapters. This chapter describes relevant work
that has used the Quranic Arabic Corpus since its initial publication. Although
several studies have cited the corpus as related work, this chapter highlights
examples that have made use of the gold-standard datasets presented in this thesis.
In Chapter 12, the main contributions of the thesis are summarized and
suggestions for future work are described.

## 11.2    Part-of-Speech Tagging

One use of the Quranic Arabic Corpus is as a dataset for machine learning. It is
attractive as a resource because it has been manually verified, and is one of the
few such resources for Arabic that is open source and freely available. In one of
the first studies of its kind for Classical Arabic, Alashqar (2012) uses the corpus
as a gold-standard dataset to compare the performance of different part-of-speech
taggers. Using the Natural Language Toolkit (NLTK) for system implementation,
he tests *n*-gram models, the Brill tagger, a Hidden Markov Model (HMM) and the
TnT tagger. In order to make the results more easily comparable to Modern
Arabic, experiments are performed using text from the corpus with and without
diacritics. 97% of the annotated data was used for training, with the remaining 3%
reserved for testing. Morphological segmentation was not considered in these experiments, so that the POS tag for each word refers to the tag associated with each word's stem. To simplify the learning problem, the Classical Arabic POS tagset (described in section 5.4) is additionally mapped to second tagset with only 9 tags. Table 11.1 below lists the results of the experiments.

The best performing tagger was the Brill tagger, with 83.2% accuracy using undiacritized text and the reduced tagset. In his conclusion, Alashqar notes that using Unicode as an orthographic representation for Classical Arabic script affects tagging accuracy, particularly for the Brill tagger. This is because tagging systems originally designed for languages such as English process Unicode diacritics in Arabic script as additional characters, increasing ambiguity during training. An alternative approach to Unicode is the new character-plus-diacritic representation for Arabic script presented in Chapter 4 of this thesis. It would be interesting to repeat these experiments using this representation to measure the resulting effect on tagging accuracy.

| Dataset | Accuracy Scores | | | | | |
|---|---|---|---|---|---|---|
| | Unigram | Bigram | Trigram | Brill | HMM | TnT |
| Diacritized | 80.0 | 80.1 | 80.0 | 36.4 | 72.5 | 64.9 |
| Undiacritized | 80.4 | 80.5 | 80.3 | 80.9 | 75.2 | 69.2 |
| Diacritized (9 tags) | 81.9 | 82.0 | 81.8 | 38.6 | 75.4 | 50.9 |
| Undiacritized (9 tags) | 82.5 | 82.3 | 82.4 | 83.2 | 77.5 | 59.0 |

Table 11.1: Accuracy scores for different Classical Arabic POS taggers.

In related work, Rabiee (2011) retrains the Stanford POS tagger using data from the Quranic Arabic Corpus. Although preferring to work with an annotated corpus for Modern Arabic, he notes that this dataset is one of the only open source tagged corpora for Arabic that has been manually verified and annotated to gold-standard level. In his research project, he uses a tagger trained against Classical Arabic to automatically annotate Modern Arabic texts.

Not many studies have been performed for the related task of automatic morphological segmentation for Quranic Arabic. However, Yusof et al. (2010) consider the stemming problem, a sub-task of full segmentation. In their approach, they propose a rule-based stemmer designed to recognize Arabic word patterns and extract stem segments. Testing on data from the Quranic Arabic Corpus, they report an average accuracy of 62.5%. In their error analysis, they conclude that the biggest challenge to their task is processing out-of-vocabulary words.

Albared et al. (2011) describe an alternative approach for statistically POS tagging the Quranic corpus, focusing adapting hidden Markov models to reduce out-of-vocabulary errors. They note that the morphological data in the corpus is a challenging to use for training statistical taggers due to a relatively high number of words appearing with low frequency. They propose smoothing methods together with a new lexical model for Classical Arabic that tags out-of-vocabulary words through linear interpolation of lexical probabilities. As with Alashqar's comparative study, they assign a single POS tag to each compound word-form, and do not consider full morphological segmentation.

In their experiments, they use 90.1% of the corpus for training, and reserve 9.9% for testing. Using this split, 14.9% of words in the test set are unknown (previously unseen). In their best performing HMM configuration, they report 85.3% tagging accuracy for unknown words, and 95% tagging accuracy overall. They are able to boost accuracy significantly using their lexical interpolation model. Their reported unknown word POS tagging accuracy is one of the best results to date for either Classical or Modern Arabic.

In related work, Khaliq and Carroll (2013) consider unsupervised learning of morphological forms, using the Quranic Arabic Corpus as a training and evaluation dataset. Working with an undiacritized version of the corpus, they train a maximum entropy classifier using orthographic features. They report an accuracy score of 73.8% for root identification of Classical Arabic word-forms, compared to an accuracy score of 63.1% for a simpler baseline system.

## 11.3    Syntactic Annotation

The Quranic Treebank is the first treebank for Classical Arabic. It contrast to work for Modern Arabic, it is also the only dependency-based Arabic treebank to annotate elliptical structures. Several recent dependency treebanks for other languages have referenced the Quranic Treebank in comparative work and employed similar solutions for annotating empty categories. Examples include Gasser (2010) for Amharic (another Semitic language related to Arabic) , Lee and Kong (2012) for Classical Chinese and Haverinen et al. (2013) for Finnish.

In related work, Seeker and Kuhn (2012) develop a new dependency treebank for German by automatically converting the TIGER treebank from a phrase-structure representation. In their annotation scheme, they explicitly include empty categories in their dependency representation, similar to the approach used for the Quranic Treebank. However, although this produces richer linguistic structures, they note that their format introduces additional complexity to statistical parsing work. This challenge was addressed by the novel integrated parser described in Chapter 9, which was intentionally designed to handle elliptical structures:

> [Elliptical structure] poses problems since today's statistical dependency parsers are not capable of handling empty nodes. Empty nodes create the problem that the number of nodes that the parser has to connect in order to arrive at a dependency structure is no longer determined by the number of tokens in the sentence. This is however one of the fundamental assumptions in dependency parsing, and the algorithms are built upon this. Recently, a parser has been proposed by Dukes and Habash (2011) that extends the transition-based paradigm for dependency parsing by adding an additional move to the parser that introduces empty nodes into the tree. As far as we know, this is the only published dependency parser so far that can handle empty nodes directly during the parsing process.

## 11.4    Quranic Pronominal Anaphora

Further work using the Quranic Arabic Corpus has focused on enhanced annotation, beyond the syntactic level. For example, the annotated corpus of Quranic pronominal anaphora (Sharaf, 2012a; Sharaf and Atwell, 2012b) relates over 24,000 pronouns in the Quran to their antecedents. Sharaf notes that identifying pronominal anaphora in both Modern and Classical Arabic is more complex compared to English due to Arabic's rich morphology. Pronouns can occur as individual words, but frequently appear as clitics attached to nouns and verbs as suffixes. To simplify the annotation process, the anaphora corpus uses the annotated morphological segmentation from the Quranic corpus to identify tagged pronouns.

   As described in Chapter 5, pronouns in the corpus are tagged using the PRON, DEM and REL tags. In the methodology described by Sharaf and Atwell (2012b) only the PRON tag for personal pronouns is used. Demonstrative (DEM) and relative (REL) pronouns are excluded from their annotation effort (approximately 15% of all pronouns) as these are few in number and have antecedents which are often non-anaphoric. They conclude that anaphoric tagging for the Quran is challenging due its stylistic use of language. They report that the distance between pronouns and their antecedents can be large. Only 2,309 pronouns (17.5% of all pronouns with antecedents) were related to the previous noun, with many preferring to antecedents up to 200 words away, and some as far as 33 verses away. Similar to the Quranic Arabic Corpus, their annotated dataset is made free available online.

## 11.5    Prosodic Analysis

Brierley et al. (2012) describe a novel approach to prosodic analysis for Arabic by introducing a boundary annotation scheme based on the traditional recitation mark-up (*tajwīd*) found in the Quran. Using compulsory and recommended recitation stops found in the Quranic script, they build a prosodic dataset which is then merged with the POS-tagged data from the Quranic Arabic Corpus, to train a tagger that chunks sentences with prosodic boundaries. Interestingly, they report that their Classical Arabic tagger produces break marks that are similar to the corresponding punctuation marks found in English translations.

## 11.6    Knowledge Representation

The Quranic Arabic Corpus provides a highly accurate version of the Arabic text of the Quran, sourced from the Tanzil project. This data is made available through JQuranTree, a new computational interface presented in Chapter 4. Several projects have used this interface to access the text of the Quran for verse similarity work. For example, Ali (2012) uses JQuranTree to construct a lexical graph where words are nodes and edges correspond to distinct bigrams. This word graph is used to build an automatic subject index to cluster related verses.

In a different study, Sharaf and Atwell (2012c) describe QurSim, an annotated dataset for the Quran. They use morphological data from the Quranic Arabic Corpus (lexical roots) together with traditional sources of exegesis to construct a verse similarity index. They note that because the Quran often describes a common topic across many different verses, resources that annotate verse similarity may be of use to researchers who want to easily access information related to a single theme.

Other knowledge representation projects have focused on formal ontologies. For example, Zaidi et al. (2012) attempt to construct an ontology automatically using lexical collocations from the Quran. One challenge they discuss is the relatively small size of the Quranic corpus as a resource for lexical semantics. For example, many words occur as hapax legomena in the Quran. In their evaluation, they compare an automatically constructed set of concepts to the ontology manually developed in this thesis (described in section 8.4.3). They note that they require further work to produce an accurate automatic ontology of comparable quality. One recommendation they make is to supplement the linguistic data in the Quranic corpus with lexical data from other related Arabic corpora.

In a different approach, Yahya et al. (2013) use the ontology in the Quranic Arabic Corpus as part of an information retrieval system for Quranic concepts. Their system is designed to handle natural language queries in both English and Malay. They manually translate the Quranic ontology into Malay for this purpose. Other projects have also extended to the Quranic ontology. For example, Yauri et al. (2013) convert the data into the more standard Web Ontology Language and enrich the ontology by adding concepts more relevant to question answering.

Boella (2011) considers the problem of automatically relating knowledge in the Quran to the information found in hadith (the collected sayings of the prophet Muhammad). Using the lemma tags from the Quranic corpus together with a set of regular expressions, Boella describes CrossQuran, a computational system that automatically provides cross references between these two corpora.

Another knowledge-related technique is semantic role labelling. Zaghouani, Hawwari and Diab (2012) propose the Quranic Arabic PropBank: a semantic role labelling project for the Quran. In their preliminary report, they consider the task of annotating roles for the 50 most frequently occurring verbs in the Quranic Treebank. They estimate that once complete, this project will supplement the Modern Arabic PropBank with frame definitions for approximately 810 new verbs. They note that the in contrast to role labelling for Modern Arabic, which has previously used constituency structure, the dependency representation used in the Quranic Treebank may be better suited to semantic annotation:

> Having the Quranic corpus annotated using a dependency structure treebank has some advantages. First, semantic arguments can be marked explicitly on the syntactic trees, so annotations of the predicate argument structure can be more consistent with the dependency structure. Secondly, the Quranic Treebank provides a rich set of dependency relations that capture syntactic-semantic information. This facilitates possible mappings between syntactic dependents and semantic arguments.

## 11.7    Supervised Collaboration

Rebdawi et al. (2013) cite the annotation model for the Quranic Arabic Corpus as a source of inspiration for developing a related online platform for collaboratively constructing an Arabic-to-English dictionary. Similar to the roles of annotators and editors described in Chapter 7, Rebdawi et al. introduce roles of dictionary users and lexicographers in their annotation model. They also use a similar three-tier architecture for their website implementation and similarly use JSP pages in their presentation tier. Their annotation platform encourages volunteers to enrich dictionary entries under the supervision of expert lexicographers. They aim to build a highly accurate online resource for Arabic-to-English word meanings using the annotation methodology of supervised collaboration.

## 11.8    Translation Studies

In addition to annotation efforts and using its datasets for machine learning, another interesting use of the corpus is for improving translations of the Quran. Previously, as accurate morphological and syntactic data for the Quran was not available, translation work was not able to easily take advantage of techniques from corpus linguistics. The Quranic Arabic Corpus encourages such an approach. For example, Younis (2012) performs a study of translation using morphological data. Using the search tools available through the website, she cites examples of verbs with different morphological forms that have been rendered as equivalent in major English translations of the Quran. However, in Arabic the different varieties of verb forms convey often subtly different semantic information. In one example, she discusses the morphological tagging of the triliteral verbs *nazzala* (نزَّل) (tagged as POS:V II in the corpus) and *anzala* (أنْزَل) (tagged as POS:V IV). These have different forms yet are usually given the same translation 'revealed' in English, ignoring the subtle distinction between the two. Younis concludes that the new morphological tagging in the corpus may be of use for producing more accurate translations of the Quran in future.

In a related study, alQinai (2011) considers the nature of synonyms in Quranic translation, and notes that their various interpretations have led to different translations of the text. Using the Quranic Arabic Corpus, he gives examples of well-known translations of the Quran that could be improved by taking into consideration the collocation of reoccurring polysemous words. As with Younis' study, alQinai points to morphological data to highlight semantic differences.

Tabrizi and Mahmud (2013) similarly use the corpus to compare translations of the Quran. They suggest that improvements to translation could focus on entity coherence and lexical cohesion. They note that pronoun resolution and word and phrase ordering are structural issues in translation that the Quranic Treebank may help to resolve in future translations of the Quran into other languages.

## 11.9 Conclusion

From a computational linguistics perspective, the Quranic Treebank has had an impact on recent research by becoming the fourth major treebank for the Arabic language, and is used as a gold-standard dataset for benchmarking statistical taggers for Arabic. The dependency-based parser presented in this thesis has also been noted as the first of its kind for elliptical structures. In addition, recent work has suggested that the new grammatical annotations may be of use for developing more accurately constructed translations of the Quran into other languages.

From an educational perspective, although the treebank is primarily used online, it has also been used as an educational resource in an offline context. For example, Almenoar (2010) reports on using the hybrid graphs in the treebank as a visual aid for teaching Arabic learners at undergraduate level, with improved results. In addition to cited research, the Quranic corpus is widely referenced online. The website includes a page with feedback from general users and academic researchers.[35] The next chapter discusses how these suggestions could be incorporated into future work, to improve the resource for further research.

---

[35] http://corpus.quran.com/feedback.jsp

I know that great, interesting, and valuable discoveries
will be made... more interesting discoveries will be
made than I have the imagination to describe – and I am
awaiting them, full of curiosity and enthusiasm.

*– Linus Pauling*

# 12    Contributions and Future Work

## 12.1    Introduction

Computational linguistics is an interdisciplinary field that applies concepts from
computer science and linguistics to model natural language. However, natural
language is by its very nature uniquely human and deeply complex. It is often
convenient to make simplifying assumptions about the nature of language. For
example, assuming that the syntactic structure of sentences can be modelled using
dependencies between pairs of words with a single root leads to mathematically
elegant dependency trees. Parsing algorithms using this representation are in turn
more comprehensible and easier to implement than they otherwise would be.

  This thesis adopts a radically different approach to syntax. For Classical Arabic,
grammarians have had over 1,000 years to conceptualize and perfect a model for
sentence structure. Unconstrained by notions of algorithmic complexity or
computability, they focused on developing a rich linguistic framework. Instead of
starting with a preconceived mathematical structure and applying it to natural
language, this thesis instead takes an existing grammatical system as its starting
point and uses it to construct a new formal representation of syntax. This chapter
discusses the consequences of this approach and is organized as follows. Section
12.2 presents the main contributions of this work. Section 12.3 describes its
limitations and section 12.4 discusses the implications of the main findings.
Finally, section 12.5 concludes with recommendations for future research.

## 12.2 Summary of Contributions

At the outset of this thesis, three research questions were asked:

1. Can crowdsourcing be used for annotating Arabic?
2. Is a hybrid representation suitable for parsing?
3. Is statistical parsing viable for Classical Arabic?

This thesis presents novel contributions to knowledge through answering these research questions. Firstly, for the first research question of annotation:

- A new methodology of supervised collaboration for Arabic was presented, including the first evaluation for online Arabic annotation. The completed morphological layer of the corpus was found to have a high accuracy score of 98.7% compared to gold-standard grammatical reference works.

- LAMP is a new Linguistic Analysis Multimodal Platform used to access and improve annotations online. Designed to be scalable and robust, it is used for the Quranic Arabic Corpus website (http://corpus.quran.com), with over 2 million users per year.

- The website also includes novel components for visualizing dependency graphs, producing phonetic transcriptions and automatically generating grammatical summaries, as well as a morphological dictionary and a new ontology of Quranic concepts linking to named-entity annotations.

Secondly, this thesis describes a new formalism for Classical Arabic consisting of orthographic, morphological and syntactic layers. When combined, these form a novel hybrid dependency-constituency representation:

- JQuranTree is a new component for Arabic orthography, based on a novel character-plus-diacritic alternative to Unicode for accurately representing the complex Uthmani script of the Quran. This representation is also faster and more memory efficient than Unicode for Arabic text searches.

- The morphological representation presented in this thesis is the first of its kind for Classical Arabic. Based on a lexeme-plus-feature representation, it is the first annotation scheme for morphemic segmentation and part-of-speech tagging specifically designed for Classical Arabic.

- The hybrid syntactic representation is the first formal specification for either Modern or Classical Arabic that is closely aligned to traditional grammatical theory. The Quranic Treebank is the first treebank for Classical Arabic, as well as the first dependency-based treebank for either Modern or Classical Arabic that annotates hybrid and elliptical structures.

Thirdly, the syntactic representation is used in combination with a novel parser to determine if hybrid statistical parsing is achievable for Classical Arabic:

- This thesis presented HSP, a new Hybrid Statistical Parser. This is the first statistical parser for Classical Arabic, as well as the first parser for either Modern or Classical Arabic that is able to construct hybrid dependency-constituency structures. It is also the first dependency-based parser in any language for elliptical structures.

- A contribution of this thesis to parsing knowledge is that accurate hybrid parsing is achievable. For Classical Arabic, HSP was evaluated using a new ELAS (Extended Labelled Attachment Score) metric for hybrid parsing. HSP achieved an F1-score of up to 89.03%, compared with up to 87.47% for a pure dependency parsing model with post-processing.

## 12.3   Challenges and Limitations

This section discusses the main challenges found during the research. Some of these required rethinking approaches or redesigning experiments, whereas other challenges remain limitations of the study and were not addressed. Of the challenges that were solved, a difficult problem was constructing a high quality annotated corpus without funding. In contrast to the three other major treebanks for Modern Arabic, it was not possible to gain access to funds for annotating Classical Arabic within the timescales of the project. Without access to paid linguistic experts, an alternative methodology of supervised collaboration was devised. An initial experiment using Amazon Mechanical Turk (section 7.5.4) showed that annotating the Quran via crowdsourcing was possible, but that volunteer experts were needed as supervisors to guarantee accuracy for deep linguistic tagging using traditional Arabic grammar.

Adopting Arabic grammatical theory as an annotation framework also required developing a new syntactic formalism. The first part of the Quranic Treebank was initially annotated as pure dependency, inspired by recent dependency projects for Modern Arabic (Habash and Roth, 2009c; Hajič et al., 2004). However, online annotators who are familiar with traditional grammar were often confused by the dependency approach to coordination and prepositional phrase attachment. The initial lack of elliptical annotation was also problematic when attempting to reconcile the treebank to traditional sources (Salih, 2007). Introducing the hybrid representation solved these issues as it was found to be strongly preferred by online annotators because of its increased linguistic expressivity. However, from a computational perspective, it was found that the new syntactic representation would not easily work 'out of the box' with existing annotation tools and parsers. A new annotation platform for offline and online correction was developed for this purpose (Chapters 7 and 8). Machine learning experiments using treebank data also had to be redesigned after abandoning the pure dependency approach. However, despite its increased complexity, it was shown that computational tasks such as parsing are achievable using new algorithms (Chapters 9 and 10).

In contrast, a number of challenges encountered during the research were not addressed, and remain open questions. From an annotation perspective, certain verses of the Quran are challenging due to variant readings. This not only arises because of general variations of opinion, but can also occur due to more fundamental differences in grammatical analysis. An example is the contrast between Islamic Sunni and Shia schools of thought as to the correct method for ritual washing before prayer. The Sunni view is that the head and feet should be washed, whereas the Shia view is that they should be only wiped. Interestingly, these religious rulings depend on choosing different head words for a conjunctive dependency in verse (5:6) of the Quran. To simplify the annotation process, the Quranic Treebank annotates only a single reading for each verse. This decision was made independently of semantics, on the grounds that annotating multiple readings would be too time consuming for the first version of the treebank. When conflicts of opinion arise that are also backed by different gold-standard analyses from grammatical reference works, a majority of consensus is usually sought. As the first version of the treebank lacks multiple variant readings, the Sunni analysis was chosen for verse (5:6) as it is more mainstream (Sunnis form up to 90% of the Islamic population). Although variant readings are sometimes included in corpora such as the Penn POS-tagged version of the Brown Corpus (Atwell, 2008), an open question remains on how best to integrate variant readings into the treebank.

From a computational perspective, another limitation of the thesis is separate morphological and syntactic disambiguation. Recent dependency parsing work for Modern Arabic assumes a pipeline approach in which gold-standard or predicted morphological data is used as input for a statistical dependency parser (Nivre et al., 2007a; Marton et al., 2013). This contrasts with Hebrew, a related Semitic language, where state-of-the-art parsing has moved to joint morphological and syntactic disambiguation, with performance improvements over the pipeline approach (Goldberg and Elhadad, 2011). For hybrid dependency-constituency parsing, the joint disambiguation task may be more complicated, but is nonetheless still a much needed approach for both Modern and Classical Arabic. This is task is discussed further in section 12.5 as recommended future work.

## 12.4    Implications

### 12.4.1  Syntax and Semantics

Using Arabic grammatical theory as a starting point for a new syntactic formalism has implications for theoretical and computational linguistics. From a theoretical perspective, this research impacts the ongoing debate on the suitability of various syntactic representations for different natural languages. One viewpoint is that the major syntactic representations are equivalent as they differ only by focusing on different aspects of sentence structure. This thesis adopts the alternative view that different representations encode fundamentally different linguistic information. For example, it is well known in the parsing research community that the seminal Collins parser, trained using the constituency representation in the Penn Treebank, crucially uses head-finding dependency rules (Collins, 1999; Bikel 2004b). The parser requires hand written heuristics to add dependencies and enrich the representation to achieve state-of-the-art parsing accuracy. This shows that both constituency and dependency information are relevant for parsing English.

For Classical Arabic, the situation is similar. This thesis showed that a hybrid representation is more linguistically expressive than either a pure dependency or a pure constituency representation, when aligning to traditional analysis. Although not directly comparable due to different test sets and forms of language, the performance scores reported in Chapter 10 are higher than both dependency and constituency parsers for Modern Arabic (Marton et al., 2013; Green and Manning, 2010). An interesting question is whether or not Arabic grammatical theory has a universal validity and is applicable to other forms of language. Successfully applying the hybrid representation to Classical Arabic implies that it should at least extend to Modern Arabic, where it may improve parsing results and related computational tasks. It may also apply to languages such as English, as a hybrid approach more naturally represents known issues with pure dependency, such as coordination and prepositional phrase attachment (Nivre, 2005).

The work in this thesis showed that Arabic grammatical theory integrates approaches also used in modern linguistics. The concepts of structure, part-of-speech tagging, morphological segmentation, constituency analysis, governance and dependency have been widely known and developed by Arabic grammarians for over a thousand years. Although Arabic grammar is considered to be one of the origins for modern dependency theories (Versteegh, 1997b), both linguistic frameworks have developed relatively independently. As such, for both modern linguists and historical Arabic grammarians to develop a similar set of concepts is both remarkable and points to a universal conception of grammar. However, in contrast to Arabic theory, modern approaches to syntax are strongly influenced by mathematical notions of elegance, computability and formal logic. In contrast, Arabic grammarians adopt a different approach, as they are primarily concerned with analysing the correct form of speech, or 'the way of speaking' (*'ilm an-naḥw* – علم النحو), without attempting to constrain the complexity of syntactic models, or restrict grammar to simplified mathematical structures.[36] This implies that because language is complex, complex formal approaches to syntax may be required in order to achieve the linguistic expressiveness exemplified by Arabic theory.

The work in this thesis also showed that *i'rāb* deals with semantics as well as syntax. Many dependency relations in the grammar are closer to semantic roles than purely syntactic ones. For example, the many subtle distinctions of particles and their associated dependencies are often described using semantic as opposed to syntactic criteria. This implies that the representation presented in this thesis may also be a good starting point for semantic analysis. For example, tasks such as semantic role labelling are simplified in the representation compared to other approaches for Arabic such as constituency (Zaghouani et al., 2012). The way in which adverbial constructions are classified by relating to concepts of time, space and circumstance are also remarkably similar to modern efforts for semantic annotation (Xavier et al., 2005).

---

[36] The term *naḥw* (نحو) originally meant correct speech, but was later used by grammarians as a technical term to refer to grammar as a whole (Carter, 2004; Versteegh, 1995).

## 12.4.2  Computational Resources

In contrast to the implications for theoretical linguistics outlined in the previous section, more specific implications can be said for the computational results. In this thesis, the hybrid representation was applied successfully to parsing. This implies that complex representations which are more plausible on linguistic grounds can still be computationally tractable. For example, although it is known that the best non-deterministic parsers outperform transition systems, one of the main findings of this thesis is that transition systems are extensible to more complex scenarios. The concept of developing an integrated hybrid parser by adding extra state transitions may be applicable to other tasks such as integrated morphological segmentation and part-of-speech tagging. The computational work in this thesis also covered other areas. Initial automatic morphological annotation was performed by adapting an analyzer to Classical Arabic (Buckwalter, 2002). This was achieved by mapping the representation used by a Modern Arabic analyzer to the tagging scheme designed for Classical Arabic. The approach applied for morphology implies that it may also be possible to adapt other computational resources, when representations can be aligned. This is needed because Classical Arabic is a less-studied language in computational linguistics.

For treebank construction, it was found that making the proofreading process as intuitive as possible improves accuracy. Related efforts for similar computational resources for other languages may benefit from the approach for Classical Arabic. This thesis showed that encouraging communication between annotators and providing a relevant suite of tools attracts potential volunteers. An implication is that lack of funds need not be a barrier to constructing annotated corpora. Devising a suitable annotation scheme, providing guidelines with examples and motivating annotators can produce results of comparable quality to paid experts. For the Quranic Arabic Corpus, the inclusion of expert supervisors was found to be a crucial element for the annotation model. This shows that for certain tasks, a good approach to annotation may be a combination of experts as well as general crowdsourcing workers to reduce costs and ensure quality, benefiting from the best of both approaches.

## 12.5   Future Work

Looking forward, two sources of inspiration for future work are continuations of the topics explored in this thesis, as well as extending the Quranic Arabic Corpus in response to its use in recent research.

### 12.5.1  Annotation and Parsing

The Quranic Arabic Corpus includes morphological annotation which has 100% coverage, as well as a syntactic layer, the Quranic Treebank, covering 50% of the Quran. By completing the treebank as recommended further work it will be possible to have the entire grammar of the Quran annotated in machine readable form. This would potentially enable several interesting projects. For example, in Chapter 11, recent work for benchmarking Classical Arabic POS taggers was described that use the corpus as training and test data. Completing the treebank would similarly allow for benchmarking parsers for Classical Arabic, as well as more generally benchmarking parsers for hybrid grammars, using a larger gold-standard dataset.

For more general annotation, one direction in which the work in this thesis may become reusable would be to extend the annotation platform to other languages. Many of its components, such as natural language generation, the message board discussion forum, and search tools are not necessarily specific to Arabic, and may be of interest to other annotation projects.

Another recommendation for future work is to improve the hybrid statistical parser. As discussed previously, the pipeline approach to morphological and syntactic disambiguation has limitations on accuracy. For morphologically-rich languages such as Arabic, morphology and syntax are closely related. Two approaches for joint parsing are adding extra operations to the transition system, or moving to a non-deterministic model. A non-deterministic parser is likely to produce superior results because Arabic is highly morphologically ambiguous. These approaches may be effective for both Classical and Modern Arabic.

## 12.5.2   Understanding the Quran

The Quran is also interesting as a knowledge resource. At present, up to a quarter of the world adheres to the Islamic faith, with projections indicating that this proportion is expected to increase (Kettani, 2010). There is a strong interest in understanding the Quran from a significant proportion of the world's population, the majority of whom do not speak Arabic. Atwell et al. (2010) have proposed understanding the Quran as a grand challenge for computer science and artificial intelligence. Having the syntactic structure of the Quran in machine readable form may be a good starting point to help drive knowledge-related projects forward.

Initial uses of the corpus to this end have included preliminary investigations into translation accuracy (alQinai, 2011; Younis, 2012; Tabrizi and Mahmud, 2013). Further recommended work involves building advanced search tools to enable translators to have better access to corpus annotations. From a semantic perspective, efforts to build on the syntactic annotation to construct a formal semantic layer are in progress (Zaghouani et al., 2012). As demonstrated by recent work, there is also demand for extending the Quranic ontology (Zaidi et al., 2012; Yahya et al., 2013; Yauri et al., 2013). A semantically annotated corpus may allow for useful applications such a question-answering system that responds to natural language queries by quoting relevant verses from the Quran.

## 12.6   Closing Remarks

Finally, it should be noted that Modern Arabic does not benefit from the same level of computational focus as languages such as English. Much work remains to be done for many computational tasks across morphology, syntax and semantics. For Classical Arabic, computational work is virtually non-existent. However, good progress has been made in recent years with a number of projects starting to improve the state-of-the-art for the Arabic language as a whole (Habash, 2010). Nonetheless, it is clear that we are only at the beginning of an exciting time for Arabic computational research, with many interesting discoveries yet to be made.

# Appendix A: Syntactic Visualization

The Quranic Arabic Corpus website presents syntactic annotation visually as dependency graphs. These are displayed using a color scheme allowing annotators to easily distinguish different parts-of-speech and dependencies. Because hybrid syntax is a novel form of annotation for Arabic, a new computational component for visualization was implemented using Java 2D, a Java framework for creating graphical images. This appendix describes the layout algorithm used. In addition to drawing primitives, two auxiliary data structures are used:

- **Visual tree**: Although Java 2D does not provide one, the layout algorithm implements a custom scene graph known as a visual tree. In this structure, leaf nodes are primitives (lines, circles, arcs, arrowheads and text), and non-leaves are containers (elements grouped and positioned together). The tree uses a box model so that nodes are specified as a tuple $(x, y, w, h)$, where the coordinates $(x, y)$ are relative to their parent, and $w$ and $h$ denote the width and height of each bounding box respectively (Figure A1, overleaf). During rendering, these coordinates are mapped to absolute image coordinates by recursing down the tree and adding offsets. Using relative coordinates allows containing bounding boxes to be easily calculated so that a group of elements can be positioned without having to modify an entire subtree.

- **Height map**: The visual tree for hybrid graphs is constructed downwards starting from the top of the image, so that words from the sentence are followed by a section that contains arcs and phrase structure. To determine the position of arcs and phrases, a height map is updated during layout. This is a list of spans $(x, w, h)$ where $x$ and $w$ denote the position and width of each span, and $h$ is the maximum height of the image rendered so far in that interval. Here, $h = 0$ is the top of the image.

Figure A1: Visual dependency graph with and without bounding boxes.

The layout algorithm uses a two-stage approach. In the first stage, the visual tree is constructed in-memory using a combination of measure and arrange steps:

1. At the start of the layout, word elements from the sentence are measured and arranged from right-to-left. Each of these is a hierarchical element consisting of a token location number, phonetic transcription, interlinear translation, Arabic script and POS tags.

2. As only words have been arranged at this point, the height map is initiated using a single span $(0, w, h)$ where $w$ and $h$ are the width and height of the canvas after step 1.

3. In this step, node points are calculated for POS tags. These are locations in the image in absolute coordinates that will form the ends of arcs.

4. Edges are sorted and added to the tree. If an edge connects two terminals these will be node points. Otherwise, new node points are calculated using phrase nodes, positioned at $(x, y)$ where $x$ is the midpoint between the terminals spanned by the phrase, and $y$ is calculated using the height map together with a margin. Arc heights are similarly calculated. Once new arcs and phrases are added to the visual tree, the height map is updated.

5. In a post-processing step, elements in the visual tree are sorted so that arcs are drawn first to avoid these overlapping edge labels.

After the visual tree is constructed, the second stage is for it to be rendered. For the website, an image file is generated and displayed online. Dependency graphs are also rendered by the offline annotation tool to modify syntactic tagging in the treebank. In addition, this tool is also used to view syntactic output for diagnosing the parser during its development (Figure 10.1, page 212).

# Appendix B: Phonetic Transcription

A phonetic transcription of Arabic script appears on the Quranic Arabic Corpus website in dependency graphs and in the word-by-word morphological analysis pages. In comparison to Modern Arabic, which is almost always written without diacritics, the Classical Arabic script of the Quran is fully diacritized so that its exact pronunciation is specified. The transcription in the corpus is generated automatically using a computational component developed specifically for this purpose. Because the encoding is designed to be readable to general users, it is not reversible. In contrast, a lossless but harder to read system is extended Buckwalter transliteration, used for computational work (presented in section 4.4.5).

| ص | ش | س | ز | ر | ذ | د | خ | ح | ج | ث | ت | ب | أ |
|---|---|---|---|---|---|---|---|---|---|---|---|---|---|
| *ṣ* | *sh* | *s* | *z* | *r* | *dh* | *d* | *kh* | *ḥ* | *j* | *th* | *t* | *b* | *ā* |

| ي | و | ه | ن | م | ل | ك | ق | ف | غ | ع | ظ | ط | ض |
|---|---|---|---|---|---|---|---|---|---|---|---|---|---|
| *y* | *w* | *h* | *n* | *m* | *l* | *k* | *q* | *f* | *gh* | *'* | *ẓ* | *ṭ* | *ḍ* |

Figure B1: Phonetic transcription for Arabic letters.

Figure B1 shows the transcription system for Arabic letters. However, there are exceptions to the transcribed phonemes shown in this diagram, described further overleaf. The computational implementation is based on the transcription for Quranic script summarized by Jones (2005). For example, long vowels are indicated by *ā*, *ī* and *ū*. In Quranic Arabic, the diacritic *madda* may also be used to lengthen a vowel. The implementation also has additional rules to handle *hamzat waṣl*, a diacritic mark used in the Quran to indicate a non-phonemic glottal stop. This is generally transcribed as *l-* except at the start of a verse where *al-* is used.

In its algorithm, the transcription component accepts an Arabic word as input in the character-plus-diacritic representation described in Chapter 4, together with morphological annotation. A lookup table is first used to check for special words. For example, disconnected letters (tagged as POS:INL) are transcribed separately. For regular words, a set of over 200 phonetic rules is applied to each character in the script. These rules use the previous and following characters as context. Three examples are listed below:

- If the current letter is *alif* with an attached *hamzat waṣl* diacritic, the next letter is *lam* with an attached *shadda*, and if the word is not tagged as POS:DEM, POS:REL or POS:COND, then output the phoneme *al-la*.

- If the current letter is *wāw* and the next is *alif* followed by a small high rounded zero, then output *ū*. For example in verse (2:188):

$$\text{لِتَأْكُلُوا} \;\rightarrow\; litakulū$$

- If the letter *alif maqṣūra* has an attached vowelized diacritic, then output *y* together with a long vowel. Otherwise, assume the letter is silent.

As an example of the component's output, Figure B2 shows verse (2:147) with the Uthmani script and a corresponding transcription. This verse illustrates long vowels as well as different phonemes for *hamzat waṣl*:

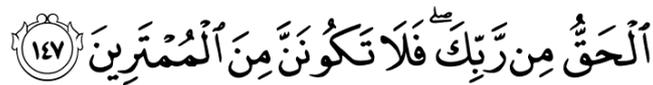

*al-ḥaqqu min rabbika falā takūnanna mina l-mum'tarīna*

Figure B2: Automatic phonetic transcription for verse (2:147).

# Appendix C: Language Generation

One use of linguistic annotation in the Quranic Arabic Corpus is to generate automatic summaries. These are more readable than formal tags and have been reported by users of the website to be easier to proofread. Grammatical summaries are produced in both English and Arabic using natural language generation. In this process, a sequence of templates are concatenated, with each template filled using morphological features. These templates are selected using the part-of-speech tags for each segment. To simplify the proofreading process, the frequently occurring determiner prefix segment *al-* (POS:DET) is not used.

An example of this process is illustrated by the compound word-form (29:69:4) *lanahdiyannahum* (لَنَهْدِيَنَّهُمْ), translated as 'We will surely guide them'. This word exhibits complex morphology with a prefix, a stem and two suffixes, using traditional segmentation rules. For summary generation, the tags for this word will be retrieved from the corpus database using feature notation:

[l:EMPH+ POS:V 1P MOOD:IMPF ROOT:hdy +n:EMPH +PRON:3MP]

Based on the segmentation implied by these tags, the following templates will be selected by the natural language generation algorithm for this example:

- The <X> word of verse <Y> is divided into <Z> morphological segments.
- <SEGMENT-LIST>
- The prefixed particle <X> is usually translated as <Y>.
- The <X> verb (<Y>) is <Z> and is in the <W> mood (<M>).
- The verb's triliteral root is <ROOT-LIST> (<ARABIC-LIST>).
- The suffixed <X> particle is known as <Y> (<Z>).
- The attached object pronoun is <X>.

   In these templates, placeholders with variable names are slots which are filled by hand written rules driven by feature tags. For the word-form (29:69:4), these templates are combined to produce the following summary:

> The fourth word of verse (29:69) is divided into 4 morphological segments. An emphatic prefix, verb, emphatic suffix and object pronoun. The prefixed particle *lām* is usually translated as 'surely' or 'indeed' and is used to add emphasis. The imperfect verb (فعل مضارع) is first person plural and is in the indicative mood (مرفوع). The verb's triliteral root is *hā dāl yā* (ه د ي). The suffixed emphatic particle is known as the *nūn* of emphasis (نون التوكيد). The attached object pronoun is third person masculine plural.